\documentclass[10pt,twocolumn,letterpaper]{article}

\usepackage[pagenumbers]{wacv} 

\definecolor{wacvblue}{rgb}{0.21,0.49,0.74}
\usepackage[pagebackref,breaklinks,colorlinks,allcolors=wacvblue]{hyperref}

\usepackage{color, colortbl}
\usepackage[first=0,last=9]{lcg}
\def\thickerhline{\noalign{\hrule height1.5pt}}

\usepackage{tabularx}
\usepackage{multirow}
\usepackage{adjustbox}
\usepackage{comment}
\usepackage{amsmath}
\usepackage{amssymb}
\usepackage{pifont}

\DeclareMathOperator*{\avg}{avg}

\newcommand{\cmark}{\ding{51}}%
\newcommand{\xmark}{\ding{55}}%

\usepackage[table,svgnames]{xcolor}
\definecolor{Gray}{gray}{0.85}
\newcolumntype{g}{>{\columncolor{Gray}}c}
\usepackage{makecell}

\def\wacvPaperID{427} 
\def\confName{WACV}
\def\confYear{2026}

\title{OPFormer: Object Pose Estimation leveraging foundation model with geometric encoding}

\author{
Artem Moroz \quad Vít Zeman \quad Martin Mikšík \quad Elizaveta Isianova \quad Miroslav David \\
Pavel Burget \quad Varun Burde \\
Czech Institute of Informatics, Robotics and Cybernetics, Czech Technical University in Prague
}

\begin{document}
\twocolumn[{
\maketitle
\begin{center}
    \captionsetup{type=figure}
    \includegraphics[width=\textwidth]{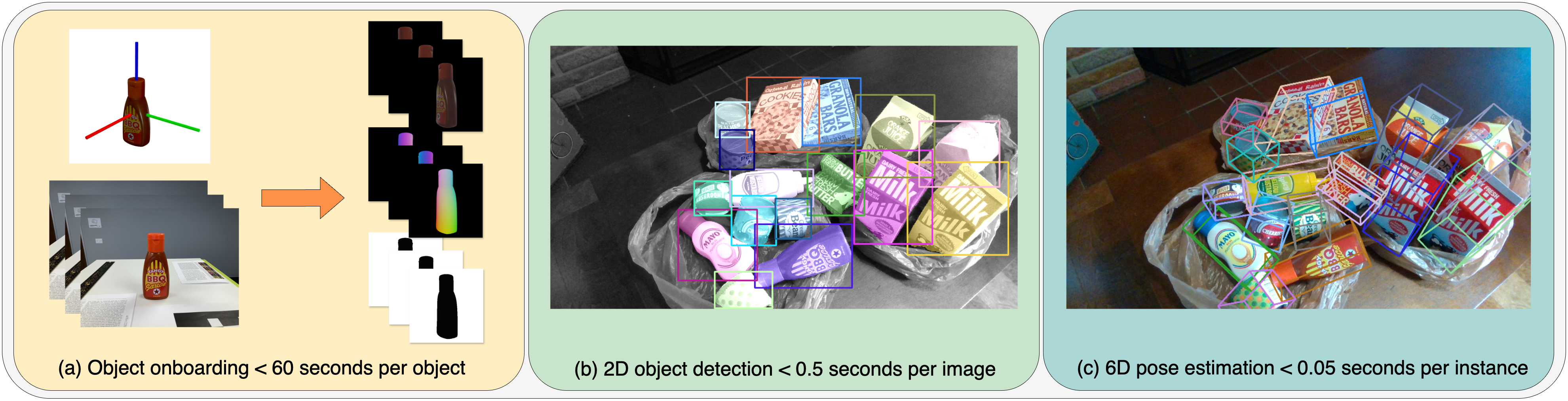}
    \captionof{figure}{We illustrate the three major blocks of our pipeline and their respective computational times: (a) object onboarding from a CAD model or a set of images takes less than a minute per object, (b) object detection with CNOS \cite{nguyen2023cnos} takes less than 0.5 seconds per image, (c) pose estimation with our methodology takes less than 0.05 seconds per instance.}
    \label{fig:Teaser}
\end{center}
}]


\begin{abstract}

We introduce a unified, end-to-end framework that seamlessly integrates object detection and pose estimation with a versatile onboarding process. Our pipeline begins with an onboarding stage that generates object representations from either traditional 3D CAD models or, in their absence, by rapidly reconstructing a high-fidelity neural representation (NeRF) from multi-view images. Given a test image, our system first employs the CNOS detector to localize target objects. For each detection, our novel pose estimation module, OPFormer, infers the precise 6D pose. The core of OPFormer is a transformer-based architecture that leverages a foundation model for robust feature extraction. It uniquely learns an object representation by jointly encoding multiple template views and enriches these features with explicit 3D geometric priors using Normalized Object Coordinate Space (NOCS). A decoder then establishes robust 2D-3D correspondences to determine the final pose. Evaluated on the challenging BOP benchmarks, our integrated system demonstrates a strong balance between accuracy and efficiency, showcasing its practical applicability in both model-based and model-free scenarios.



\end{abstract}

\section{Introduction}\label{sec:intro}


Object pose estimation from RGB or RGB-D images is a fundamental challenge in various applications such as augmented reality, robotics, and autonomous driving. The task is defined as the estimation of a rigid transformation that aligns the object's local coordinate frame with the camera's coordinate frame.

Substantial improvement in this field has been achieved by supervised deep learning-based algorithms \cite{Su2022ZebraPoseCT, Li2019CDPNCD, xiang2018posecnn, Wang_2021_GDRN, Kehl2017SSD6DMR, Labbe2020CosyPoseCM, Rad2017BB8AS, Park2019Pix2PosePC, Zakharov2019DPOD6P, Tekin2017RealTimeSS} trained for specific objects. The second branch of methods enables category-level generalization, allowing them to handle different object instances within a learned category \cite{Chen2023SecondPoseSD, Lin2024InstanceAdaptiveAG, Lin2023VINetBC, Di2022GPVPoseCO, Chen2021FSNetFS, Wang2019NormalizedOC, Chen2021SGPASP, Lin2021DualPoseNetC6, Lin2022CategoryLevel6O}. However, both object-specific and category-level methodologies exhibit limited generalizability to fully unseen objects, which makes them impractical for applications involving a wide variety of objects.

The BOP challenge \cite{hodan2020bop} introduced a task for estimating the 6D pose of novel objects, where methods are given only the 3D CAD model of a previously unseen object and have a computationally-limited onboarding period before inference. It served as a catalyst for zero-shot, instance-level methodologies \cite{nguyen2024gigaPose, foundationposewen2024, ornek2024foundpose, GenFlow, Labbe2022MegaPose6P, caraffa2024freeze, Lin2023SAM6DSA, OSOP}, whose strong generalization capabilities facilitate their deployment in real-world applications. Several works \cite{foundationposewen2024, Li2022NeRFPoseAF, sun2022onepose, he2022oneposeplusplus, liu2022gen6d} study model-free object pose estimation, a paradigm where the object's CAD model is not available beforehand.

While these advancements are promising, the field still struggles with fundamental challenges. There is a crucial need for methods that can leverage more descriptive, geometrically-aware object representations without sacrificing computational efficiency. Moreover, the dependence on pre-existing CAD models remains a significant barrier to scalability. An ideal solution must therefore deliver state-of-the-art accuracy, operate at a speed suitable for practical use, and offer the versatility to function effectively even when a 3D model is not provided.

To address these challenges, we introduce OPFormer (\textbf{O}bject \textbf{P}ose Trans\textbf{Former}), a unified methodology for both model-based and model-free 6D pose estimation of novel objects from a single RGB image. During the initial onboarding phase, the object's templates are rendered. Beyond the standard procedure of leveraging a provided CAD model, we also investigate an alternative pathway, which involves training a Neural Radiance Field (NeRF) \cite{Mildenhall2020NeRF} on multiview images to learn a neural implicit representation, which is subsequently used for template rendering. Given a test RGB image, we first use the CNOS \cite{nguyen2023cnos} 2D detector to generate a perspective crop of the target object. We treat object pose estimation as a correspondence search between the test image crop and a set of object templates. To facilitate this search, we obtain the object's cross-template embedding using a frozen DINOv2 vision foundation model, a proposed weight adapter, and a template encoder with 3D-aware positional encoding, derived from Normalized Object Coordinate Space \cite{Wang2019NormalizedOC}. Template embeddings are then passed to a transformer decoder, which establishes matches with the test image to obtain semi-dense correspondences. The final 6D pose is computed from the filtered correspondences by solving a Perspective-n-Point (PnP) problem.

Our work presents the following key contributions:
\begin{enumerate}
\item We introduce a novel methodology for 6D pose estimation of unseen objects from a single RGB image, built upon an enhanced transformer-based architecture that uses 3D positional encoding and learned inter-template relations for descriptive embeddings, and is validated by a detailed ablation study.
\item Our modular approach utilizes a 3D reconstruction pipeline for rapid object onboarding from multi-view images, which enhances the system's versatility in applications where objects are not known a priori.
\item We conducted a comprehensive evaluation on the challenging BOP-Classic-Core and BOP-H3 datasets using strict BOP metrics, which demonstrates our method's competitive performance and a strong balance between accuracy and runtime. 
\end{enumerate}

\begin{figure*}
    \centering
    \includegraphics[width=\linewidth]{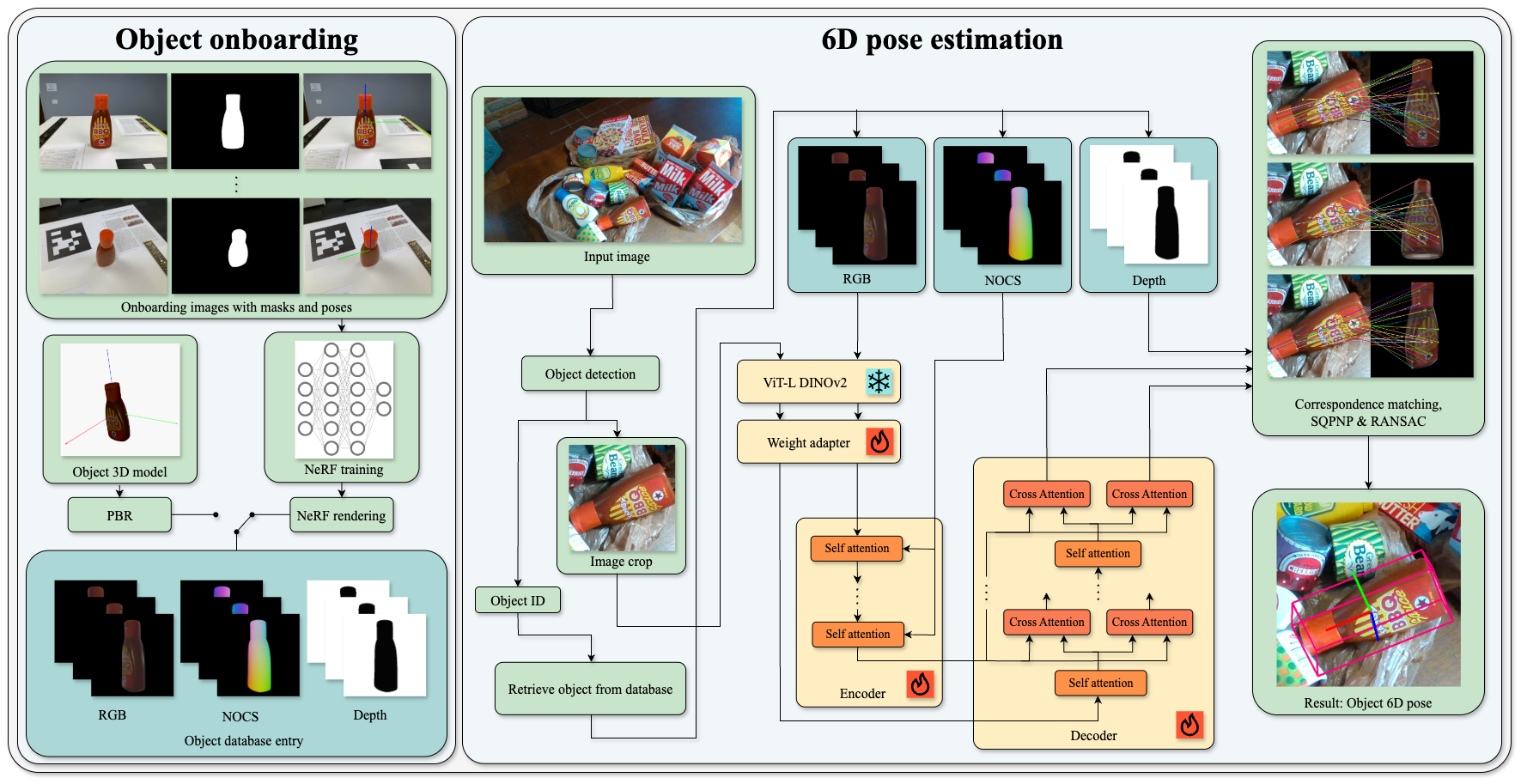}
    \caption{Our onboarding pipeline functions with either a CAD model or a set of images with accurate pose annotation. Given a set of
    multi-view images of an object, we train a NeRF using the Instant Neural Graphics Primitives (iNGP) framework. 
    RGB, depth, and NOCS templates are rendered from either CAD model or the trained NeRF.
    The pose estimation framework subsequently utilizes these templates along with the cropped test image. The RGB templates and test image crop are initially passed through the ViT-L DINOv2 block to extract the features. Template features are then passed to the encoder block, along with NOCS maps, to obtain positionally embedded template representation. The decoder determines relations between template and test image encodings via cross-attention. The feature matching stage establishes 2D-3D correspondences, and estimates 6D object pose by solving the PnP problem.}
    \label{fig:posePipeline}
\vspace{-0.35cm}
\end{figure*}

\section{Related work}\label{sec:related}

Object pose estimation methodologies can be broadly classified based on their generalization capabilities. The primary distinction lies between methods that require training on specific object instances and those that can infer the pose of previously unseen objects, often leveraging a category-level understanding or zero-shot approaches.
\newline
\textbf{Instance-Level Pose Estimation}. Networks in this category are trained to learn the specific geometry and appearance of individual objects. These approaches typically follow one of two main strategies: either directly regressing the 6D pose parameters \cite{xiang2018posecnn, Deep-6DPose}, or predicting 2D keypoints, either sparse \cite{Tekin2017RealTimeSS, Rad2017BB8AS} or dense \cite{Zakharov2019DPOD6P, Wang_2021_GDRN, Park2019Pix2PosePC, Li2019CDPNCD}, which are then used with a Perspective-n-Point (PnP) algorithm to solve for the final pose. Many state-of-the-art methods have refined this correspondence-based approach. For instance, GDR-Net \cite{Wang_2021_GDRN} employs a dynamic zoom-in mechanism and a ResNet-based architecture \cite{resnet} to predict dense correspondence maps for efficient 2D-3D matching. Similarly, ZebraPose \cite{Su2022ZebraPoseCT} uses a hierarchical approach to encode object surfaces by assigning discrete codes to each vertex, which are then used by a PnP solver. Other methods, such as \cite{Haugaard2021SurfEmbDA}, focus on learning dense 2D-3D correspondences across an object's surface via a contrastive loss.
\newline
\textbf{Category-Level and Zero-Shot Pose Estimation}. This second category of methods is designed for scalability and does not require object-specific training, making them suitable for applications with a wide variety of objects. It includes approaches that generalize to novel instances within known categories \cite{Chen2020LearningCS, Chen2024SecondPoseSC, Lin2024InstanceAdaptiveAG, Wang_2019_CVPR} as well as category-agnostic or "zero-shot" methods that work on entirely new objects. These approaches can be further divided into traditional template-based techniques and more recent methods that leverage large-scale foundation models.
\newline
\textbf{Template-Based Methods}. Historically, this category was dominated by classical template-based methods \cite{Payet2011FromCT, Ulrich2012CombiningSA, Muoz2016Fast6P}. These techniques relied on a database of templates generated by rendering a 3D CAD model from various viewpoints. During inference, a target image is compared against each template to find the best match based on a similarity score, thereby estimating the pose. More modern approaches continue to follow a template-based, render-and-compare paradigm \cite{Labbe2022MegaPose6P, Li2018DeepIMDI, nguyen2024gigaPose}. The work in \cite{nguyen2024gigaPose}, for example, efficiently samples templates and uses a nearest-neighbor search in a feature space to find the best match, offering the advantage of fast inference. MegaPose \cite{Labbe2022MegaPose6P} also leverages a render-and-compare pipeline, where a ResNet-based architecture classifies the best-matching template. In contrast to these methods that process templates independently, our approach, OPFormer, employs a transformer encoder that learns an object representation by attending across all template views simultaneously. This rich representation is then leveraged by a multi-stage matching process that uses features from several decoder depths and a neighbor-based view selection to ensure robust correspondences with runtime efficiency. In \cite{OSOP}, templates and input images are represented as feature tensors to retrieve 2D-2D correspondences, which are then lifted to 3D to solve the pose with PnP. Another line of research \cite{GenFlow} focuses on predicting the optical flow between the rendered and observed images to iteratively refine the pose.
\newline
\textbf{Foundation Model-Based and Model-Free Methods}. Recent advancements in this area increasingly leverage the powerful representations learned by large-scale, pre-trained foundation models and reduce the reliance on pre-existing CAD models. With the advent of large-scale datasets and synthetic data generation tools, the field shifted towards leveraging foundation models. One significant direction involves generating a 3D representation from reference images before matching. Methods like Gen6D \cite{liu2022gen6d}, OnePose \cite{sun2022onepose}, and OnePose++ \cite{he2022oneposeplusplus} use structure-from-motion on multi-view video to build a point cloud for pose estimation. Different from these methods, which often generate sparse point clouds, our unified framework also supports a model-free pipeline that uses Neural Radiance Fields (NeRF) \cite{Mildenhall2020NeRF} to generate dense, high-fidelity templates from multi-view images. To reduce data requirements, FS6D \cite{he2022fs6d} tackles few-shot estimation, while others like Oryon \cite{corsetti2024oryon} and Any6D \cite{lee2025any6d} push the boundary to using just a single RGB-D reference image, often employing multi-modal features or novel alignment strategies. Other methods directly leverage foundation models for robust feature extraction and matching. FoundPose \cite{ornek2024foundpose} utilizes the rich feature space of DINOv2 \cite{oquab2023dinov2} to create efficient template representations using bag-of-words retrieval techniques, with the final pose solved via PnP. While we also leverage DINOv2 features as a starting point, OPFormer introduces a trainable architecture to significantly enhance them. First, a custom weight adapter aggregates features from all 24 layers of the ViT backbone using dilated convolutions to create a richer descriptor. Then, our transformer architecture explicitly encodes 3D geometry using a novel 3D rotary positional embedding (RoPE) derived from the object's normalized coordinate space. Similarly, \cite{ZS6D} leverages a pre-trained ViT for extracting descriptors, and SAM-6D \cite{Lin2023SAM6DSA} employs the Segment Anything Model for zero-shot pose estimation by first generating object proposals and then performing dense correspondence matching. FoundationPose \cite{foundationposewen2024} employs a transformer-based architecture on both RGB and depth data, using a contrastive loss to better handle object symmetries. The work in \cite{caraffa2024freeze} demonstrates that combining 3D geometric foundation features with 2D vision foundation features can create robust representations for 3D registration without extensive training. More recently, Co-op \cite{moon2025co} introduced a correspondence-based framework that uses a hybrid representation for coarse estimation and a probabilistic flow model for refinement, achieving high accuracy with a small number of templates.

\section{Methodology}\label{sec:methodology}
\cref{fig:posePipeline} shows the visualization of the proposed pipeline in details.
Given a single RGB test image, 2D object detection represented as bounding box, camera intrinsic parameters, and either a CAD model or a multi-view image set of an object, our method addresses the challenge of 6D pose estimation of previously unseen objects.

\subsection{Object onboarding}
As a result of the onboarding stage, we obtain a rendered set of RGB templates $T = \{T_1, ..., T_N\}$, corresponding depth images $D = \{D_1, ..., D_N\}$, and NOCS (Normalized Object Coordinate Space) images $C = \{C_1, ..., C_N\}$, all of resolution $(H, W)$ for each particular object.
 NOCS images have 3 channels expressing $X$, $Y$, and $ Z$ normalized coordinates in the object frame.
 
\subsubsection{Model-based setup}
\label{sec:model-based-templates}
Given a 3D CAD/Mesh model, we first establish a canonical object frame. 
We compute the center and diagonal of the object's oriented 3D bounding box. 
The model is then normalized by translating its centroid to the origin and scaling it to fit within a unit sphere. 
To generate templates, we position virtual cameras on the vertices of a subdivided icosahedron centered at the origin, with each camera's principal axis pointing towards the center.

\subsubsection{Model-free setup}
\label{sec:model-free-templates}
In the absence of a CAD model, we use a provided set of multi-view RGB images with known poses to reconstruct the object's geometry.
We use Neural Radiance Fields (NeRF) \cite{Mildenhall2020NeRF}, specifically the efficient implementation from Instant-NGP \cite{mueller2022instant}, to create a high-fidelity 3D representation. This choice is supported by prior work \cite{Burde2024ComparativeEO, Burde2024ObjectPE}, which shows that NeRF-based renderings surpasses CAD models in pose estimation accuracy. 
The NeRF model is trained and used to render the required templates in under a minute on a single NVIDIA RTX 4090 GPU. 
The same normalization and camera placement strategy as in Sec. \ref{sec:model-based-templates} is applied, with NOCS maps derived from the reconstructed geometry.

\begin{figure}
    \centering
    \includegraphics[width=\linewidth]{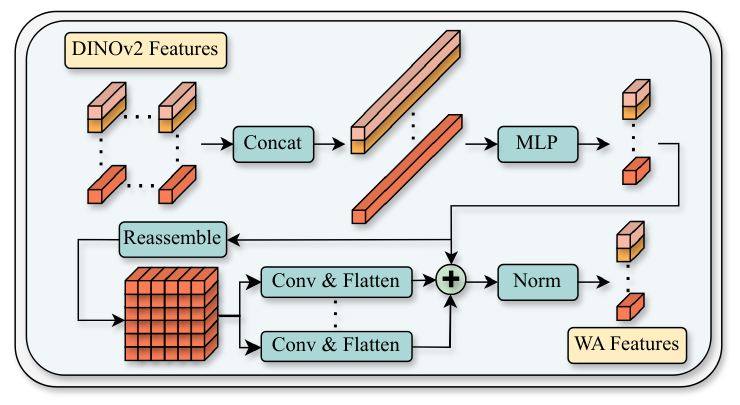}
    \caption{Architecture of the weight adapter (WA) module.
    Multi-layer patch descriptors from a DINOv2 model are dimensionally reduced, then processed by dilated convolutions. The outputs are then summed and normalized to generate the final feature set.}
    \label{fig:adapter}
\vspace{-0.35cm}
\end{figure}

\subsection{6D pose estimation}
\subsubsection{Feature extraction}
We use frozen ViT-L DINOv2 \cite{Dosovitskiy2020AnII, oquab2023dinov2} with registers \cite{Darcet2023VisionTN} to obtain an image encoding that contains geometric and semantic information for subsequent similarity search. To further enhance patch descriptors, we propose a convolution-based weight adapter that aggregates information from multiple stages of the vision transformer.

Given a test image crop $I$ of size $(\hat{H}, \hat{W})$ and $N$ RGB templates of size $(H, W)$, we register patch descriptors $I_p \in \mathbb{R}^{\hat{H}_p \times \hat{W}_p \times d}$ and $T_p \in \mathbb{R}^{N\times H_p \times W_p \times d}$ respectively. When the image of spatial resolution $(H, W)$ is processed by the feature extraction backbone, we obtain a $H_p \times W_p$-long sequence of patch descriptors, where $H/H_p$ = $W/W_p$ = 14 is a ViT DINOv2 patch size. 

The proposed weight adapter first extracts patch descriptors from all 24 layers of the ViT-L DINOv2 model. 
For each image patch, it concatenates these descriptors along their feature dimension to create a single feature vector. 
A subsequent linear layer reduces the feature dimension to meet feasible computational requirements. 
In the next step, the sequence of patch descriptors is reshaped to a 3D tensor with a spatial resolution $(H_p, W_p)$ and is processed by 4 parallel convolution layers with varying dilation factor to acquire diverse local context. 
In the final step, the outputs of these convolutional layers are flattened, summed, and the final set of descriptors is normalized by layer norm. 
Architecture of weight adapter is illustrated in \cref{fig:adapter}.

\subsubsection{Encoder}
Subsequent to the weight adapter we build a transformer encoder, which encodes only template features. 
Our encoder is designed to learn relationships that exist between different templates. It achieves this by flattening the patch descriptors from all templates into a single sequence and processing them all collectively.
We enhance the template patch descriptors with 3D positional embedding by incorporating NOCS images $C$. 
We downsize every $C_{i}$ by a factor of 14 and flatten in order to align them with dimensions of $\bar{T}$: $\bar{C} \in \mathbb{R}^{M\times 3}$, where $M = N\times H_p \times W_p$.
Our implementation extends the RoPE \cite{Su2021RoFormerET} to the 3D case. 

Considering a query vectors $\textbf{q}_m \in\mathbb{R}^{d}$ and a key vectors $\textbf{k}_m \in\mathbb{R}^{d}$ in the self-attention module, we split their \textit{d}-dimensional space into \textit{d}/6 sub-spaces, each with dimension 6. For a corresponding triplet $(x_m, y_m, z_m)$ in the NOCS image we define a block diagonal matrix $\textbf{R}_{m} \in \mathbb{R}^{d\times d}$, composed of $2 \times 2$ rotation matrices $\textbf{R}_{\theta_{x, k}}$, $\textbf{R}_{\theta_{y, k}}$, $\textbf{R}_{\theta_{z, k}}$, $\theta_{t, k} = \pi t 10000 ^ {-2(k-1)/d},  t \in \{x_m, y_m, z_m\},$ and $ k = 1, ..., d/6$. Further, a query vectors $\textbf{q}_m$ and a key vectors $\textbf{k}_m$ are rotated by  $\textbf{R}_m$ before calculating attention. The total calculation of self-attention is defined by \cref{rotaty_matrix}, \cref{self-attention}, \cref{attention-similarity}.

\begin{equation}
    \textbf{R}_m = \text{diag}[\textbf{R}_{\theta_{x, 1}}, \textbf{R}_{\theta_{y, 1}}, \textbf{R}_{\theta_{z, 1}}, ..., \textbf{R}_{\theta_{x,\frac{d}{6}}}, \textbf{R}_{\theta_{y,\frac{d}{6}}}, \textbf{R}_{\theta_{z,\frac{d}{6}}}]
\label{rotaty_matrix}
\end{equation}

\begin{equation}
    \text{Att}(Q, K, V)_m = \frac{\sum_{j=1}^{M}{ \text{sim}(\textbf{R}_m \textbf{q}_m, \textbf{R}_j\textbf{k}_j) \textbf{v}_j }}
    {\sum_{j=1}^{M}{ \text{sim}(\textbf{R}_m \textbf{q}_m, \textbf{R}_j\textbf{k}_j) }}
    \label{self-attention}
\end{equation}

\begin{equation}
    \text{sim}(\hat{\textbf{q}}_m, \hat{\textbf{k}}_j) = \text{exp}( \hat{\textbf{q}}_m^T \hat{\textbf{k}}_j / \sqrt{d})
    \label{attention-similarity}
\end{equation}

The encoder is comprised from four blocks, each integrating multi-head self-attention and a SwiGLU feed-forward network \cite{Shazeer2020GLUVI}, with layer normalization applied after each component.

\subsubsection{Decoder}
The proposed decoder module comprises four layers, each performing self-attention over the test image descriptor vectors, followed by two cross-attention operations, from test image (as queries) to templates and from templates (as queries) to test image. The self-attention blocks are enhanced with 2D RoPE \cite{Heo2024RotaryPE}, while no positional embeddings are used in the cross-attention layers. The output vectors are L2-normalized before the correspondence matching stage.

\subsubsection{Correspondence matching and 6D pose estimation}

2D-3D correspondences are obtained by matching patch descriptors from the test image and the templates. To improve robustness, we utilize descriptors from the final three decoder layers (specifically 2, 3, and 4). 
For each patch in the test image, we calculate its cosine similarity with respect to all template patches.
A primary candidate view is then determined via a voting process: for each test image patch, we find its best-matching patch across all templates, and the reference template that accumulates the most of these top matches is selected. This chosen view is subsequently augmented with its six neighboring views.
We retrieve correspondences from each of the seven selected templates independently. 
To enforce a mutual nearest-neighbor constraint, we apply a dual softmax with a threshold to filter out ambiguous correspondences. 
The object's 3D points are extracted from the corresponding depth maps D, and the 6D pose is estimated using a Perspective-n-Point (PnP) algorithm within a RANSAC loop.
\subsection{Training procedure}

Our model is trained using contrastive learning on a large-scale synthetic dataset of 2 million images rendered from Google-Scanned-Objects (GSO) \cite{Downs2022GoogleSO} and ShapeNet \cite{Chang2015ShapeNetAI} models, following the protocol of MegaPose \cite{Labbe2022MegaPose6P}. For each object, we generate 162 templates.

In each training step, we randomly choose one object category, select a subset of templates $T = \{ T_{1}, ...,  T_{K} \}$, and sample a training image \textit{I}. 
We then extract equally-sized square patches, $T_p$ from the templates and $I_p$ from the image.
Our training objective $\mathcal{L}$ is a weighted sum of two loss terms, $\mathcal{L}_{1}$ and $\mathcal{L}_{2}$, derived from two contrastive sampling methods that match patches from $T_p$ with patches from $I_p$.

\begin{enumerate}
    \item \textbf{Template-to-Image Loss ($\mathcal{L}_{1}$):} An anchor patch $a$ is sampled from $T_p$. The single corresponding positive patch $p$ is from $I_p$, and negative patches $n$ are sampled from the set $I_p \setminus \{p\}$.
    
    \item \textbf{Image-to-Template Loss ($\mathcal{L}_{2}$):} An anchor patch $a$ is sampled from $I_p$. The set of positive patches $P = \{ p_{1}, ..., p_{j} \}$ is a subset of $T_p$, accounting for multiple correct matches. Negative patches $n$ are sampled from the set $T_p \setminus P$.
\end{enumerate}

Both loss terms are implemented as focal loss \cite{Lin2017FocalLF}. For each of the $N$ sampled anchors, the loss classifies one positive pair $(a_i, p_i)$ against $M$ negative pairs $\{(a_i, n_{ij})\}_{j=1}^{M}$.
Patch descriptors $\textbf{a}_{i}$, $\textbf{p}_{i}$ and $ \textbf{n}_{ij}$, corresponding to $a_{i}$, $p_{i}$, and $ n_{ij}$, are matched using cosine similarity with scaling parameter $\tau$. We implement the loss function using the equations \cref{training_loss}, \cref{NC}, \cref{similarity}: 
\begin{equation}
    \mathcal{L} = \lambda_1 \mathcal{L}_{1} + \lambda_2 \mathcal{L}_{2}
    \label{training_loss}
\end{equation}
\begin{equation}
    \mathcal{L}_l = - \sum^{N}_{i=1}{(1 - s_i) ^ {\gamma}\ln (s_i)},  \hspace{0.3cm} l \in \{1, 2\}
    \label{NC}
\end{equation}
\begin{equation}
    s_i = \frac{e^{(\textbf{a}_i \cdot \textbf{p}_i) \backslash \tau}}{e^{(\textbf{a}_i\cdot \textbf{p}_i) \backslash \tau} + \sum^{M}_{j=1}{e^{(\textbf{a}_i \cdot \textbf{n}_{ij}) \backslash \tau }}}
    \label{similarity}
\end{equation}

The hyperparameters are set as follows: temperature $\tau = 0.01$, loss weighting coefficients $\lambda_1 = 1$, $\lambda_2 = 1$, and the focal loss focusing parameter $\gamma = 1$. The model is trained for 20 million iterations using AdamW optimizer and cosine annealing scheduler. The linear warm-up stage lasts 50 thousand iterations.
The entire training process took approximately 4 days on 4 NVIDIA A100 GPU cards.

\begin{table*}[!ht]
\centering
\begin{adjustbox}{width=\linewidth} 
\begin{tabular}{c r l c c c c c c c c c c}
\thickerhline
\textbf{Method} & \multicolumn{2}{c}{\hspace{.3cm}Pose refinement \hspace{.15cm}\# hyps.} & \textbf{LM-O} & \textbf{TLESS} & \textbf{TUD-L} & \textbf{IC-BIN} & \textbf{ITODD} & \textbf{HB} & \textbf{YCB-V} & \textbf{Average} & \textbf{Time (s)} \\
\thickerhline
\rowcolor{gray!25} OPFormer & \multicolumn{2}{c}{\hspace{1.75cm}\xmark} & \underline{57.2} & \underline{47.1} & \textbf{68.9} & \underline{45.6} & \underline{38.7} & \underline{75.0} & \textbf{65.1} & \underline{56.8} & \underline{0.554}\\

Co-op & \multicolumn{2}{c}{\hspace{1.75cm}\xmark} & \textbf{59.7} & \textbf{59.2} & \underline{64.2} & \textbf{45.8} & \textbf{39.1} & \textbf{78.1} & \underline{62.6} & \textbf{58.4} & 0.979\\
FoundPose & \multicolumn{2}{c}{\hspace{1.75cm}\xmark} & 39.7 & 33.8 & 46.9 & 23.9 & 20.4 & 50.8 & 45.2 & 37.3 & 1.690 \\
GigaPose & \multicolumn{2}{c}{\hspace{1.75cm}\xmark} & 29.6 & 26.4 & 30.0 & 22.3 & 17.5 & 34.1 & 27.8 & 26.8 & \textbf{0.384} \\
GenFlow & \multicolumn{2}{c}{\hspace{1.75cm}\xmark} & 25.0 & 21.5 & 30.0 & 16.8 & 15.4 & 28.3 & 27.7 & 23.5 & 3.839 \\
MegaPose & \multicolumn{2}{c}{\hspace{1.75cm}\xmark} & 22.9 & 17.7 & 25.8 & 15.2 & 10.8 & 25.1 & 28.1 & 20.8 & 15.465 \\
\thickerhline

\rowcolor{gray!25} OPFormer &  MegaPose  & 1 & \underline{60.4} & 51.9 & \underline{69.5} & \underline{49.7} & 41.2 & \underline{76.6} & \textbf{68.7} & \underline{59.7} & \textbf{1.783}\\
Co-op & Co-op & 1 & \textbf{64.2} & \textbf{63.5} & \textbf{71.7} & \textbf{51.2} & \textbf{47.3} & \textbf{83.2} & \underline{67.0} & \textbf{64.0} & \underline{1.852}\\
FoundPose & FeatRef & 1 & 39.5 & 39.6 & 56.7 & 28.3 & 26.2 & 58.5 & 49.7 & 42.6 & 2.648 \\
FoundPose & MegaPose  & 1 & 55.4 & 51.0 & 63.3 & 43.0 & 34.6 & 69.5 & 66.1 & 54.7 & 4.398 \\
FoundPose & FeatRef+Megapose  & 1 & 55.6 & 51.1 & 63.3 & 43.3 & 35.7 & 69.7 & 66.1 & 55.0 & 6.385 \\
GigaPose & MegaPose  & 1 & 55.7 & 54.1 & 58.0 & 45.0 & 37.6 & 69.3 & 63.2 & 54.7 & 2.301 \\
GigaPose & GenFlow  & 1 & 59.5 & \underline{55.0} & 60.7 & 47.8 & \underline{41.3} & 72.2 & 60.8 & 56.8 & 2.213 \\
MegaPose & MegaPose  & 1 & 49.9 & 47.7 & 65.3 & 36.7 & 31.5 & 65.4 & 60.1 & 50.9 & 31.724 \\
\thickerhline

\rowcolor{gray!25}OPFormer & MegaPose & 5 & 60.4 & 52.8 & \underline{70.3} & 49.7 & 42.8 & \underline{76.7} & 68.7 & 60.2 & \textbf{2.338} \\
Co-op & Co-op & 5 & \textbf{65.5} & \textbf{64.8} & \textbf{72.9} & \textbf{54.4} & \textbf{49.1} & \textbf{85.0} & \underline{68.9}  & \textbf{65.8} & \underline{4.190}\\
FoundPose & FeatRef+Megapose  & 5 & 61.0 & 57.0 & 49.3 & 47.9 & 40.7 & 72.3 & \textbf{69.0} & 59.6 & 20.523 \\
GigaPose & GenFlow & 5 & \underline{63.1} & \underline{58.2} & 66.4 & \underline{49.8} & \underline{45.3} & 75.6 & 65.2 & \underline{60.5} & 10.637 \\
GigaPose & MegaPose & 5 & 59.8 & 56.5 & 63.1 & 47.3 & 39.7 & 72.2 & 66.1 & 57.8 & 7.682 \\
MegaPose & MegaPose & 5 & 56.0 & 50.7 & 68.4 & 41.4 & 33.8 & 70.4 & 62.1 & 54.7 & 47.386 \\
GenFlow & GenFlow & 5 & 56.3 & 52.3 & 68.4 & 45.3 & 39.5 & 73.9 & 63.3 & 57.0 & 20.890 \\
\thickerhline
\end{tabular}
\end{adjustbox}
\caption{Average Recall (AR) scores on the seven core BOP datasets for model-based 6D localization task. The best result is highlighted in \textbf{bold}, and the second best is \underline{underlined}. The results are shown from top to bottom: coarse estimation only, refinement with a single hypothesis, and refinement with five hypotheses. The table was created based on BOP challenge leaderboards. To provide comparable runtime benchmark, we evaluated the fastest competing model, GigaPose, on our hardware setup, ensuring the identical hardware conditions.  The observed 0.501-second runtime for GigaPose confirms our method's efficiency, being only 10 \% slower than the current most efficient methodology.}
\label{tab:BOP23}
\vspace{-0.35cm}
\end{table*}

\section{Experiments}\label{sec:results}

\subsection{BOP Challenges}\label{sec:BOPCHallenge}
\textbf{Tasks.}
We evaluate the performance of OPFormer according to the BOP Challenges 2019-2023 (BOP23)~\cite{hodan2024bopchallenge2023detection} and 2024 (BOP24) and it's tasks\footnote{\url{https://bop.felk.cvut.cz/tasks/}}, namely older Model-based 6D Localization of Unseen Objects and newer Model-based6D  Detection of Unseen Objects and Model-free 6D Detection of Unseen Objects.
\newline
\textbf{Datasets}.
For the 2023 challenge, we evaluated our method on all datasets from the BOP-Classic-Core challenge, namely LM-O \cite{LM-O}, T-LESS~\cite{hodan2017tless}, ITODD~\cite{ITODD}, HB~\cite{HB}, YCB-V~\cite{xiang2018posecnn}, IC-BIN~\cite{Doumanoglou2015Recovering6O}, and TUD-L~\cite{Hodan_2018_ECCV}. 
We also evaluate on the new challenging datasets HOPEv2~\cite{Tyree2022HOPE}, HANDAL~\cite{Guo2023HANDAL}, and HOT3D~\cite{banerjee2024hot3d}, collectively named BOP-H3, introduced in BOP24.
\newline
\textbf{Metrics}.
We present our results using the error metrics defined by the BOP challenges as defined in \cite{hodan2020bop}. 
The BOP23 task evaluates the average recall (AR) of the Visible Surface Discrepancy (VSD), Maximum Symmetry-Aware Surface Distance (MSSD), and Maximum Symmetry-Aware Projection Distance (MSPD) metrics.
For BOP 24, the results are computed as the average precision (AP) of MSSD and MSPD. For a detailed description of the metrics, please see \cref{sec:A_metrics}.

\subsection{Implementation details}
We use a resolution of $(420, 420)$ pixels for both test image and templates. We employ BlenderProc \cite{Denninger2023} for photorealistic rendering of templates.
For the inference stage, we set the subdivision frequency of the icosahedron to 2, resulting in a total of 42 different camera coordinate frames. 
3D rotary positional embedding requires the dimension of the patch descriptors to be multiple of 6; therefore, we set $d$ = 576.
The number of attention heads in both the self-attention and cross-attention modules is set to 8, with each head having an inner dimension of 72. We utilize the efficient implementation of attention from xFormers library \cite{xFormers2022}.
We use the OpenCV implementation of PnP RANSAC with the SQPnP \cite{Terzakis2020ACF} solver for which we set the number of iterations to 800 and the re-projection error to 14 pixels.
We use the default 2D detection method CNOS \cite{nguyen2023cnos}. 
For the tasks where object CAD models are provided, we use them to generate the templates both for the CNOS \cite{nguyen2023cnos} and the pose estimation method. 
In the 6D object localization task we select the $n$ highest-scored CNOS predictions per object category in each image, where $n$ corresponds to the known number of object instances.
For the 6D object detection task, where the object categories and the number of instances are unknown, we consider only CNOS detections with a confidence score above 0.4.
Static onboarding data from model-free 6D detection task are used to create an implicit representation of the objects with iNGP \cite{mueller2022instant}, from which we then render the templates as explained in \cref{sec:model-free-templates}.  We use available segmentation masks for static onboarding images to remove the background. The CNOS detection pipeline directly uses the provided video frames as reference object templates. 
We apply MegaPose \cite{Labbe2022MegaPose6P} refiner on top of the proposed OPFormer. For the model-free task we replace Panda3D renderer, originally used by MegaPose, with the iNGP renderer. 
We chose the number of templates, as well as the template and test image resolutions, based on the empirical results presented in \cref{sec:Additional_ablation_studies}.

\begin{figure*}
  \centering
  \subfloat{\includegraphics[height=.15\textwidth]{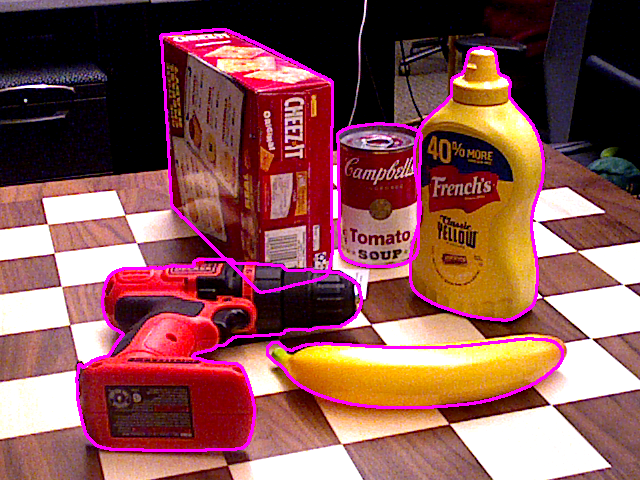}}\quad
  \subfloat{\includegraphics[height=.15\textwidth]{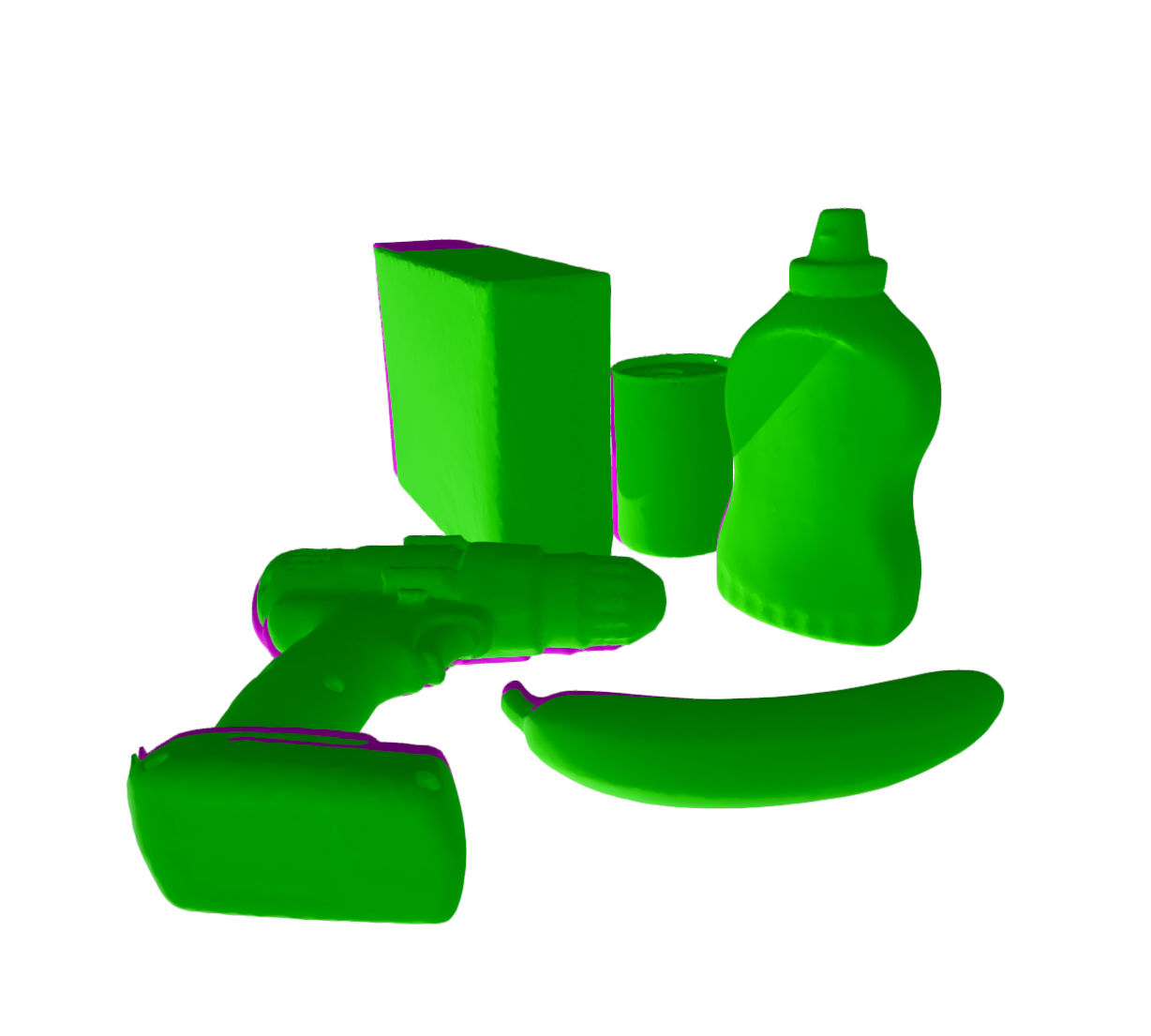}}\quad
  \subfloat{\includegraphics[height=.15\textwidth]{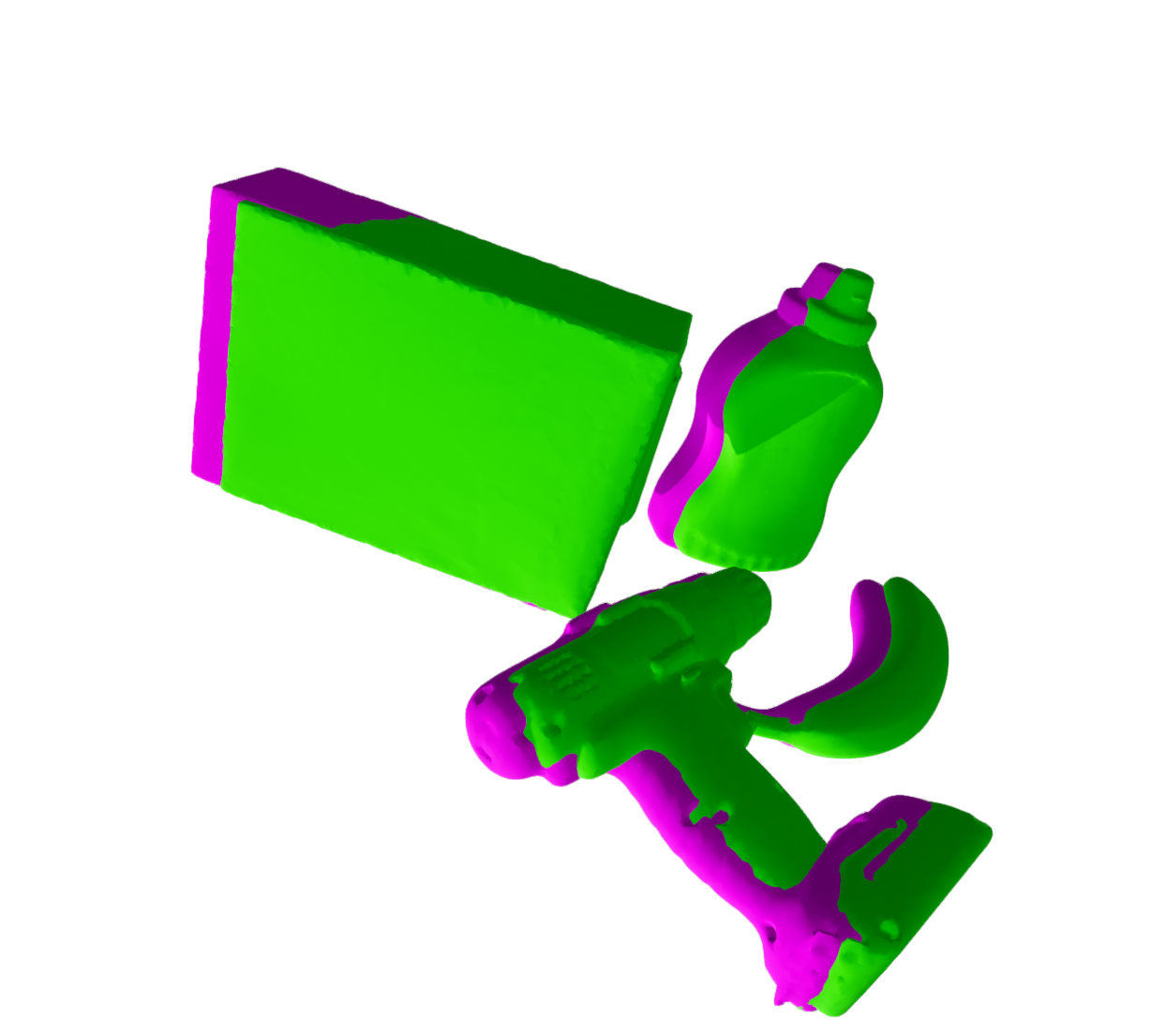}}\quad
  \subfloat{\includegraphics[height=.15\textwidth]{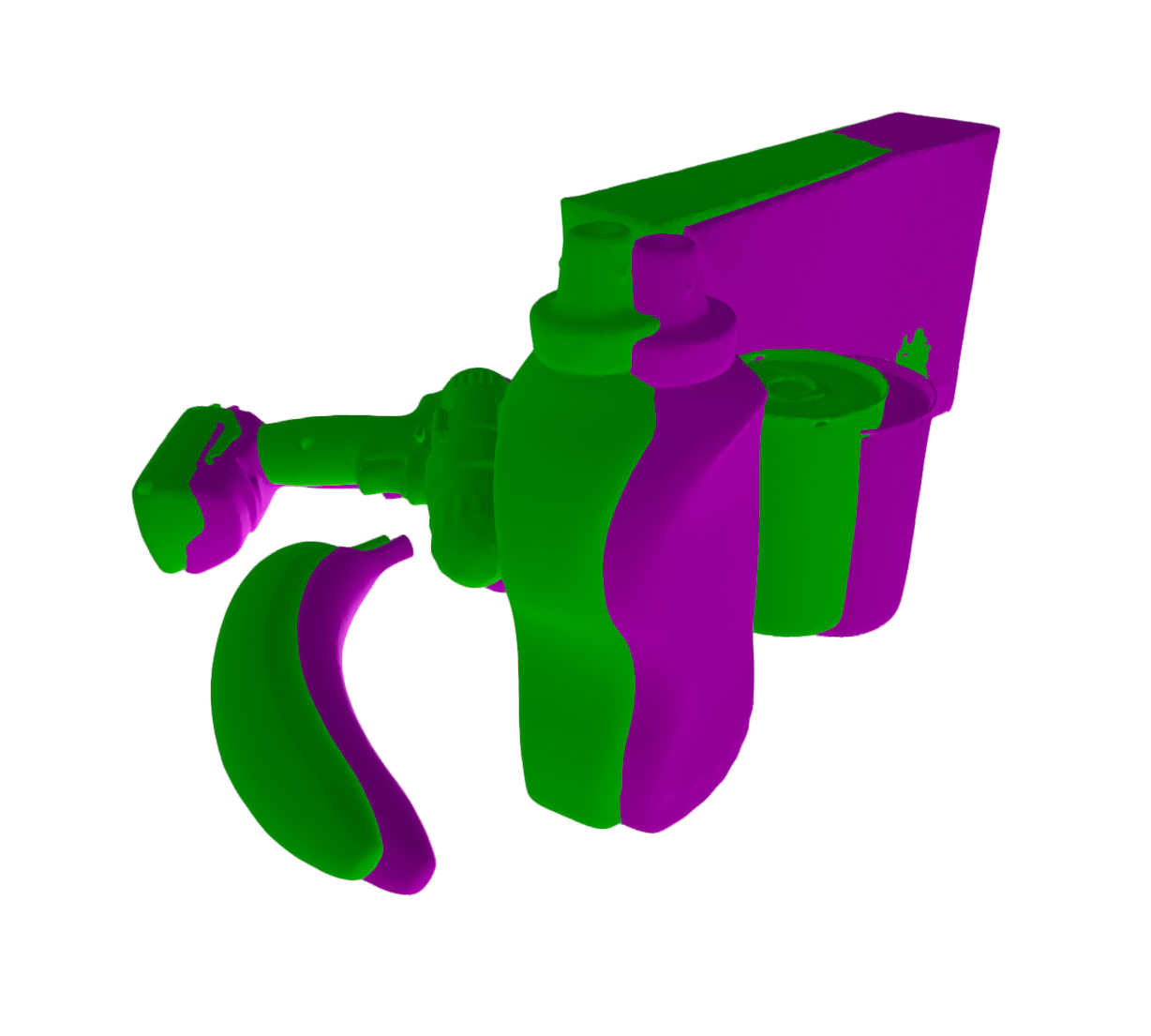}}\\
  
  \subfloat{\includegraphics[height=.15\textwidth]{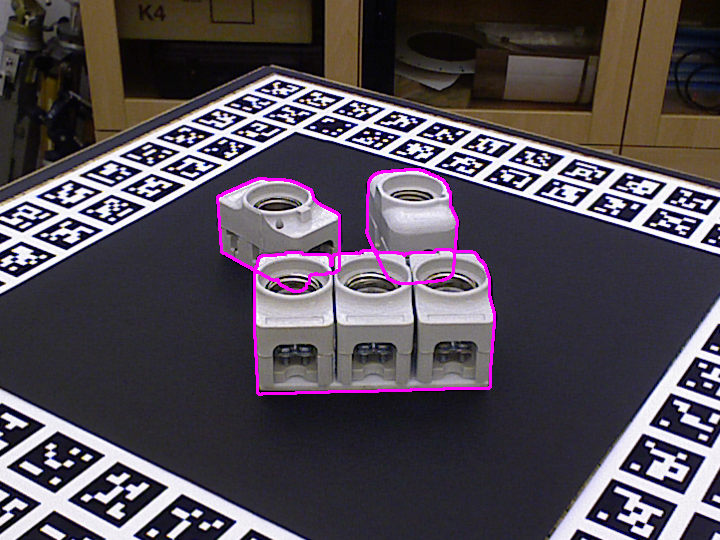}}\quad
  \subfloat{\includegraphics[height=.15\textwidth]{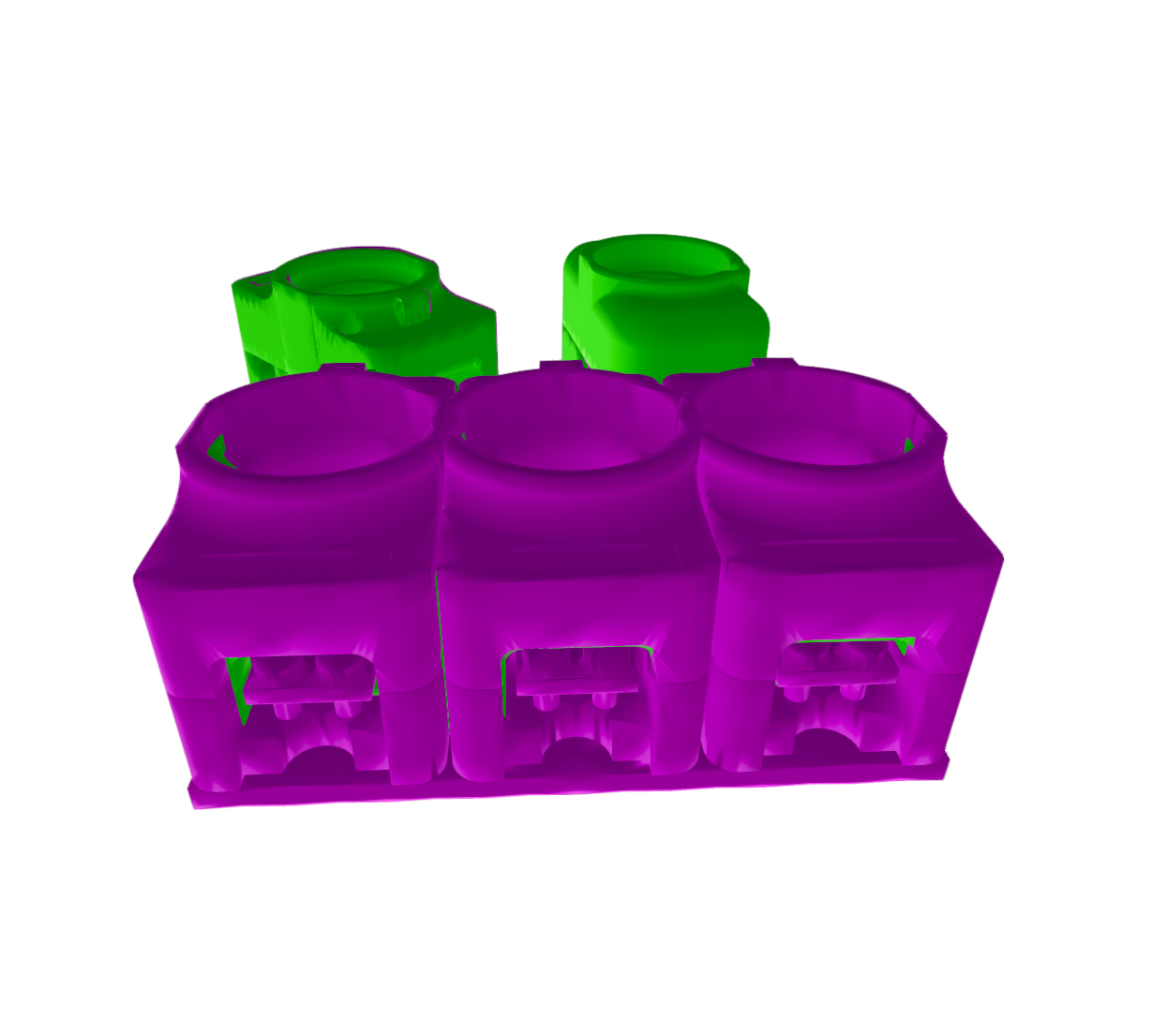}}\quad
  \subfloat{\includegraphics[height=.15\textwidth]{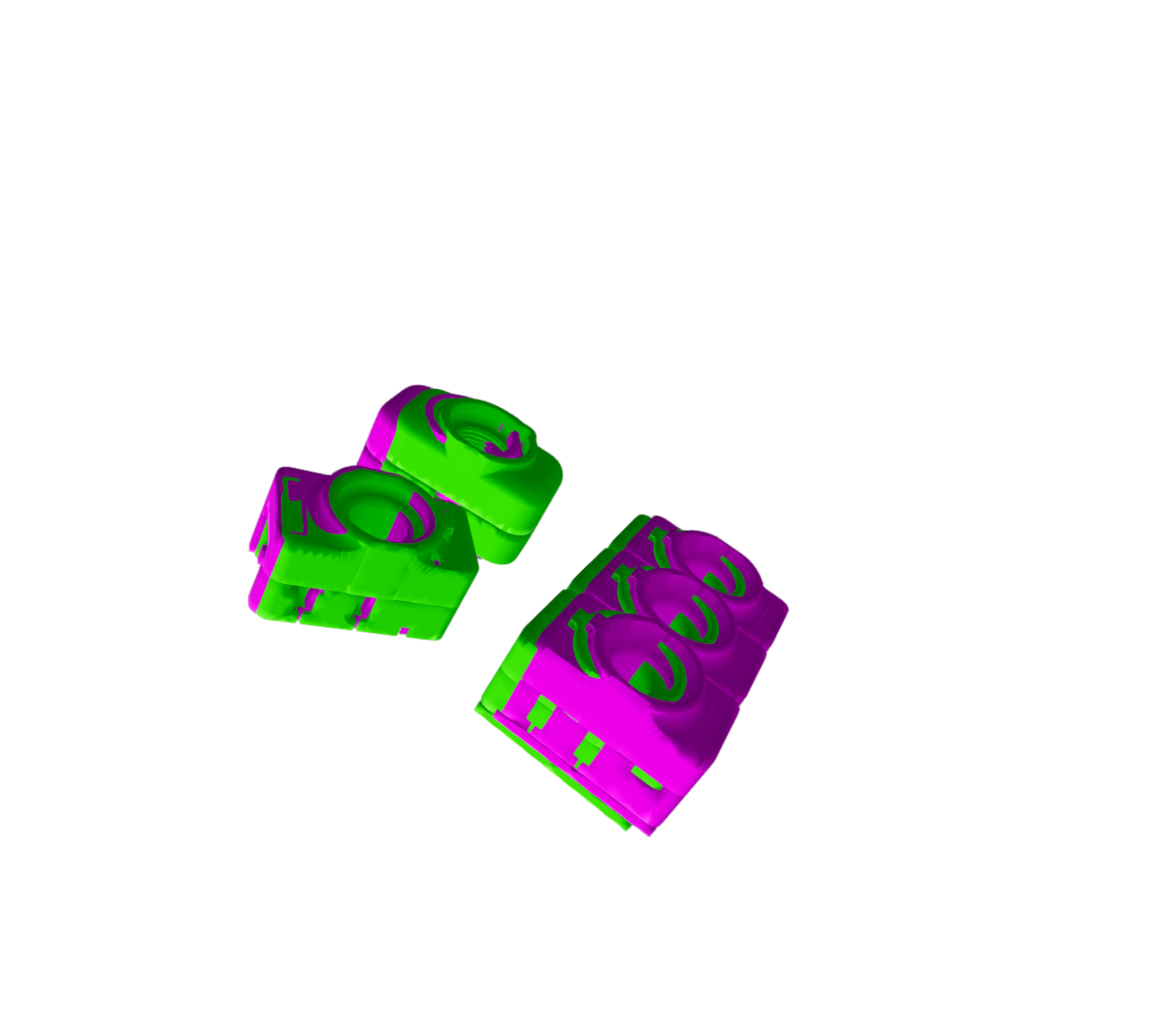}}\quad
  \subfloat{\includegraphics[height=.15\textwidth]{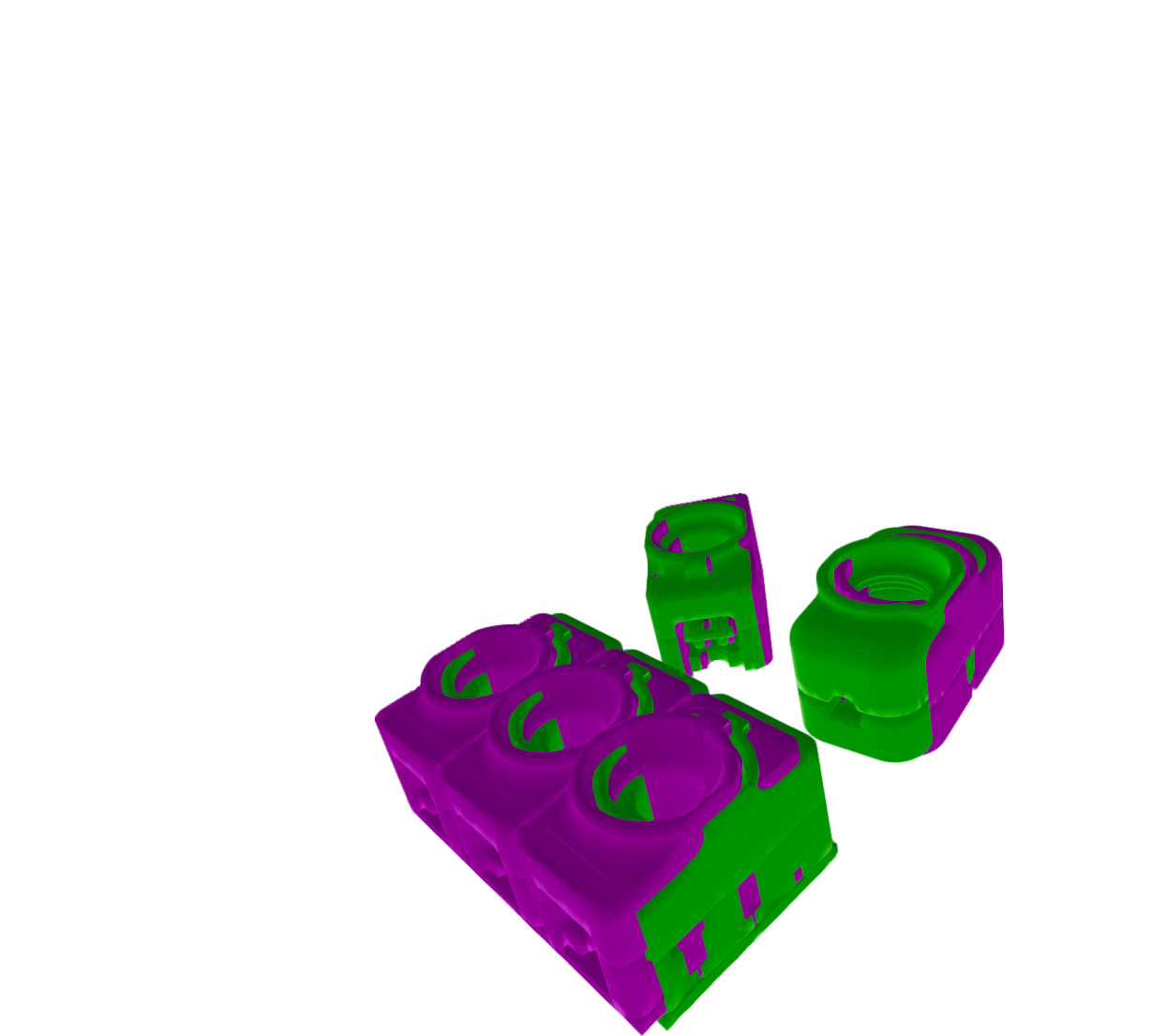}}\\
  
  
  
  \caption{Visualization of the \textcolor{green}{ground-truth} (green) and \textcolor{magenta}{estimated} (magenta) 6D poses is presented on two datasets (from the first row to last): YCB-V and T-LESS. The first column displays the test image with a contour highlight of the projection made from the predicted pose. The subsequent three columns depicts the corresponding 3D views from different viewing angles. The first 3D view is taken from approximately the same viewing angle as the the image.
  Additional visualization
  is provided in \cref{sec:A_visualization}.
}
  \label{fig:3DPredictions}
\vspace{-0.45cm}
\end{figure*}

\begin{table}
\centering
\begin{adjustbox}{width=\linewidth}
\begin{tabular}{c c c c c c c c}
\thickerhline
 Task & Method & Refiner  & HOT3D & HOPEv2 & HANDAL & Average & Time (s)\\
\thickerhline
\rowcolor{gray!25}\cellcolor{white} & OPFormer & \xmark  & \textbf{31.6} & \textbf{37.0} & \textbf{36.7} & \textbf{35.1} & 1.049 \\
& GigaPose & \xmark &  	7.2&	16.7&	4.1& 9.4& 0.869\\ \cline{2-8} 
\rowcolor{gray!25} \cellcolor{white}& OPFormer & MegaPose & \textbf{30.3} & \textbf{37.9} & \textbf{38.8} & \textbf{35.7} & 2.817  \\
\cellcolor{white}\multirow{-4}{*}{\makecell{Model-\\based}} & GigaPose & GenFlow & 26.8 &	41.1	& 25.6 & 31.2 & 5.271\\
\thickerhline
\rowcolor{gray!25}\cellcolor{white} Model- &  & \xmark &  - & 36.6 & 30.9 & - & -\\\cline{3-8}
\rowcolor{gray!25}\cellcolor{white}free& \multirow{-2}{*}{OPFormer}& MegaPose &  - & 35.6 & 32.4 & - & - \\ 
\thickerhline
\end{tabular}
\end{adjustbox}
\caption{Average Precision (AP) scores on BOP-H3 datasets for model-based and model-free 6D detection task. The best result is highlighted in \textbf{bold}. Pose refinement was applied to a single hypothesis from coarse prediction. We ommited the model-free task from HOT3D due to its deviations from the rest of the BOP datasets, requiring significant fisheye image undistortion incompatible with our direct application of NeRF-based reconstruction. The table was created based on BOP challenge leaderboards.}
\label{tab:BOP24}
\vspace{-0.45cm}
\end{table}

\subsection{Results}\label{sec:results-res}
\textbf{6D pose localization task}.
\cref{tab:BOP23} summarizes the performance of OPFormer on the 6D pose localization task, comparing it against recent state-of-the-art methods. The results highlight our method's excellent balance between high accuracy and computational efficiency. In the coarse estimation setting, OPFormer demonstrates highly competitive performance, particularly on the TUD-L \cite{Hodan_2018_ECCV} and YCB-V \cite{xiang2018posecnn} datasets, where it achieves state-of-the-art results. It achieves the second-best average recall (56.8 AR), closely following the state-of-the-art method Co-op \cite{moon2025co}, while being more runtime-efficient, even when using comparable capable hardware for inference (RTX 3090 vs. A40). This combination of top-tier accuracy and low latency makes OPFormer a standout solution, as it significantly outperforms other established methods like FoundPose \cite{ornek2024foundpose}, GigaPose \cite{nguyen2024gigaPose}, GenFlow \cite{GenFlow}, and MegaPose \cite{Labbe2022MegaPose6P}. When augmented with pose refinement, OPFormer continues to showcase its efficiency. Using a single MegaPose refinement hypothesis, it again secures a top-two accuracy position (59.7 AR) while being the fastest among the leading refined methods. Even in the more intensive 5-hypothesis setting, OPFormer remains remarkably efficient. It achieves an accuracy (60.2 AR) nearly identical to the GigaPose+GenFlow \cite{nguyen2024gigaPose,GenFlow} combination (60.5 AR), yet with an inference time that is over multiple times faster. In summary, while Co-op currently sets the state of the art for pure accuracy, OPFormer distinguishes itself by offering a superior trade-off between performance and speed. It delivers near-SOTA accuracy at a fraction of the computational cost, establishing it as a highly practical and efficient solution for real-world deployment where both accuracy and inference speed are critical. The OPFormer's result of the 6D localization of unseen object task are visualized in \cref{fig:3DPredictions}.
\newline
\textbf{ 6D object detection for model-based and model-free}. For the 6D object detection tasks, OPFormer demonstrates a dominant performance, as detailed in \cref{tab:BOP24}. To the best of our knowledge, no results have been previously published for model-free setup; therefore, our unified pipeline, which integrates an object onboarding process, establishes a new state-of-the-art. In the model-based detection setting, OPFormer significantly outperforms the next-best method, GigaPose \cite{nguyen2024gigaPose}. During coarse estimation (without refinement), our method achieves an average precision (AP) of 35.1, nearly four times higher than GigaPose's 9.4 AP, with a negligible difference in inference time. This advantage is maintained after refinement; OPFormer with a single MegaPose \cite{Labbe2022MegaPose6P} hypothesis achieves a 35.7 AP, surpassing the GigaPose+GenFlow \cite{nguyen2024gigaPose,GenFlow} combination while also being almost twice as fast. When comparing performance across datasets, we observe similar results on HOPEv2 and HANDAL, but performance drops on the HOT3D dataset. We hypothesize that this discrepancy is due to the presence of grayscale test images and inaccuracies introduced during the undistortion of fisheye images. Our approach achieves comparable performance on both model-based and model-free tasks, demonstrating its effectiveness even in the more challenging model-free setting. \cref{sec:A_BOPdetails} provides more detailed quantitative results of our method.

\subsection{Ablation experiments}\label{sec:Ablation}

\begin{table}
\begin{adjustbox}{width=\linewidth} 
 \begin{tabular}{ccccccccccccc} 
 \thickerhline 
 \rotatebox[origin=c]{90}{ Model number } & \rotatebox[origin=c]{90}{Weight adapter} & \rotatebox[origin=c]{90}{ Template encoder } & \rotatebox[origin=c]{90}{ NOCS Rotary PE } & \rotatebox[origin=c]{90}{ Two-way decoder } & \rotatebox[origin=c]{90}{ SQPnP } & \rotatebox[origin=c]{90}{ Neighb. templates } & \rotatebox[origin=c]{90}{ Segmentation } & \rotatebox[origin=c]{90}{LM-O} & \rotatebox[origin=c]{90}{YCB-V} & \rotatebox[origin=c]{90}{T-LESS} & \rotatebox[origin=c]{90}{Average}\\ 
 \thickerhline 
 1 & \cmark & \cmark & \cmark & \cmark & \cmark & \cmark & \xmark & \textbf{57.2} & \underline{65.1} & \underline{47.1} & \textbf{56.5} \\ 
 \thickerhline
\rowcolor{green!5} 2 & \xmark & \cmark & \xmark & \xmark & \cmark & \cmark & \xmark & 55.5 & 62.1 & 40.2 & 52.6 \\ 
 \hline 
\rowcolor{green!5}3 & \cmark & \xmark & \xmark & \xmark & \cmark & \cmark & \xmark & 54.8 & 61.4 & 43.0 & 53.0 \\  \hline 
\rowcolor{green!5} 4 & \cmark & \cmark & \xmark & \xmark & \cmark & \cmark & \xmark & 56.6 & 63.6 & 45.3 & 55.2 \\ \hline 
\rowcolor{green!5} 5 & \cmark & \cmark & \cmark & \xmark & \cmark & \cmark & \xmark & \underline{57.1} & 64.5 & 46.2 & \underline{55.9} \\ 
 \hline 
\rowcolor{blue!5} 6 & \cmark & \cmark & \cmark & \cmark & \xmark & \cmark & \xmark & 56.2 & 62.5 & 46.0 & 54.9 \\ 
 \hline 
\rowcolor{blue!5}  7 & \cmark & \cmark & \cmark & \cmark & \cmark & \xmark & \xmark & 56.9 & \textbf{65.5} & 45.1 & 55.8\\ 
 \hline
\rowcolor{blue!5} 8 & \cmark & \cmark & \cmark & \cmark & \cmark & \cmark & \cmark & 55.0 & 63.3 & \textbf{48.3} & 55.5\\ 
 \thickerhline 
 
\end{tabular}
\end{adjustbox}
\caption{Ablation studies evaluate the architecture and inference settings of our 6D pose estimation solution without refinement.
Experiments are conducted on three BOP23 core datasets: LM-O, YCB-V and T-LESS. 
Model 1 is presented in the \cref{sec:methodology} and used for reporting our results in \cref{tab:BOP23} and \ref{tab:BOP24}.}\label{tab:Ablation}
\vspace{-0.45cm}
\end{table}

We compare eight varied model setups, with the results presented in \cref{tab:Ablation}. Further ablation studies on inference settings are available in \cref{sec:Additional_ablation_studies} and OPFormer's limitations are discussed in \cref{sec:A_Limits}.
\newline
\textbf{Architecture Design}.
We assess the importance of model components such as the weight adapter, template encoder, NOCS-based positional embedding, and the two-way decoder. Model 2 relies on patch descriptors extracted from layer 18 of the ViT-L, as proposed by \cite{ornek2024foundpose}, uses a Set Abstraction Layer \cite{Qi2017PointNetDH} for 3D positional embedding, as explored in \cite{Lin2023SAM6DSA}, and employs a one-way decoder. This decoder does not calculate cross-attention from template descriptors (as queries) to image descriptors; instead, it directly uses template descriptors from the encoder.
A comparison between Models 2 and 4 indicates that the proposed weight adapter provides a moderate performance increase, particularly on the T-LESS dataset, which contains textureless objects. We attribute this improvement to a better positional understanding obtained by combining features from deep and shallow ViT-L layers \cite{Fu2023DTLDT}.
Ablating the template encoder in Model 3 (in contrast to Model 4) and directly using weight adapter descriptors degrades performance noticeably, indicating the encoder's important role in learning a superior feature representation space.
We observe a slight improvement when replacing the Set Abstraction Layer with the proposed NOCS-based rotary positional embedding (Models 4 and 5). Notably, our implementation does not require any trainable parameters, unlike the one used in \cite{Lin2023SAM6DSA}. The two-way decoder, as shown by the comparison between Models 1 and 5, boosts performance in our similarity search task by considering mutual bidirectional context from both feature sequences. 
\newline
\textbf{Inference Settings}.
The SQPnP solver (Model 1) is compared to the more frequently used EPnP \cite{EPnP} solver (Model 6). 
SQPnP provides more accurate results, as it is proven to be more robust with respect to noisy correspondences \cite{Terzakis2020ACF}.
Taking six neighboring vertex templates improves performance, which is demonstrated by the comparison between Models 1 and 7.
Furthermore, we analyze the impact of segmentation masks by comparing Model 6 to Model 1. 
We observe that segmentation can boost performance on some datasets (T-LESS) while degrading results on others (YCB-V, LM-O). 
We hypothesize that this occurs due to mask imperfections, which may obscure parts of objects and decrease the number of valid patch correspondences.

\section{Conclusion}\label{sec:conclusion}


In this work, we have introduced OPFormer, a unified pipeline for 6D pose estimation of unseen objects that successfully addresses the critical challenges of generalization, efficiency, and data dependency. Our key innovation lies in a novel transformer architecture that learns enhanced object representation via inter-template encoder with 3D-aware positional embedding. The uniqueness of our approach stems from its ability to learn a object representation and its flexibility in onboarding new objects, either from a 3D CAD model or a rapid NeRF reconstruction. This eliminates the need for object-specific training, making our system immediately applicable to dynamic real-world scenarios such as robotic bin-picking and augmented reality. Our pipeline has not only demonstrated state-of-the-art performance on the BOP 2023 challenge but also established the initial baselines for both model-based and model-free tasks on two of the three new BOP 2024 datasets. 

\section*{Acknowledgement}
This work was financed by the EU Horizon 2020 project RICAIP (grant agreement No. 857306); supported by the Ministry of Education, Youth and Sports of the Czech Republic through the e-INFRA CZ (ID:90254) and the Grant Agency of the Czech Technical University in Prague (grant No.SGS23/172/OHK3/3T/13); and co-funded by European Union under the project Robotics and advanced industrial production - ROBOPROX (reg. no. CZ.02.01.01/00/22\_008/0004590).
\clearpage

{
    \small
    \bibliographystyle{ieeenat_fullname}
    \bibliography{main}

\begin{thebibliography}{78}
\providecommand{\natexlab}[1]{#1}
\providecommand{\url}[1]{\texttt{#1}}
\expandafter\ifx\csname urlstyle\endcsname\relax
  \providecommand{\doi}[1]{doi: #1}\else
  \providecommand{\doi}{doi: \begingroup \urlstyle{rm}\Url}\fi

\bibitem[Ausserlechner et~al.(2024)Ausserlechner, Haberger, Thalhammer, Weibel, and Vincze]{ZS6D}
Philipp Ausserlechner, David Haberger, Stefan Thalhammer, Jean-Baptiste Weibel, and Markus Vincze.
\newblock Zs6d: Zero-shot 6d object pose estimation using vision transformers.
\newblock In \emph{2024 IEEE International Conference on Robotics and Automation (ICRA)}, pages 463--469, 2024.

\bibitem[Banerjee et~al.(2025)Banerjee, Shkodrani, Moulon, Hampali, Han, Zhang, Zhang, Fountain, Miller, Basol, Newcombe, Wang, Engel, and Hodan]{banerjee2024hot3d}
Prithviraj Banerjee, Sindi Shkodrani, Pierre Moulon, Shreyas Hampali, Shangchen Han, Fan Zhang, Linguang Zhang, Jade Fountain, Edward Miller, Selen Basol, Richard Newcombe, Robert Wang, Jakob~Julian Engel, and Tomas Hodan.
\newblock {HOT3D}: Hand and object tracking in {3D} from egocentric multi-view videos.
\newblock \emph{CVPR}, 2025.

\bibitem[Brachmann et~al.(2014)Brachmann, Krull, Michel, Gumhold, Shotton, and Rother]{LM-O}
Eric Brachmann, Alexander Krull, Frank Michel, Stefan Gumhold, Jamie Shotton, and Carsten Rother.
\newblock Learning 6d object pose estimation using 3d object coordinates.
\newblock In \emph{Computer Vision -- ECCV 2014}, pages 536--551, Cham, 2014. Springer International Publishing.

\bibitem[Burde et~al.(2024{\natexlab{a}})Burde, Benbihi, Burget, and Sattler]{Burde2024ComparativeEO}
Varun Burde, Assia Benbihi, Pavel Burget, and Torsten Sattler.
\newblock Comparative evaluation of 3d reconstruction methods for object pose estimation.
\newblock \emph{2025 IEEE/CVF Winter Conference on Applications of Computer Vision (WACV)}, pages 7669--7681, 2024{\natexlab{a}}.

\bibitem[Burde et~al.(2024{\natexlab{b}})Burde, Moroz, Zeman, and Burget]{Burde2024ObjectPE}
Varun Burde, Artem Moroz, Vit Zeman, and Pavel Burget.
\newblock Object pose estimation using implicit representation for transparent objects.
\newblock In \emph{ECCV Workshops}, 2024{\natexlab{b}}.

\bibitem[Caraffa et~al.(2024)Caraffa, Boscaini, Hamza, and Poiesi]{caraffa2024freeze}
Andrea Caraffa, Davide Boscaini, Amir Hamza, and Fabio Poiesi.
\newblock Freeze: Training-free zero-shot 6d pose estimation with geometric and vision foundation models.
\newblock \emph{European Conference on Computer Vision (ECCV)}, 2024.

\bibitem[Chang et~al.(2015)Chang, Funkhouser, Guibas, Hanrahan, Huang, Li, Savarese, Savva, Song, Su, Xiao, Yi, and Yu]{Chang2015ShapeNetAI}
Angel~X. Chang, Thomas~A. Funkhouser, Leonidas~J. Guibas, Pat Hanrahan, Qi-Xing Huang, Zimo Li, Silvio Savarese, Manolis Savva, Shuran Song, Hao Su, Jianxiong Xiao, L. Yi, and Fisher Yu.
\newblock Shapenet: An information-rich 3d model repository.
\newblock \emph{ArXiv}, abs/1512.03012, 2015.

\bibitem[Chen and Dou(2021)]{Chen2021SGPASP}
Kai Chen and Qi Dou.
\newblock Sgpa: Structure-guided prior adaptation for category-level 6d object pose estimation.
\newblock \emph{2021 IEEE/CVF International Conference on Computer Vision (ICCV)}, pages 2753--2762, 2021.

\bibitem[Chen et~al.(2021)Chen, Jia, Chang, Duan, Shen, and Leonardis]{Chen2021FSNetFS}
Wei Chen, Xi Jia, Hyung~Jin Chang, Jinming Duan, Linlin Shen, and Ale{\vs} Leonardis.
\newblock Fs-net: Fast shape-based network for category-level 6d object pose estimation with decoupled rotation mechanism.
\newblock \emph{2021 IEEE/CVF Conference on Computer Vision and Pattern Recognition (CVPR)}, pages 1581--1590, 2021.

\bibitem[Chen et~al.(2023)Chen, Di, Zhai, Manhardt, Zhang, Zhang, Tombari, Navab, and Busam]{Chen2023SecondPoseSD}
Yamei Chen, Yan Di, Guangyao Zhai, Fabian Manhardt, Chenyangguang Zhang, Ruida Zhang, Federico Tombari, Nassir Navab, and Benjamin Busam.
\newblock Secondpose: Se(3)-consistent dual-stream feature fusion for category-level pose estimation.
\newblock \emph{2024 IEEE/CVF Conference on Computer Vision and Pattern Recognition (CVPR)}, pages 9959--9969, 2023.

\bibitem[Chen et~al.(2024)Chen, Di, Zhai, Manhardt, Zhang, Zhang, Tombari, Navab, and Busam]{Chen2024SecondPoseSC}
Yamei Chen, Yan Di, Guangyao Zhai, Fabian Manhardt, Chenyangguang Zhang, Ruida Zhang, Federico Tombari, Nassir Navab, and Benjamin Busam.
\newblock Secondpose: Se(3)-consistent dual-stream feature fusion for category-level pose estimation.
\newblock In \emph{2024 IEEE/CVF Conference on Computer Vision and Pattern Recognition (CVPR)}, pages 9959--9969, 2024.

\bibitem[Corsetti et~al.(2024)Corsetti, Boscaini, Oh, Cavallaro, and Poiesi]{corsetti2024oryon}
Jaime Corsetti, Davide Boscaini, Changjae Oh, Andrea Cavallaro, and Fabio Poiesi.
\newblock Open-vocabulary object 6d pose estimation.
\newblock In \emph{2024 IEEE/CVF Conference on Computer Vision and Pattern Recognition (CVPR)}, pages 18071--18080, 2024.

\bibitem[Darcet et~al.(2024)Darcet, Oquab, Mairal, and Bojanowski]{Darcet2023VisionTN}
Timoth{'e}{\'e} Darcet, Maxime Oquab, Julien Mairal, and Piotr Bojanowski.
\newblock Vision transformers need registers.
\newblock In \emph{International Conference on Learning Representations}, 2024.

\bibitem[Denninger et~al.(2023)Denninger, Winkelbauer, Sundermeyer, Boerdijk, Knauer, Strobl, Humt, and Triebel]{Denninger2023}
Maximilian Denninger, Dominik Winkelbauer, Martin Sundermeyer, Wout Boerdijk, Markus Knauer, Klaus~H. Strobl, Matthias Humt, and Rudolph Triebel.
\newblock Blenderproc2: A procedural pipeline for photorealistic rendering.
\newblock \emph{Journal of Open Source Software}, 8\penalty0 (82):\penalty0 4901, 2023.

\bibitem[Di et~al.(2022)Di, Zhang, Lou, Manhardt, Ji, Navab, and Tombari]{Di2022GPVPoseCO}
Yan Di, Ruida Zhang, Zhiqiang Lou, Fabian Manhardt, Xiangyang Ji, Nassir Navab, and Federico Tombari.
\newblock Gpv-pose: Category-level object pose estimation via geometry-guided point-wise voting.
\newblock \emph{2022 IEEE/CVF Conference on Computer Vision and Pattern Recognition (CVPR)}, pages 6771--6781, 2022.

\bibitem[Do et~al.(2018)Do, Cai, Pham, and Reid]{Deep-6DPose}
Thanh-Toan Do, Ming Cai, Trung Pham, and Ian Reid.
\newblock Deep-6dpose: Recovering 6d object pose from a single rgb image.
\newblock \emph{ArXiv}, 2018.

\bibitem[Dosovitskiy et~al.(2021)Dosovitskiy, Beyer, Kolesnikov, Weissenborn, Zhai, Unterthiner, Dehghani, Minderer, Heigold, Gelly, Uszkoreit, and Houlsby]{Dosovitskiy2020AnII}
Alexey Dosovitskiy, Lucas Beyer, Alexander Kolesnikov, Dirk Weissenborn, Xiaohua Zhai, Thomas Unterthiner, Mostafa Dehghani, Matthias Minderer, Georg Heigold, Sylvain Gelly, Jakob Uszkoreit, and Neil Houlsby.
\newblock An image is worth 16x16 words: Transformers for image recognition at scale.
\newblock In \emph{International Conference on Learning Representations}, 2021.

\bibitem[Doumanoglou et~al.(2015)Doumanoglou, Kouskouridas, Malassiotis, and Kim]{Doumanoglou2015Recovering6O}
Andreas Doumanoglou, Rigas Kouskouridas, Sotiris Malassiotis, and Tae-Kyun Kim.
\newblock Recovering 6d object pose and predicting next-best-view in the crowd.
\newblock \emph{2016 IEEE Conference on Computer Vision and Pattern Recognition (CVPR)}, pages 3583--3592, 2015.

\bibitem[Downs et~al.(2022)Downs, Francis, Koenig, Kinman, Hickman, Reymann, McHugh, and Vanhoucke]{Downs2022GoogleSO}
Laura Downs, Anthony Francis, Nate Koenig, Brandon Kinman, Ryan~Michael Hickman, Krista Reymann, Thomas~Barlow McHugh, and Vincent Vanhoucke.
\newblock Google scanned objects: A high-quality dataset of 3d scanned household items.
\newblock \emph{2022 International Conference on Robotics and Automation (ICRA)}, pages 2553--2560, 2022.

\bibitem[Drost et~al.(2017)Drost, Ulrich, Bergmann, Härtinger, and Steger]{ITODD}
Bertram Drost, Markus Ulrich, Paul Bergmann, Philipp Härtinger, and Carsten Steger.
\newblock Introducing mvtec itodd — a dataset for 3d object recognition in industry.
\newblock In \emph{2017 IEEE International Conference on Computer Vision Workshops (ICCVW)}, pages 2200--2208, 2017.

\bibitem[Fu et~al.(2023)Fu, Zhu, and Wu]{Fu2023DTLDT}
Minghao Fu, Ke Zhu, and Jianxin Wu.
\newblock Dtl: Disentangled transfer learning for visual recognition.
\newblock In \emph{AAAI Conference on Artificial Intelligence}, 2023.

\bibitem[Guo et~al.(2023)Guo, Wen, Yuan, Tremblay, Tyree, Smith, and Birchfield]{Guo2023HANDAL}
Andrew Guo, Bowen Wen, Jianhe Yuan, Jonathan Tremblay, Stephen Tyree, Jeffrey Smith, and Stan Birchfield.
\newblock Handal: A dataset of real-world manipulable object categories with pose annotations, affordances, and reconstructions.
\newblock In \emph{2023 IEEE/RSJ International Conference on Intelligent Robots and Systems (IROS)}, pages 11428--11435, 2023.

\bibitem[Haugaard and Buch(2021)]{Haugaard2021SurfEmbDA}
Rasmus~Laurvig Haugaard and Anders~Glent Buch.
\newblock Surfemb: Dense and continuous correspondence distributions for object pose estimation with learnt surface embeddings.
\newblock \emph{2022 IEEE/CVF Conference on Computer Vision and Pattern Recognition (CVPR)}, pages 6739--6748, 2021.

\bibitem[He et~al.(2015)He, Zhang, Ren, and Sun]{resnet}
Kaiming He, X. Zhang, Shaoqing Ren, and Jian Sun.
\newblock Deep residual learning for image recognition.
\newblock \emph{2016 IEEE Conference on Computer Vision and Pattern Recognition (CVPR)}, pages 770--778, 2015.

\bibitem[He et~al.(2022)He, Sun, Wang, Huang, Bao, and Zhou]{he2022oneposeplusplus}
Xingyi He, Jiaming Sun, Yuang Wang, Di Huang, Hujun Bao, and Xiaowei Zhou.
\newblock Onepose++: Keypoint-free one-shot object pose estimation without {CAD} models.
\newblock In \emph{Advances in Neural Information Processing Systems}, 2022.

\bibitem[Heo et~al.(2024)Heo, Park, Han, and Yun]{Heo2024RotaryPE}
Byeongho Heo, Song Park, Dongyoon Han, and Sangdoo Yun.
\newblock Rotary position embedding for vision transformer.
\newblock In \emph{European Conference on Computer Vision}, 2024.

\bibitem[Hoda{\v{n}} et~al.(2017)Hoda{\v{n}}, Haluza, Obdr{\v{z}}{\'a}lek, Matas, Lourakis, and Zabulis]{hodan2017tless}
Tom{\'a}{\v{s}} Hoda{\v{n}}, Pavel Haluza, {\v{S}}t{\v{e}}p{\'a}n Obdr{\v{z}}{\'a}lek, Ji{\v{r}}{\'\i} Matas, Manolis Lourakis, and Xenophon Zabulis.
\newblock {T-LESS}: An {RGB-D} dataset for {6D} pose estimation of texture-less objects.
\newblock \emph{IEEE Winter Conference on Applications of Computer Vision (WACV)}, 2017.

\bibitem[Hodan et~al.(2018)Hodan, Michel, Brachmann, Kehl, GlentBuch, Kraft, Drost, Vidal, Ihrke, Zabulis, Sahin, Manhardt, Tombari, Kim, Matas, and Rother]{Hodan_2018_ECCV}
Tomas Hodan, Frank Michel, Eric Brachmann, Wadim Kehl, Anders GlentBuch, Dirk Kraft, Bertram Drost, Joel Vidal, Stephan Ihrke, Xenophon Zabulis, Caner Sahin, Fabian Manhardt, Federico Tombari, Tae-Kyun Kim, Jiri Matas, and Carsten Rother.
\newblock Bop: Benchmark for 6d object pose estimation.
\newblock In \emph{Proceedings of the European Conference on Computer Vision (ECCV)}, 2018.

\bibitem[Hoda{\v{n}} et~al.(2020)Hoda{\v{n}}, Sundermeyer, Drost, Labb{\'e}, Brachmann, Michel, Rother, and Matas]{hodan2020bop}
Tom{\'a}{\v{s}} Hoda{\v{n}}, Martin Sundermeyer, Bertram Drost, Yann Labb{\'e}, Eric Brachmann, Frank Michel, Carsten Rother, and Ji{\v{r}}{\'i} Matas.
\newblock {BOP} challenge 2020 on {6D} object localization.
\newblock \emph{European Conference on Computer Vision Workshops (ECCVW)}, 2020.

\bibitem[Hodan et~al.(2024)Hodan, Sundermeyer, Labbe, Nguyen, Wang, Brachmann, Drost, Lepetit, Rother, and Matas]{hodan2024bopchallenge2023detection}
Tomas Hodan, Martin Sundermeyer, Yann Labbe, Van~Nguyen Nguyen, Gu Wang, Eric Brachmann, Bertram Drost, Vincent Lepetit, Carsten Rother, and Jiri Matas.
\newblock Bop challenge 2023 on detection, segmentation and pose estimation of seen and unseen rigid objects, 2024.

\bibitem[Kaskman et~al.(2019)Kaskman, Zakharov, Shugurov, and Ilic]{HB}
Roman Kaskman, Sergey Zakharov, Ivan Shugurov, and Slobodan Ilic.
\newblock Homebreweddb: Rgb-d dataset for 6d pose estimation of 3d objects.
\newblock In \emph{2019 IEEE/CVF International Conference on Computer Vision Workshop (ICCVW)}, pages 2767--2776, 2019.

\bibitem[Kehl et~al.(2017)Kehl, Manhardt, Tombari, Ilic, and Navab]{Kehl2017SSD6DMR}
Wadim Kehl, Fabian Manhardt, Federico Tombari, Slobodan Ilic, and Nassir Navab.
\newblock Ssd-6d: Making rgb-based 3d detection and 6d pose estimation great again.
\newblock \emph{2017 IEEE International Conference on Computer Vision (ICCV)}, pages 1530--1538, 2017.

\bibitem[Labb'e et~al.(2020)Labb'e, Carpentier, Aubry, and Sivic]{Labbe2020CosyPoseCM}
Yann Labb'e, Justin Carpentier, Mathieu Aubry, and Josef Sivic.
\newblock Cosypose: Consistent multi-view multi-object 6d pose estimation.
\newblock In \emph{European Conference on Computer Vision}, 2020.

\bibitem[Labb'e et~al.(2022)Labb'e, Manuelli, Mousavian, Tyree, Birchfield, Tremblay, Carpentier, Aubry, Fox, and Sivic]{Labbe2022MegaPose6P}
Yann Labb'e, Lucas Manuelli, Arsalan Mousavian, Stephen Tyree, Stan Birchfield, Jonathan Tremblay, Justin Carpentier, Mathieu Aubry, Dieter Fox, and Josef Sivic.
\newblock Megapose: 6d pose estimation of novel objects via render \& compare.
\newblock In \emph{Conference on Robot Learning}, 2022.

\bibitem[Lee et~al.(2025)Lee, Wen, Kang, Kang, Kweon, and Yoon]{lee2025any6d}
Taeyeop Lee, Bowen Wen, Minjun Kang, Gyuree Kang, In~So Kweon, and Kuk-Jin Yoon.
\newblock {Any6D}: Model-free 6d pose estimation of novel objects.
\newblock In \emph{Proceedings of the Computer Vision and Pattern Recognition Conference (CVPR)}, 2025.

\bibitem[Lefaudeux et~al.(2022)Lefaudeux, Massa, Liskovich, Xiong, Caggiano, Naren, Xu, Hu, Tintore, Zhang, Labatut, Haziza, Wehrstedt, Reizenstein, and Sizov]{xFormers2022}
Benjamin Lefaudeux, Francisco Massa, Diana Liskovich, Wenhan Xiong, Vittorio Caggiano, Sean Naren, Min Xu, Jieru Hu, Marta Tintore, Susan Zhang, Patrick Labatut, Daniel Haziza, Luca Wehrstedt, Jeremy Reizenstein, and Grigory Sizov.
\newblock xformers: A modular and hackable transformer modelling library.
\newblock \url{https://github.com/facebookresearch/xformers}, 2022.

\bibitem[Li et~al.(2022)Li, Yu, Shugurov, Busam, Yang, and Ilic]{Li2022NeRFPoseAF}
Fu Li, Hao Yu, Ivan~S. Shugurov, Benjamin Busam, Shaowu Yang, and Slobodan Ilic.
\newblock Nerf-pose: A first-reconstruct-then-regress approach for weakly-supervised 6d object pose estimation.
\newblock \emph{2023 IEEE/CVF International Conference on Computer Vision Workshops (ICCVW)}, pages 2115--2125, 2022.

\bibitem[Li et~al.(2018)Li, Wang, Ji, Xiang, and Fox]{Li2018DeepIMDI}
Yi Li, Gu Wang, Xiangyang Ji, Yu Xiang, and Dieter Fox.
\newblock Deepim: Deep iterative matching for 6d pose estimation.
\newblock \emph{International Journal of Computer Vision}, 128:\penalty0 657 -- 678, 2018.

\bibitem[Li et~al.(2019)Li, Wang, and Ji]{Li2019CDPNCD}
Zhigang Li, Gu Wang, and Xiangyang Ji.
\newblock Cdpn: Coordinates-based disentangled pose network for real-time rgb-based 6-dof object pose estimation.
\newblock \emph{2019 IEEE/CVF International Conference on Computer Vision (ICCV)}, pages 7677--7686, 2019.

\bibitem[Lin et~al.(2021)Lin, Wei, Li, Xu, Jia, and Li]{Lin2021DualPoseNetC6}
Jiehong Lin, Zewei Wei, Zhihao Li, Songcen Xu, Kui Jia, and Yuanqing Li.
\newblock Dualposenet: Category-level 6d object pose and size estimation using dual pose network with refined learning of pose consistency.
\newblock \emph{2021 IEEE/CVF International Conference on Computer Vision (ICCV)}, pages 3540--3549, 2021.

\bibitem[Lin et~al.(2022)Lin, Wei, Ding, and Jia]{Lin2022CategoryLevel6O}
Jiehong Lin, Zewei Wei, Changxing Ding, and Kui Jia.
\newblock Category-level 6d object pose and size estimation using self-supervised deep prior deformation networks.
\newblock In \emph{European Conference on Computer Vision}, 2022.

\bibitem[Lin et~al.(2023{\natexlab{a}})Lin, Liu, Lu, and Jia]{Lin2023SAM6DSA}
Jiehong Lin, Lihua Liu, Dekun Lu, and Kui Jia.
\newblock Sam-6d: Segment anything model meets zero-shot 6d object pose estimation.
\newblock \emph{2024 IEEE/CVF Conference on Computer Vision and Pattern Recognition (CVPR)}, pages 27906--27916, 2023{\natexlab{a}}.

\bibitem[Lin et~al.(2023{\natexlab{b}})Lin, Wei, Zhang, and Jia]{Lin2023VINetBC}
Jiehong Lin, Zewei Wei, Yabin Zhang, and Kui Jia.
\newblock Vi-net: Boosting category-level 6d object pose estimation via learning decoupled rotations on the spherical representations.
\newblock \emph{2023 IEEE/CVF International Conference on Computer Vision (ICCV)}, pages 13955--13965, 2023{\natexlab{b}}.

\bibitem[Lin et~al.(2017)Lin, Goyal, Girshick, He, and Doll{\'a}r]{Lin2017FocalLF}
Tsung-Yi Lin, Priya Goyal, Ross~B. Girshick, Kaiming He, and Piotr Doll{\'a}r.
\newblock Focal loss for dense object detection.
\newblock \emph{IEEE Transactions on Pattern Analysis and Machine Intelligence}, 42:\penalty0 318--327, 2017.

\bibitem[Lin et~al.(2024)Lin, Yang, Gao, and Zhang]{Lin2024InstanceAdaptiveAG}
Xiao Lin, Wenfei Yang, Yuan Gao, and Tianzhu Zhang.
\newblock Instance-adaptive and geometric-aware keypoint learning for category-level 6d object pose estimation.
\newblock \emph{2024 IEEE/CVF Conference on Computer Vision and Pattern Recognition (CVPR)}, pages 21040--21049, 2024.

\bibitem[Liu et~al.(2022{\natexlab{a}})Liu, Wen, Peng, Lin, Long, Komura, and Wang]{liu2022gen6d}
Yuan Liu, Yilin Wen, Sida Peng, Cheng Lin, Xiaoxiao Long, Taku Komura, and Wenping Wang.
\newblock Gen6d: Generalizable model-free 6-dof object pose estimation from rgb images.
\newblock In \emph{ECCV}, 2022{\natexlab{a}}.

\bibitem[Liu et~al.(2022{\natexlab{b}})Liu, Wen, Peng, Lin, Long, Komura, and Wang]{Chen2020LearningCS}
Yuan Liu, Yilin Wen, Sida Peng, Chu-Hsing Lin, Xiaoxiao Long, Taku Komura, and Wenping Wang.
\newblock Gen6d: Generalizable model-free 6-dof object pose estimation from rgb images.
\newblock In \emph{European Conference on Computer Vision}, 2022{\natexlab{b}}.

\bibitem[Lu et~al.(2024)Lu, Jaykumar, Guo, Ruozzi, and Xiang]{Lu2024AdaptingPV}
Ya Lu, Jishnu Jaykumar, Yunhui Guo, Nicholas Ruozzi, and Yu Xiang.
\newblock Adapting pre-trained vision models for novel instance detection and segmentation.
\newblock \emph{ArXiv}, abs/2405.17859, 2024.

\bibitem[Mildenhall et~al.(2020)Mildenhall, Srinivasan, Tancik, Barron, Ramamoorthi, and Ng]{Mildenhall2020NeRF}
Ben Mildenhall, Pratul~P. Srinivasan, Matthew Tancik, Jonathan~T. Barron, Ravi Ramamoorthi, and Ren Ng.
\newblock Nerf.
\newblock \emph{Communications of the ACM}, 65:\penalty0 99 -- 106, 2020.

\bibitem[Moon et~al.(2024)Moon, Son, Hur, and Kim]{GenFlow}
Sungphill Moon, Hyeontae Son, Dongcheol Hur, and Sangwook Kim.
\newblock Genflow: Generalizable recurrent flow for 6d pose refinement of novel objects.
\newblock In \emph{2024 IEEE/CVF Conference on Computer Vision and Pattern Recognition (CVPR)}, pages 10039--10049, 2024.

\bibitem[Moon et~al.(2025)Moon, Son, Hur, and Kim]{moon2025co}
Sungphill Moon, Hyeontae Son, Dongcheol Hur, and Sangwook Kim.
\newblock Co-op: Correspondence-based novel object pose estimation.
\newblock In \emph{Proceedings of the IEEE/CVF Conference on Computer Vision and Pattern Recognition}, 2025.

\bibitem[M\"uller et~al.(2022)M\"uller, Evans, Schied, and Keller]{mueller2022instant}
Thomas M\"uller, Alex Evans, Christoph Schied, and Alexander Keller.
\newblock Instant neural graphics primitives with a multiresolution hash encoding.
\newblock \emph{ACM Trans. Graph.}, 41\penalty0 (4):\penalty0 102:1--102:15, 2022.

\bibitem[Mu{\~n}oz et~al.(2016)Mu{\~n}oz, Konishi, Beltran, Murino, and Bue]{Muoz2016Fast6P}
Enrique Mu{\~n}oz, Yoshinori Konishi, C. Beltran, Vittorio Murino, and Alessio~Del Bue.
\newblock Fast 6d pose from a single rgb image using cascaded forests templates.
\newblock \emph{2016 IEEE/RSJ International Conference on Intelligent Robots and Systems (IROS)}, pages 4062--4069, 2016.

\bibitem[Nguyen et~al.(2023)Nguyen, Groueix, Ponimatkin, Lepetit, and Hodan]{nguyen2023cnos}
Van~Nguyen Nguyen, Thibault Groueix, Georgy Ponimatkin, Vincent Lepetit, and Tomas Hodan.
\newblock Cnos: A strong baseline for cad-based novel object segmentation.
\newblock In \emph{Proceedings of the IEEE/CVF International Conference on Computer Vision}, pages 2134--2140, 2023.

\bibitem[Nguyen et~al.(2024)Nguyen, Groueix, Salzmann, and Lepetit]{nguyen2024gigaPose}
Van~Nguyen Nguyen, Thibault Groueix, Mathieu Salzmann, and Vincent Lepetit.
\newblock Gigapose: Fast and robust novel object pose estimation via one correspondence.
\newblock In \emph{Proceedings of the IEEE/CVF Conference on Computer Vision and Pattern Recognition}, 2024.

\bibitem[Oquab et~al.(2023)Oquab, Darcet, Moutakanni, Vo, Szafraniec, Khalidov, Fernandez, Haziza, Massa, El-Nouby, Howes, Huang, Xu, Sharma, Li, Galuba, Rabbat, Assran, Ballas, Synnaeve, Misra, Jegou, Mairal, Labatut, Joulin, and Bojanowski]{oquab2023dinov2}
Maxime Oquab, Timothée Darcet, Theo Moutakanni, Huy~V. Vo, Marc Szafraniec, Vasil Khalidov, Pierre Fernandez, Daniel Haziza, Francisco Massa, Alaaeldin El-Nouby, Russell Howes, Po-Yao Huang, Hu Xu, Vasu Sharma, Shang-Wen Li, Wojciech Galuba, Mike Rabbat, Mido Assran, Nicolas Ballas, Gabriel Synnaeve, Ishan Misra, Herve Jegou, Julien Mairal, Patrick Labatut, Armand Joulin, and Piotr Bojanowski.
\newblock Dinov2: Learning robust visual features without supervision, 2023.

\bibitem[{\"O}rnek et~al.(2024){\"O}rnek, Labb\'e, Tekin, Ma, Keskin, Forster, and Hoda{\v{n}}]{ornek2024foundpose}
Evin~P{\i}nar {\"O}rnek, Yann Labb\'e, Bugra Tekin, Lingni Ma, Cem Keskin, Christian Forster, and Tom{\'a}{\v{s}} Hoda{\v{n}}.
\newblock Foundpose: Unseen object pose estimation with foundation features.
\newblock \emph{European Conference on Computer Vision (ECCV)}, 2024.

\bibitem[Park et~al.(2019)Park, Patten, and Vincze]{Park2019Pix2PosePC}
Kiru Park, Timothy Patten, and Markus Vincze.
\newblock Pix2pose: Pixel-wise coordinate regression of objects for 6d pose estimation.
\newblock \emph{2019 IEEE/CVF International Conference on Computer Vision (ICCV)}, pages 7667--7676, 2019.

\bibitem[Payet and Todorovic(2011)]{Payet2011FromCT}
Nadia Payet and Sinisa Todorovic.
\newblock From contours to 3d object detection and pose estimation.
\newblock \emph{2011 International Conference on Computer Vision}, pages 983--990, 2011.

\bibitem[Qi et~al.(2017)Qi, Yi, Su, and Guibas]{Qi2017PointNetDH}
C. Qi, L. Yi, Hao Su, and Leonidas~J. Guibas.
\newblock Pointnet++: Deep hierarchical feature learning on point sets in a metric space.
\newblock \emph{ArXiv}, abs/1706.02413, 2017.

\bibitem[Rad and Lepetit(2017)]{Rad2017BB8AS}
Mahdi Rad and Vincent Lepetit.
\newblock Bb8: A scalable, accurate, robust to partial occlusion method for predicting the 3d poses of challenging objects without using depth.
\newblock \emph{2017 IEEE International Conference on Computer Vision (ICCV)}, pages 3848--3856, 2017.

\bibitem[Shazeer(2020)]{Shazeer2020GLUVI}
Noam~M. Shazeer.
\newblock Glu variants improve transformer.
\newblock \emph{ArXiv}, abs/2002.05202, 2020.

\bibitem[Shugurov et~al.(2022)Shugurov, Li, Busam, and Ilic]{OSOP}
Ivan Shugurov, Fu Li, Benjamin Busam, and Slobodan Ilic.
\newblock Osop: A multi-stage one shot object pose estimation framework.
\newblock In \emph{2022 IEEE/CVF Conference on Computer Vision and Pattern Recognition (CVPR)}, pages 6825--6834, 2022.

\bibitem[Su et~al.(2021)Su, Lu, Pan, Wen, and Liu]{Su2021RoFormerET}
Jianlin Su, Yu Lu, Shengfeng Pan, Bo Wen, and Yunfeng Liu.
\newblock Roformer: Enhanced transformer with rotary position embedding.
\newblock \emph{ArXiv}, abs/2104.09864, 2021.

\bibitem[Su et~al.(2022)Su, Saleh, Fetzer, Rambach, Navab, Busam, Stricker, and Tombari]{Su2022ZebraPoseCT}
Yongzhi Su, Mahdi Saleh, Torben Fetzer, Jason~Raphael Rambach, Nassir Navab, Benjamin Busam, Didier Stricker, and Federico Tombari.
\newblock Zebrapose: Coarse to fine surface encoding for 6dof object pose estimation.
\newblock \emph{2022 IEEE/CVF Conference on Computer Vision and Pattern Recognition (CVPR)}, pages 6728--6738, 2022.

\bibitem[Sun et~al.(2022)Sun, Wang, Zhang, He, Zhao, Zhang, and Zhou]{sun2022onepose}
Jiaming Sun, Zihao Wang, Siyu Zhang, Xingyi He, Hongcheng Zhao, Guofeng Zhang, and Xiaowei Zhou.
\newblock {OnePose}: One-shot object pose estimation without {CAD} models.
\newblock \emph{CVPR}, 2022.

\bibitem[Tekin et~al.(2017)Tekin, Sinha, and Fua]{Tekin2017RealTimeSS}
Bugra Tekin, Sudipta~N. Sinha, and Pascal~V. Fua.
\newblock Real-time seamless single shot 6d object pose prediction.
\newblock \emph{2018 IEEE/CVF Conference on Computer Vision and Pattern Recognition}, pages 292--301, 2017.

\bibitem[Terzakis and Lourakis(2020)]{Terzakis2020ACF}
George Terzakis and Manolis I.~A. Lourakis.
\newblock A consistently fast and globally optimal solution to the perspective-n-point problem.
\newblock In \emph{European Conference on Computer Vision}, 2020.

\bibitem[Tyree et~al.(2022)Tyree, Tremblay, To, Cheng, Mosier, Smith, and Birchfield]{Tyree2022HOPE}
Stephen Tyree, Jonathan Tremblay, Thang To, Jia Cheng, Terry Mosier, Jeffrey Smith, and Stan Birchfield.
\newblock 6-dof pose estimation of household objects for robotic manipulation: An accessible dataset and benchmark.
\newblock In \emph{2022 IEEE/RSJ International Conference on Intelligent Robots and Systems (IROS)}, pages 13081--13088, 2022.

\bibitem[Ulrich et~al.(2012)Ulrich, Wiedemann, and Steger]{Ulrich2012CombiningSA}
Markus Ulrich, Christian Wiedemann, and Carsten Steger.
\newblock Combining scale-space and similarity-based aspect graphs for fast 3d object recognition.
\newblock \emph{IEEE Transactions on Pattern Analysis and Machine Intelligence}, 34:\penalty0 1902--1914, 2012.

\bibitem[Vincent et~al.(2009)Vincent, Francesc, and Pascal]{EPnP}
Lepetit Vincent, Moreno-Noguer Francesc, and Fua Pascal.
\newblock Epnp: An accurate o(n) solution to the pnp problem.
\newblock \emph{2009 International Journal of Computer Vision}, 2009.

\bibitem[Wang et~al.(2021)Wang, Manhardt, Tombari, and Ji]{Wang_2021_GDRN}
Gu Wang, Fabian Manhardt, Federico Tombari, and Xiangyang Ji.
\newblock {GDR-Net}: Geometry-guided direct regression network for monocular 6d object pose estimation.
\newblock In \emph{IEEE/CVF Conference on Computer Vision and Pattern Recognition (CVPR)}, pages 16611--16621, 2021.

\bibitem[Wang et~al.(2019{\natexlab{a}})Wang, Sridhar, Huang, Valentin, Song, and Guibas]{Wang_2019_CVPR}
He Wang, Srinath Sridhar, Jingwei Huang, Julien Valentin, Shuran Song, and Leonidas~J. Guibas.
\newblock Normalized object coordinate space for category-level 6d object pose and size estimation.
\newblock In \emph{The IEEE Conference on Computer Vision and Pattern Recognition (CVPR)}, 2019{\natexlab{a}}.

\bibitem[Wang et~al.(2019{\natexlab{b}})Wang, Sridhar, Huang, Valentin, Song, and Guibas]{Wang2019NormalizedOC}
He Wang, Srinath Sridhar, Jingwei Huang, Julien P.~C. Valentin, Shuran Song, and Leonidas~J. Guibas.
\newblock Normalized object coordinate space for category-level 6d object pose and size estimation.
\newblock \emph{2019 IEEE/CVF Conference on Computer Vision and Pattern Recognition (CVPR)}, pages 2637--2646, 2019{\natexlab{b}}.

\bibitem[Wen et~al.(2023)Wen, Yang, Kautz, and Birchfield]{foundationposewen2024}
Bowen Wen, Wei Yang, Jan Kautz, and Stanley~T. Birchfield.
\newblock Foundationpose: Unified 6d pose estimation and tracking of novel objects.
\newblock \emph{2024 IEEE/CVF Conference on Computer Vision and Pattern Recognition (CVPR)}, pages 17868--17879, 2023.

\bibitem[Xiang et~al.(2018)Xiang, Schmidt, Narayanan, and Fox]{xiang2018posecnn}
Yu Xiang, Tanner Schmidt, Venkatraman Narayanan, and Dieter Fox.
\newblock {PoseCNN}: A convolutional neural network for {6D} object pose estimation in cluttered scenes.
\newblock In \emph{Robotics: Science and Systems (RSS)}, 2018.

\bibitem[Yisheng et~al.(2022)Yisheng, Yao, Haoqiang, Qifeng, and Jian]{he2022fs6d}
He Yisheng, Wang Yao, Fan Haoqiang, Chen Qifeng, and Sun Jian.
\newblock Fs6d: Few-shot 6d pose estimation of novel objects.
\newblock \emph{CVPR}, 2022.

\bibitem[Zakharov et~al.(2019)Zakharov, Shugurov, and Ilic]{Zakharov2019DPOD6P}
Sergey Zakharov, Ivan~S. Shugurov, and Slobodan Ilic.
\newblock Dpod: 6d pose object detector and refiner.
\newblock \emph{2019 IEEE/CVF International Conference on Computer Vision (ICCV)}, pages 1941--1950, 2019.

\end{thebibliography}
}

 \clearpage
\appendix
\setcounter{page}{1}
\maketitlesupplementary

\section{Ablation studies}\label{sec:Additional_ablation_studies}

\subsection{Number of reference templates}
\begin{table}[ht]
    \centering
    \begin{adjustbox}{width=\linewidth}
    \begin{tabular}{cgccccc}
    \thickerhline
Model number & 1 & 7 & 9 & 10 & 11 & 12\\ \thickerhline
Num. templates & 42 & 42 & 12 & 12 & 162 & 162 \\
Neighb. templates & \cmark & \xmark & \xmark & \cmark & \xmark & \cmark \\
\hline
LM-O & 57.2 & 56.9 & 55.7 & 53.9 & \underline{57.3} & \textbf{57.8} \\ 
YCB-V &  65.1 & 65.5 & 65.0 & 60.4 & \underline{65.8} & \textbf{66.5} \\
T-LESS &  \underline{47.1} & 45.1 & 42.8 & 42.1 & 45.4 & \textbf{47.4} \\ \hline
Avg &  \underline{56.5} & 55.8 & 54.5 & 52.1 & 56.2 & \textbf{57.2}\\
Time [s] & 0.59 & 0.52 & 0.47 & 0.57 & 0.76 & 0.81\\
\thickerhline
    \end{tabular}
    \end{adjustbox}
    \caption{Impact of the inference setting with varied number of templates and neighborhood templates. Models 1 and 7 are taken over from \cref{tab:Ablation}.}
    \label{tab:A_Grid}
\end{table}

In \cref{tab:A_Grid} we explore impact of inference setting while considering the number of templates based on the subdivision of icosahedron and inclusion of the prediction's six neighboring templates.
The models 1 and 7 are transferred from \cref{tab:Ablation}.
We observe that our model benefits from an increase number of templates at cost of additional computational resources. 
Furthermore, adding the 6 neighborhood templates increases performance if the template sampling is sufficiently dense.

\subsection{Image resolution}
\begin{table}[ht]
    \centering
    \begin{adjustbox}{width=\linewidth}
    \begin{tabular}{cgcccccccc}
    \thickerhline
Model number & 1 & 13 & 14 & 15 & 16 & 17 & 18 & 19 & 20\\
\thickerhline
Template image size & 420 & 420 & 420 & 280 & 280 & 280 & 140 & 140 & 140\\
Test image size & 420 & 280 & 140 & 420 & 280 & 140 & 420 & 280 & 140 \\
\hline
LM-O & 57.2 & \textbf{57.5} & 54.4 & 56.6 & 56.6 & 53.4 & 51.6 & 51.5 & 48.7 \\ 
YCB-V &  65.1 & \textbf{66.2} & 64.4 & 63.4 & 65.1 & 64.2 & 60.9 & 60.8 & 57.3 \\
T-LESS &  \textbf{47.1} & 45.6 & 37.1 & 46.6 & 45.1 & 37.7 & 38.3 & 37.9 & 32.4 \\ \hline
Avg & \textbf{56.5} & 56.4 & 52.0 & 55.5 & 55.6 & 51.8 & 50.3 & 50.1 & 46.1 \\
Time [s] & 0.59 & 0.47 & 0.43 & 0.51 & 0.46 & 0.42 & 0.49 & 0.45 & 0.42 \\
\thickerhline
    \end{tabular}
    \end{adjustbox}
    \caption{Impact of the template image resolution and test image resolution on the coarse 6D pose estimation. Models 1 is taken over from \cref{tab:Ablation}}
    \label{tab:resolutions}
\end{table}

We study the influence of the template and test image resolutions. The results of this experiment are shown in \cref{tab:resolutions}. 
For this experiment, resolutions were selected as multiples of the DinoV2 patch size (14 pixels). 
The resolution is defined as $r = 14s, s \in \{10, 20, 30\}$. 
For texture-less objects present in the T-LESS dataset, higher input resolution is crucial because performance suffers greatly with low-resolution images.
Even with the lowest resolution setting for both test image and reference templates, our method achieves higher average recall when compared to \cite{ornek2024foundpose, nguyen2024gigaPose, GenFlow, Labbe2022MegaPose6P} with coarse-only strategy. 
Further quantitative results, which we do not include into the table, show that performance plateaus at a resolution of $(420, 420)$ and slightly degrades at higher resolutions (e.g., $(700, 700)$).
We attribute this to two factors. 
First, at very high resolutions, a single 14×14 patch captures a smaller and less descriptive region of an image than it would at lower resolutions.
Second, due to limited computational resources, our model was trained on resolutions only up to $(420, 420)$ and therefore does not generalize to the longer token sequences generated by these out-of-distribution inputs.

\subsection{2D detection}\label{sec:A_2D}
\begin{table}[ht]
\centering
\begin{adjustbox}{width=\linewidth} 
\begin{tabular}{cccccc} 
\thickerhline 
2D detection source  & LM-O & YCB-V & T-LESS & Avg & Time \\
\thickerhline
Ground truth & \textbf{68.3} & \textbf{69.7} & \textbf{70.6} &\textbf{ 69.5}  & - \\\hline
\rowcolor{gray!25}CNOS FastSAM \cite{nguyen2023cnos} & 57.2 & 65.1 & 47.1 & 56.6 & \underline{0.59}\\
NIDS \cite{Lu2024AdaptingPV} & \underline{58.3} & \underline{66.3} & \underline{52.3} & \underline{58.9} & 0.76\\
SAM6D FastSAM \cite{Lin2023SAM6DSA} & 57.6 & 65.3 & 48.2 & 57.0 & \textbf{0.57}\\
\thickerhline 
\end{tabular}
\end{adjustbox}
\caption{Impact of the 2D detection's quality on the coarse 6D pose estimation of the OPFormer. The CNOS FastSAM 2D detector was used for evaluation in the rest of the experiments.}
\label{tab:A_2Ddet}
\end{table}

\cref{tab:A_2Ddet} presents the impact of different 2D detection sources on OPFormer's performance. We compare results from several proposed methods \cite{nguyen2023cnos, Lin2023SAM6DSA, Lu2024AdaptingPV} against the performance achieved using ground-truth bounding boxes.
Erroneous 2D detections are a primary cause of pose estimation failures, leading to a significant drop in Average Recall (AR) scores. This issue is particularly pronounced on the T-LESS dataset, where some objects have a similar appearance, causing the detector to frequently mismatch object categories.
When comparing 2D detection methodologies, the NIDS detector \cite{Lu2024AdaptingPV} yields the highest detection accuracy, which in turn improves the final 6D pose estimation. However, this accuracy comes with a trade-off: an approximately $30 \%$ reduction in inference speed.

\section{Limitations}\label{sec:A_Limits}
Our analysis identifies two primary failure modes. First, the most significant performance degradation occurs when the 2D detector fails to produce a bounding box or misclassifies the object category. We provide a discussion and a quantitative analysis of this effect in \cref{sec:A_2D} (\cref{tab:A_2Ddet}). Second, the model faces challenges with symmetric or texture-less objects. For such objects, the descriptor vectors for symmetric parts are highly similar, which leads to ambiguous or incorrect feature matching.
Additionally, as the other RGB-based methods the prediction precisions degrades depending on the input image size resolution as the input size changes from down-sampling to up-sampling, which inherently introduces noise and artifacts.

The quality of the NeRF-based 3D reconstruction is highly sensitive to the input of multi-view images with known poses. Deficiencies in this input data, such as an insufficient number of views, poor image quality, inaccurate camera poses, or faulty background segmentation, prevent the model from learning a high-fidelity representation. This directly results in the generation of flawed RGB, NOCS, and depth templates. Since our method's accuracy relies on establishing robust 2D-3D correspondences between a test image and these templates, relying on inaccurate templates inevitably leads to poor feature matching and a significantly degraded final 6D pose estimation. 
\section{BOP results in detail}\label{sec:A_BOPdetails}
\begin{table}[ht]
\centering
\begin{adjustbox}{width=\linewidth}
\begin{tabular}{cccccc}
\thickerhline
Dataset & Refiner  & $\mathrm{AR_{VSD}}$ & $\mathrm{AR_{MSPD}}$ & $\mathrm{AR_{MSSD}}$ & $\mathrm{AR}$\\
\thickerhline                      
\multirow{3}{*}{LM-O} & \xmark      & 43.3     & 73.1     & 55.3     & 57.2     \\
&  1 hyps.                          & 47.4     & 72.9     & 60.9     & 60.4     \\
&  5 hyps.                          & 47.4     & 73.1     & 60.8    & 60.4     \\
\hline
\multirow{3}{*}{T-LESS} & \xmark    & 41.9     & 55.5     & 44.1     & 47.1     \\
&  1 hyps.                           & 48.8     & 56.8     & 50.0     & 51.9     \\
&  5 hyps.                          & 49.6     & 58.0     & 50.9    & 52.8     \\
\hline
\multirow{3}{*}{TUD-L} & \xmark     & 54.9     & 85.3     & 66.5     & 68.9     \\
&  1 hyps.                           & 55.9     & 85.6     & 67.0     & 69.5     \\
&  5 hyps.                          & 56.8     & 86.3     & 68.0    & 70.3     \\
\hline
\multirow{3}{*}{IC-BIN} & \xmark    & 40.7     & 52.7     & 43.3     & 45.6     \\
&  1 hyps.                           & 44.9     & 55.5     & 48.8     & 49.7     \\
&  5 hyps.                          & 44.5     & 55.7     & 48.8    & 49.7     \\
\hline
\multirow{3}{*}{ITODD} & \xmark     & 30.1     & 52.7     & 33.3     & 38.7     \\
&  1 hyps.                           & 33.6     & 52.3     & 37.6     & 41.2     \\ 
&  5 hyps.                          & 35.1     & 54.6     & 38.9    & 42.8     \\
\hline
\multirow{3}{*}{HB} & \xmark        & 70.7     &  80.3     & 74.1     & 75.0     \\
&  1 hyps.                           & 72.9     & 80.9     & 76.1     & 76.6     \\
&  5 hyps.                          & 72.9     & 81.0     & 76.2    & 76.7     \\
\hline
\multirow{3}{*}{YCB-V} & \xmark     & 54.9     & 78.4     & 62.1     & 65.1     \\
&  1 hyps.                           & 59.1     & 80.9     & 66.2     & 68.7     \\
&  5 hyps.                          & 59.0     & 81.0     & 66.2    & 68.7     \\

\thickerhline
\end{tabular}
\end{adjustbox}
\caption{Detailed scores of OPFormer from \cref{tab:BOP23} for 7 BOP-Core-Classic datasets.}
\label{tab:A_BOP23}
\end{table}

\begin{table}
\centering
\begin{adjustbox}{width=\linewidth}
\begin{tabular}{cccccc}
\thickerhline
Dataset & Task & Refiner  & $\mathrm{AP_{MSPD}}$ & $\mathrm{AP_{MSSD}}$ & $\mathrm{AP}$\\
\thickerhline
\multirow{4}{*}{HOPEv2} & \multirow{2}{*}{Model-based} & \xmark  & 42.5 & 31.4 & 37.0\\
&  & \cmark &  42.6 &  33.3 & 37.9 \\ \cline{2-6} 
& \multirow{2}{*}{Model-free} & \xmark & 47.1 & 26.0 & 36.6  \\
&  & \cmark &  46.5 &  24.8 & 35.6\\
\thickerhline
\multirow{4}{*}{HANDAL} & \multirow{2}{*}{Model-based} & \xmark  & 41.1 & 32.4 & 36.7\\
& & \cmark &  41.9 & 35.7 & 38.8 \\ \cline{2-6} 
& \multirow{2}{*}{Model-free} & \xmark & 37.2 & 24.5 & 30.9  \\
& & \cmark &  37.8 &  27.0 & 32.4 \\
\thickerhline
\multirow{2}{*}{HOT3D} & \multirow{2}{*}{Model-based} & \xmark  &  35.2 &  28.0 & 31.6 \\
&  & \cmark &  34.2 &  26.4 & 30.3 \\ 
\thickerhline
\end{tabular}
\end{adjustbox}
\caption{Detailed scores of OPFormer from \cref{tab:BOP24} on the BOP-H3 datasets in model-based and model-free 6D detection tasks.}
\label{tab:A_BOP24}
\end{table}

\cref{tab:A_BOP23} presents scores that complement those in \cref{tab:BOP23}, detailing result for the initial coarse estimation and subsequent pose refinements. As refinement we deploy MegaPose render-and-compare refiner on both single best hypothesis and the top five hypotheses.
Furthermore, \cref{tab:A_BOP24} provides detailed results to \cref{tab:BOP24}, showing the coarse estimation and refinement of single best hypothesis.
Both tables provide the results based on the error metrics described in \cref{sec:BOPCHallenge} and \cref{sec:A_metrics}.

\section{Metric description} \label{sec:A_metrics}
The objective of the evaluation metrics is to calculate the precision of the prediction, given the estimated pose $\mathbf{\hat{P}}$ and the ground truth $\mathbf{\Bar{P}}$ for a model $\mathcal{M}$ with set of vertices $ \mathcal{V}_\mathcal{M}$ and set of symmetries $S_\mathcal{M}$  in image~$I$.
For our evaluation we follow the BOP metrics as described by \cite{hodan2020bop}. 
 The calculation of both their Average Recall and Average Precision is performed as follows
\begin{align}
\mathrm{AR} &= (\mathrm{AR_{VSD}} +\mathrm{AR_{MSPD}} + \mathrm{AR_{MSSD}}) /3, \label{eq:BOP_AR}\\
\mathrm{AP} &= (\mathrm{AP_{MSPD}} + \mathrm{AP_{MSSD}}) /2. \label{eq:BOP_AP}
\end{align}
The errors for each metric are calculated as 
\begin{equation}\label{eq:VSD}
\begin{adjustbox}{width=\linewidth}
$
    e_{\mathrm{VSD}}(\hat{D},\Bar{D},\hat{V},\Bar{V},\tau) = \avg_{p\in\hat{V}\cup\Bar{V}}\begin{cases}
        0 & \text{if } p \in \hat{V}\cap\Bar{V} \\ &  \wedge \mid\hat{D}(p)-\Bar{D}(p)\mid < \tau,\\
        1 & \mathrm{othervise},
    \end{cases}$
\end{adjustbox}
\end{equation}
\begin{equation}\label{eq:MSSD}
\begin{adjustbox}{width=\linewidth}$
    e_{\mathrm{MSSD}}(\mathbf{\hat{P}, \Bar{P}},S_\mathcal{M}, \mathcal{V_M})= 
        \min_{\mathbf{S}\in S_\mathcal{M}} \max_{\mathbf{x}\in \mathcal{V_M}}  \lVert \mathbf{\hat{P}x} - \mathbf{\Bar{P}Sx} \rVert_2,
$\end{adjustbox}
\end{equation}
\begin{equation}\label{eq:MSPD}
\begin{adjustbox}{width=\linewidth}
$    e_{\mathrm{MSPD}}(\mathbf{\hat{P}}, \mathbf{\Bar{P}}, S_\mathcal{M}, \mathcal{V_M}) =
        \min_{\mathbf{S}\in S_\mathcal{M}} \max_{\mathbf{x}\in \mathcal{V_M}}  \lVert \mathrm{proj}(\mathbf{\hat{P}x}) - \mathrm{proj}(\mathbf{\Bar{P}Sx}) \rVert_2. 
$\end{adjustbox}
\end{equation}
The \textbf{VSD} (Visible Surface Discrepancy) metric is focused on aliment of the visible part of the model, using the  the visibility masks $\hat{V}$ and $\Bar{V}$ and the distance maps $\hat{D}$ and $\Bar{D}$, the image distance map $D_I$, and the misalignment tolerance $\tau$.
The \textbf{MSSD} (Maximum Symmetry-Aware Surface Distance) is a strict metric that calculates maximum distance between corresponding points of the estimated and ground truth surfaces, while considering possible object symmetries.
The \textbf{MSPD} (Maximum Symmetry-Aware Projection Distance) measures the pixel distance between the estimated and ground-truth image projections, while also considering the symmetries.

In contrast, other metrics, like rotation and translation error, do not consider the model's geometric properties, as they only account for the 6D pose $\mathbf{\Bar{P}}$ and $\mathbf{\hat{P}}$.  3DIoU only considers the geometry of the model's bounding box which can lead to misclassifications of continuously symmetric objects or, in cases of point or plane symmetries, where 3D bounding boxes can be identical but the model orientations differ.

ADD and ADI consider an object's geometric properties, but due to their calculation of average vertex distance, they are highly dependent on the mesh sampling. ADI accounts for symmetries by measuring the distance to the nearest neighbor; however, this approach becomes problematic with objects exhibiting discrete symmetries.

\section{Additional visualization} \label{sec:A_visualization}



\begin{figure*}
    \centering
    \subfloat{\includegraphics[width=.4\textwidth]{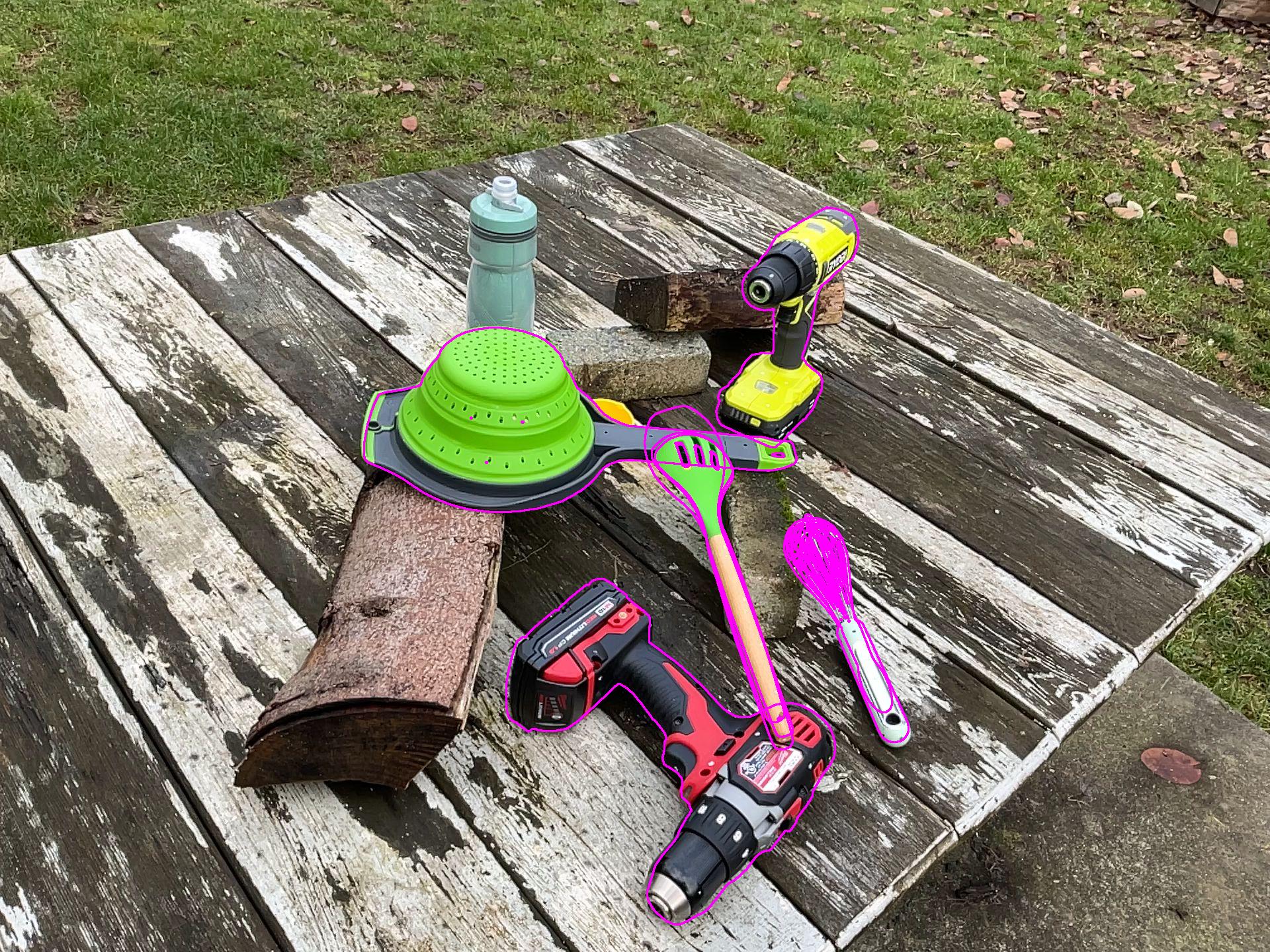}}\quad
    \subfloat{\includegraphics[width=.4\textwidth]{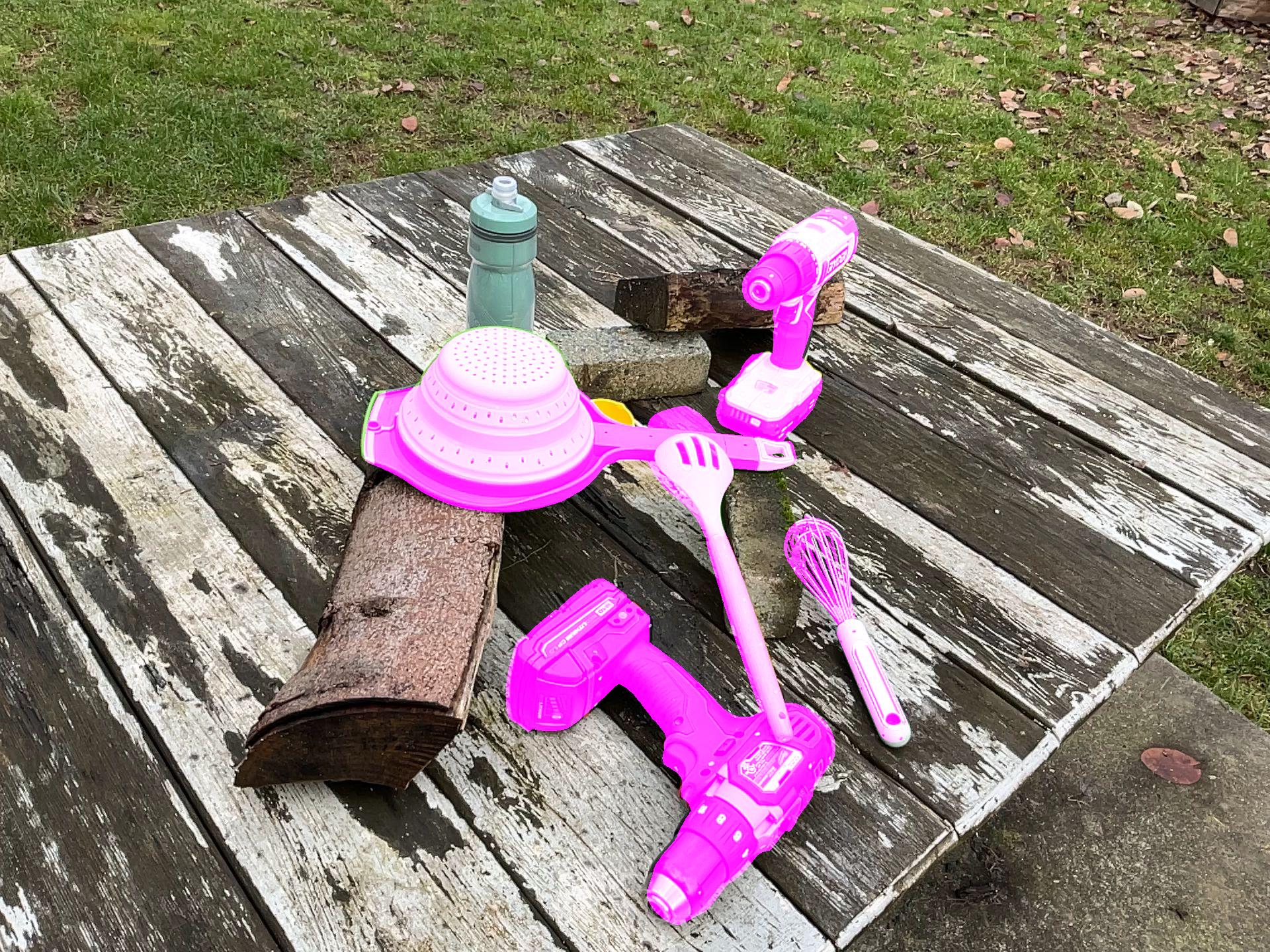}}\\

    \subfloat{\includegraphics[width=.4\textwidth]{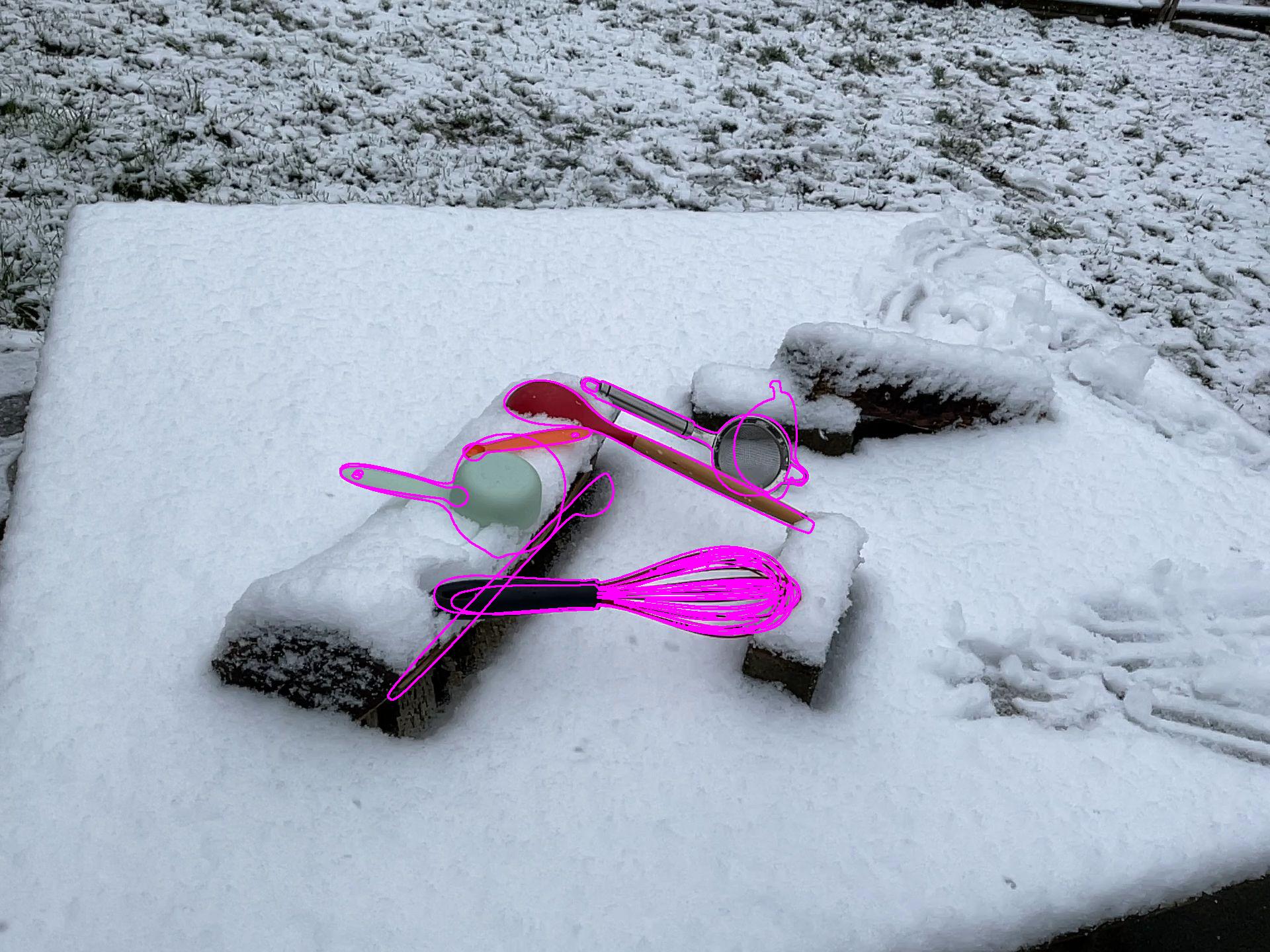}}\quad
    \subfloat{\includegraphics[width=.4\textwidth]{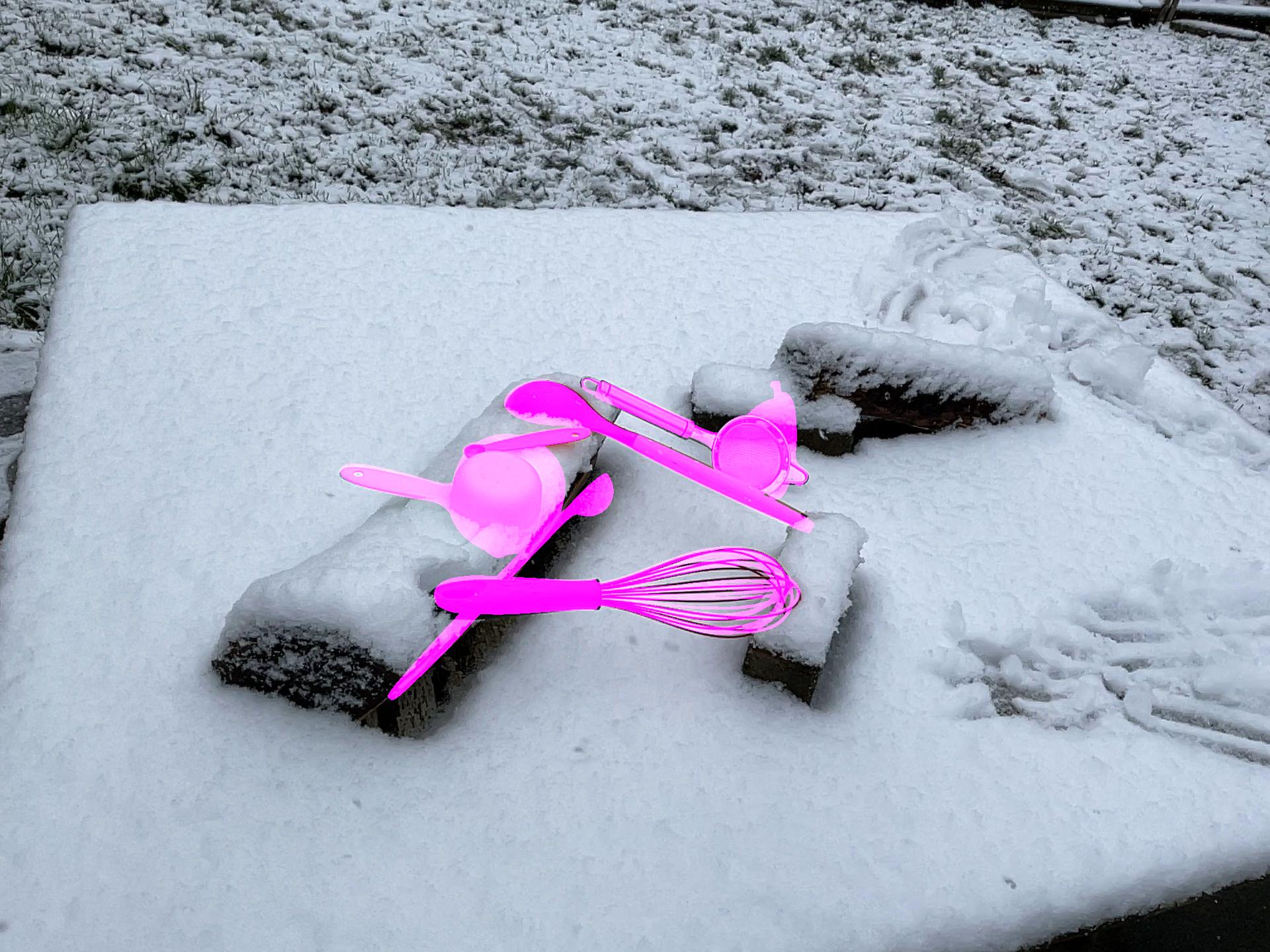}}\\

    \subfloat{\includegraphics[width=.4\textwidth]{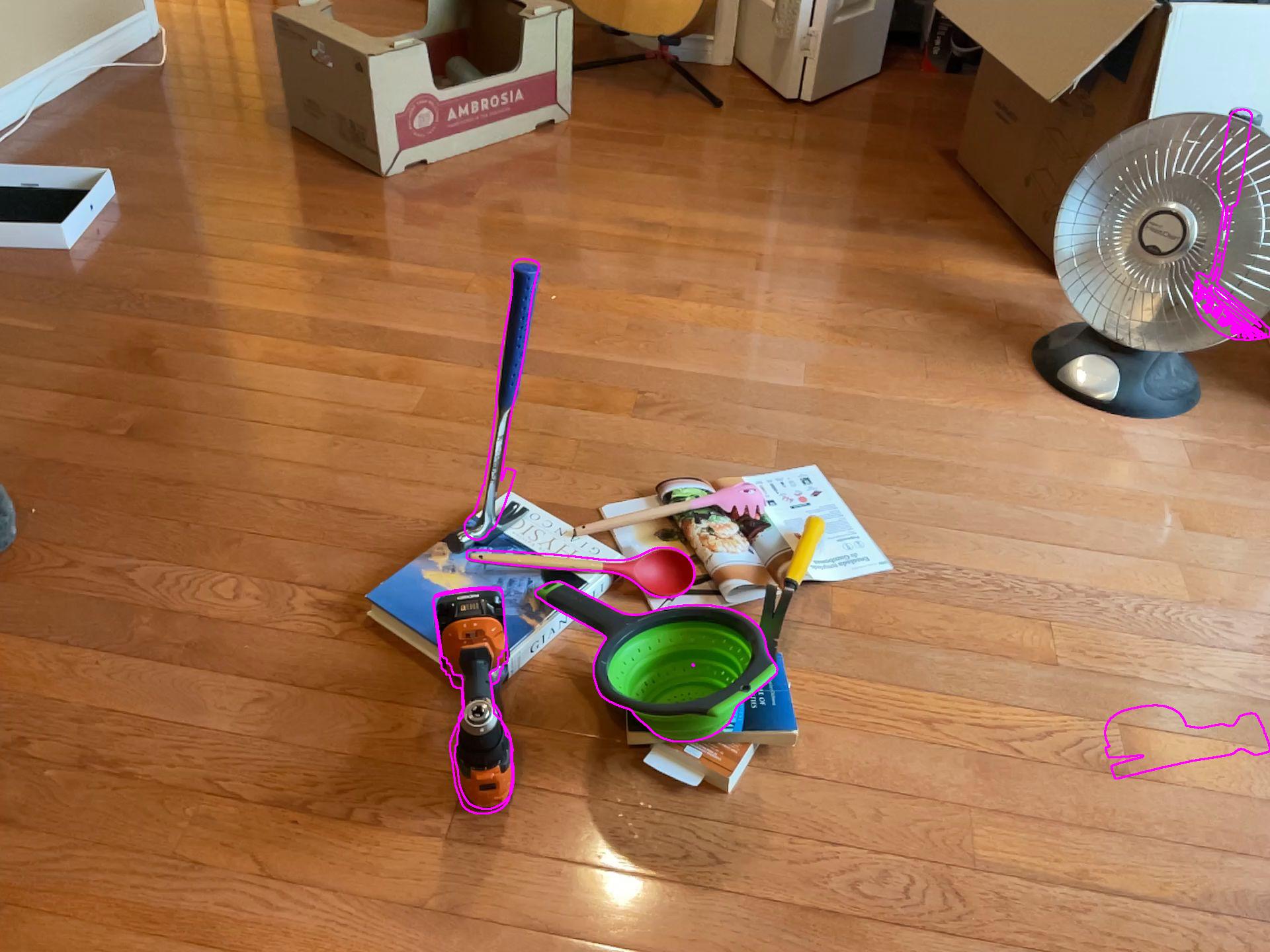}}\quad
    \subfloat{\includegraphics[width=.4\textwidth]{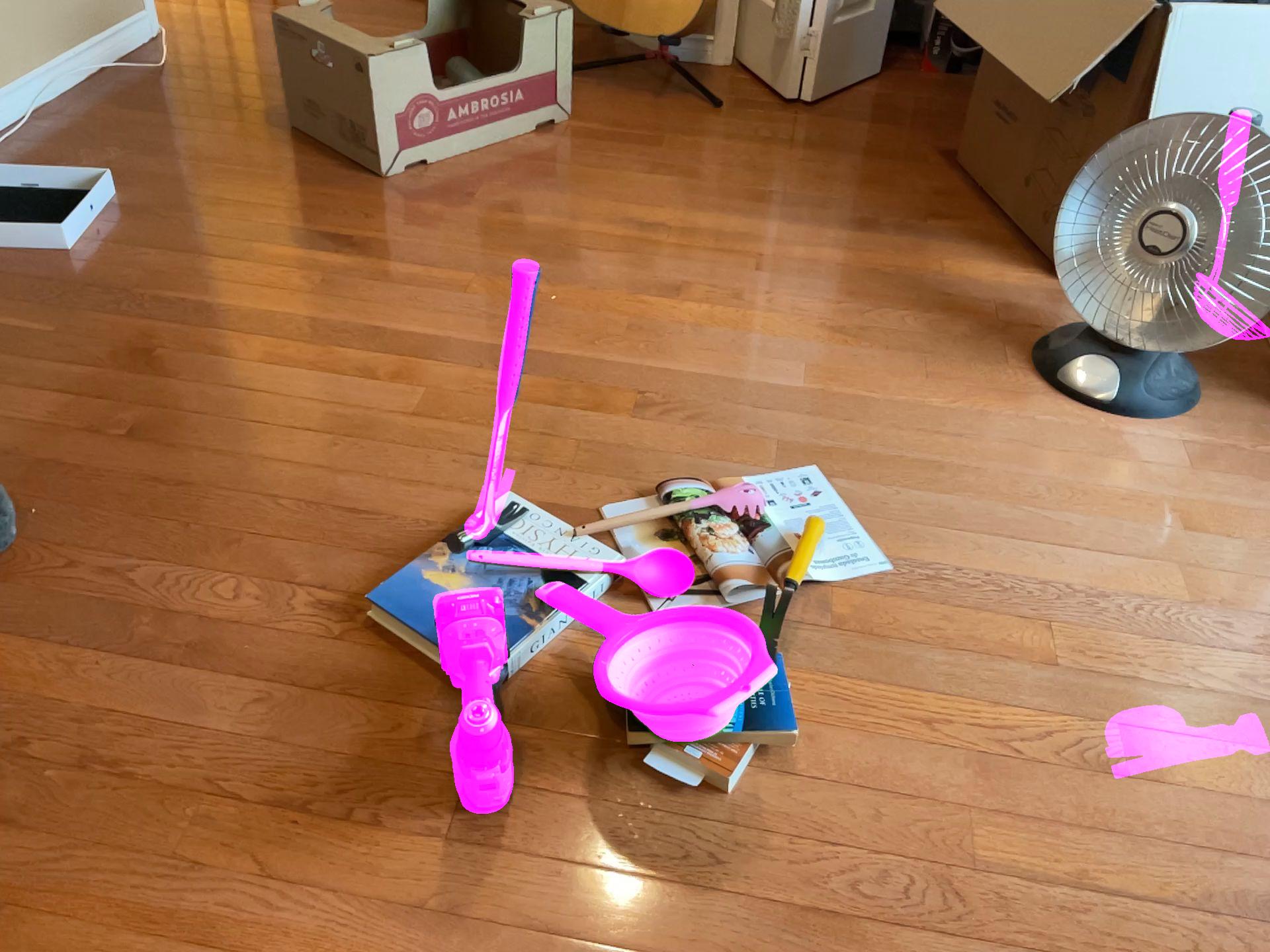}}\\

    \subfloat{\includegraphics[width=.4\textwidth]{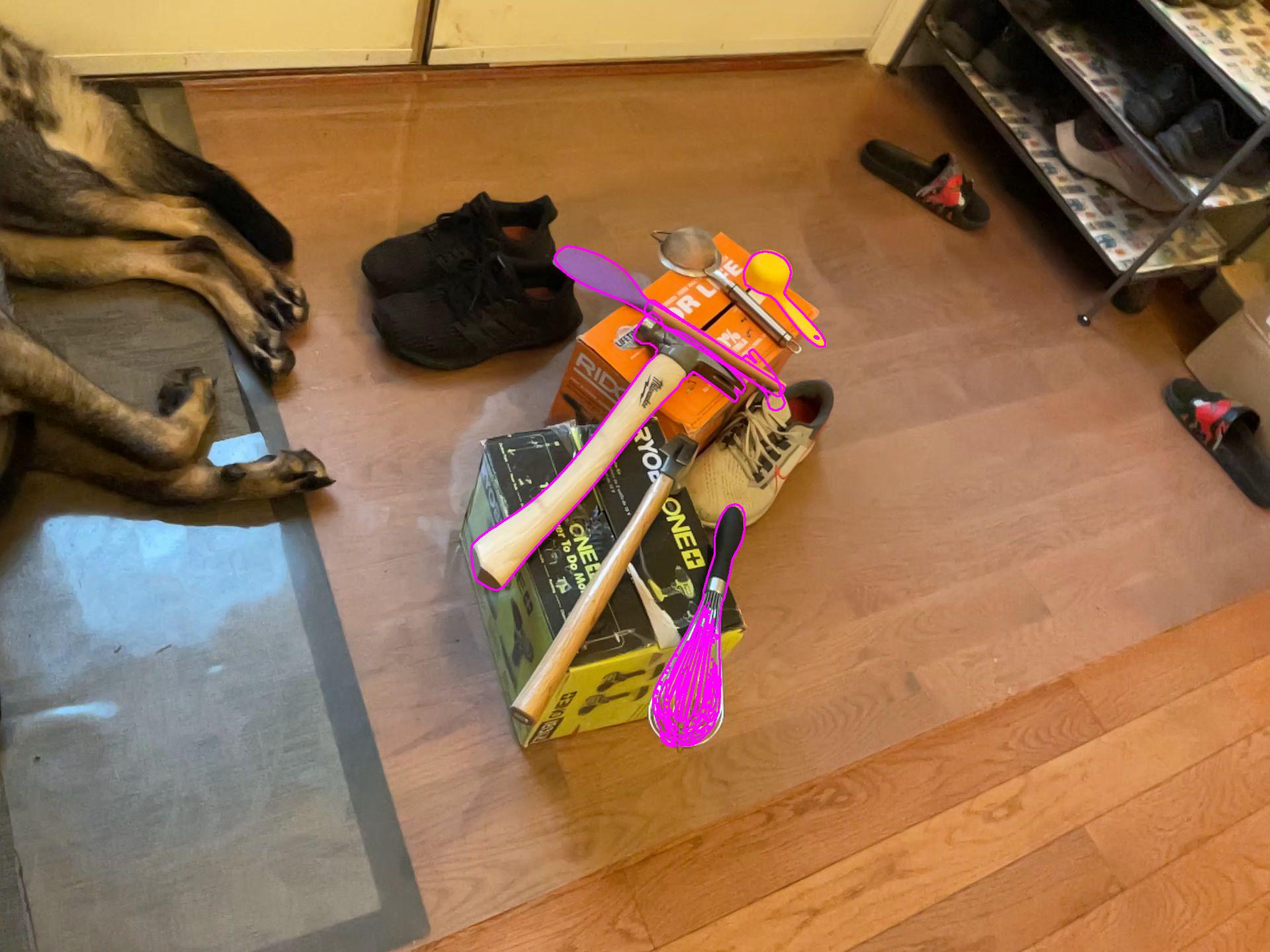}}\quad
    \subfloat{\includegraphics[width=.4\textwidth]{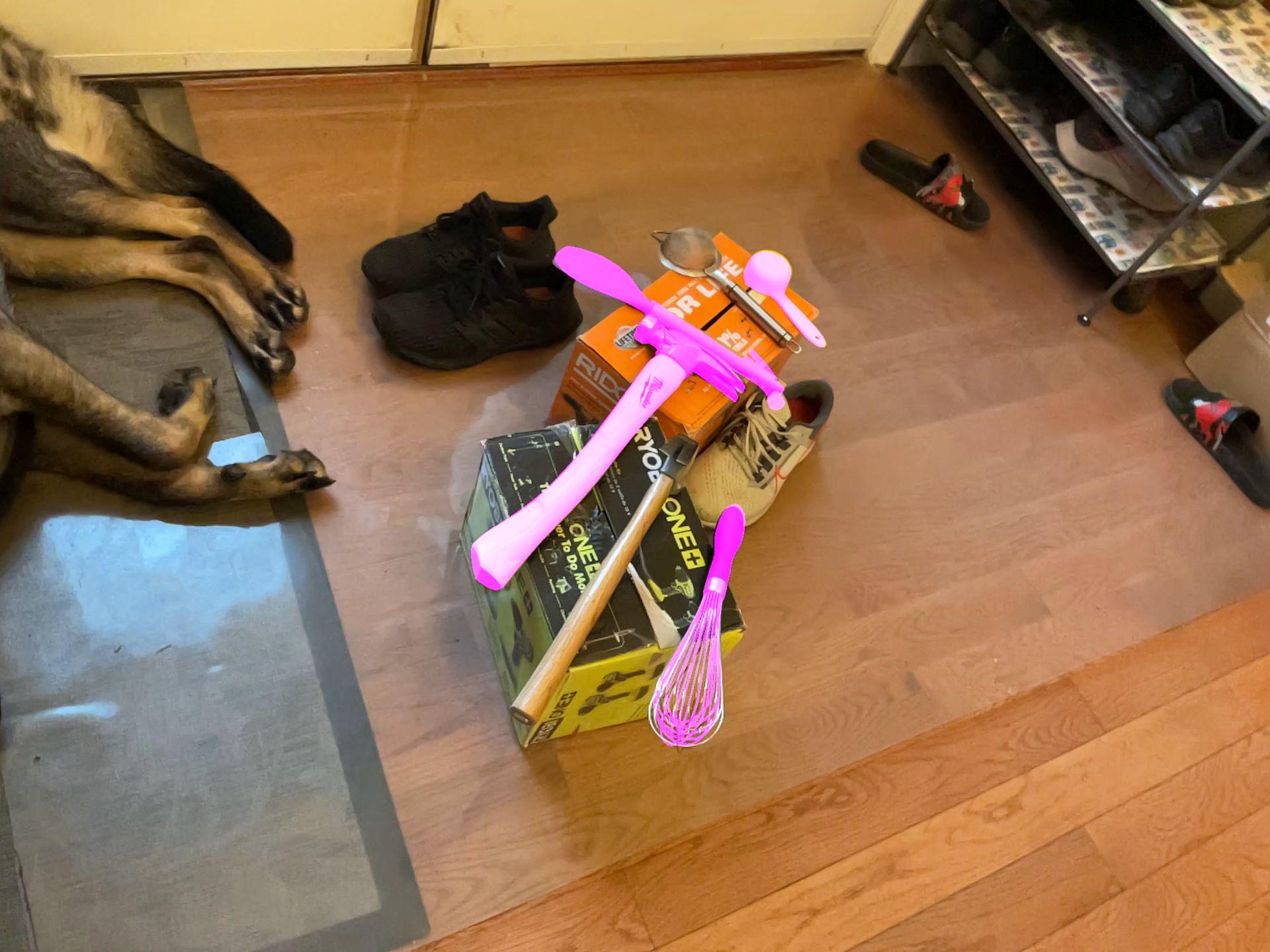}}\\

    \caption{HANDAL dataset visualization of 2D projection made with the predicted 6D poses. In the left column is visualized contour highlight and in the right the projected masks of the prediction}
    \label{fig:A0_HANDAL}
\end{figure*}

\begin{figure*}
  \centering
  \subfloat{\includegraphics[height=.2\textwidth]{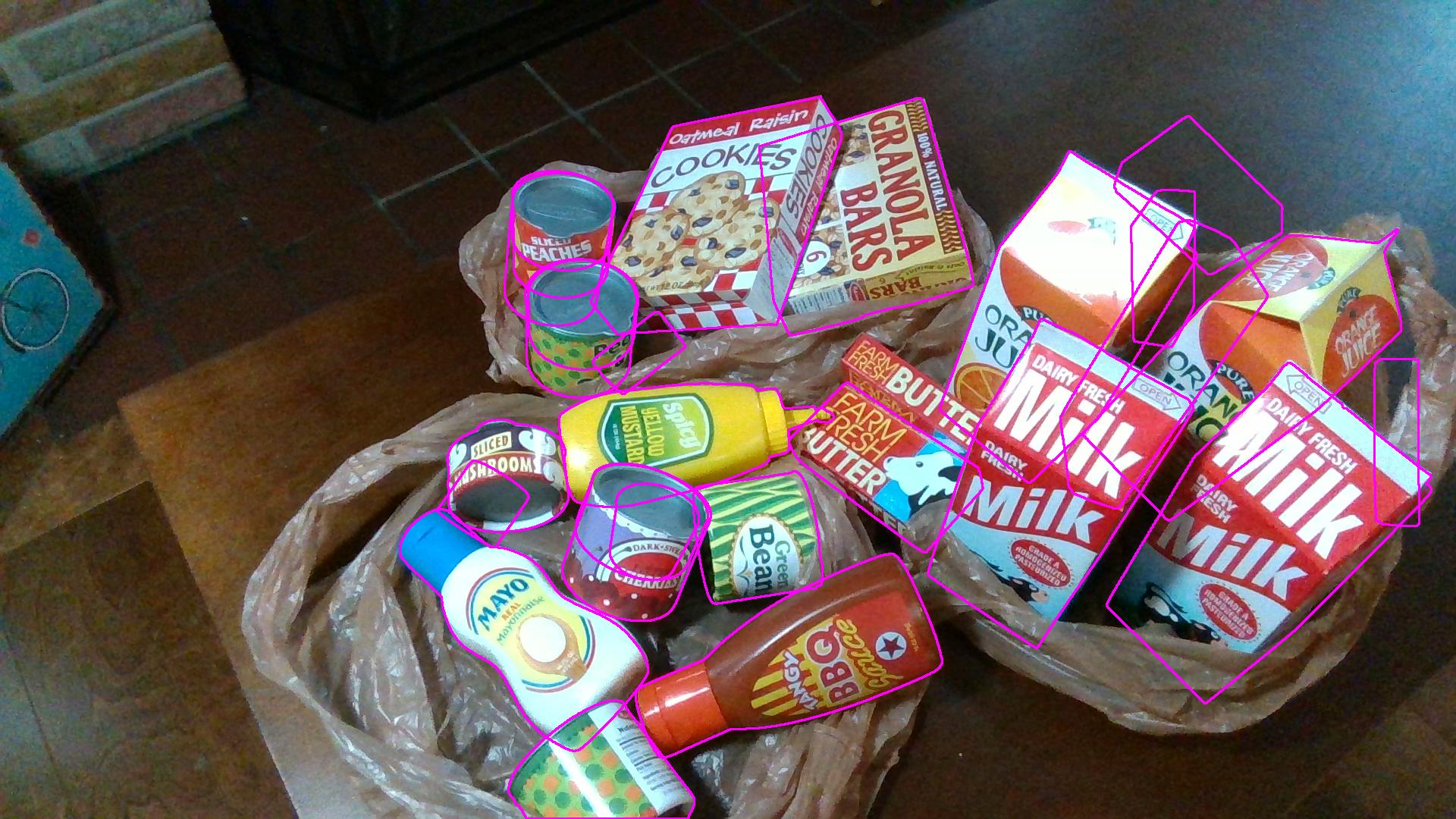}}\quad
  \subfloat{\includegraphics[height=.2\textwidth]{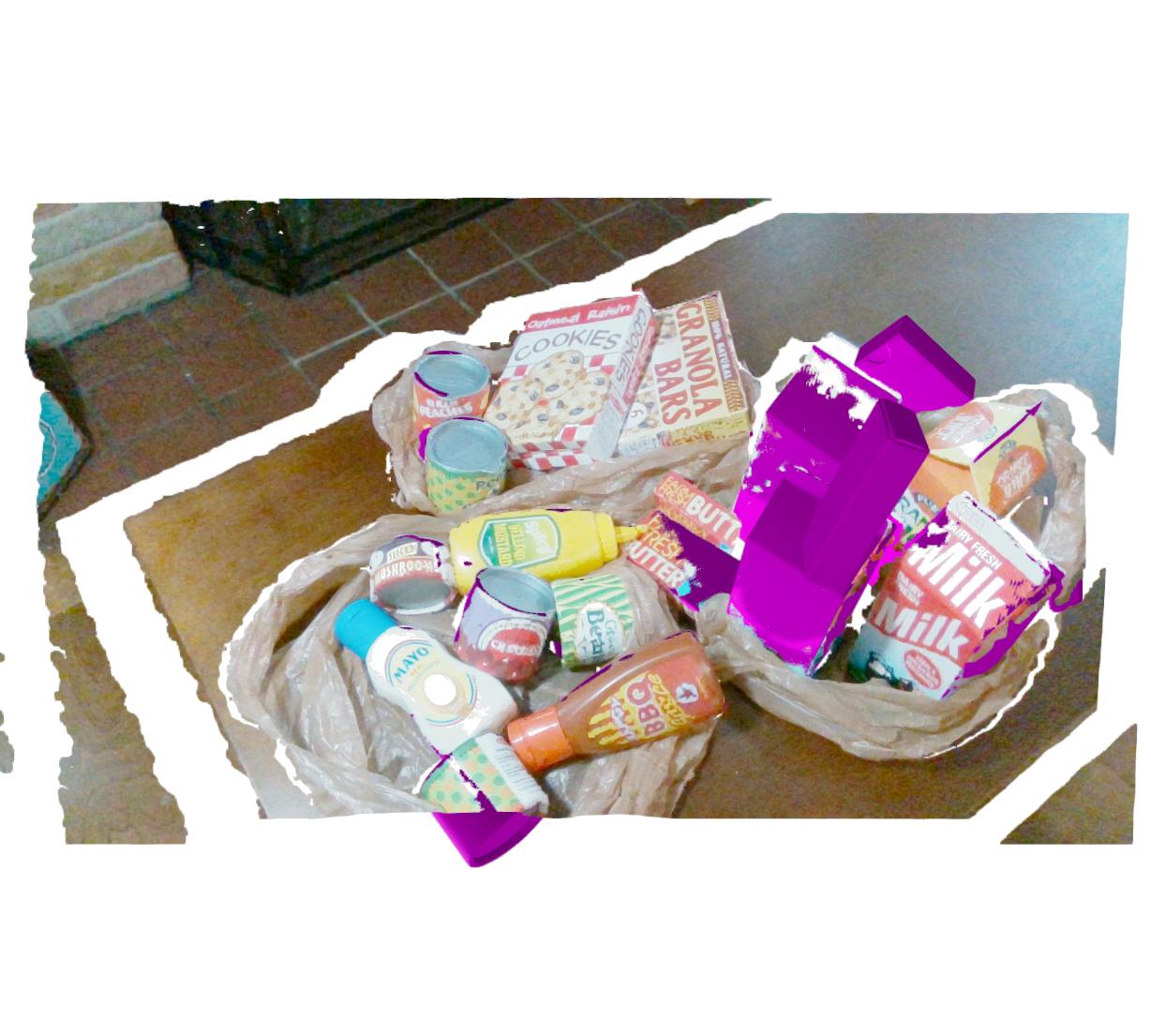}}\quad
  \subfloat{\includegraphics[height=.2\textwidth]{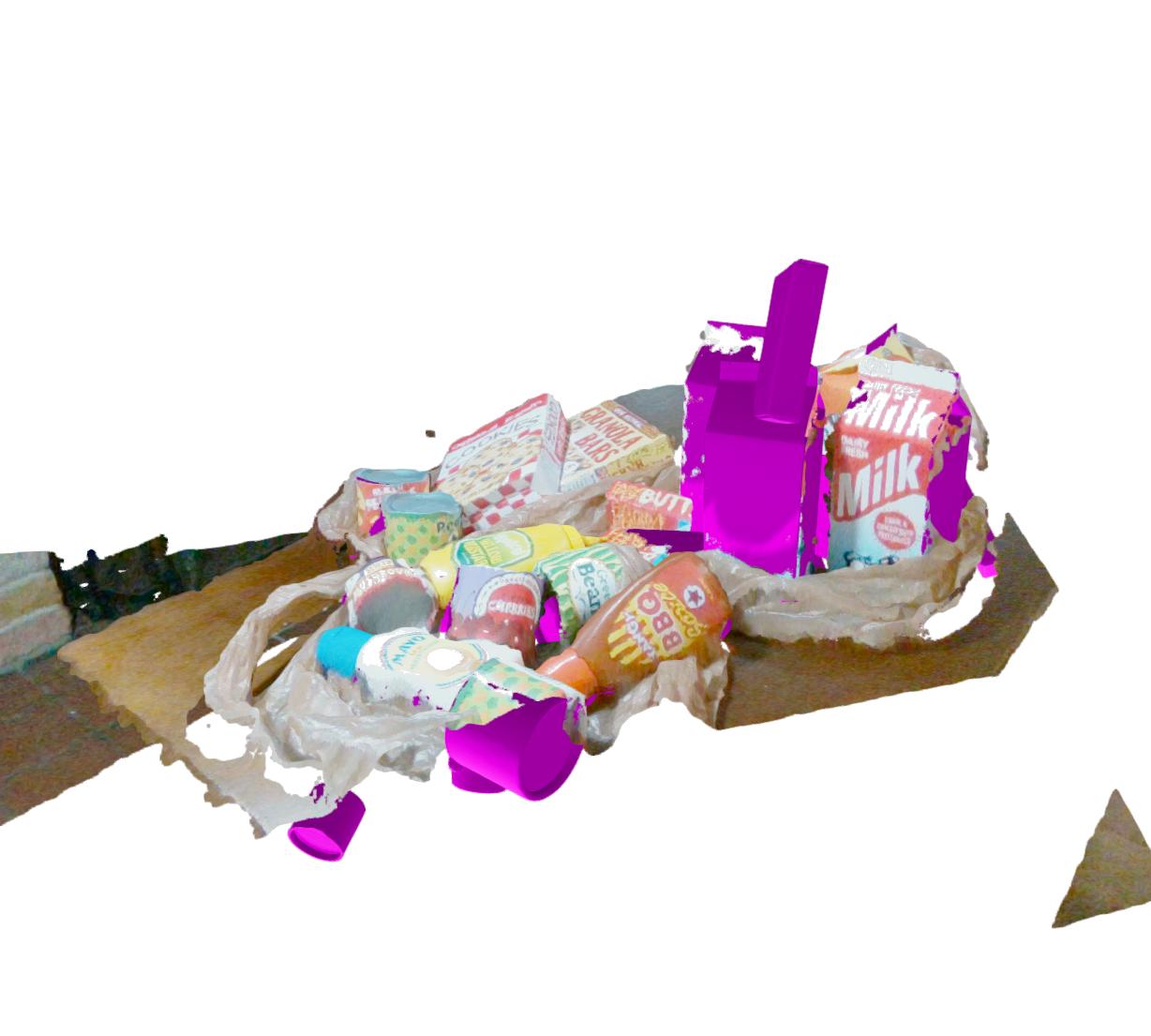}}\\
  
  \subfloat{\includegraphics[height=.2\textwidth]{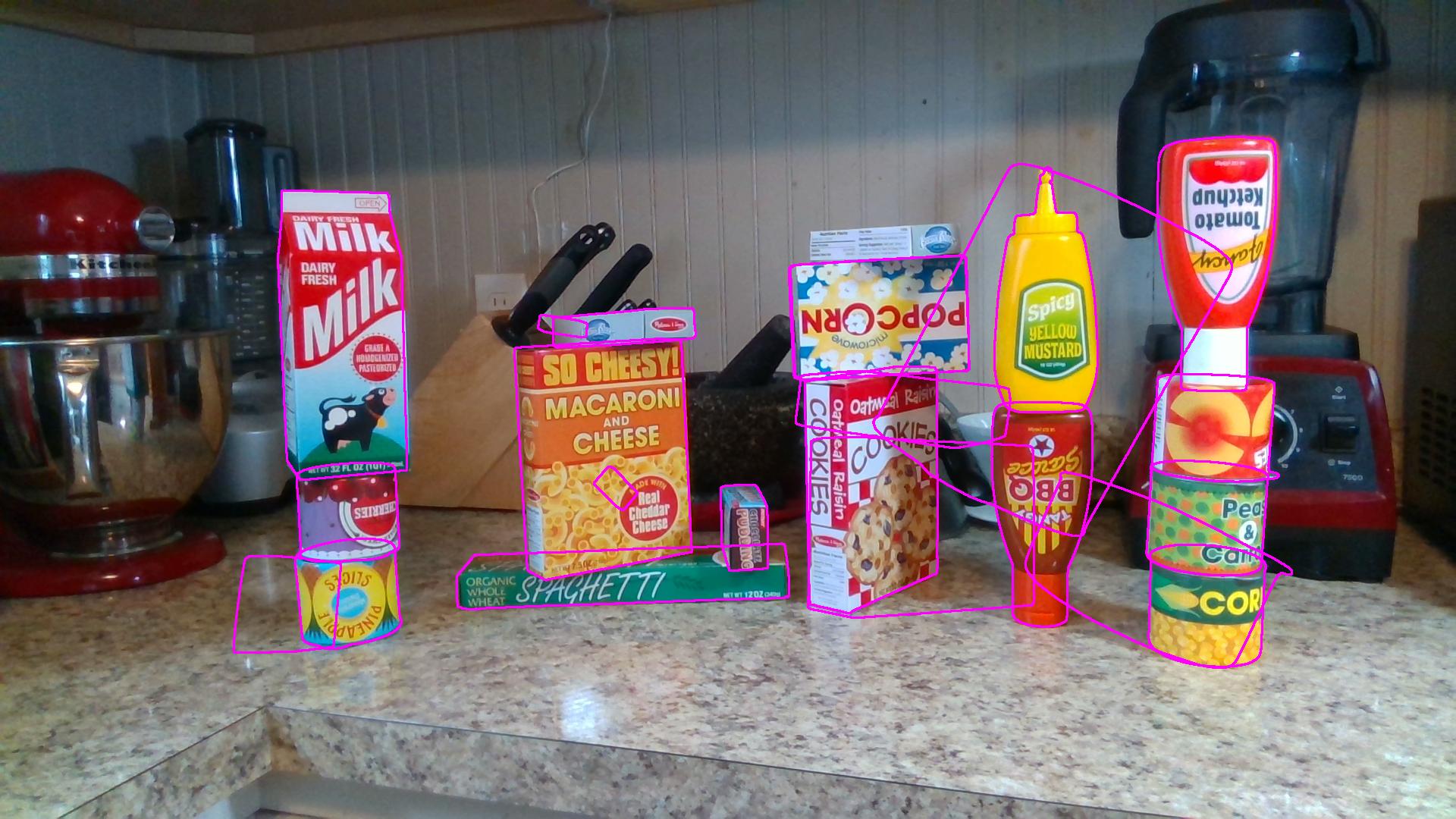}}\quad
  \subfloat{\includegraphics[height=.2\textwidth]{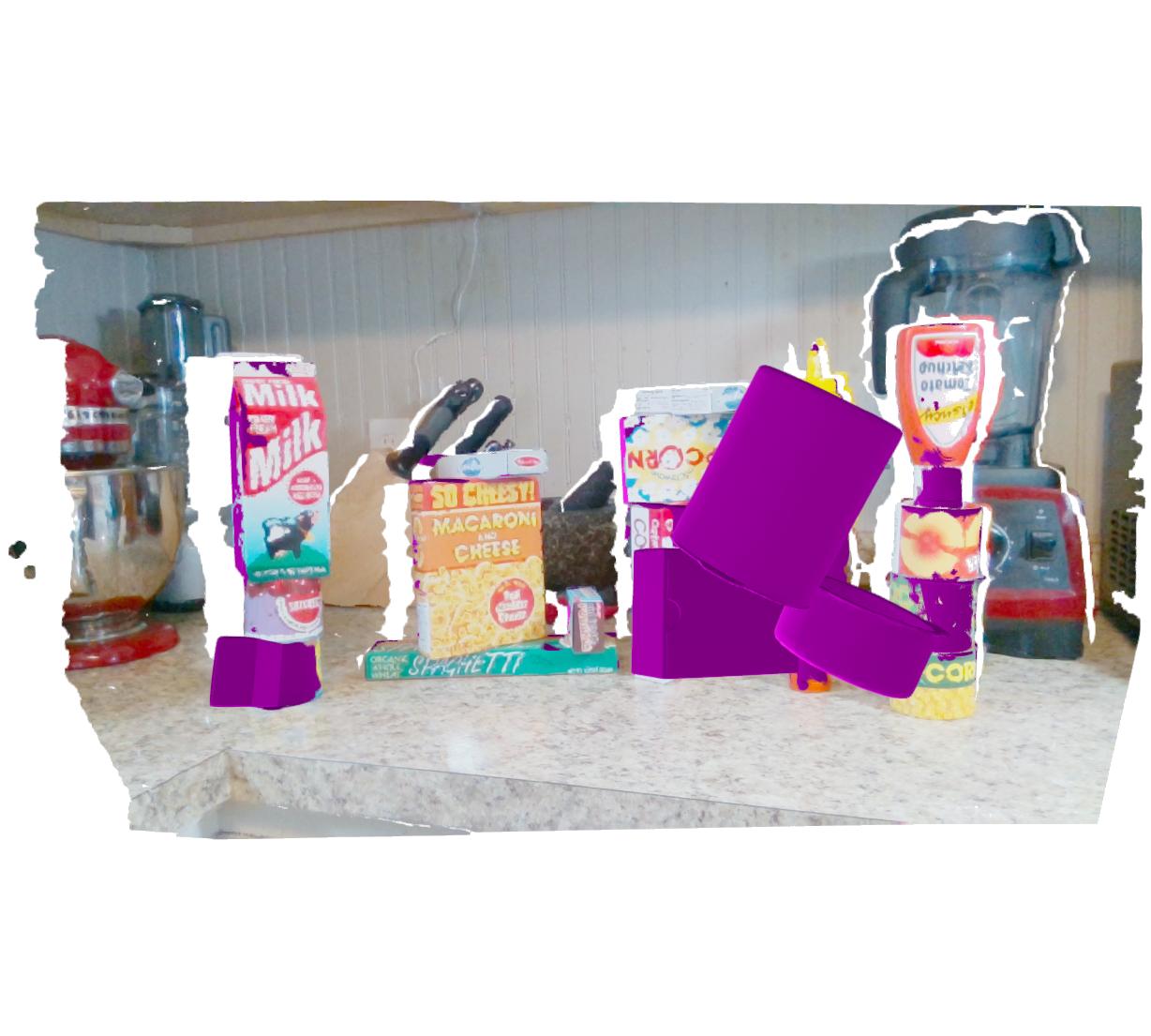}}\quad
  \subfloat{\includegraphics[height=.2\textwidth]{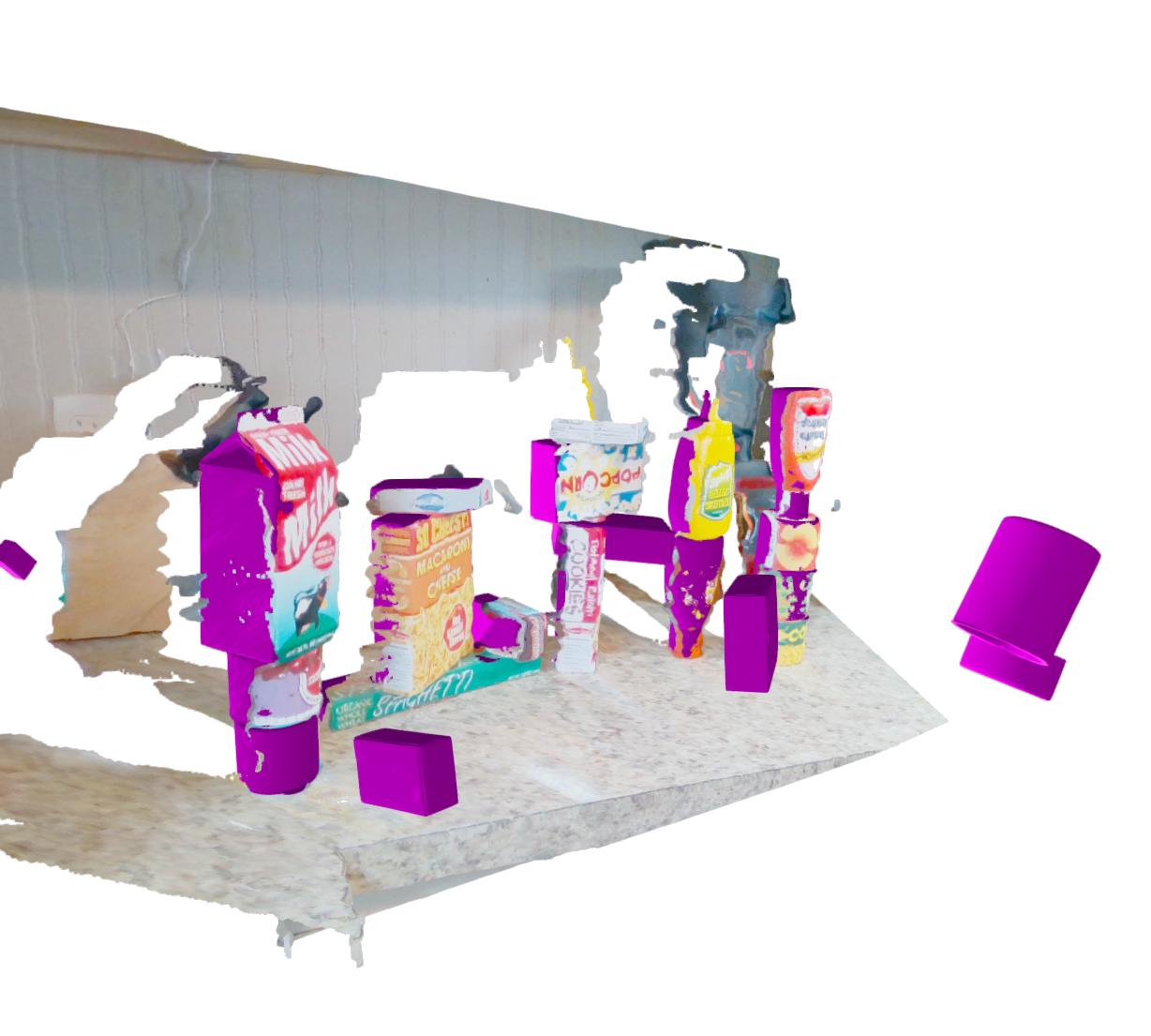}}\\

  \subfloat{\includegraphics[height=.2\textwidth]{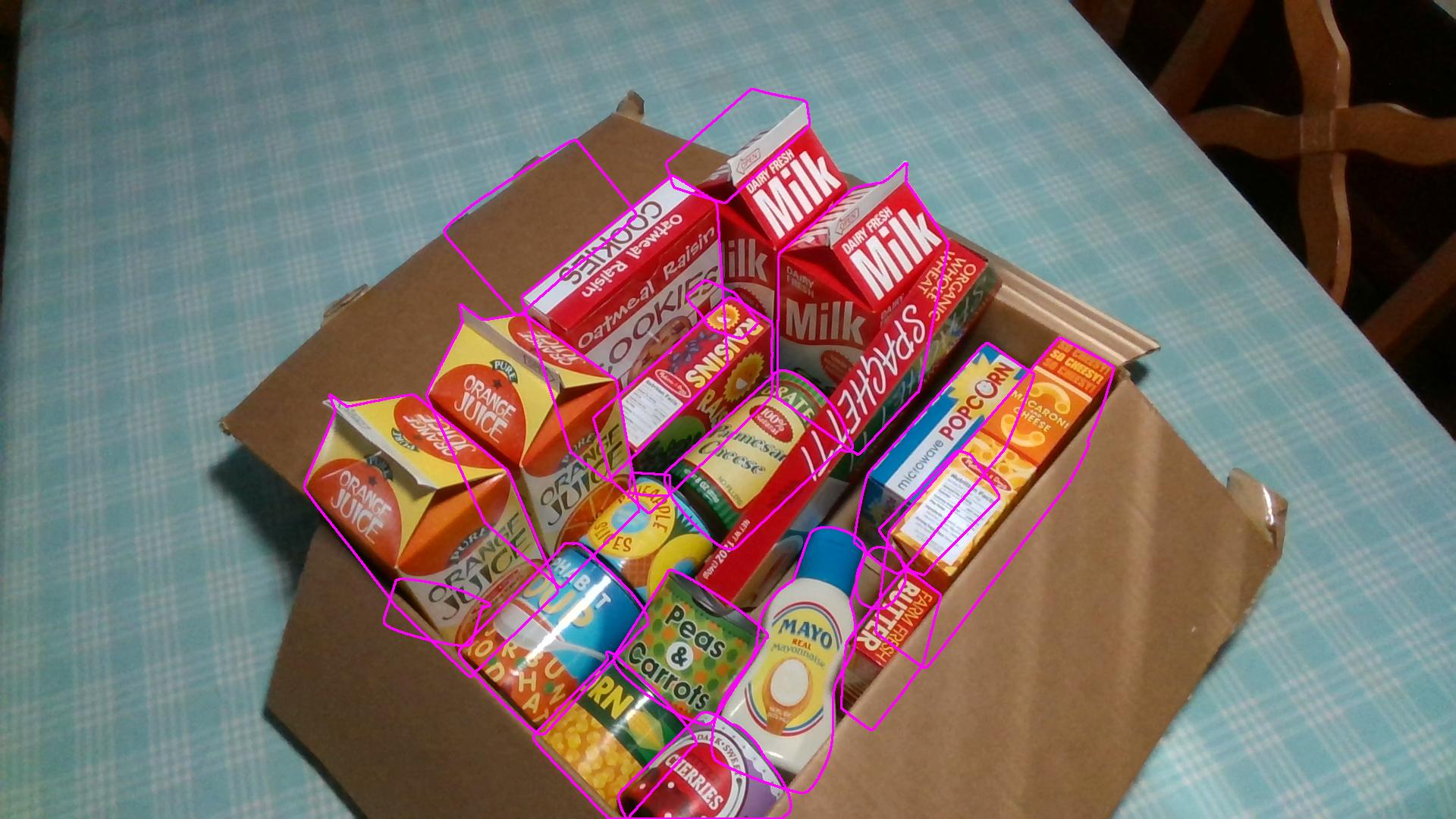}}\quad
  \subfloat{\includegraphics[height=.2\textwidth]{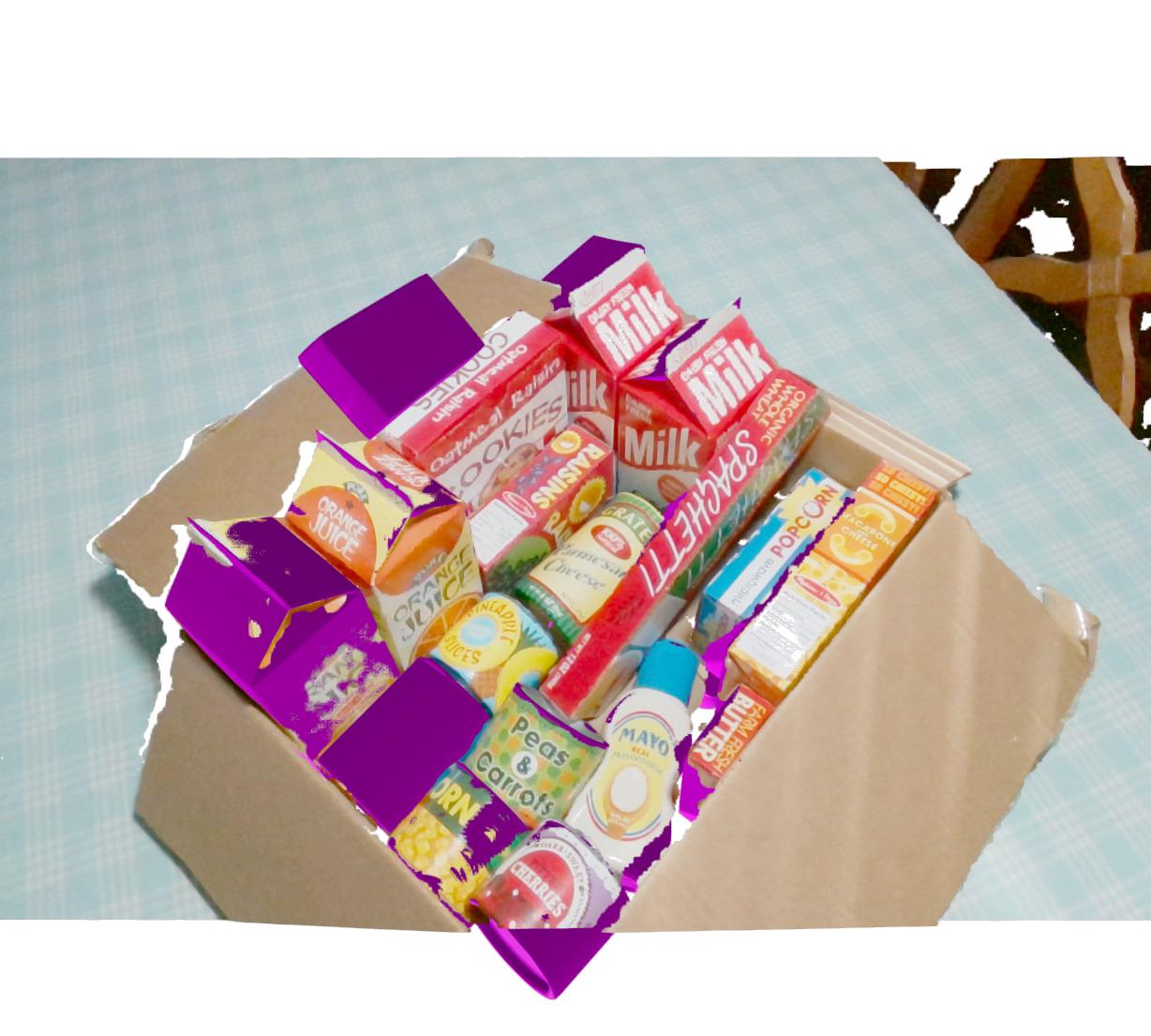}}\quad
  \subfloat{\includegraphics[height=.2\textwidth]{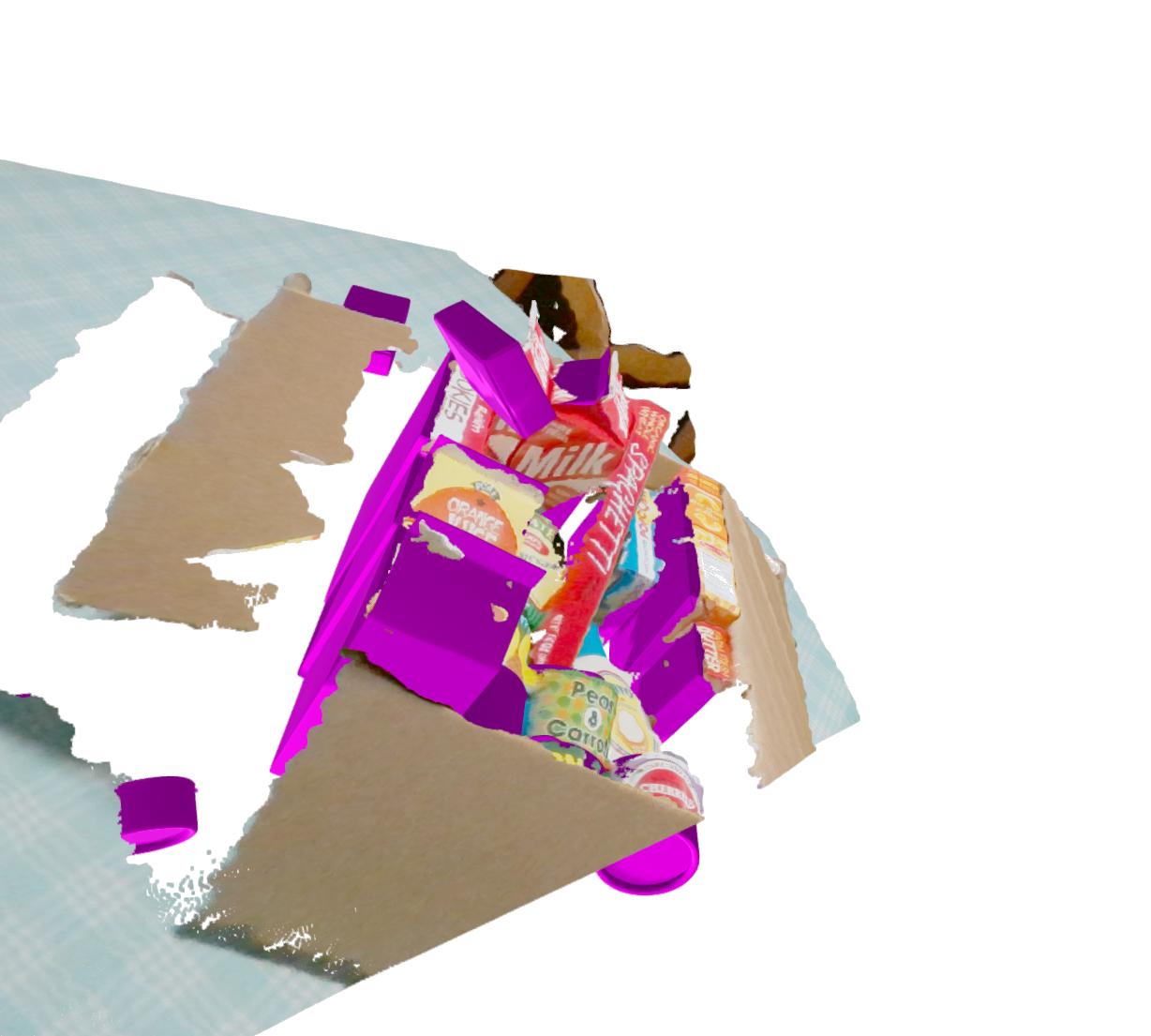}}\\

  \subfloat{\includegraphics[height=.2\textwidth]{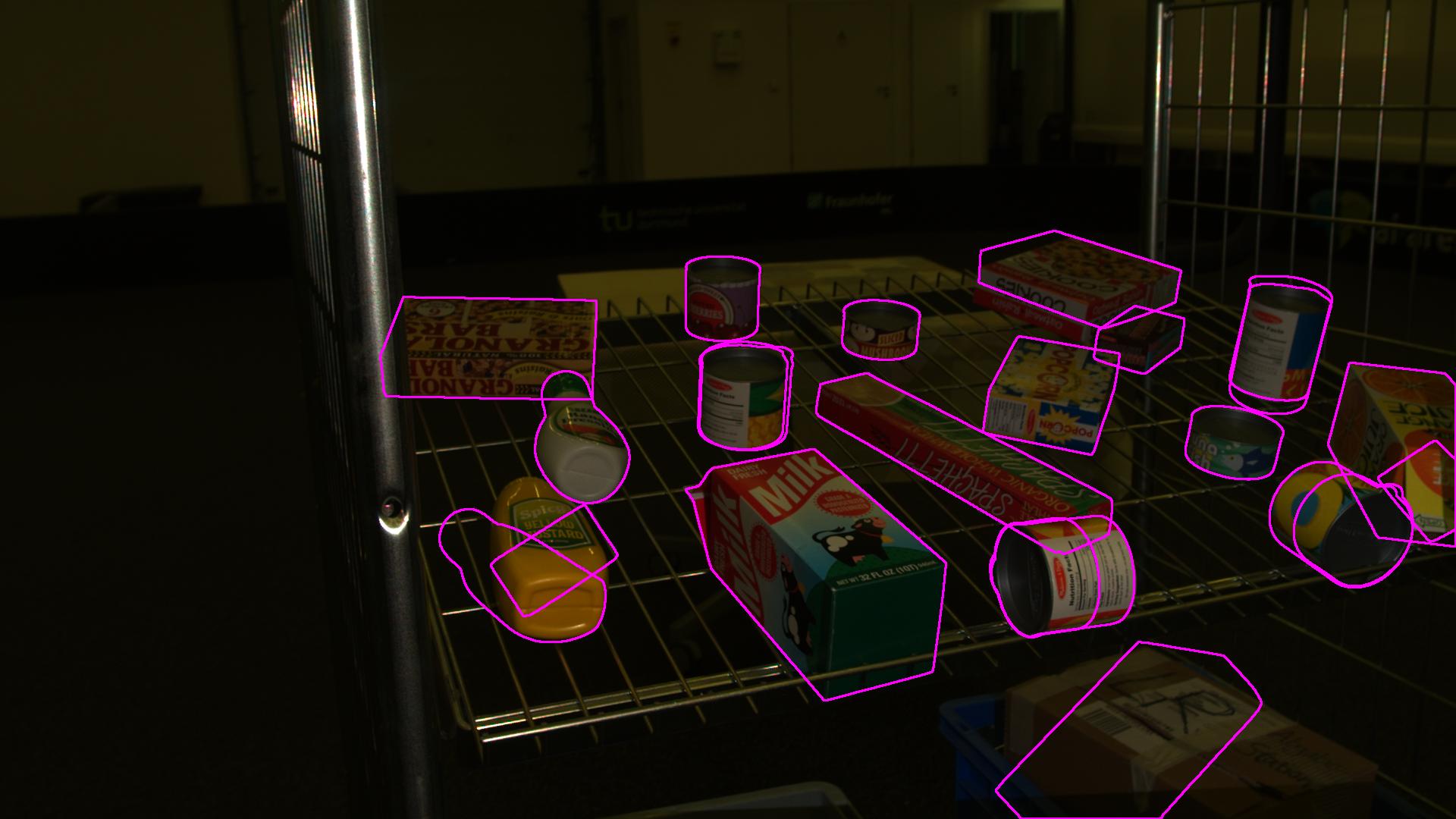}}\quad
  \subfloat{\includegraphics[height=.2\textwidth]{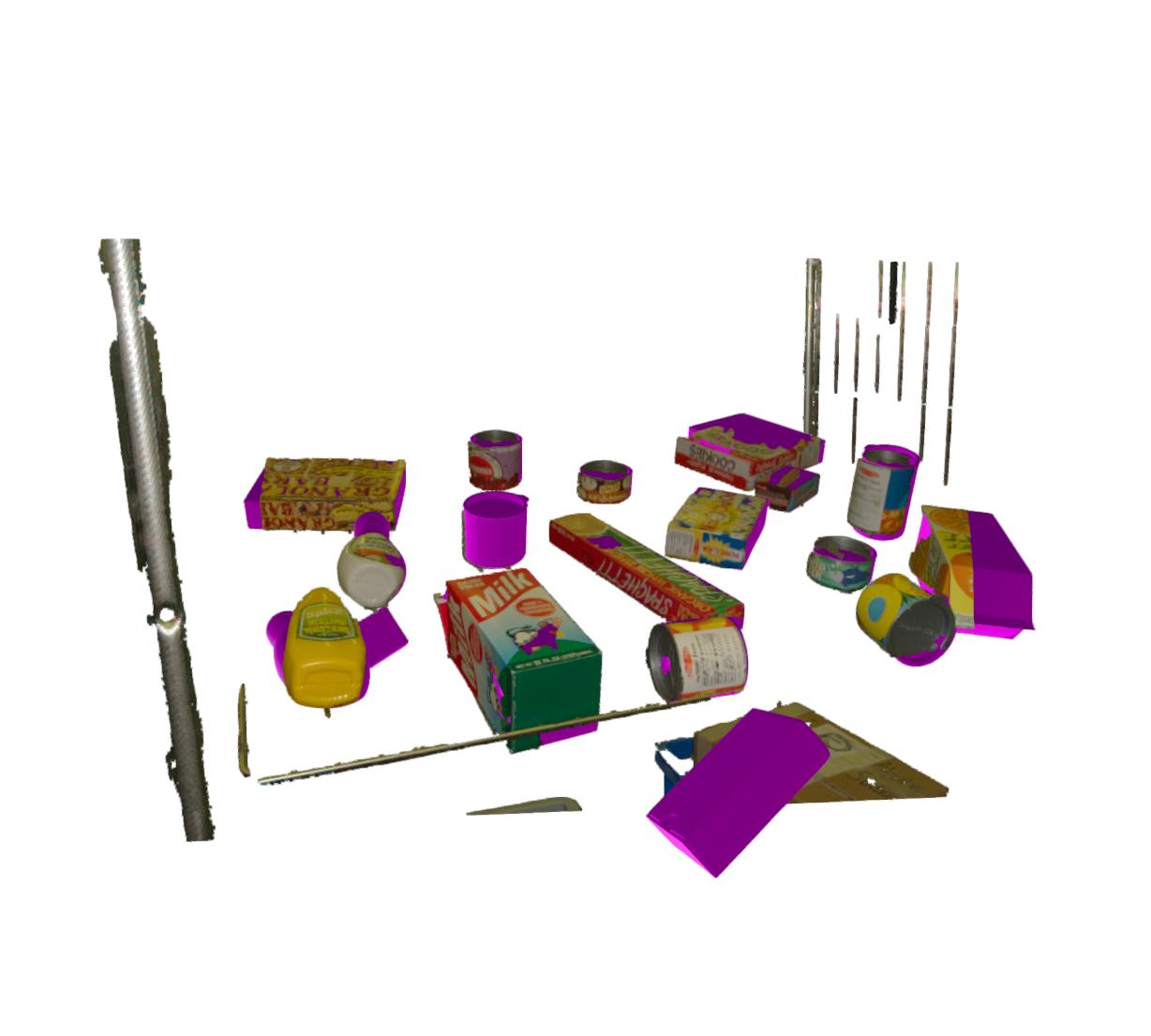}}\quad
  \subfloat{\includegraphics[height=.2\textwidth]{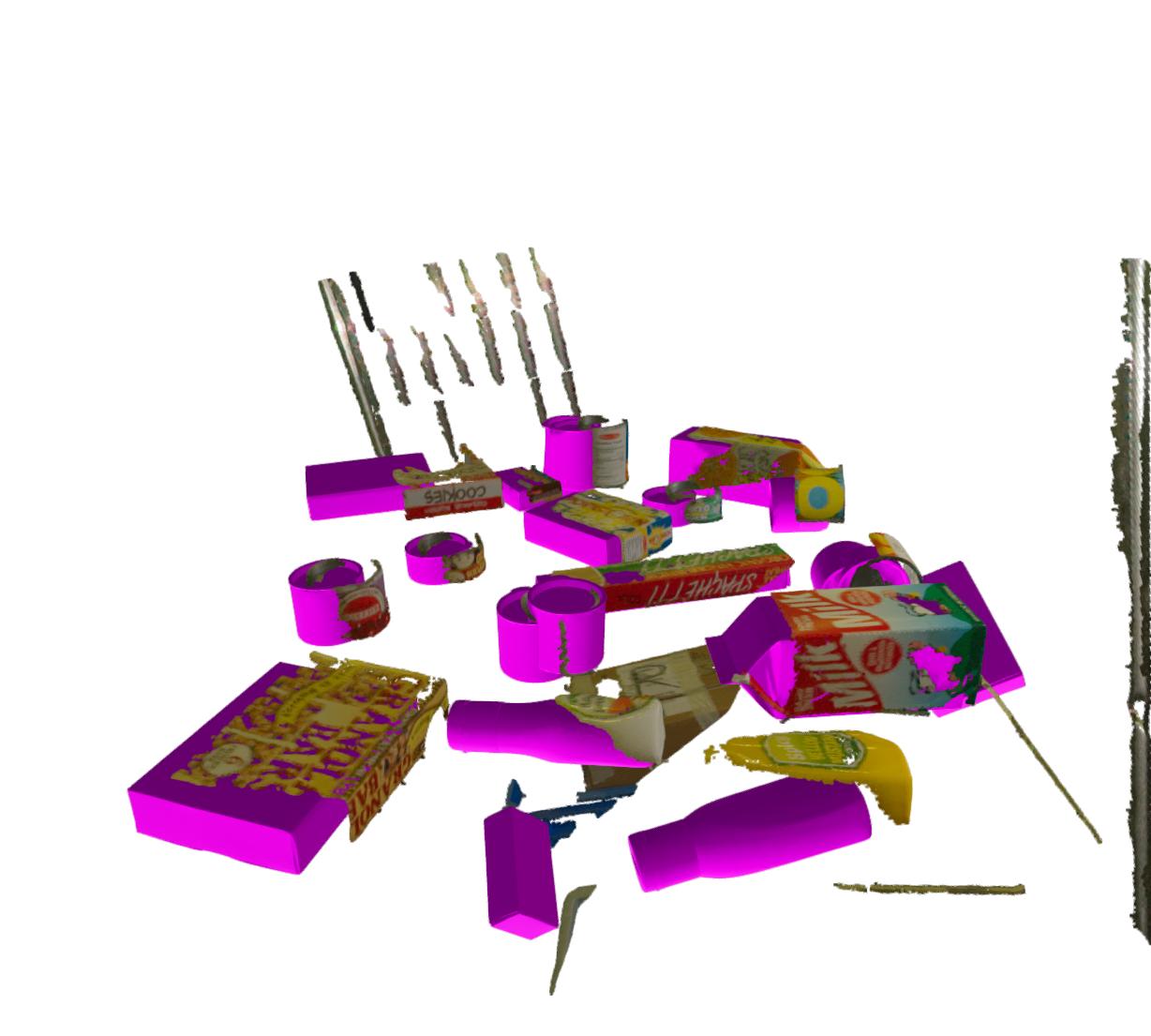}}\\

  \subfloat{\includegraphics[height=.2\textwidth]{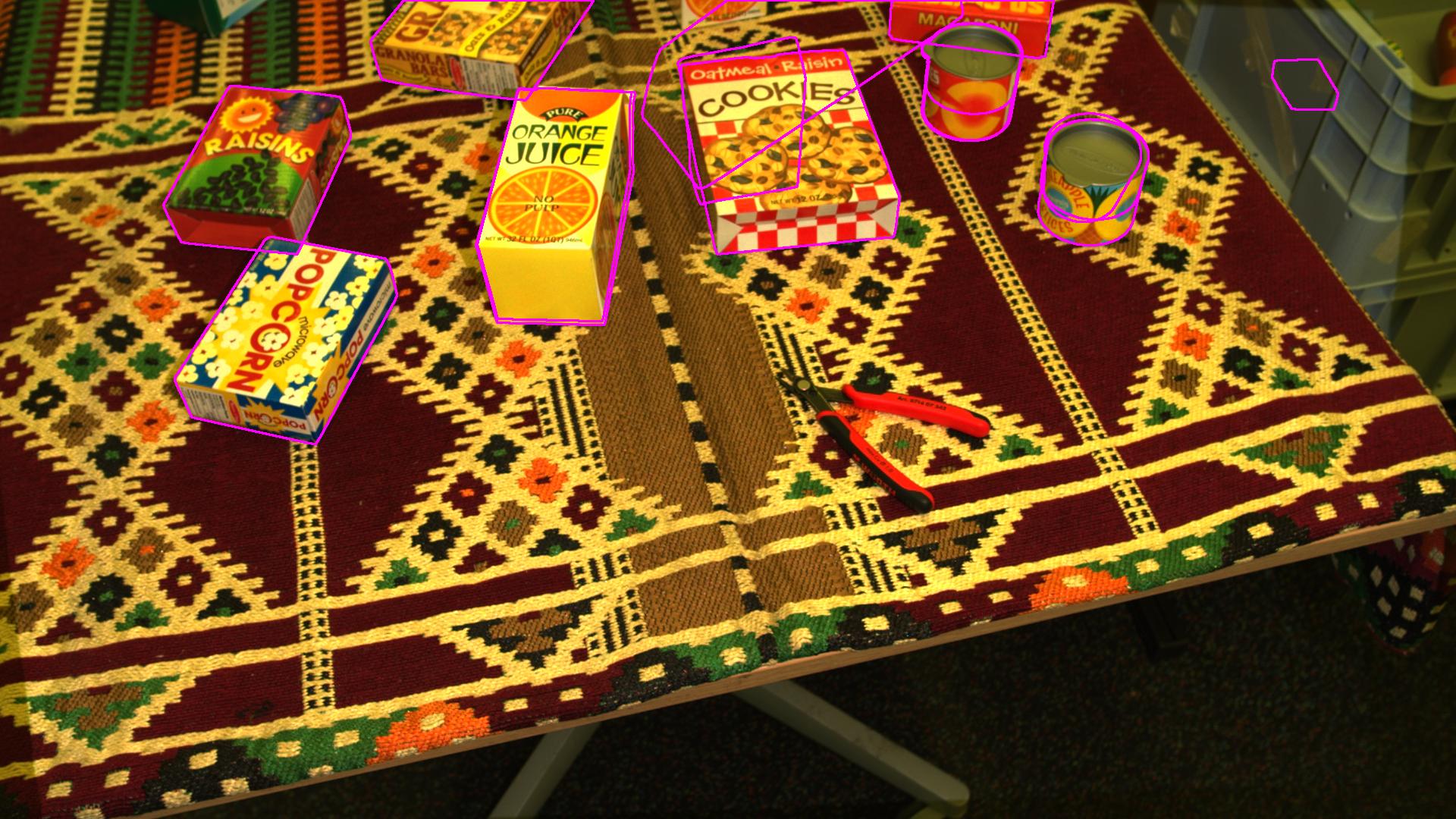}}\quad
  \subfloat{\includegraphics[height=.2\textwidth]{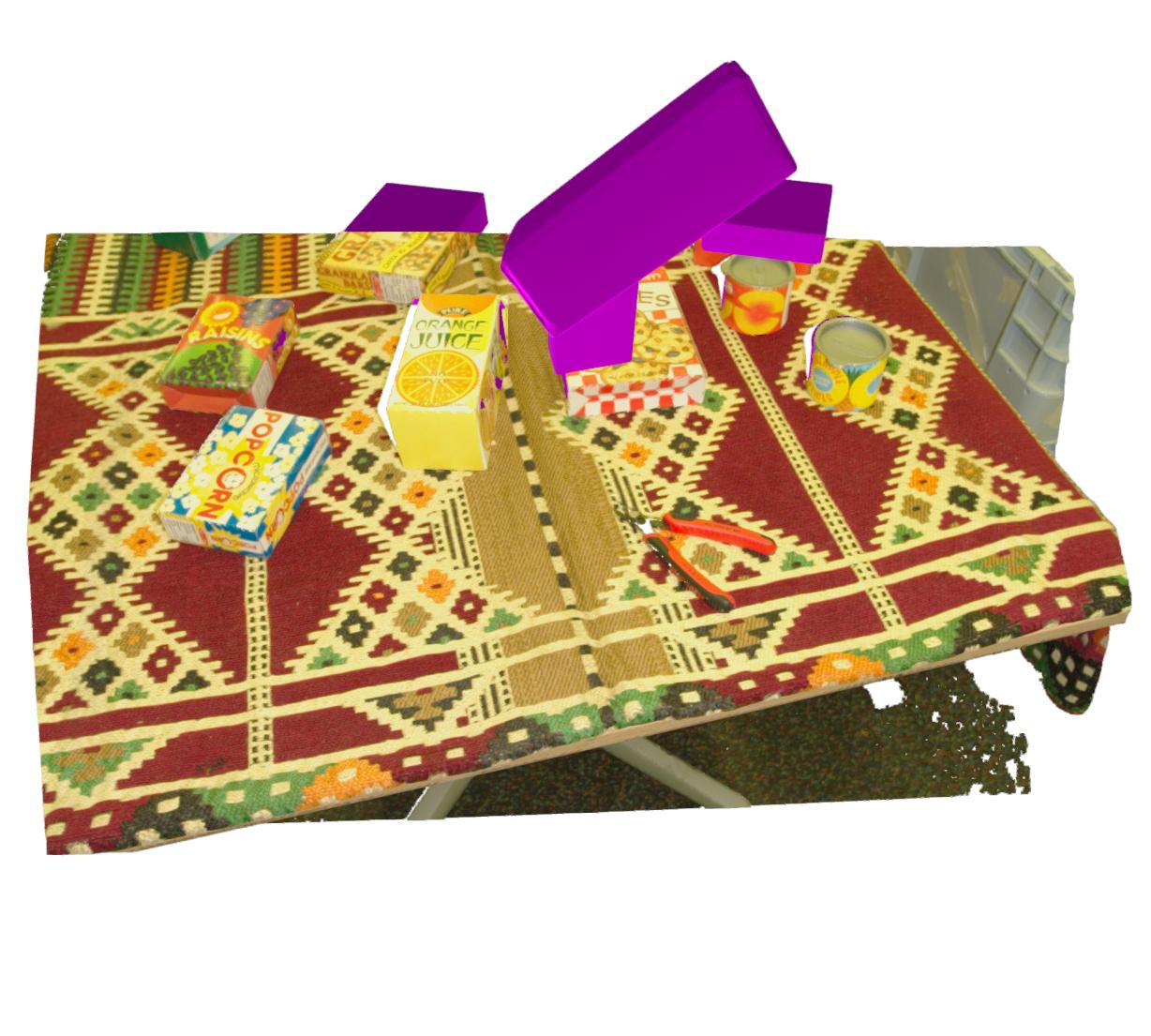}}\quad
  \subfloat{\includegraphics[height=.2\textwidth]{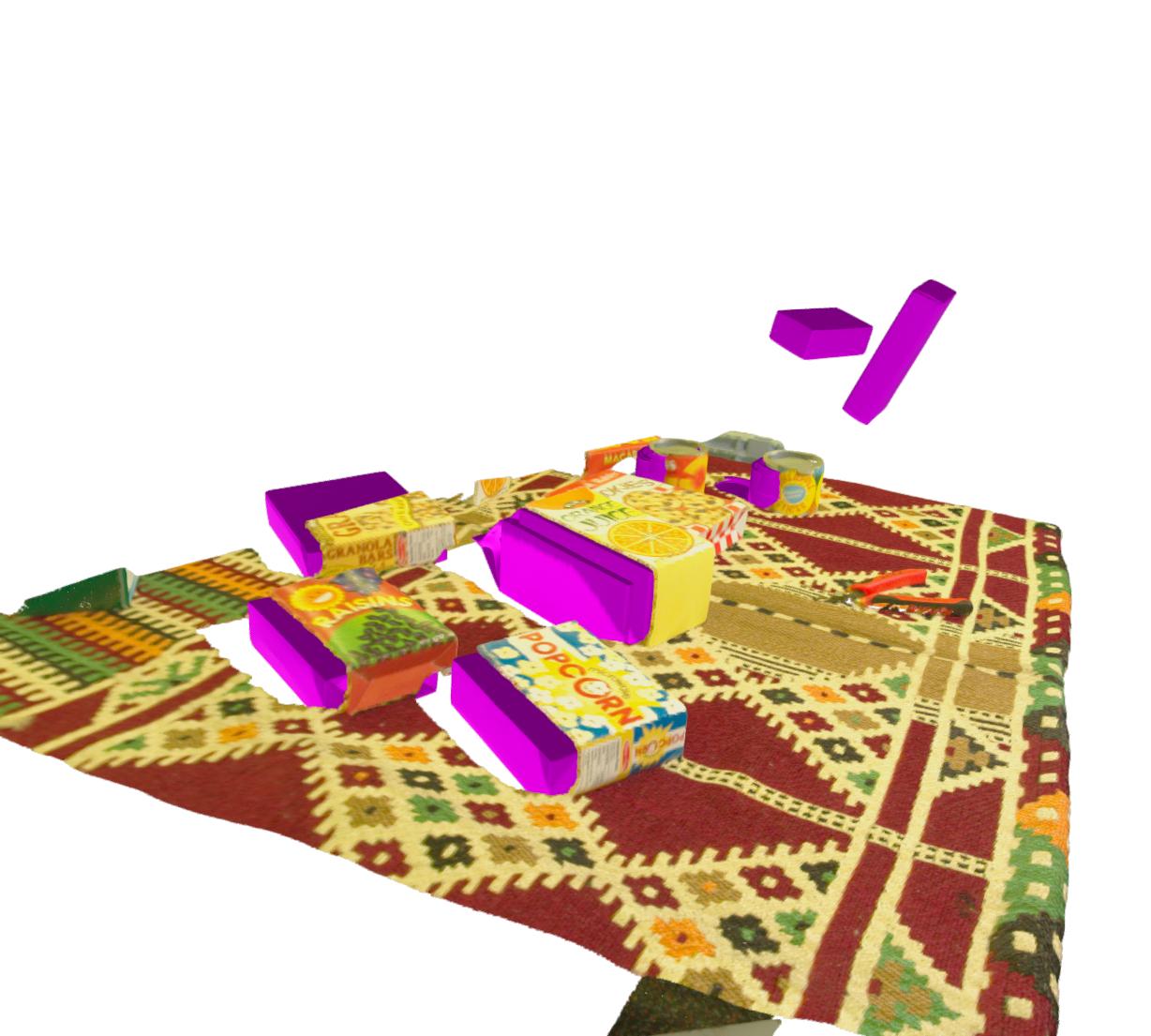}}
  
  \caption{HOPEv2 dataset visualization with \textcolor{magenta}{estimated} (magenta) 6D pose of the meshes and point cloud. The first column shows the test image with a contour of the projection made by the predicted pose. The other two columns show the corresponding 3D view from different viewing angles. The first is captured from approximately the same viewing angle as the image was taken.}
  \label{fig:A0_HOPEv2}
\end{figure*}

\begin{figure*}
  \centering
  \subfloat{\includegraphics[height=.2\textwidth]{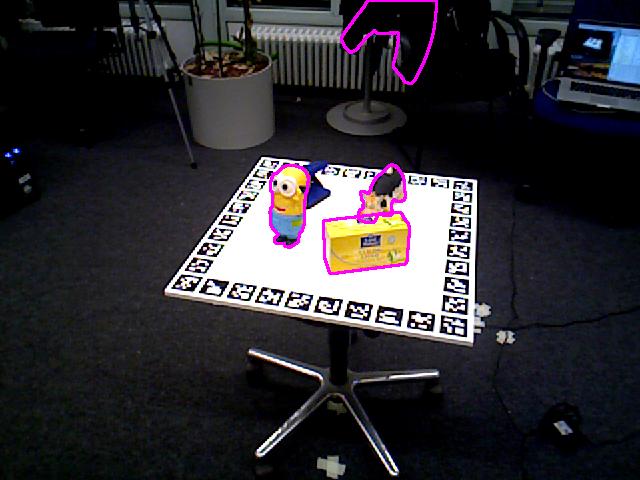}}\quad
  \subfloat{\includegraphics[height=.2\textwidth]{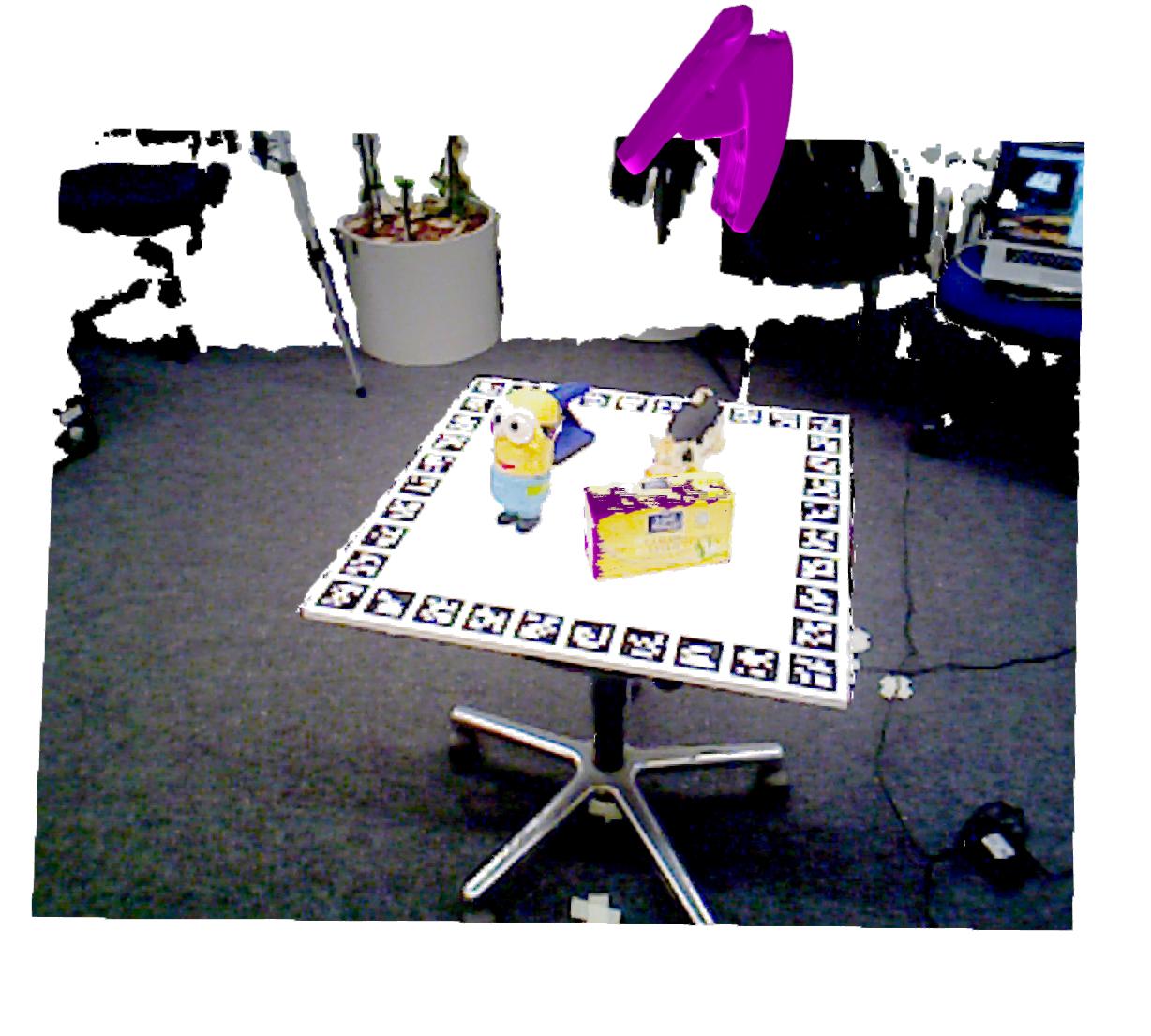}}\quad
  \subfloat{\includegraphics[height=.2\textwidth]{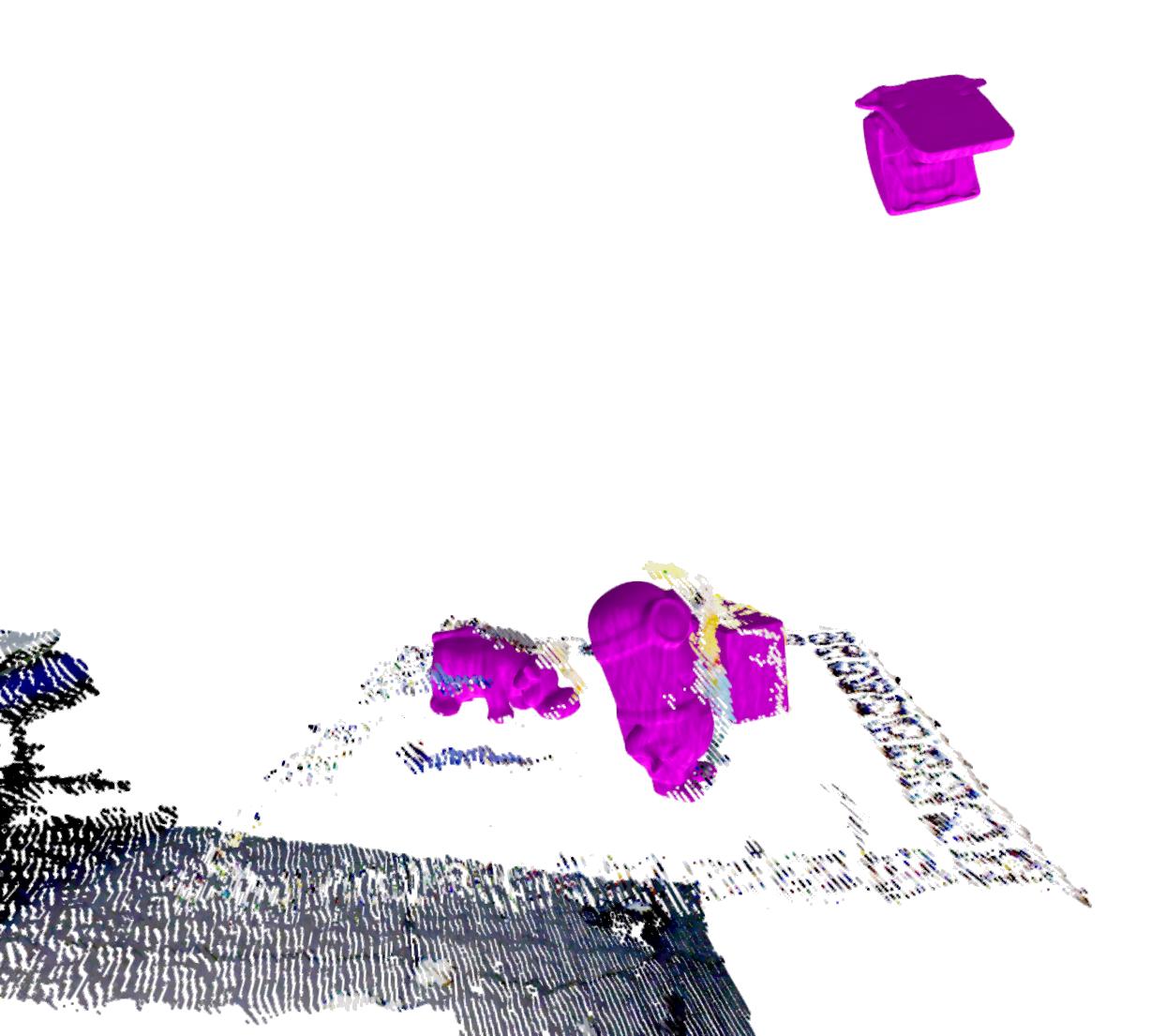}}\quad
  \subfloat{\includegraphics[height=.2\textwidth]{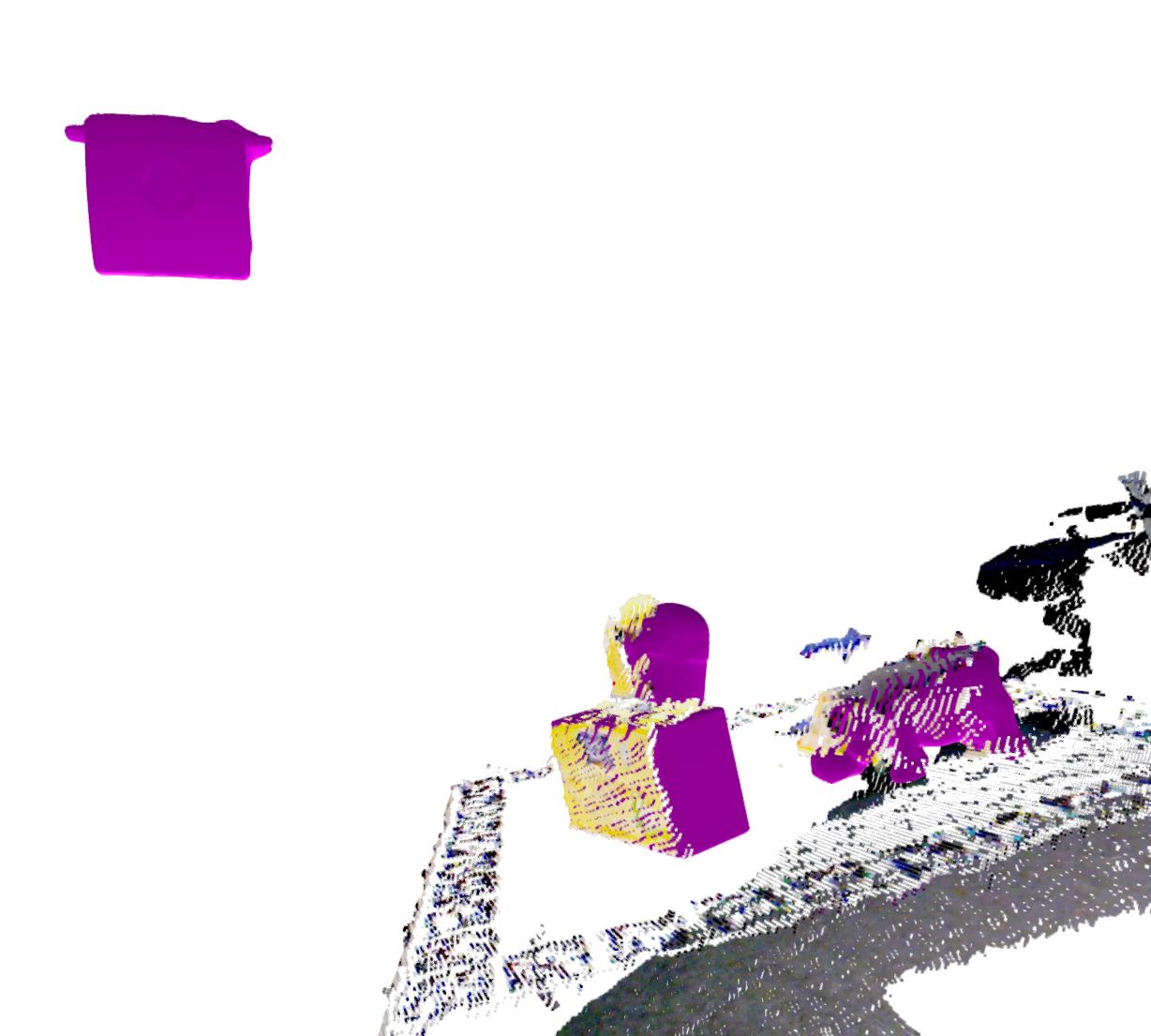}}\\
  
  \subfloat{\includegraphics[height=.2\textwidth]{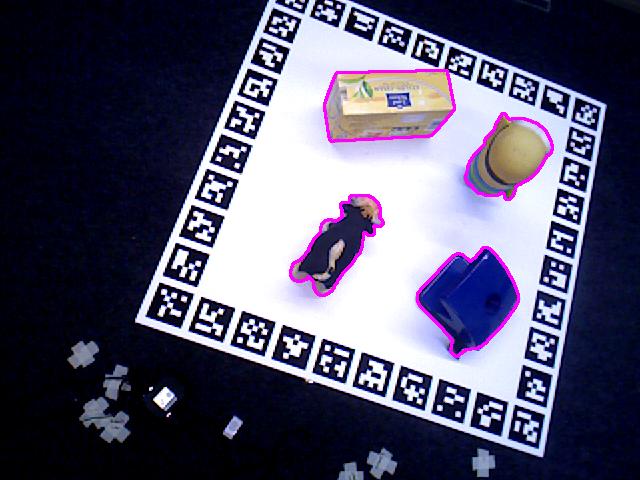}}\quad
  \subfloat{\includegraphics[height=.2\textwidth]{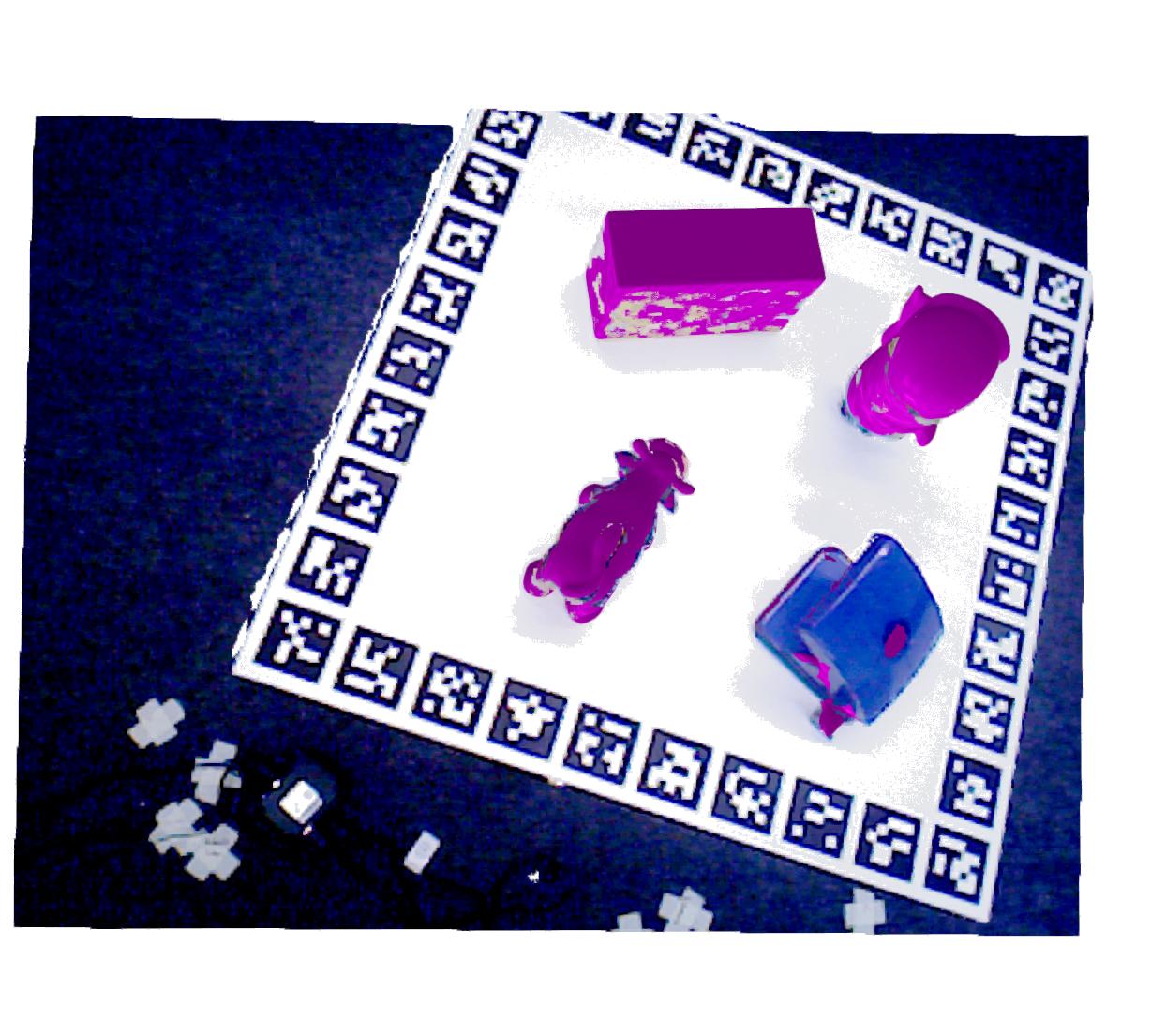}}\quad
  \subfloat{\includegraphics[height=.2\textwidth]{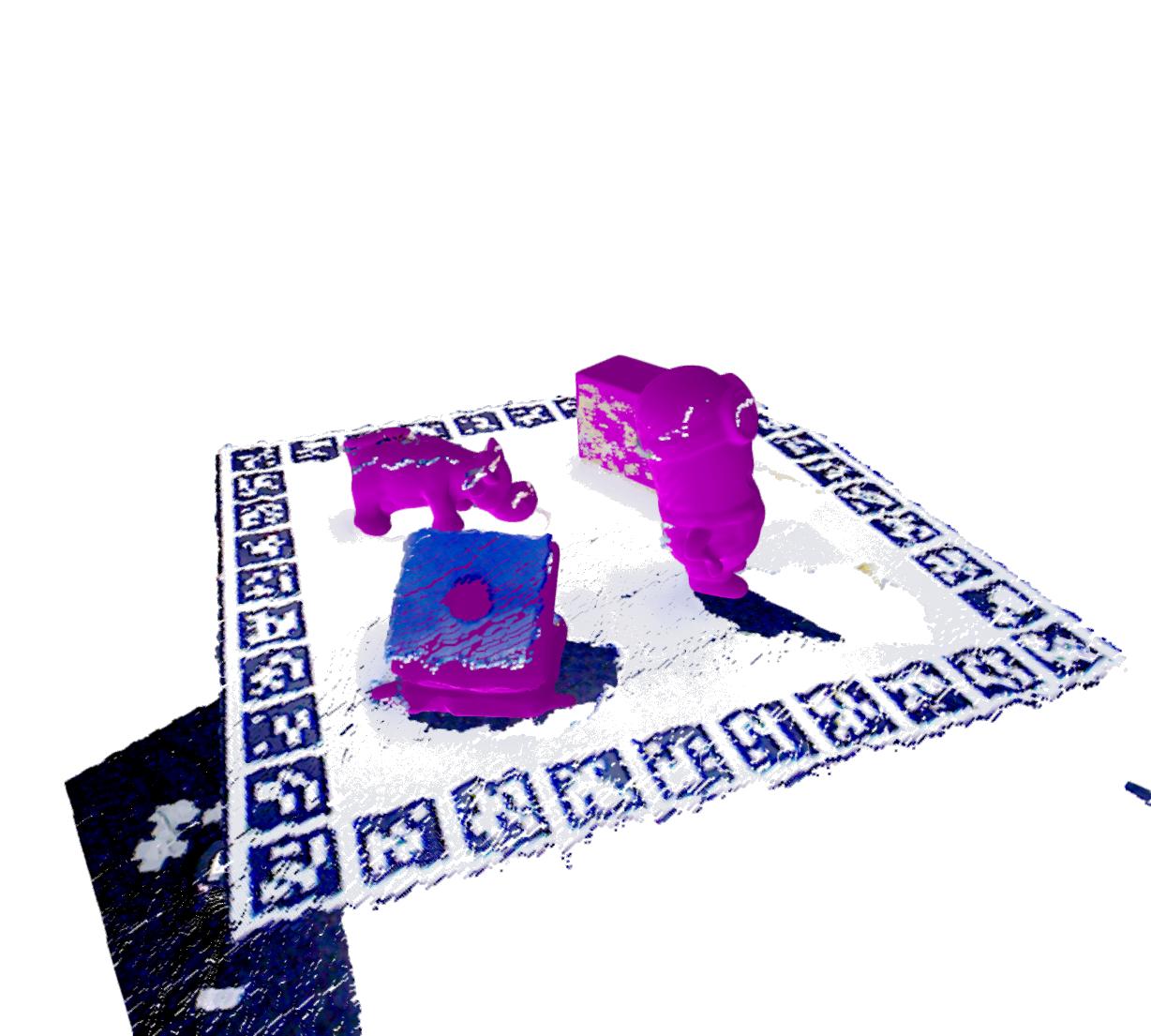}}\quad
  \subfloat{\includegraphics[height=.2\textwidth]{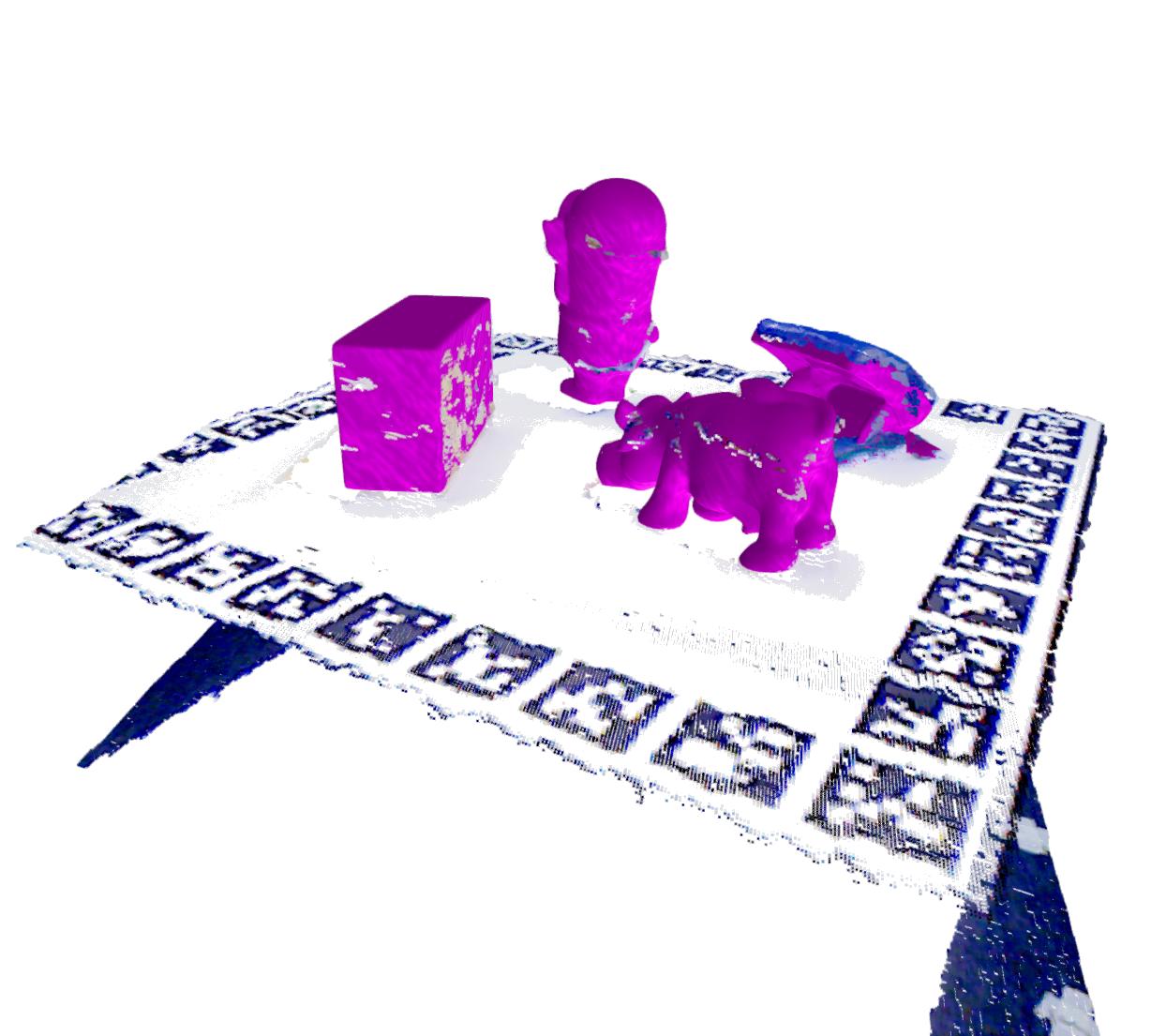}}\\
  
  \subfloat{\includegraphics[height=.2\textwidth]{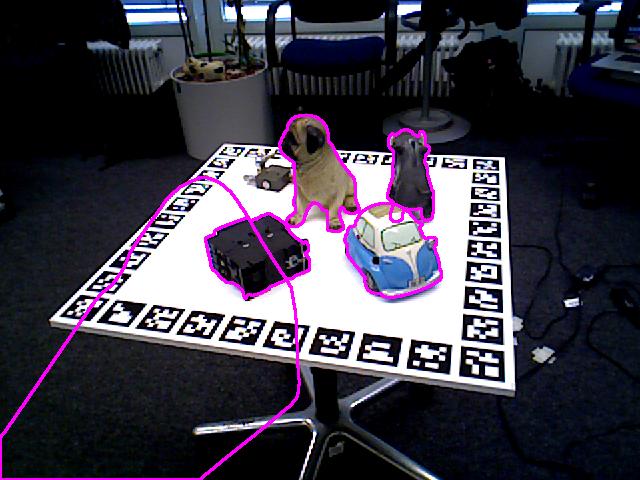}}\quad
  \subfloat{\includegraphics[height=.2\textwidth]{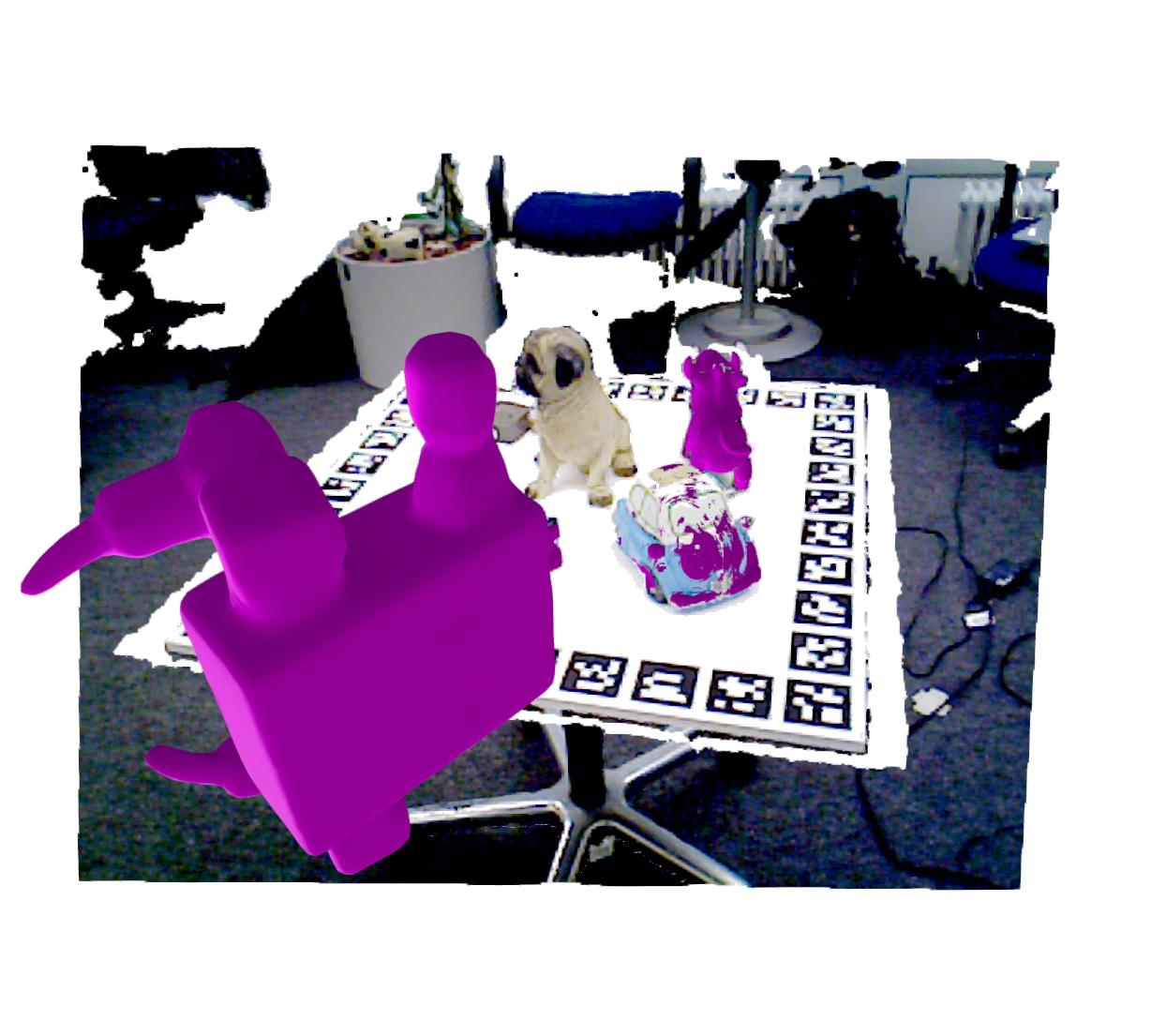}}\quad
  \subfloat{\includegraphics[height=.2\textwidth]{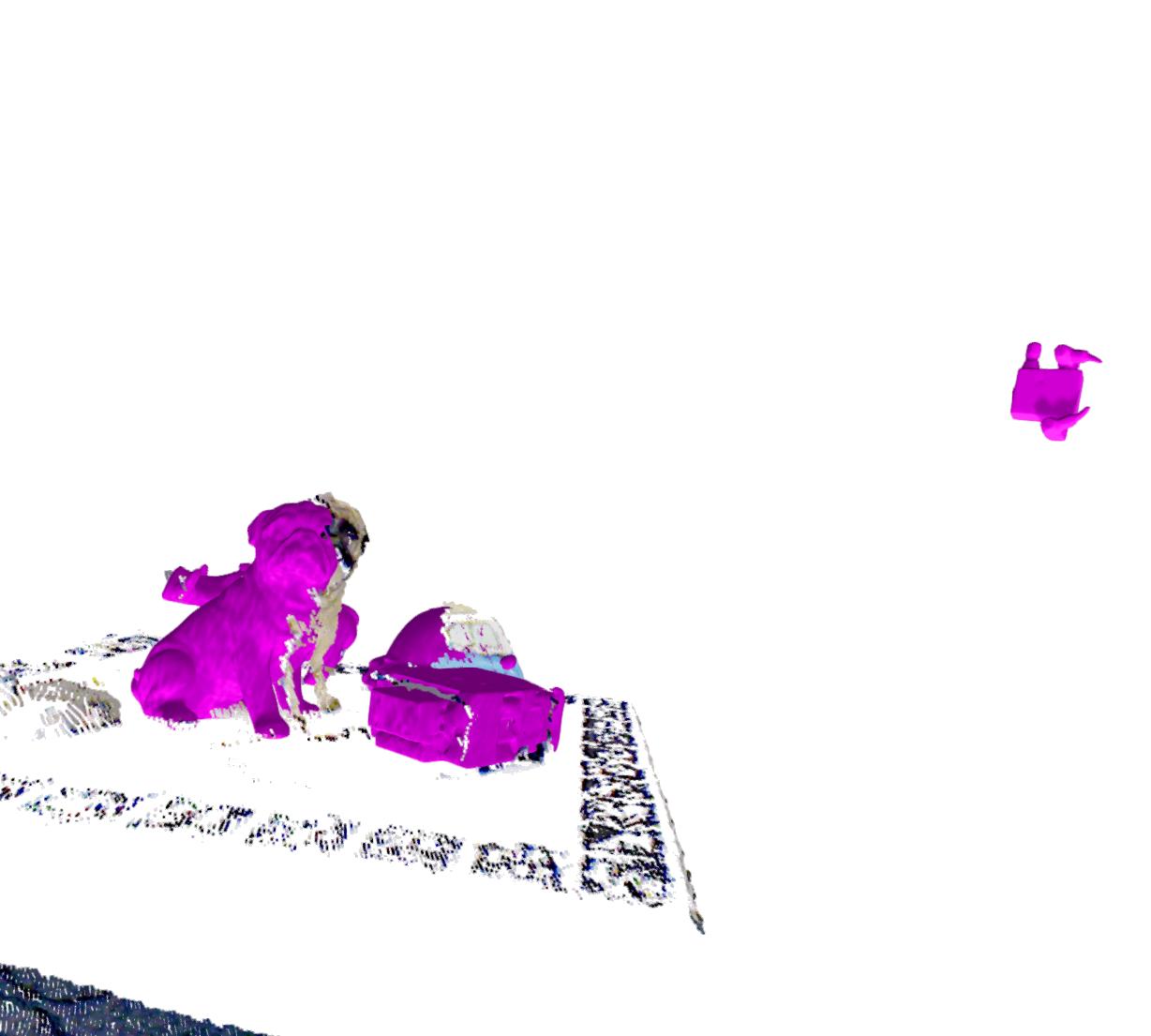}}\quad
  \subfloat{\includegraphics[height=.2\textwidth]{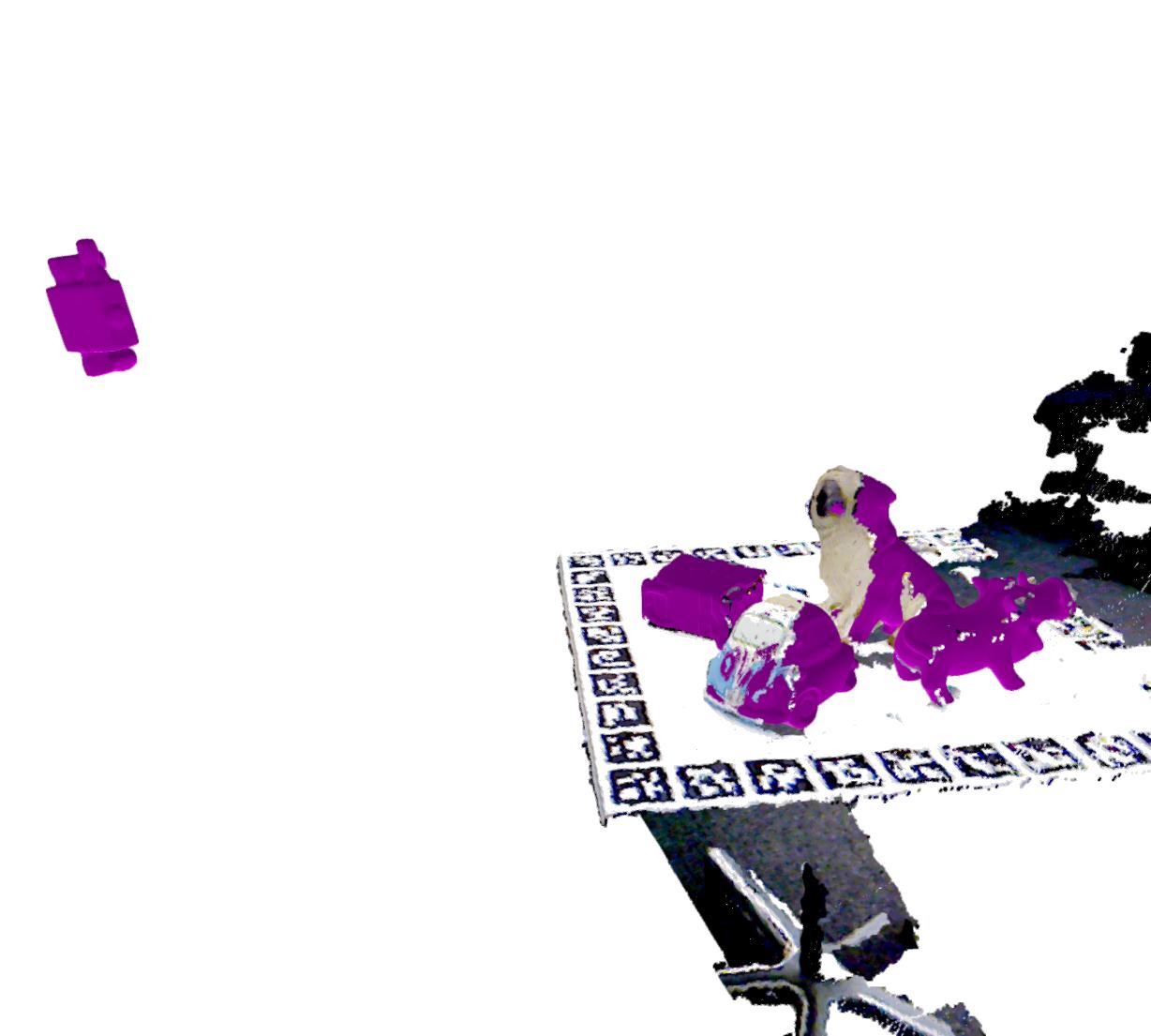}}\\
  
  \subfloat{\includegraphics[height=.2\textwidth]{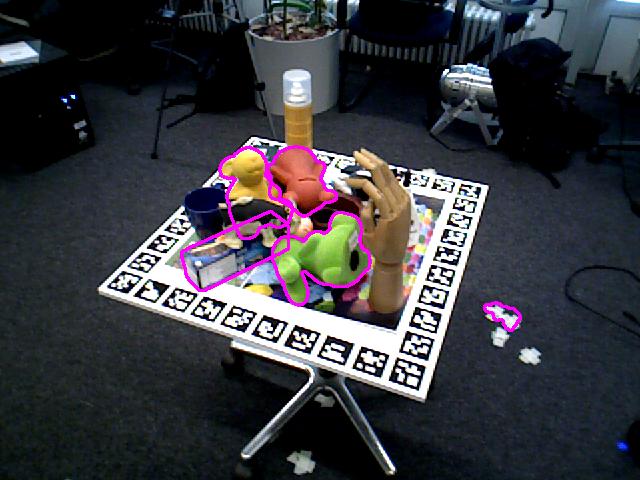}}\quad
  \subfloat{\includegraphics[height=.2\textwidth]{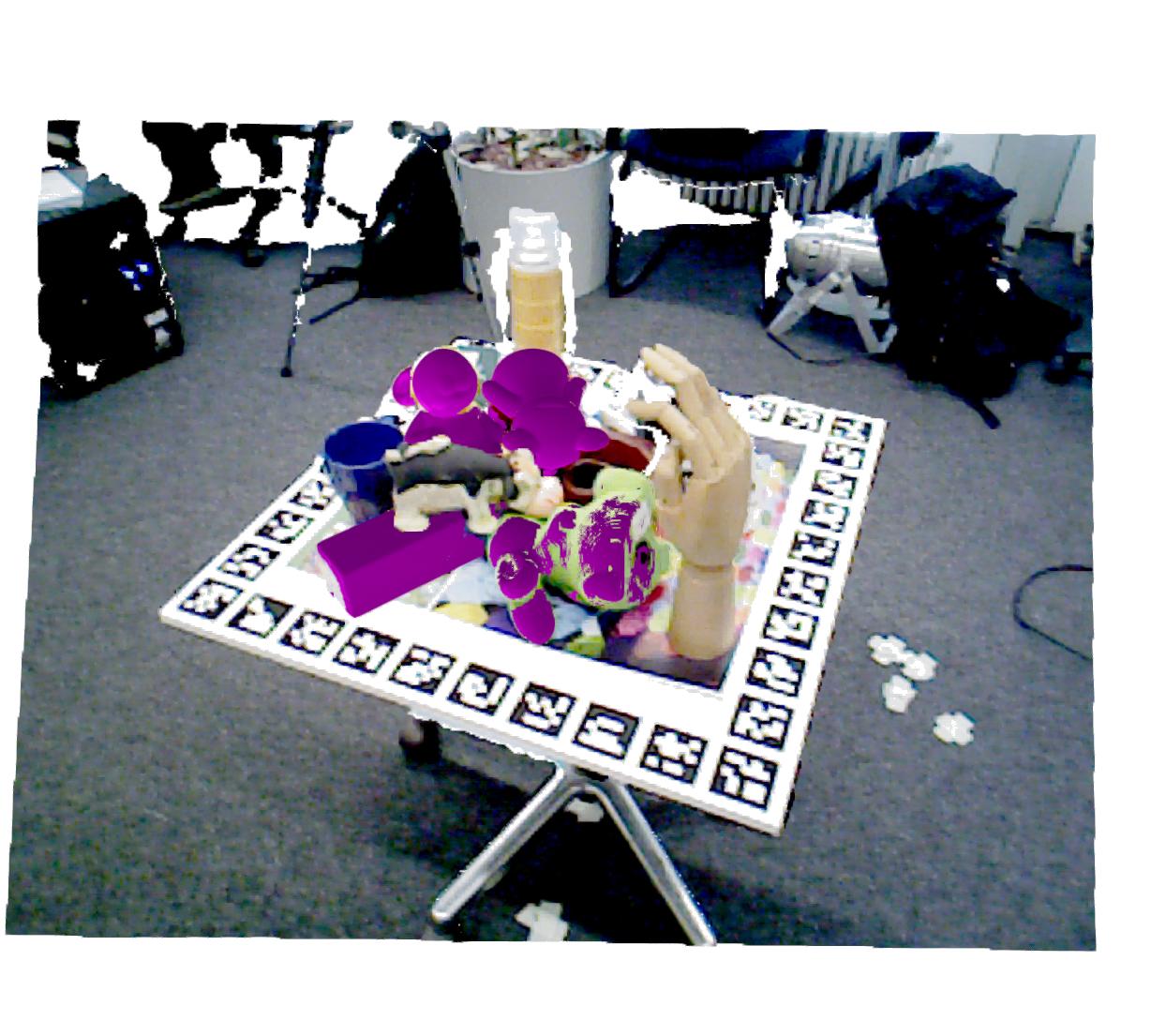}}\quad
  \subfloat{\includegraphics[height=.2\textwidth]{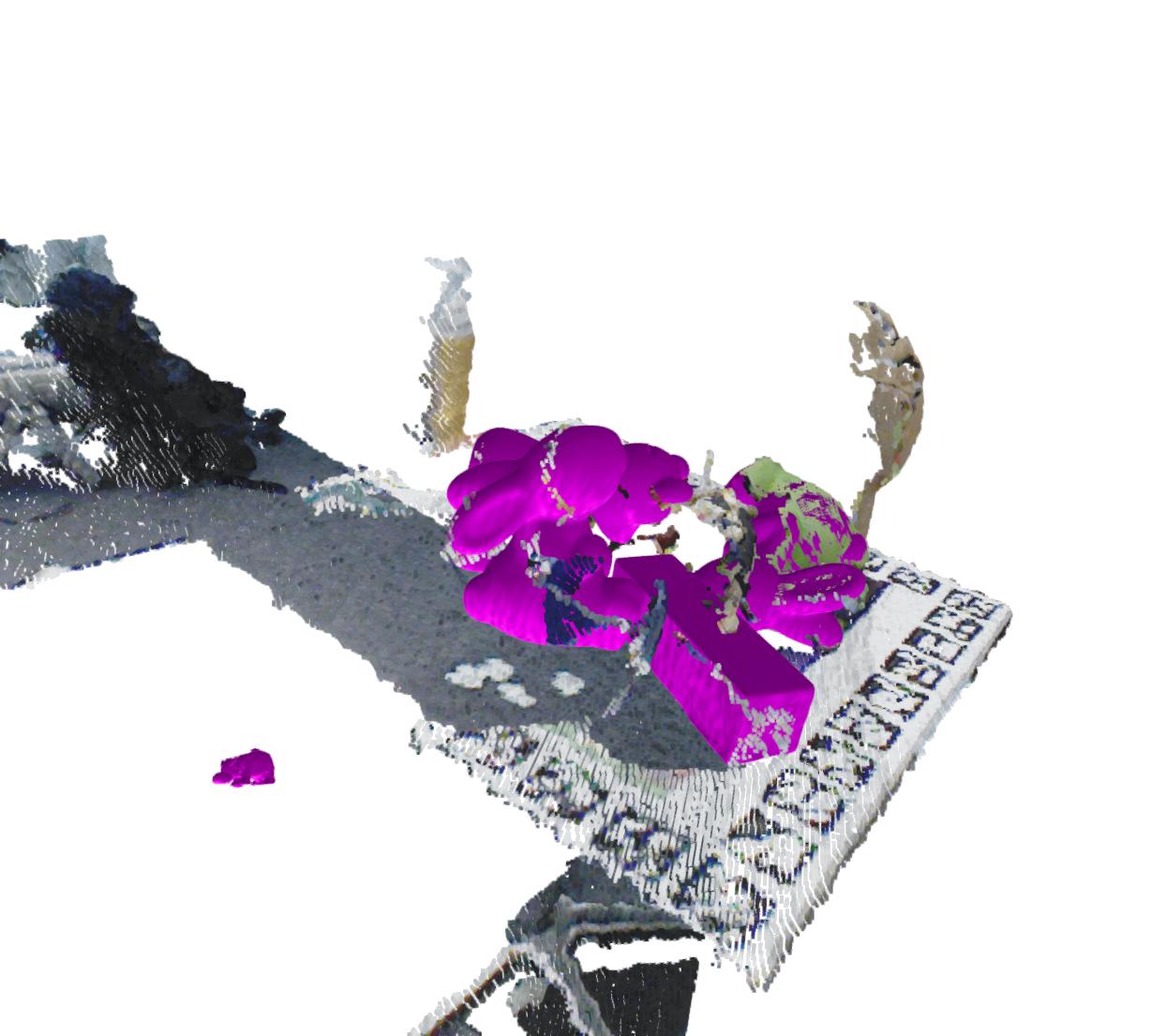}}\quad
  \subfloat{\includegraphics[height=.2\textwidth]{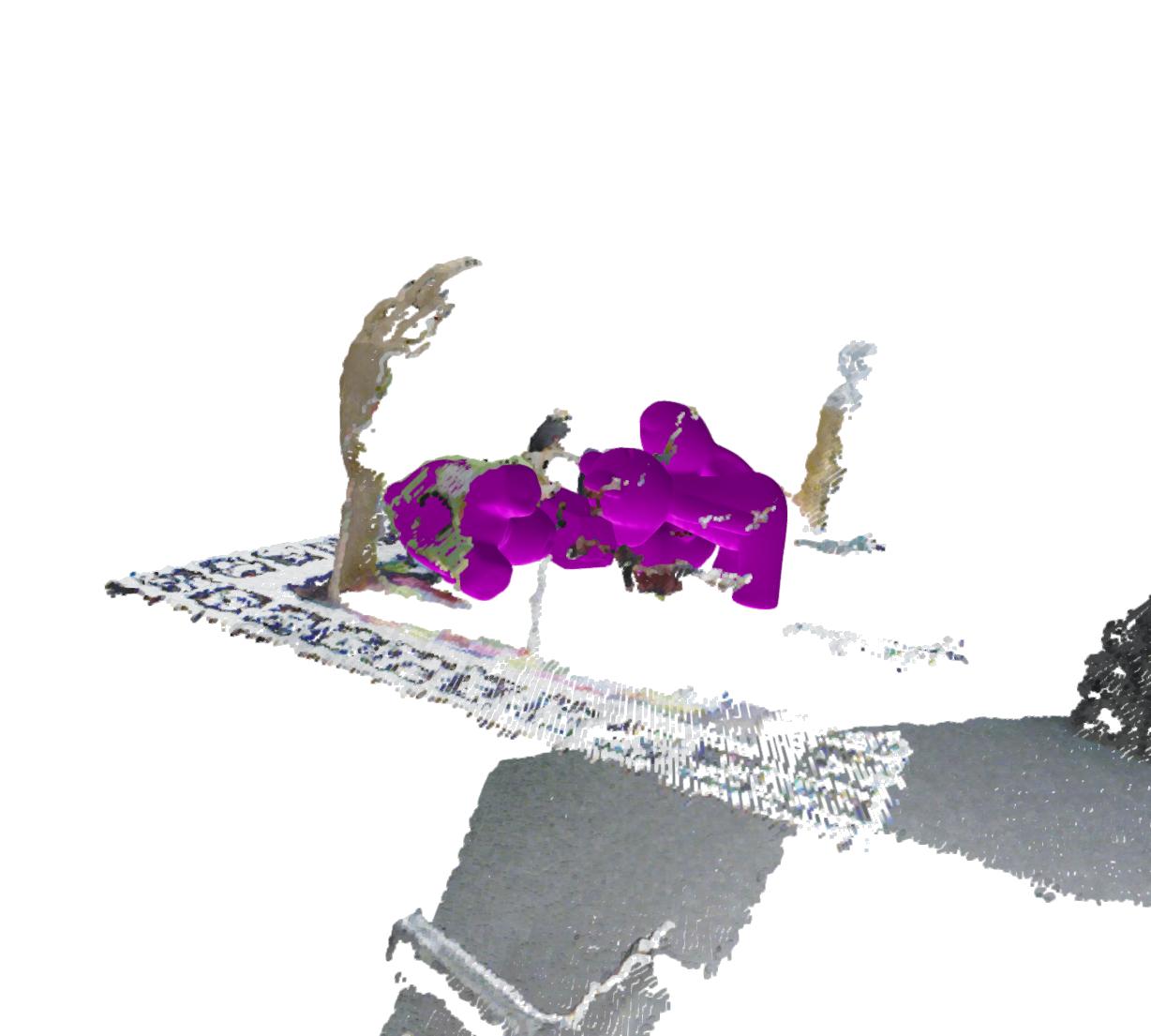}}\\
  
  \subfloat{\includegraphics[height=.2\textwidth]{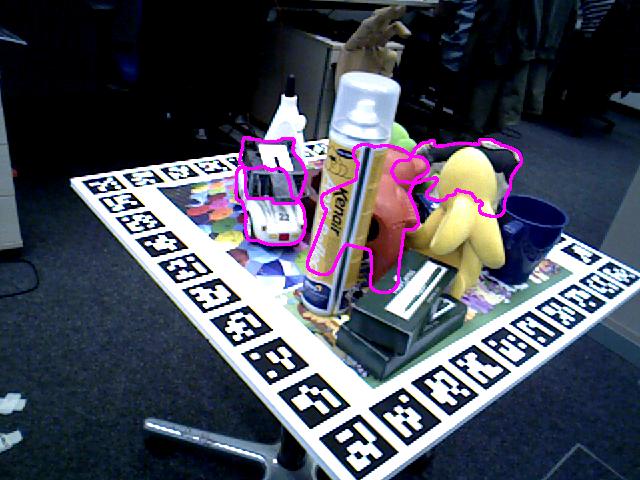}}\quad
  \subfloat{\includegraphics[height=.2\textwidth]{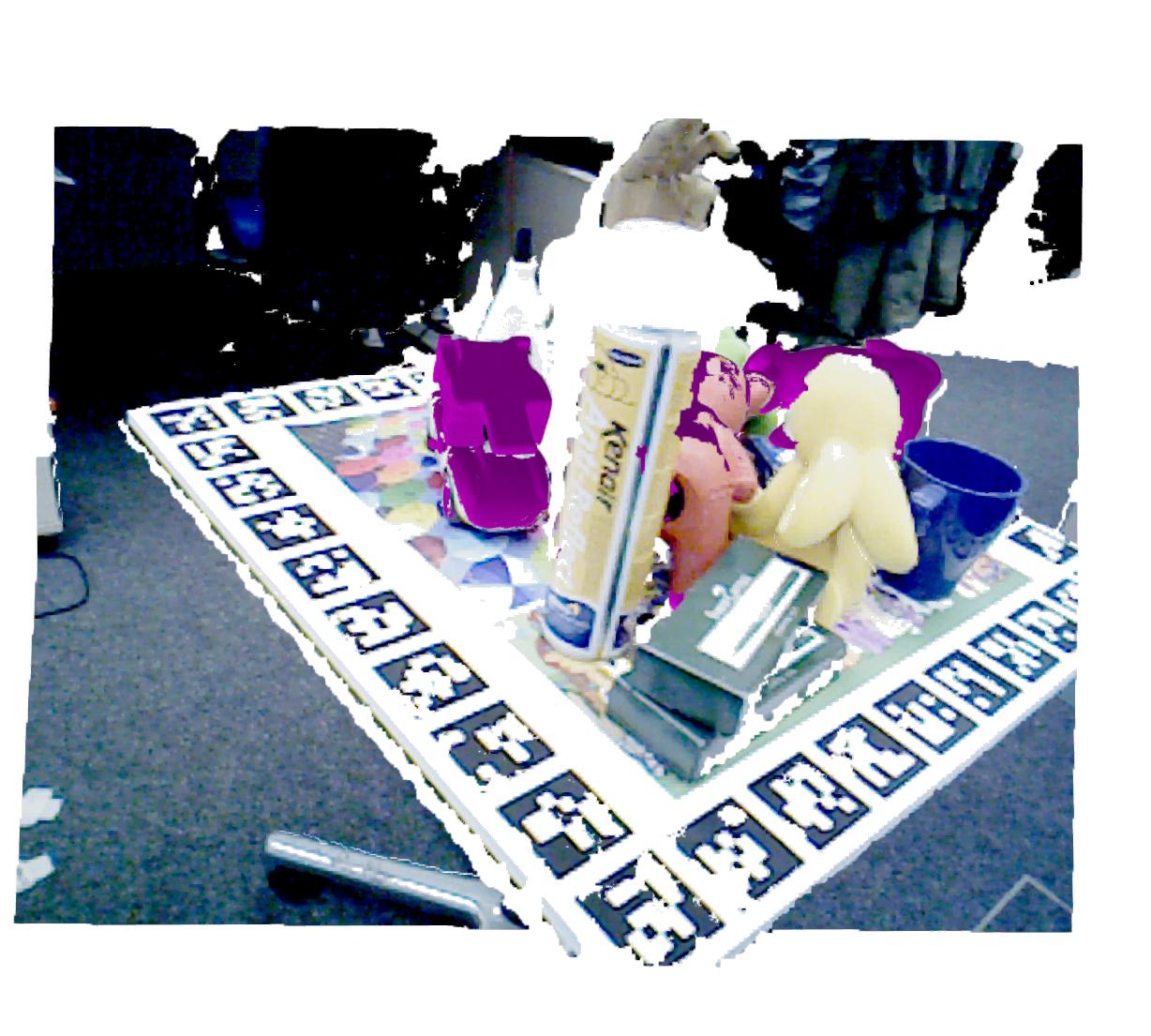}}\quad
  \subfloat{\includegraphics[height=.2\textwidth]{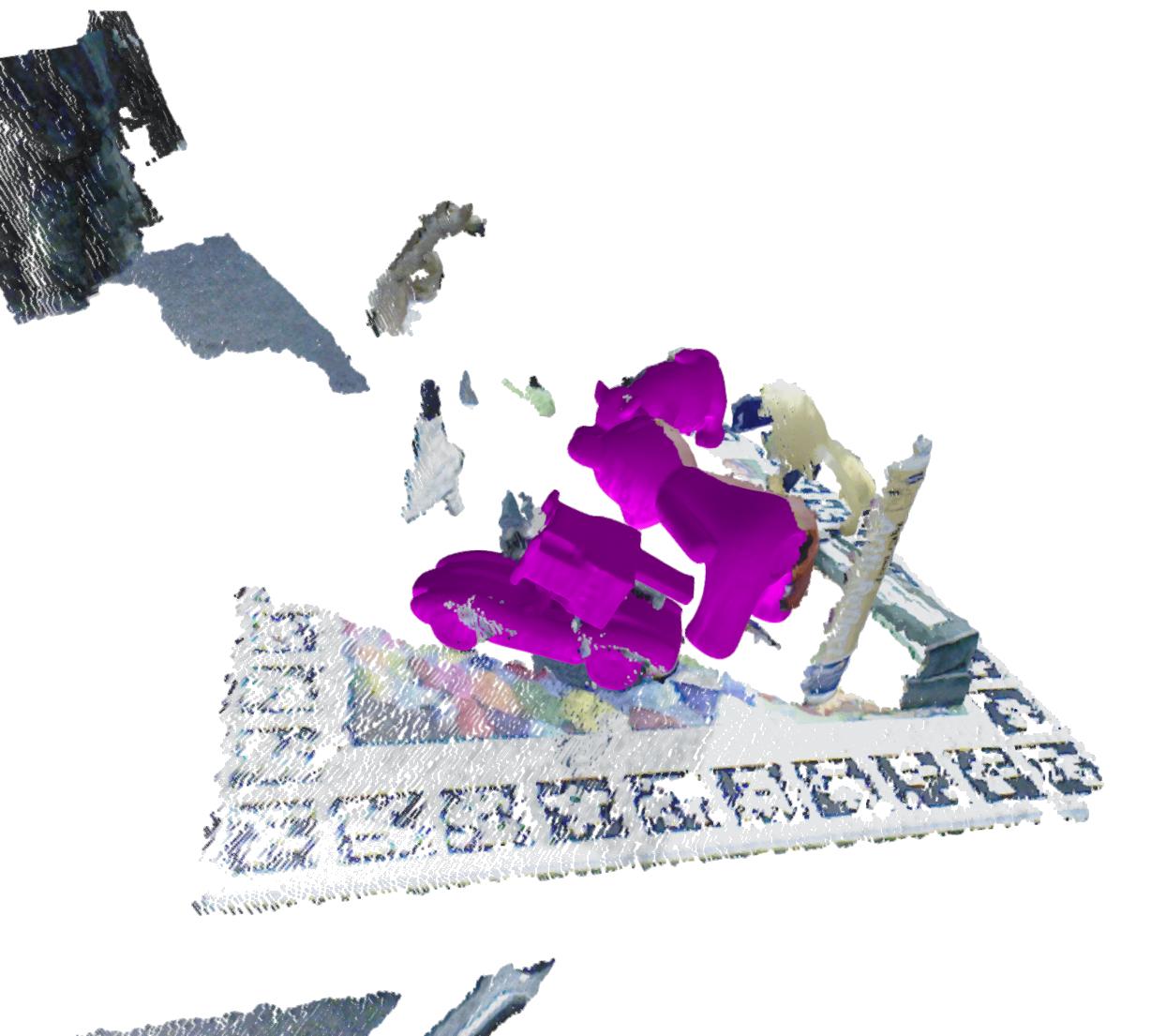}}\quad
  \subfloat{\includegraphics[height=.2\textwidth]{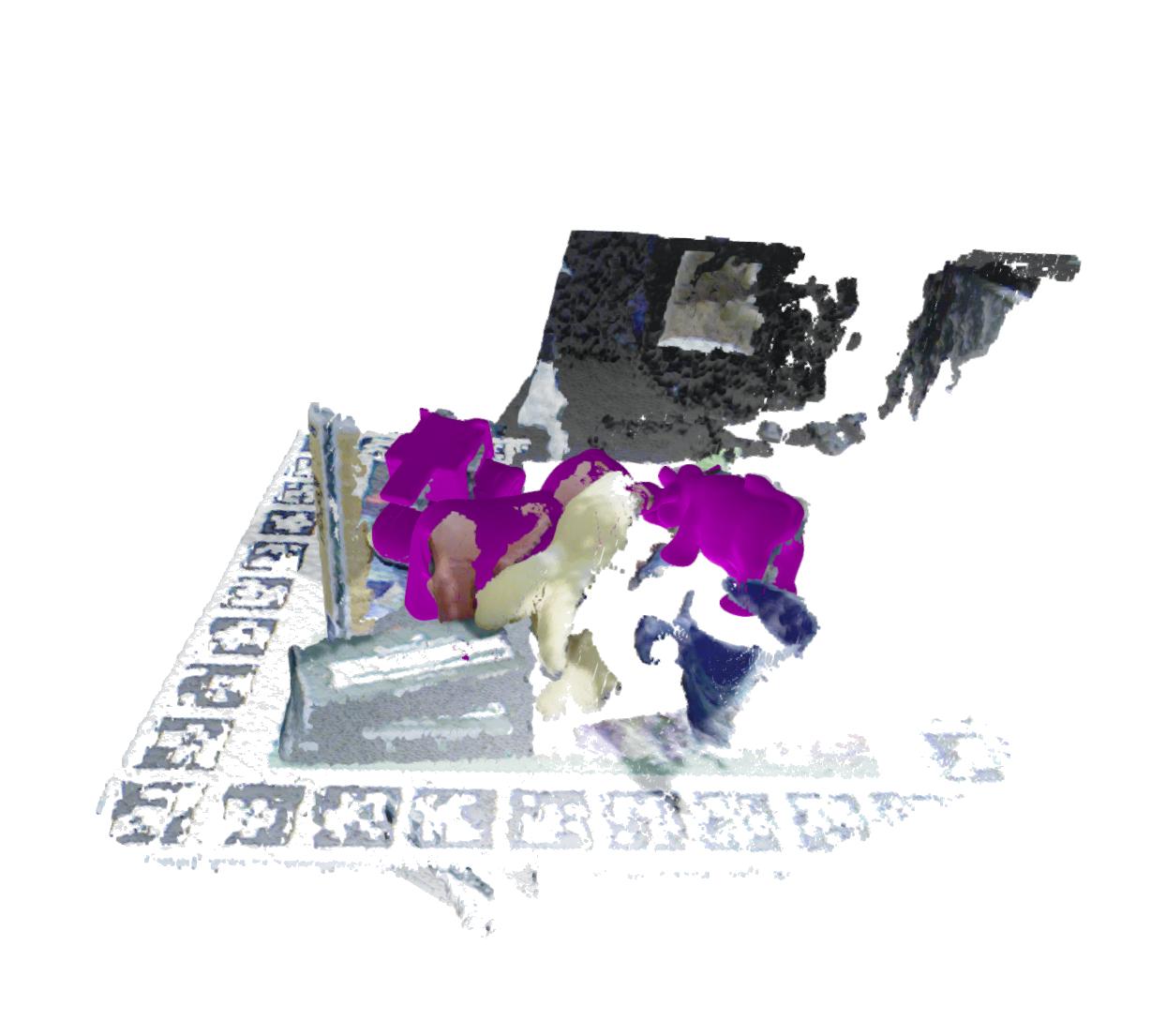}}\\
  
  \caption{HomebrewedDB (HB) dataset visualization with \textcolor{magenta}{estimated} (magenta) 6D pose of the meshes and point cloud. The first column shows the test image with a contour of the projection made by the predicted pose. The other three columns show the corresponding 3D view from different viewing angles. The first is taken from approximately the same viewing angle as the image was taken.}
  \label{fig:A0_HB}
\end{figure*}

\begin{figure*}
  \centering
  \subfloat{\includegraphics[height=.2\textwidth]{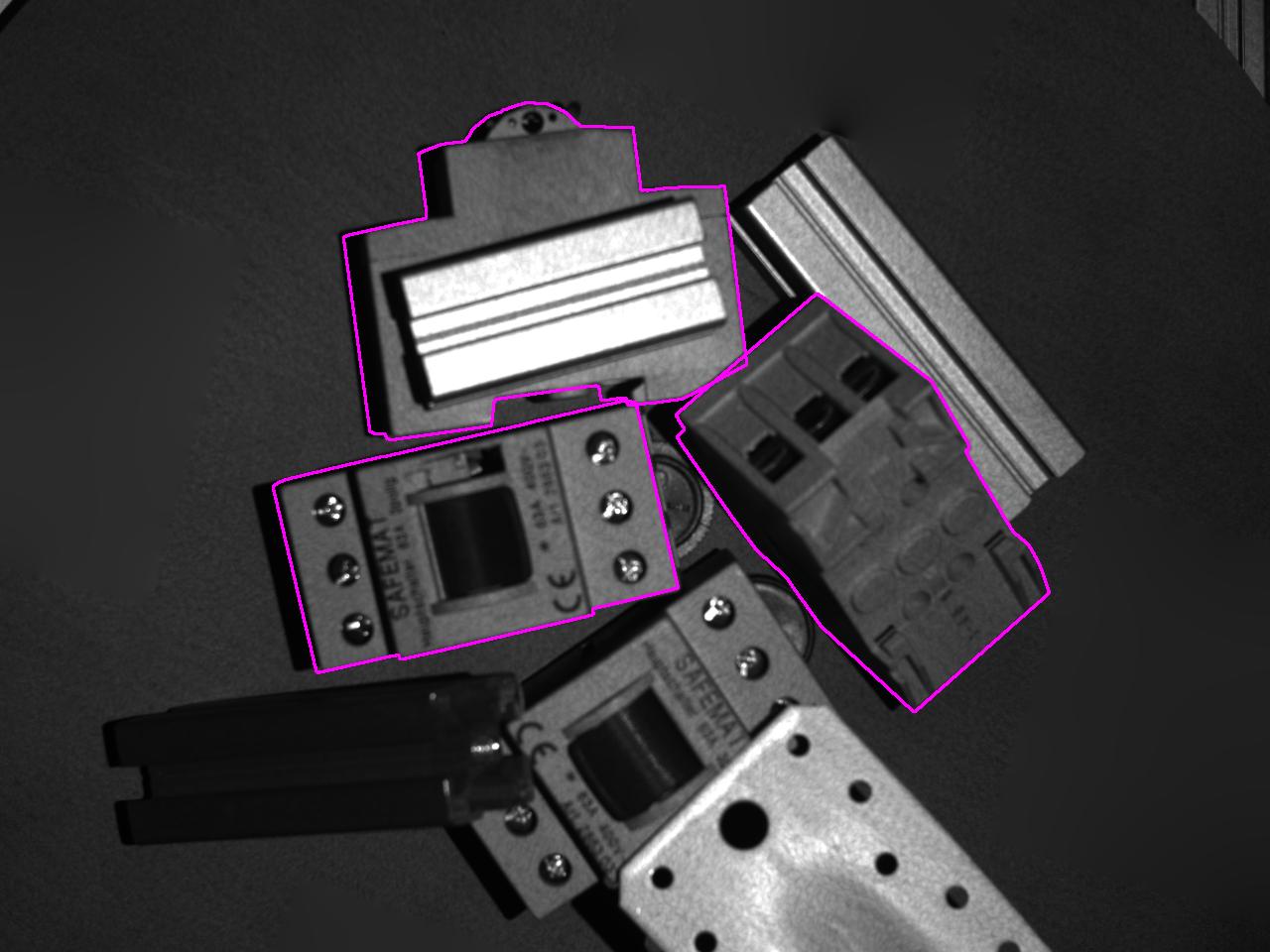}}\quad
  \subfloat{\includegraphics[height=.2\textwidth]{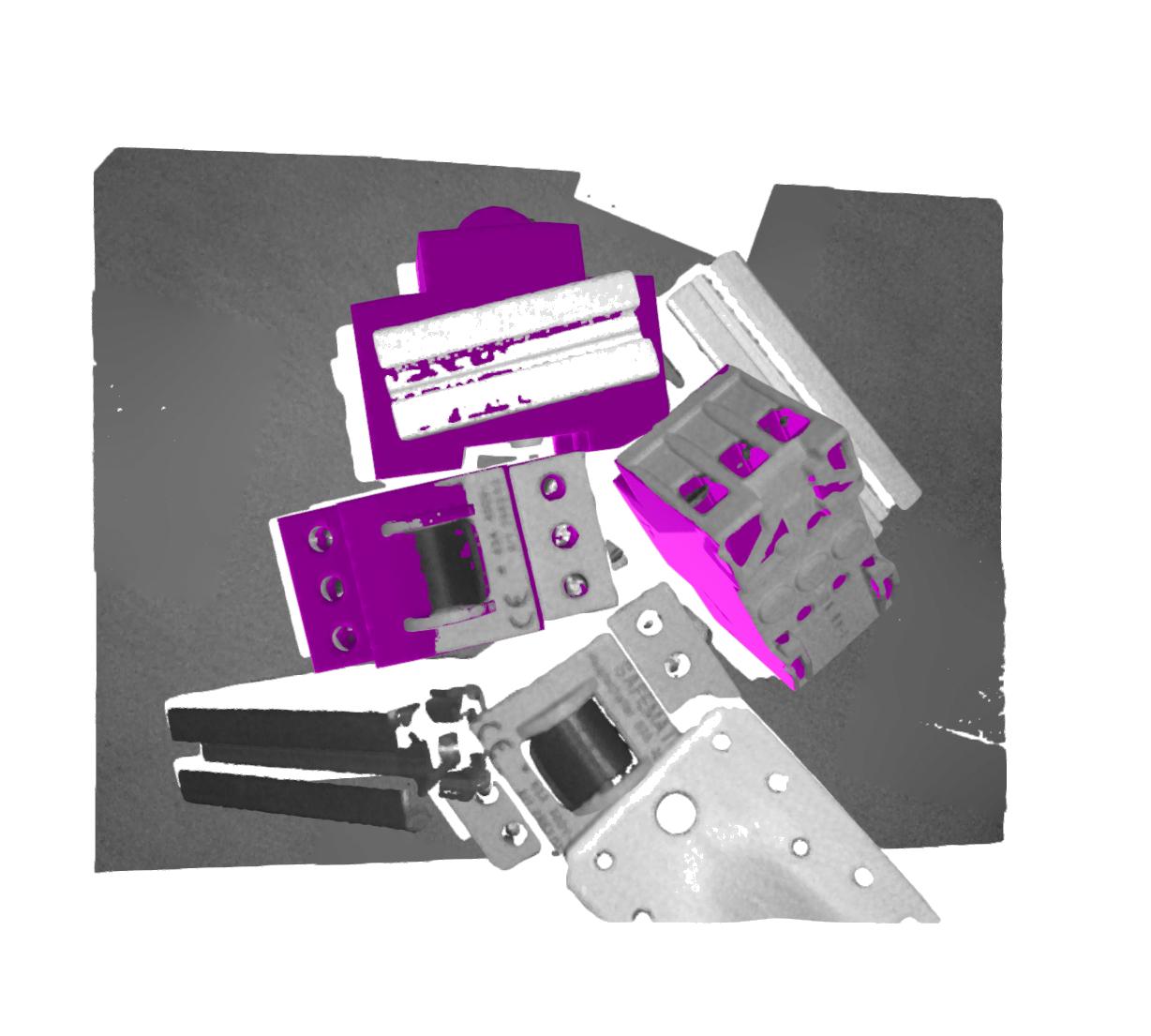}}\quad
  \subfloat{\includegraphics[height=.2\textwidth]{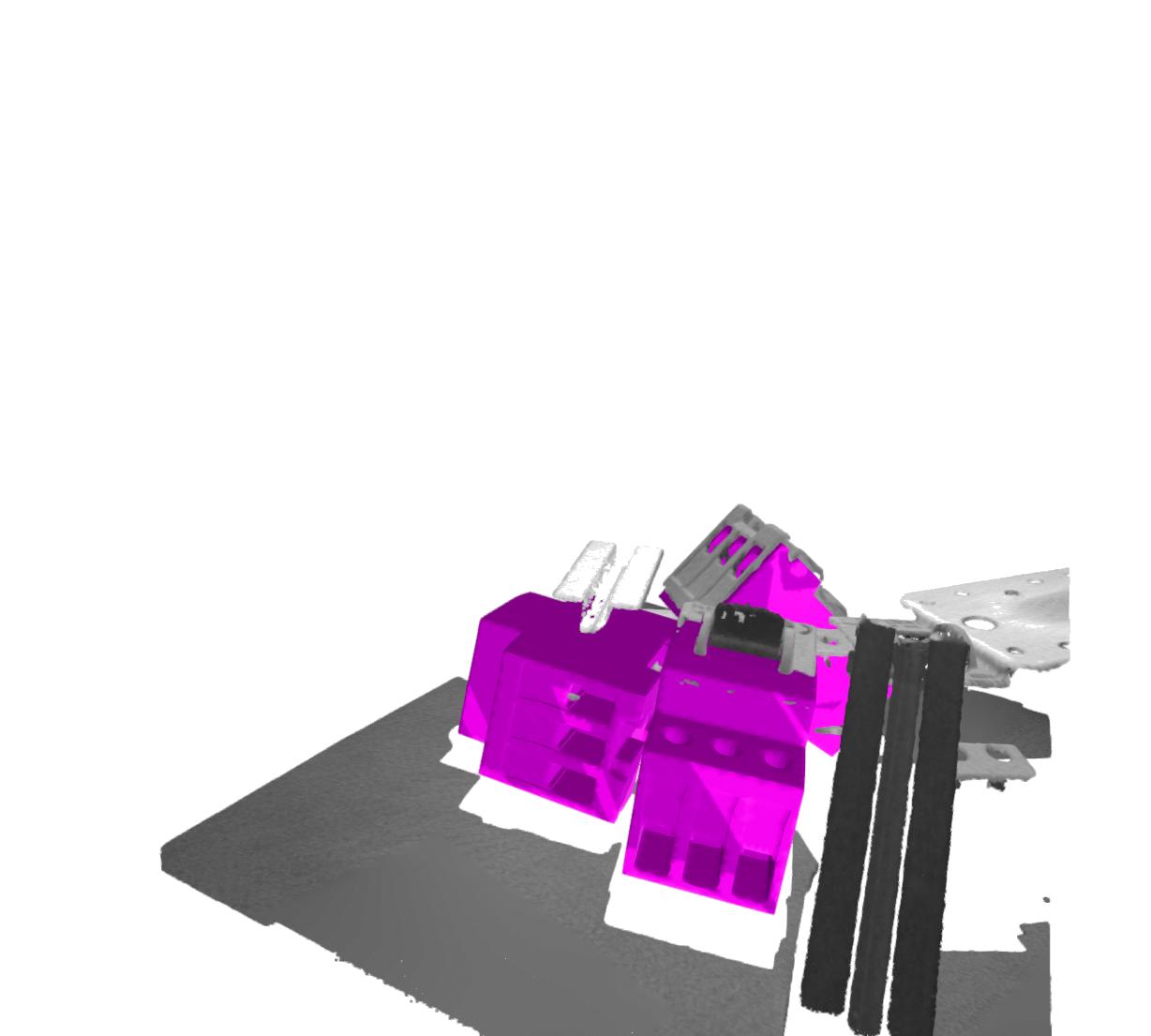}}\quad
  \subfloat{\includegraphics[height=.2\textwidth]{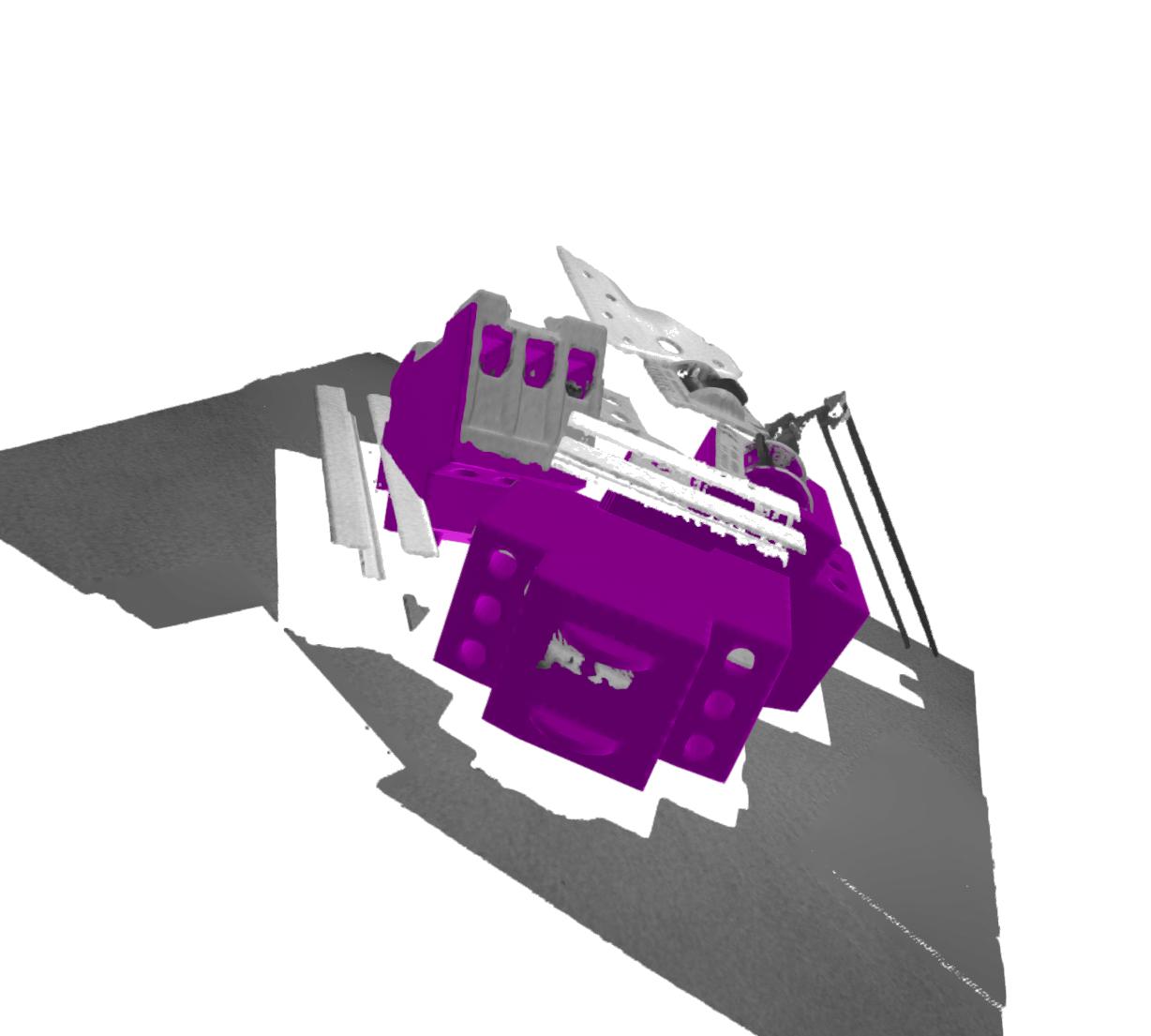}}\\
  
  \subfloat{\includegraphics[height=.2\textwidth]{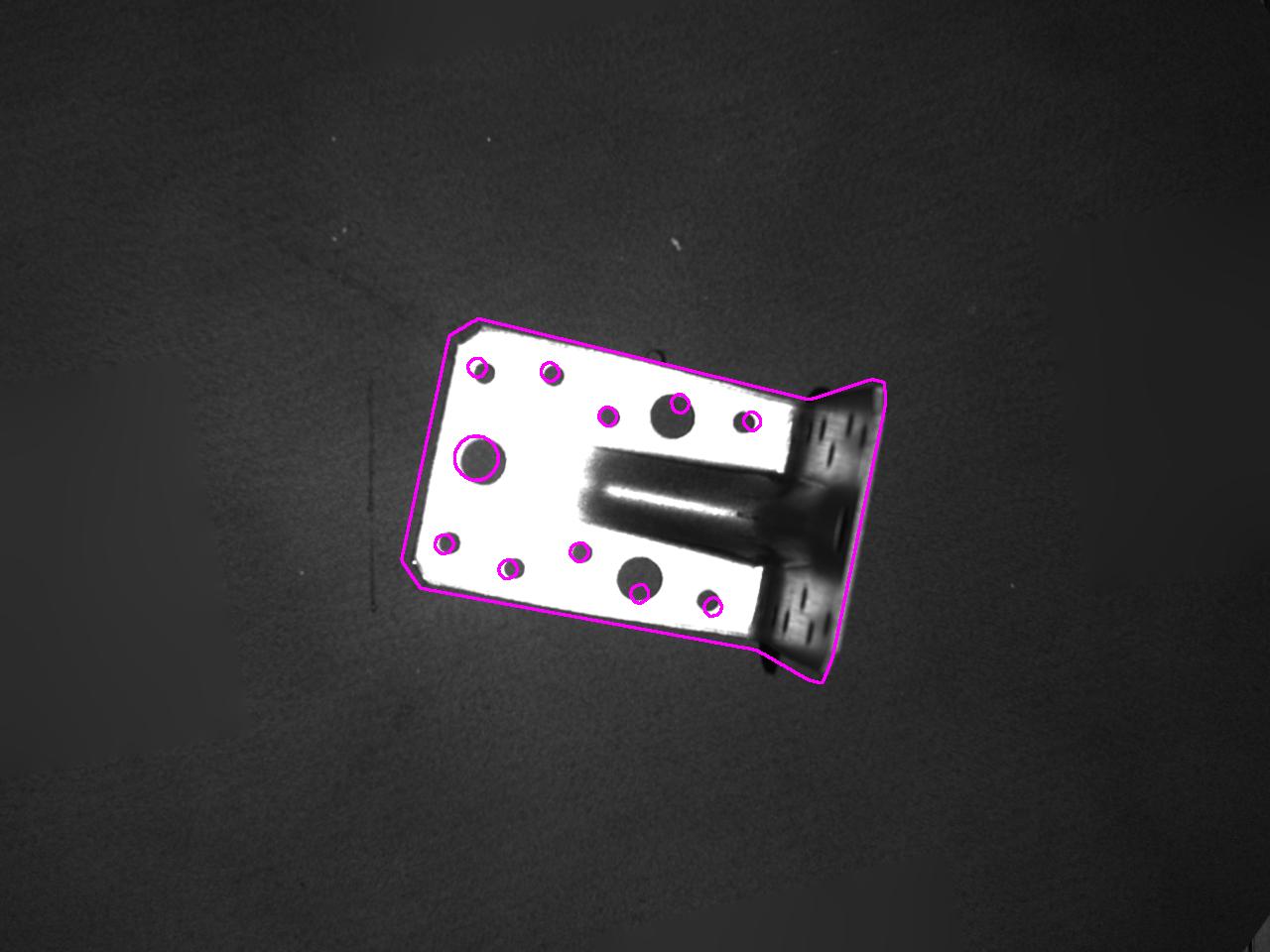}}\quad
  \subfloat{\includegraphics[height=.2\textwidth]{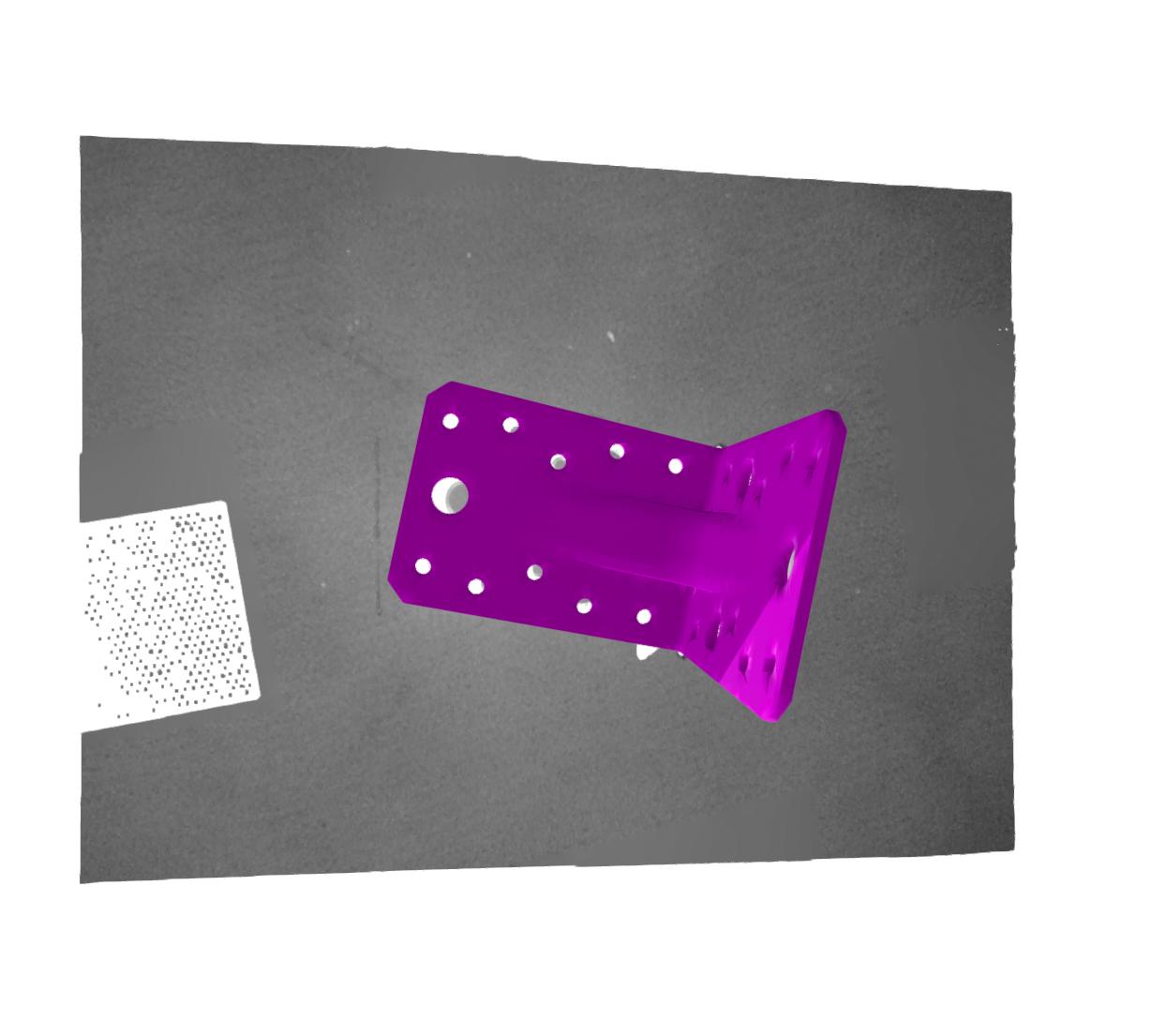}}\quad
  \subfloat{\includegraphics[height=.2\textwidth]{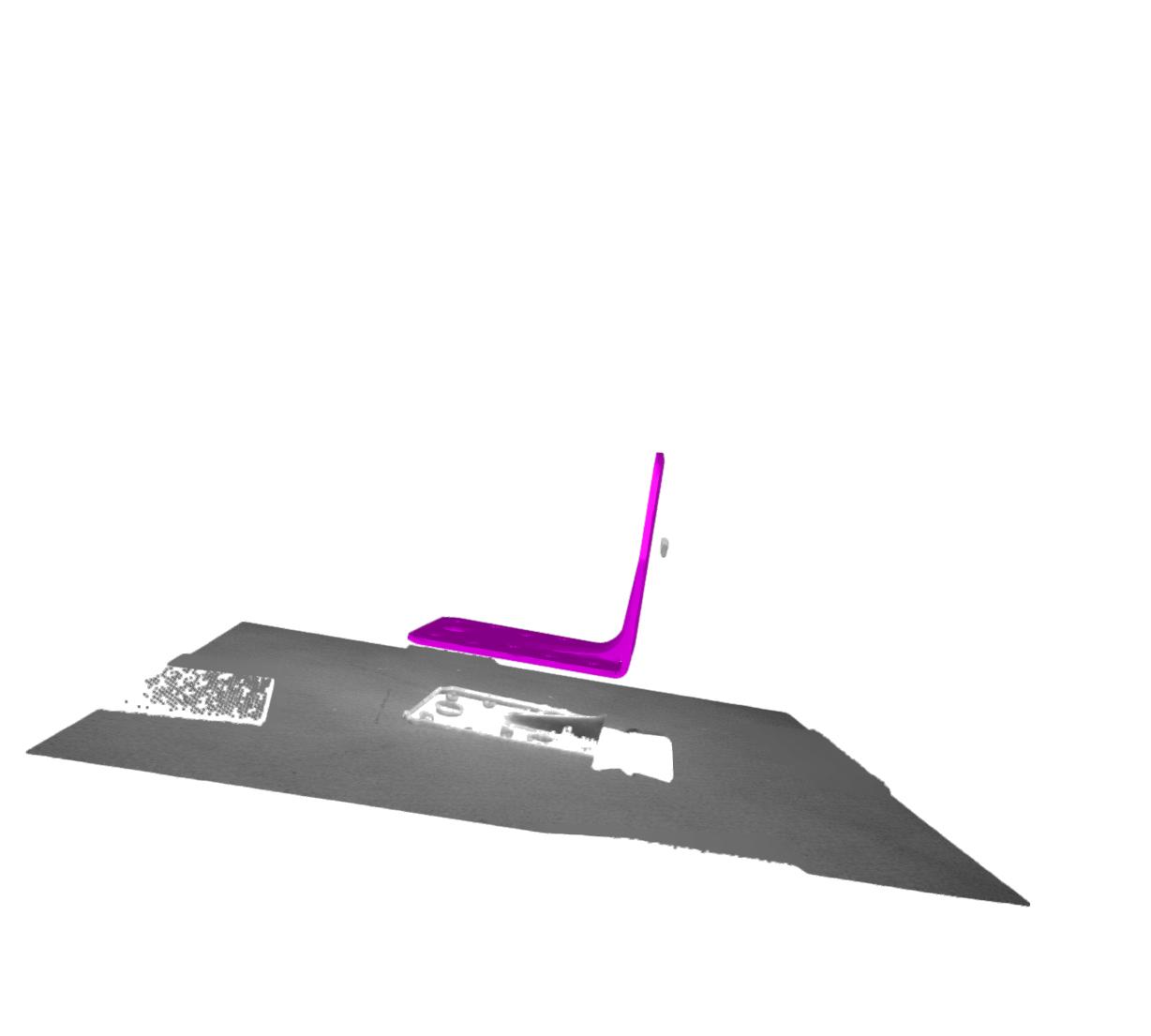}}\quad
  \subfloat{\includegraphics[height=.2\textwidth]{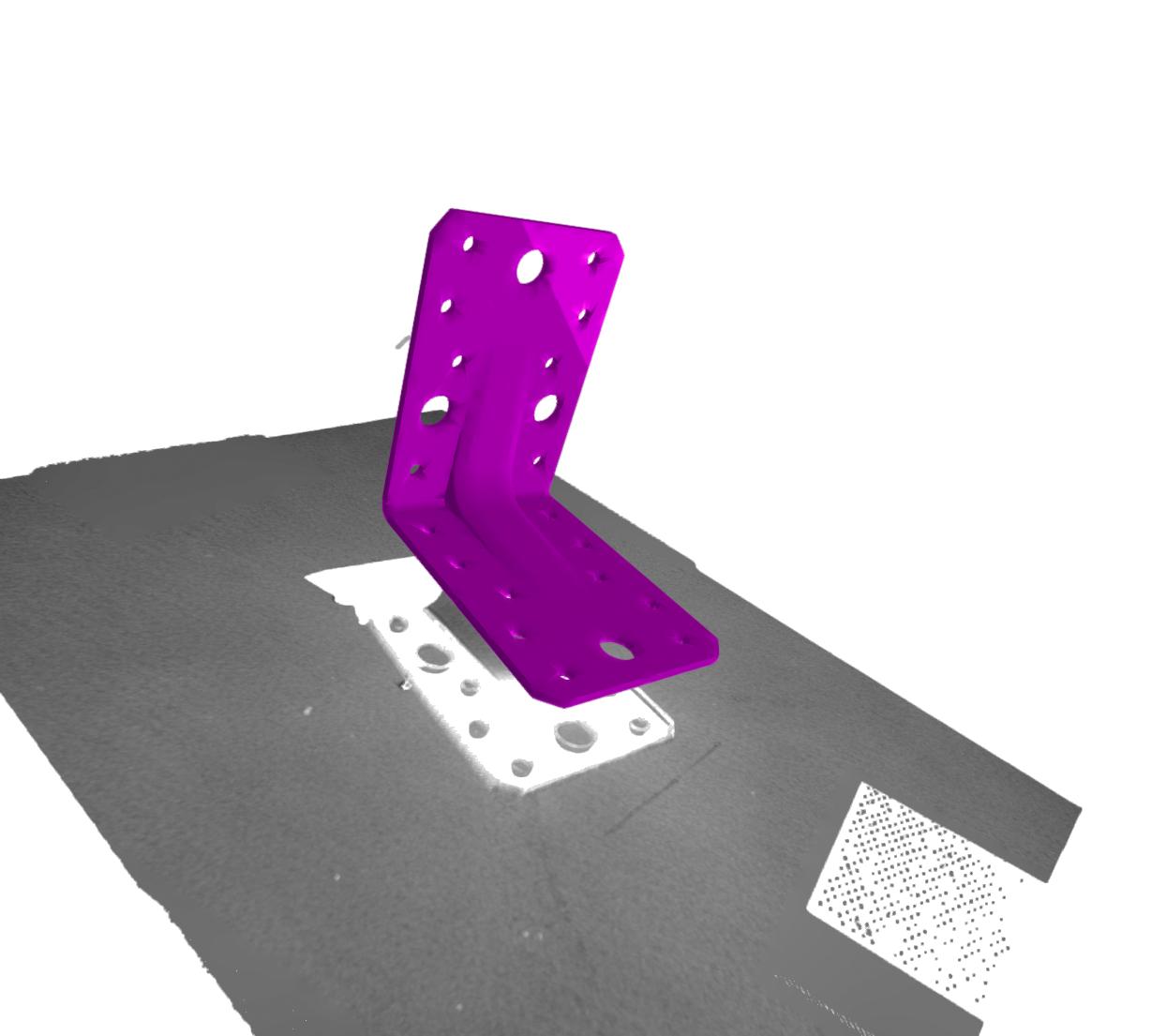}}\\
  
  \subfloat{\includegraphics[height=.2\textwidth]{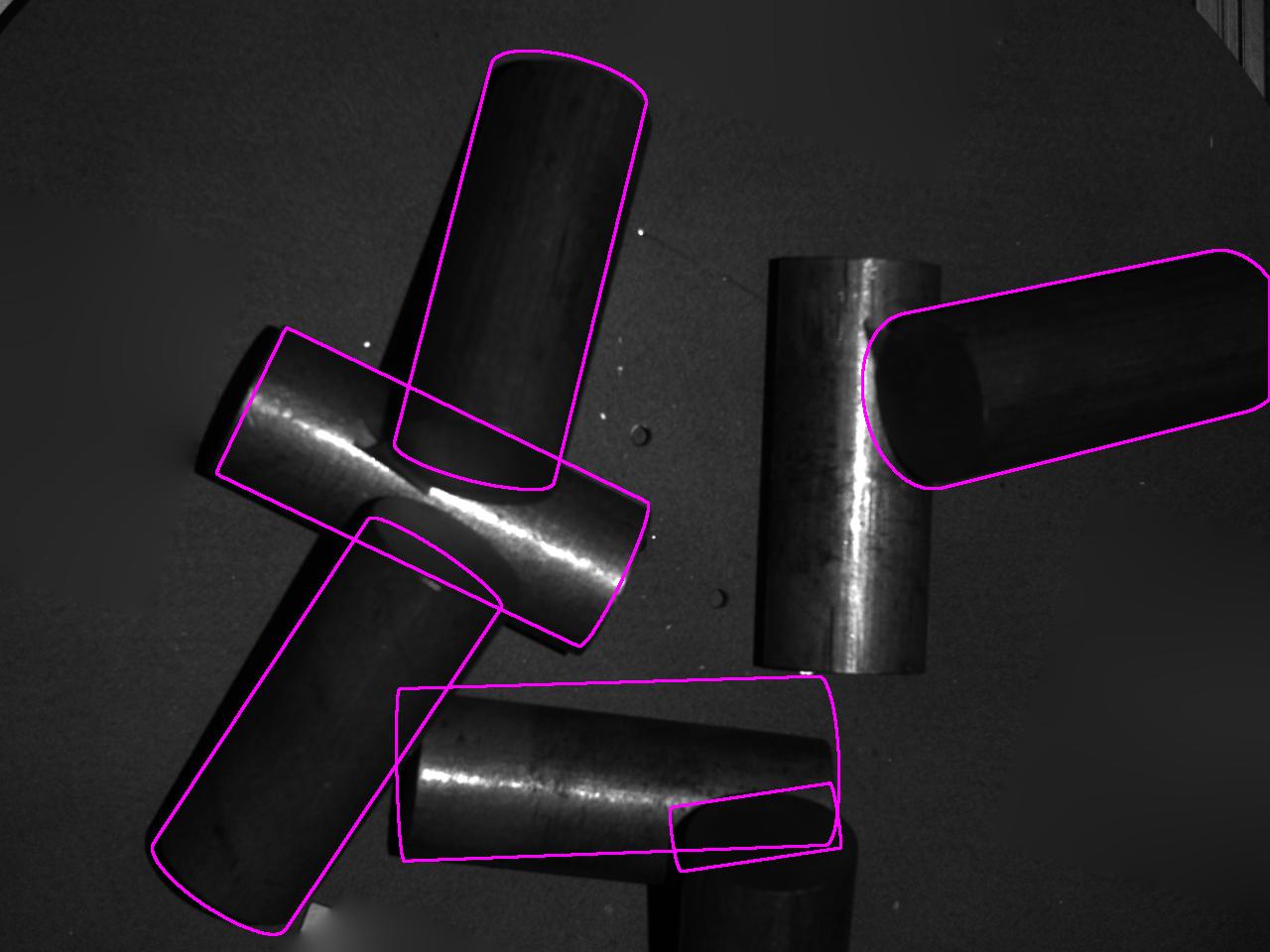}}\quad
  \subfloat{\includegraphics[height=.2\textwidth]{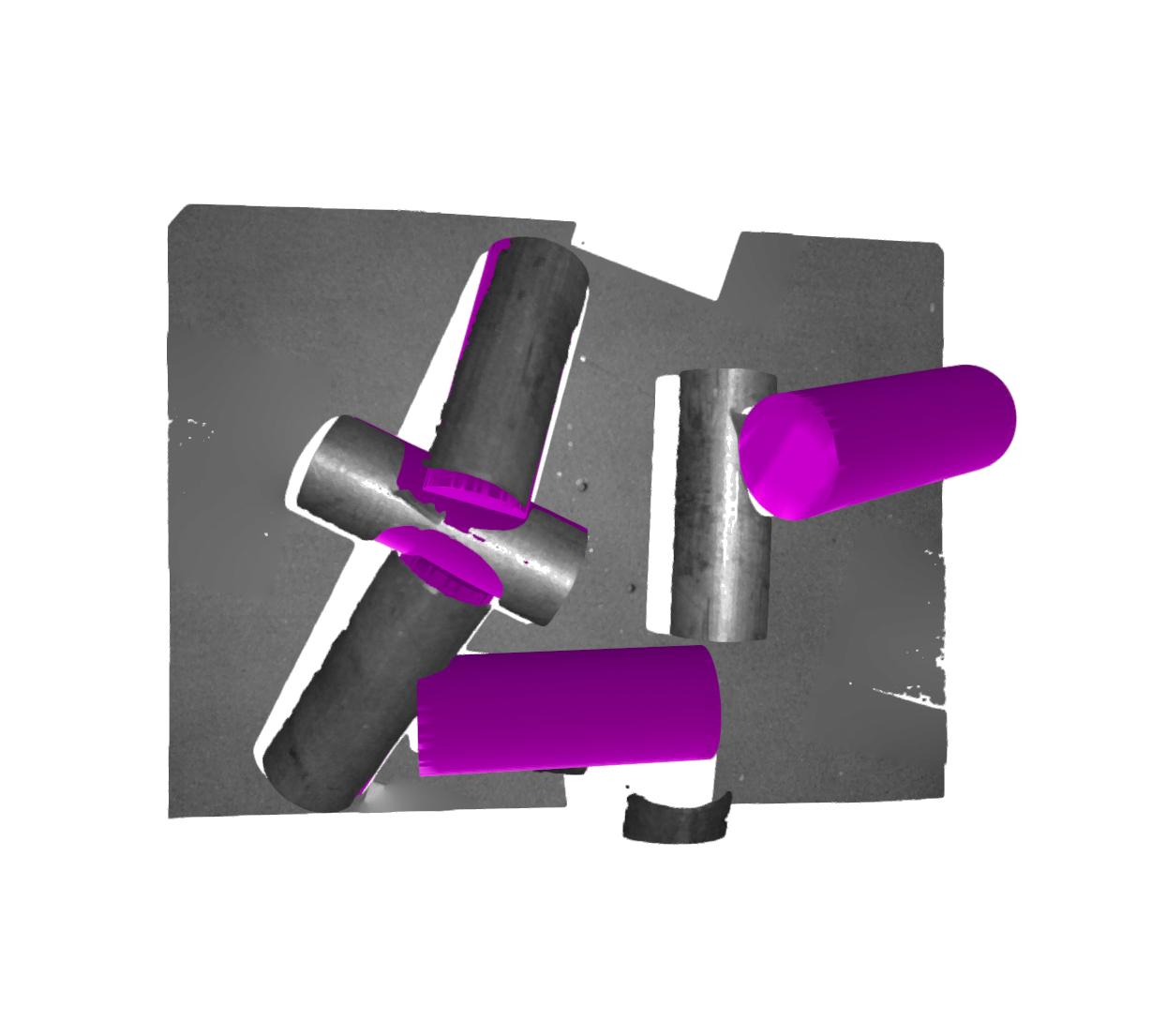}}\quad
  \subfloat{\includegraphics[height=.2\textwidth]{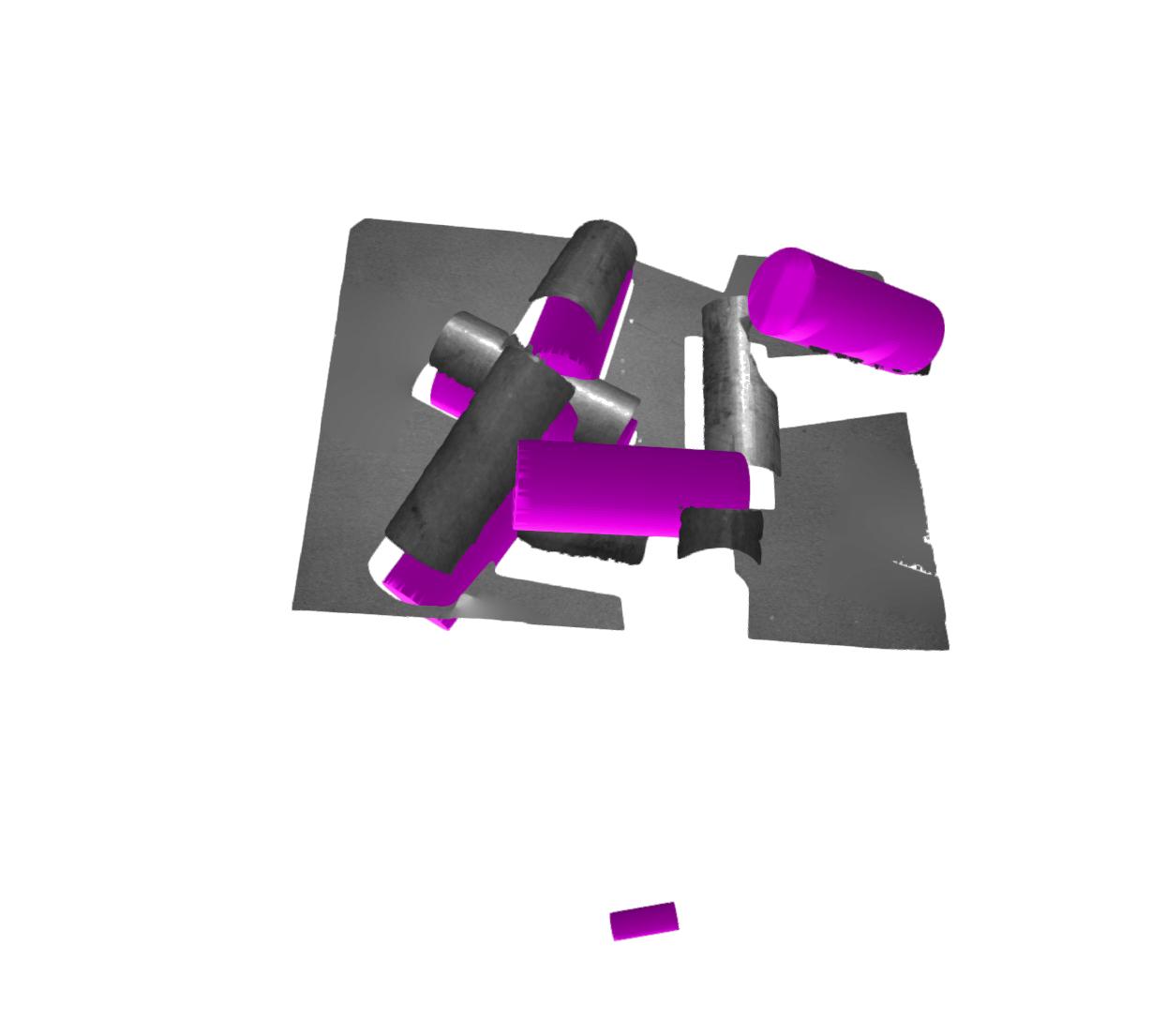}}\quad
  \subfloat{\includegraphics[height=.2\textwidth]{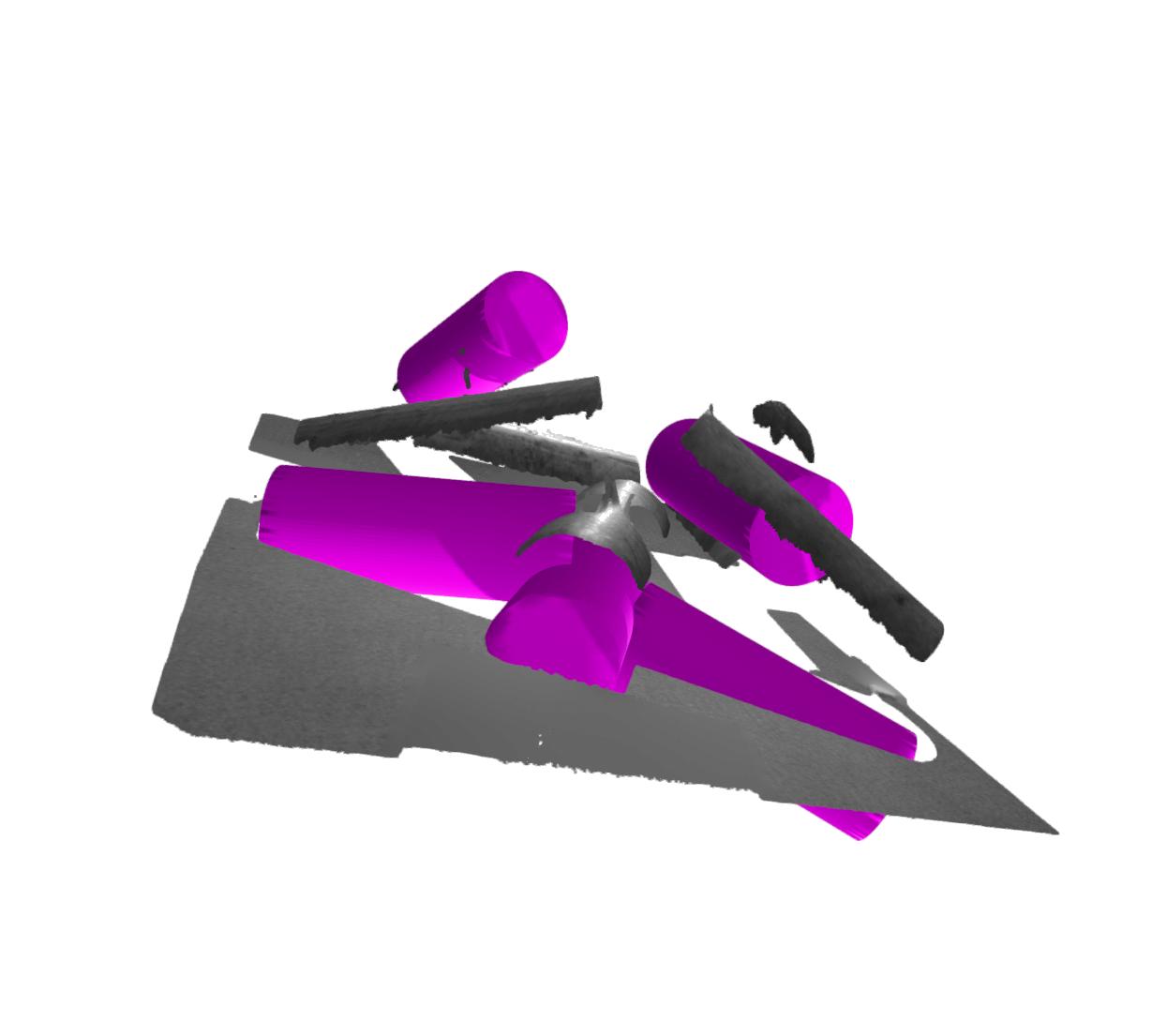}}\\
  
  \subfloat{\includegraphics[height=.2\textwidth]{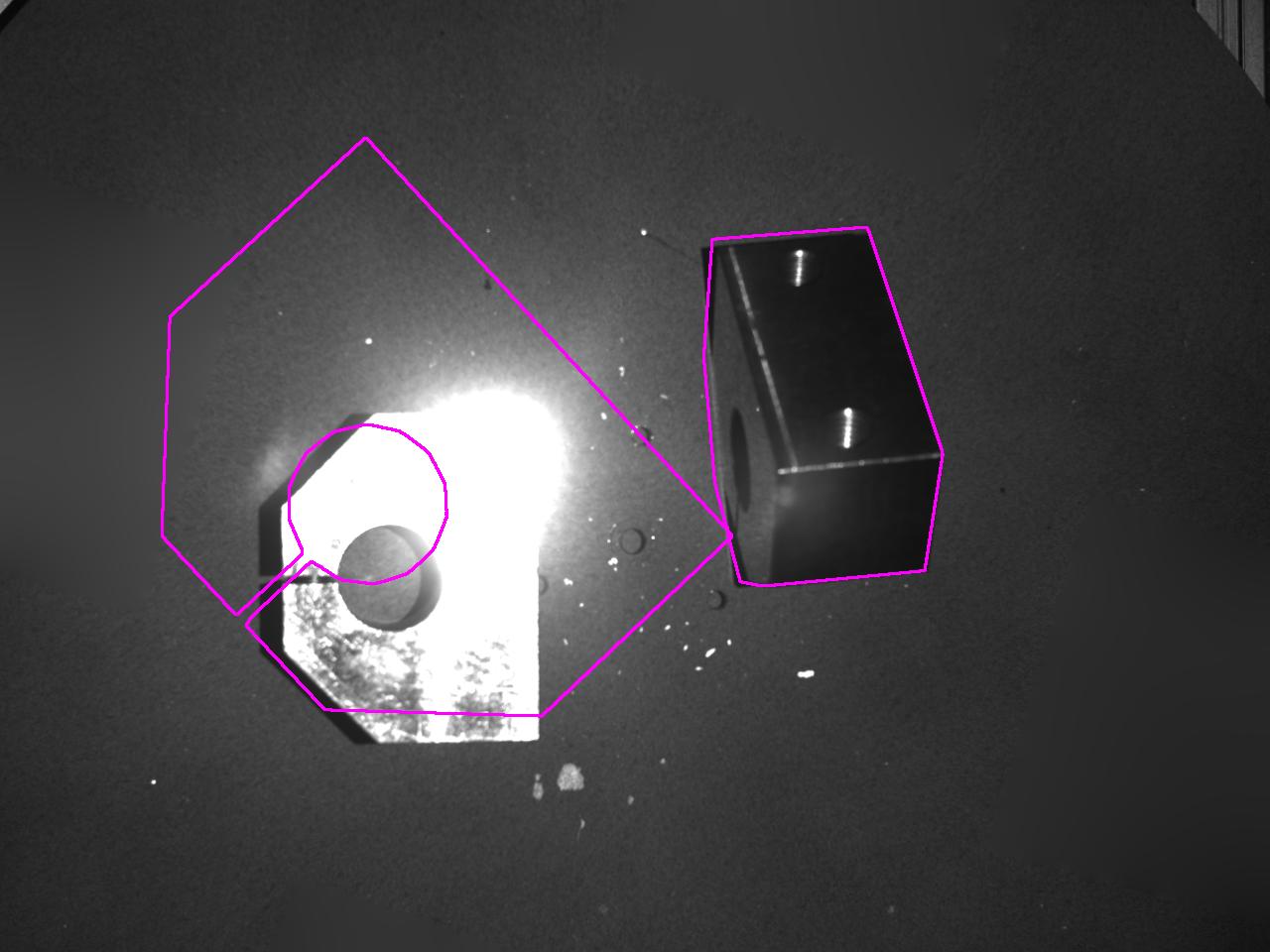}}\quad
  \subfloat{\includegraphics[height=.2\textwidth]{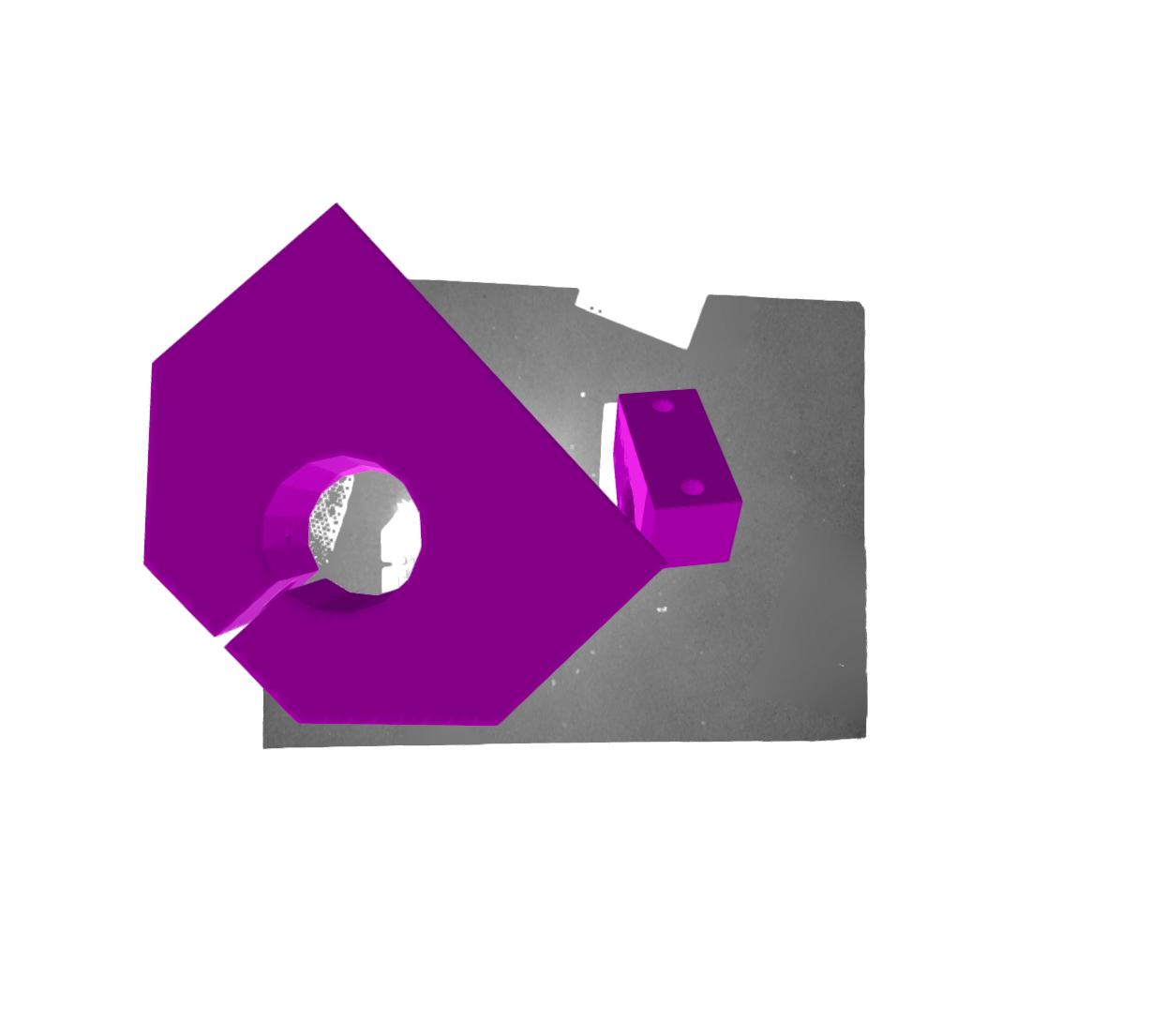}}\quad
  \subfloat{\includegraphics[height=.2\textwidth]{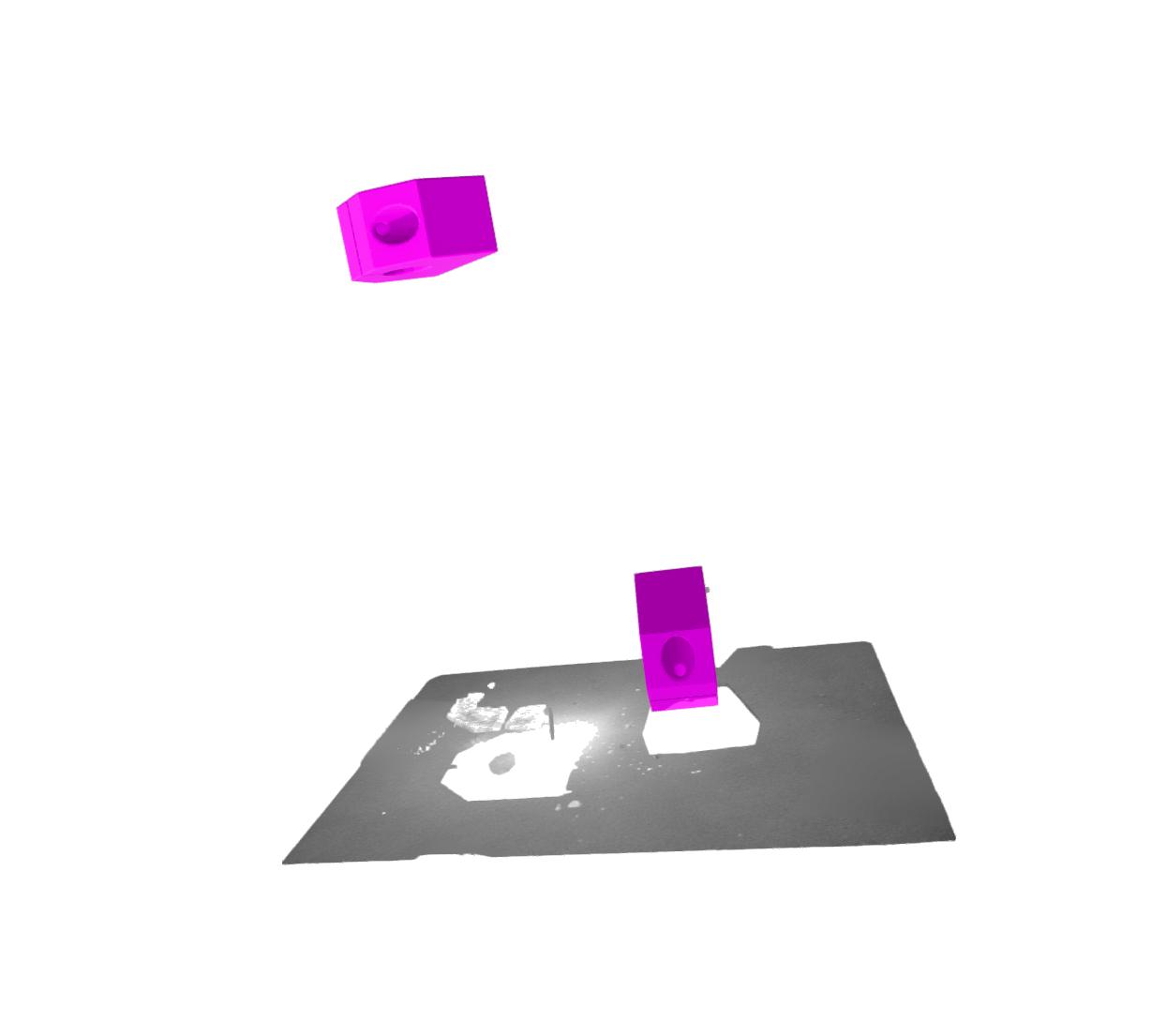}}\quad
  \subfloat{\includegraphics[height=.2\textwidth]{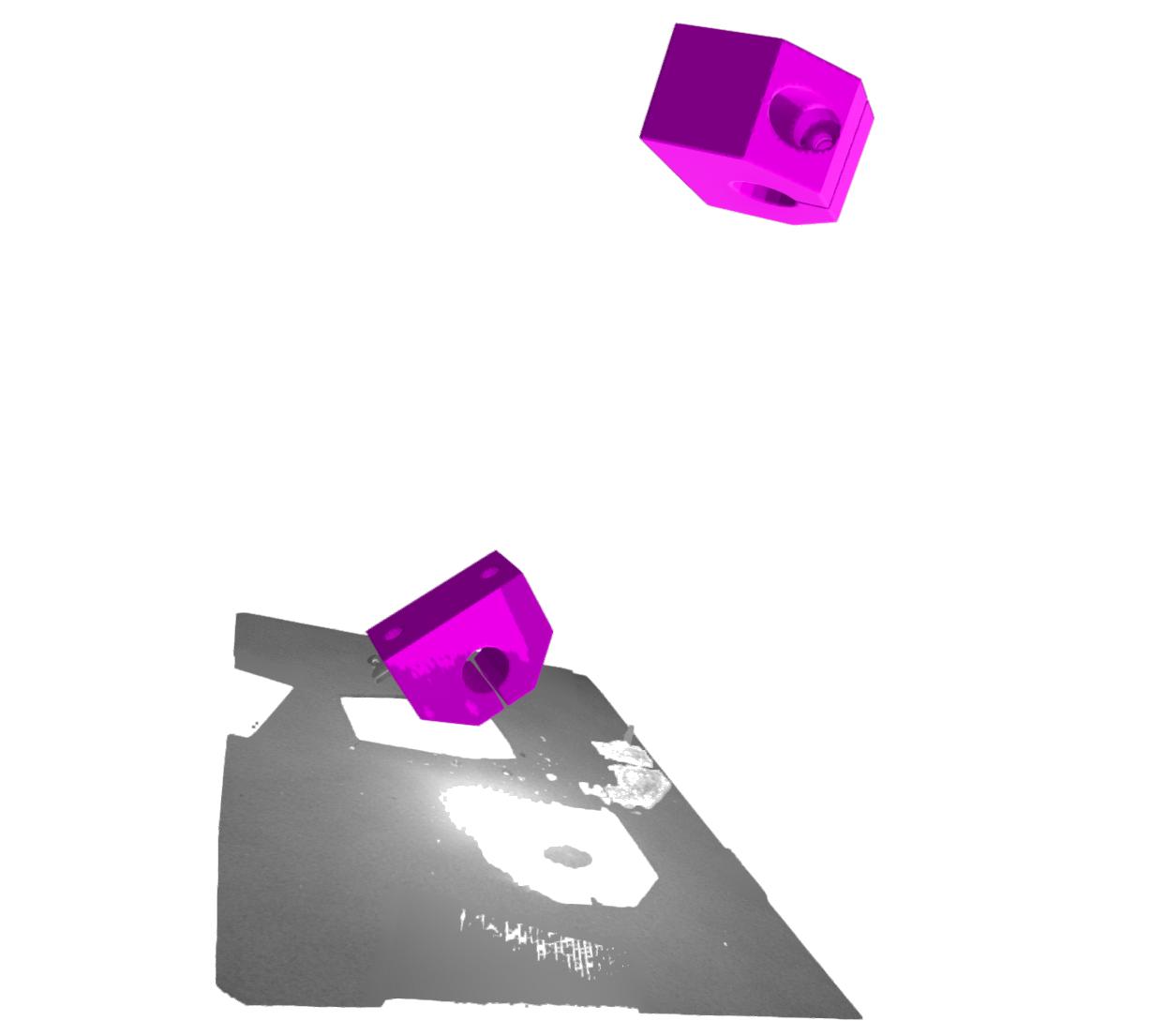}}\\
  
  \subfloat{\includegraphics[height=.2\textwidth]{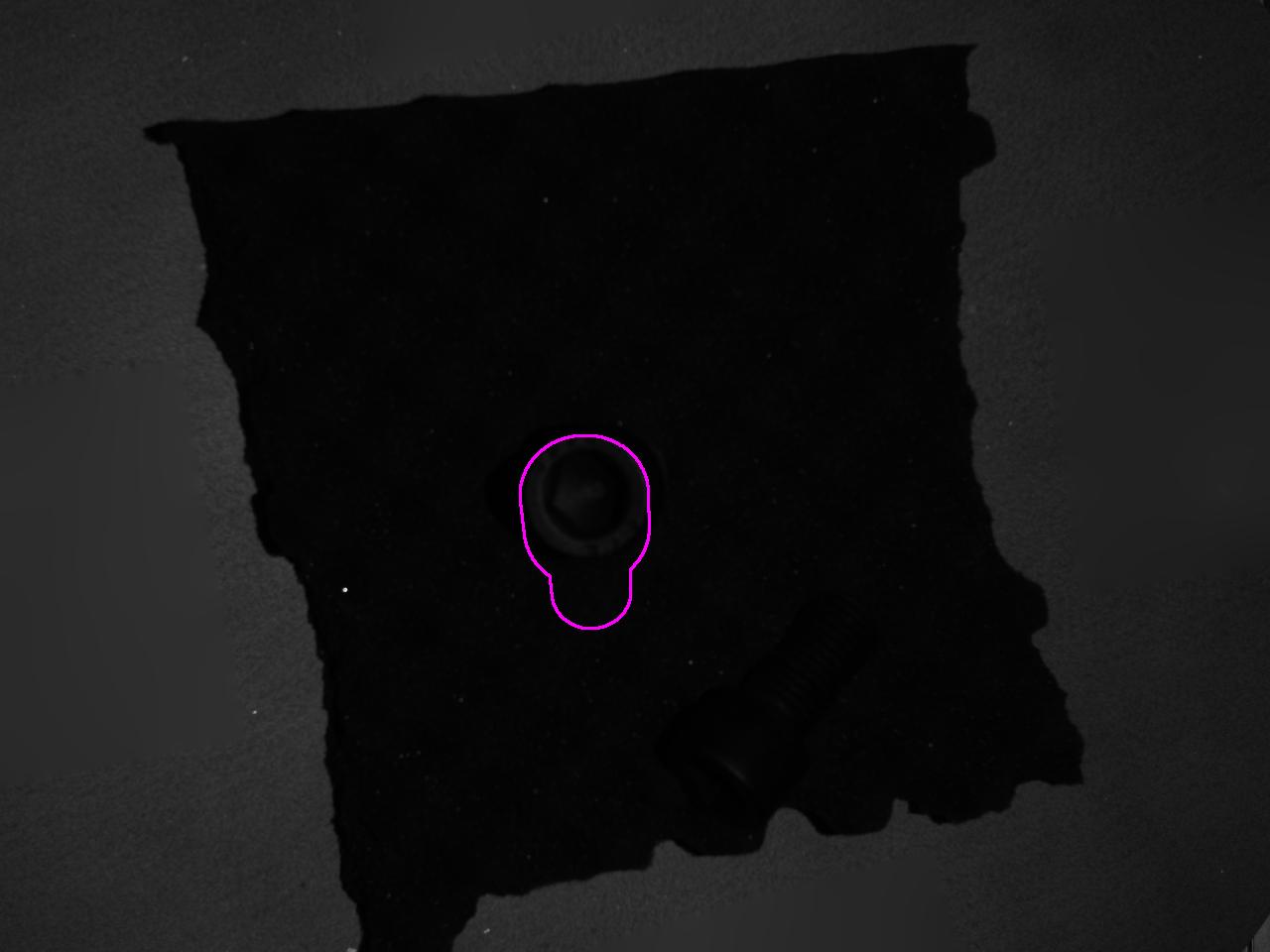}}\quad
  \subfloat{\includegraphics[height=.2\textwidth]{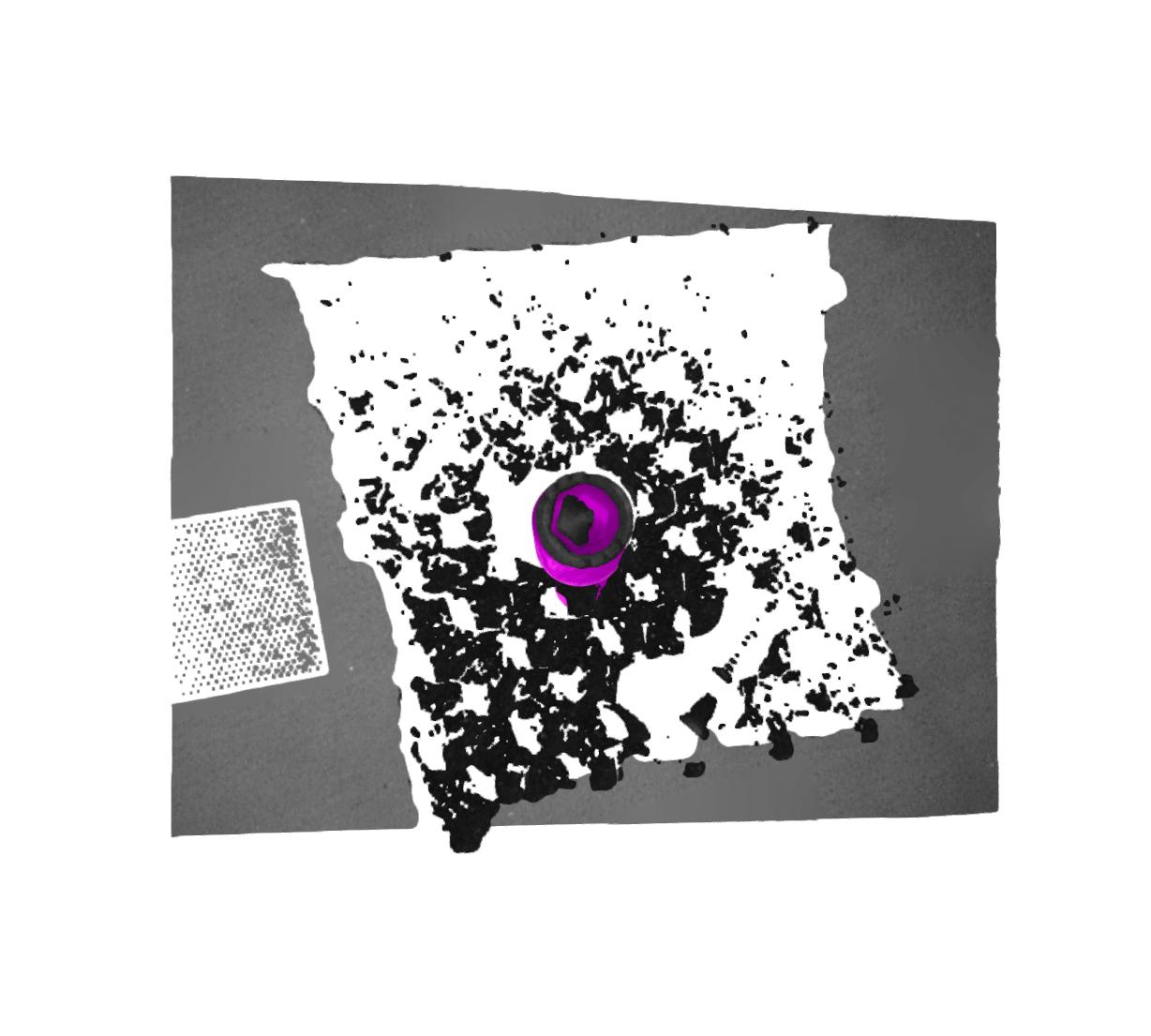}}\quad
  \subfloat{\includegraphics[height=.2\textwidth]{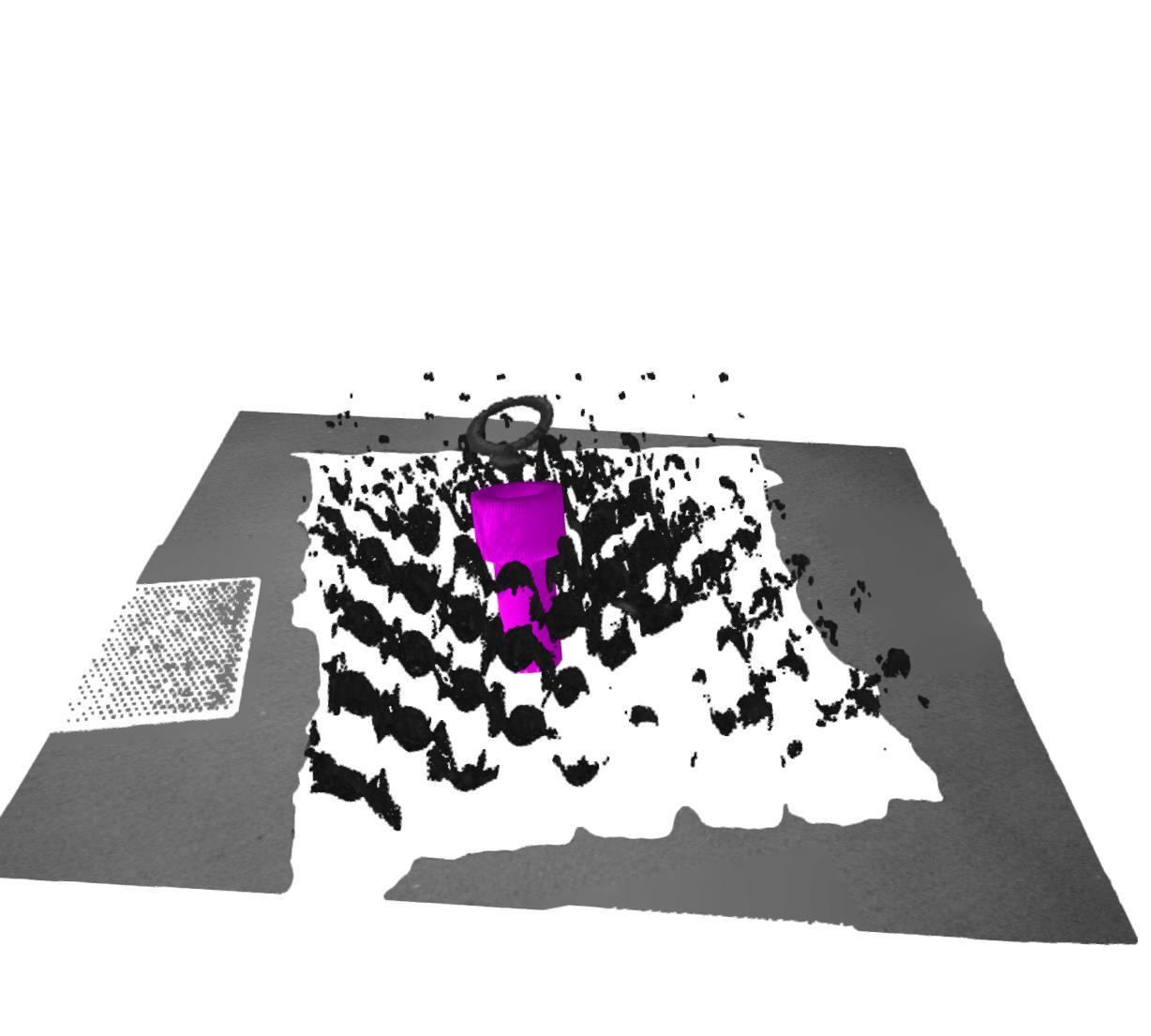}}\quad
  \subfloat{\includegraphics[height=.2\textwidth]{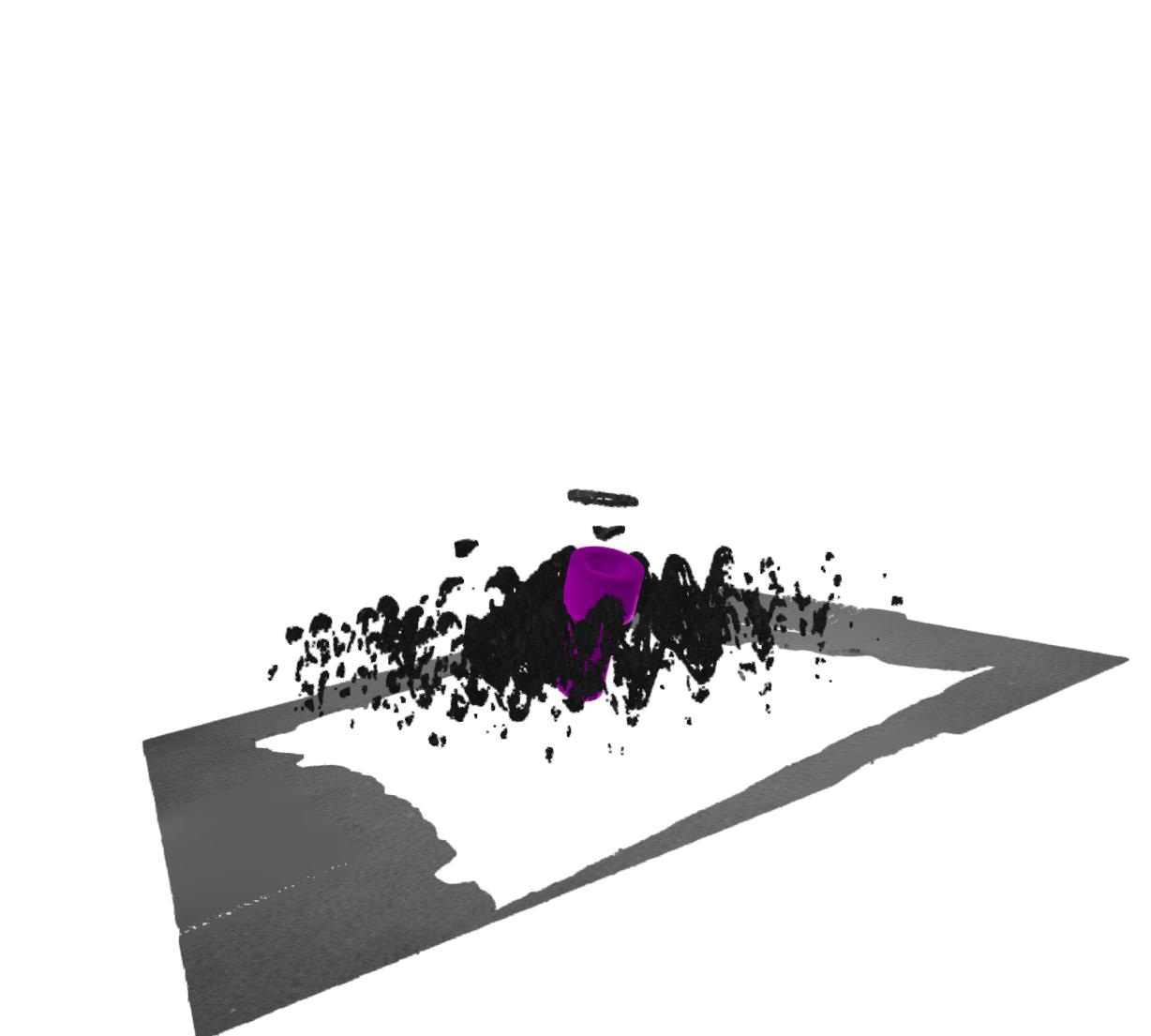}}\\
  
  \caption{ITODD dataset visualization with \textcolor{magenta}{estimated} (magenta) 6D pose of the meshes and point cloud. The first column shows the test image with a contour of the projection made by the predicted pose. The other three columns show the corresponding 3D view from different viewing angles. The first is taken from approximately the same viewing angle as the image was taken.}
  \label{fig:A0_ITODD}
\end{figure*}

\begin{figure*}
  \centering
  \subfloat{\includegraphics[height=.2\textwidth]{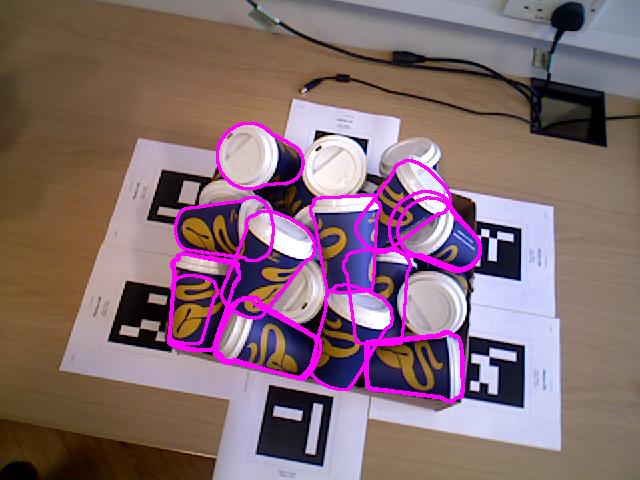}}\quad
  \subfloat{\includegraphics[height=.2\textwidth]{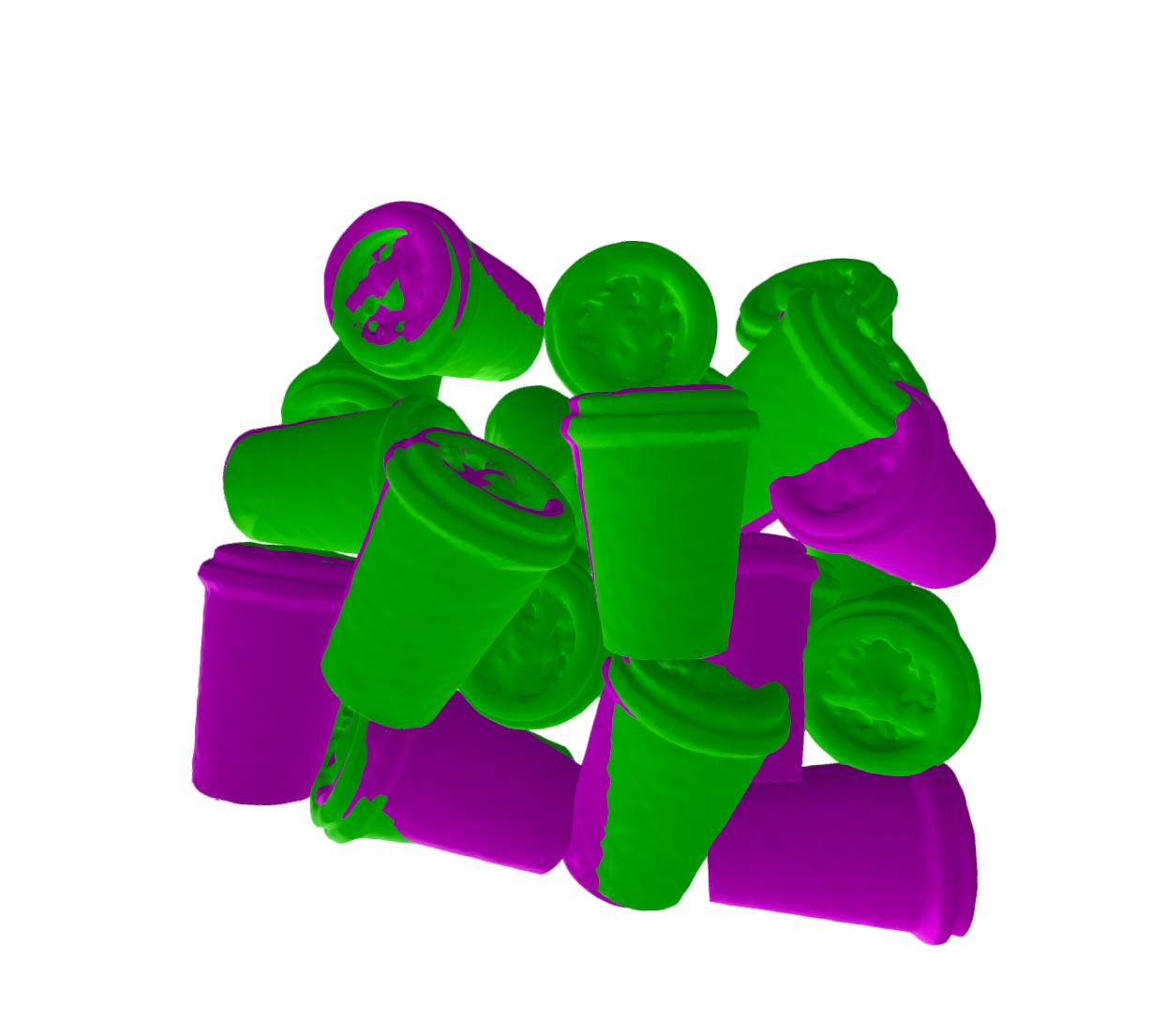}}\quad
  \subfloat{\includegraphics[height=.2\textwidth]{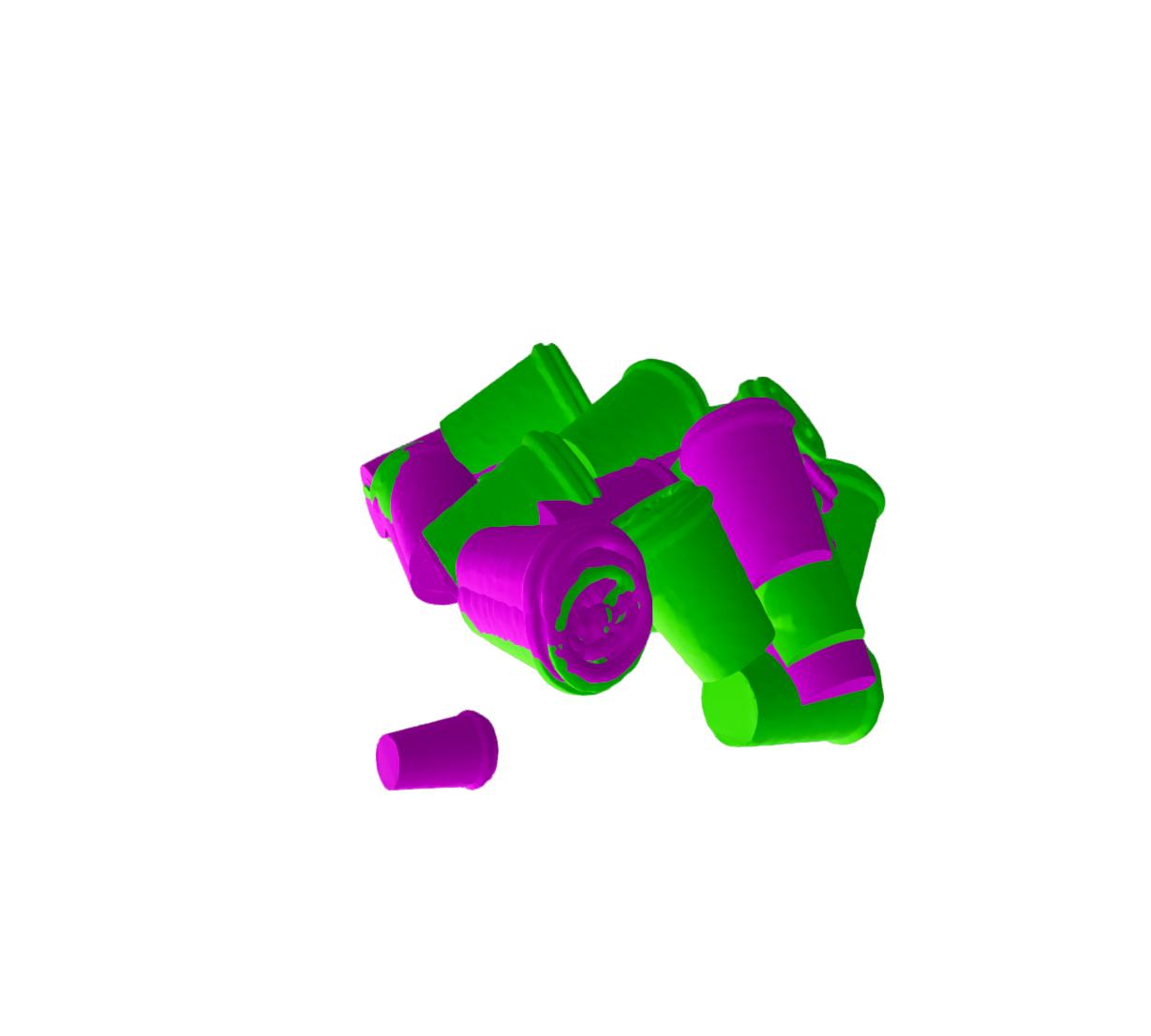}}\quad
  \subfloat{\includegraphics[height=.2\textwidth]{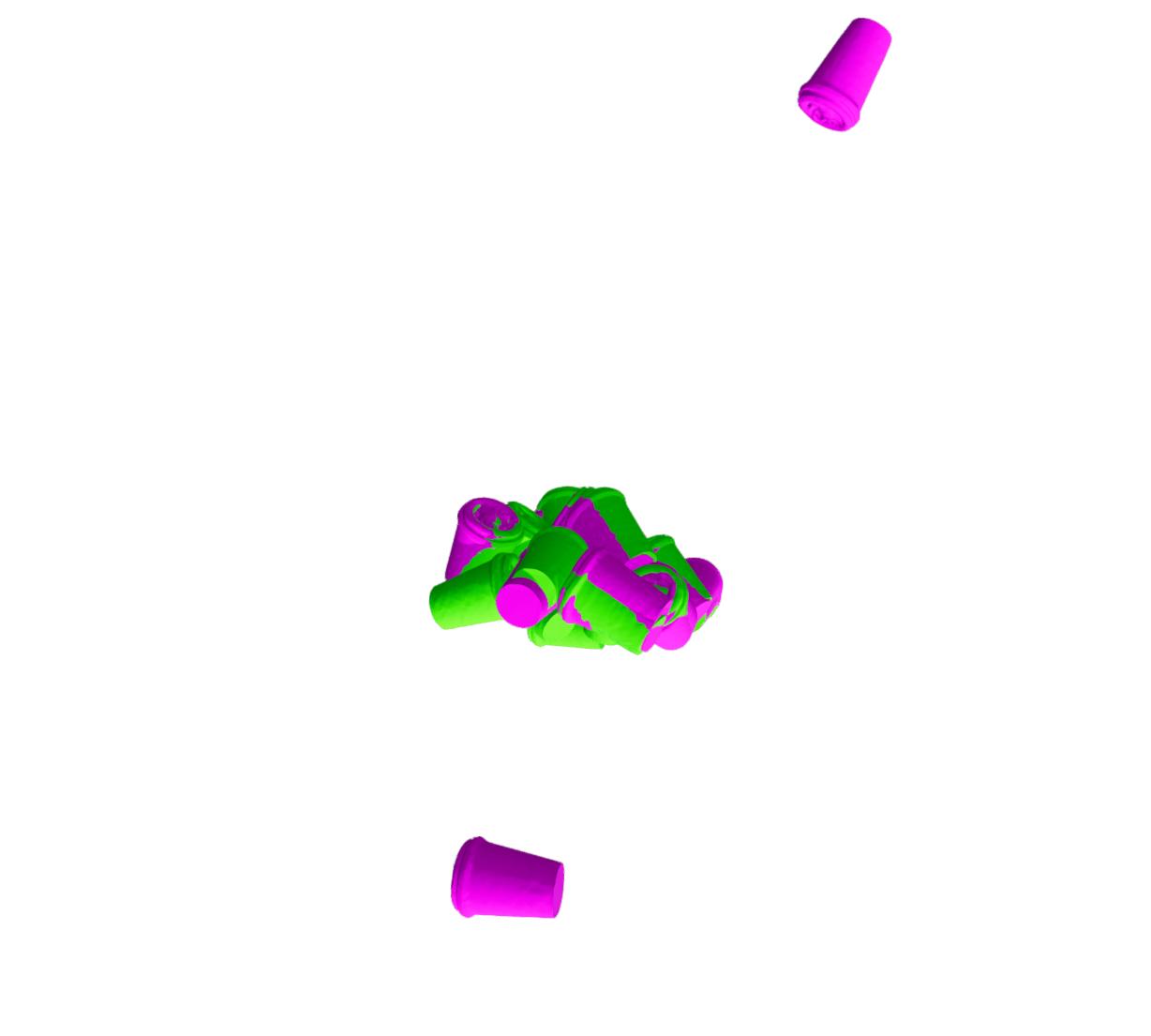}}\\
  
  \subfloat{\includegraphics[height=.2\textwidth]{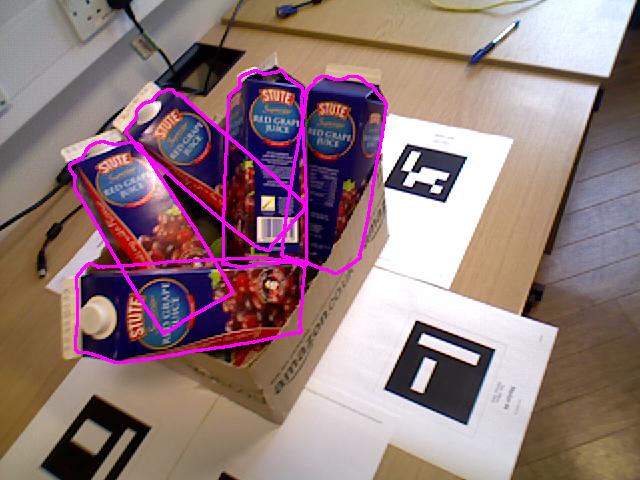}}\quad
  \subfloat{\includegraphics[height=.2\textwidth]{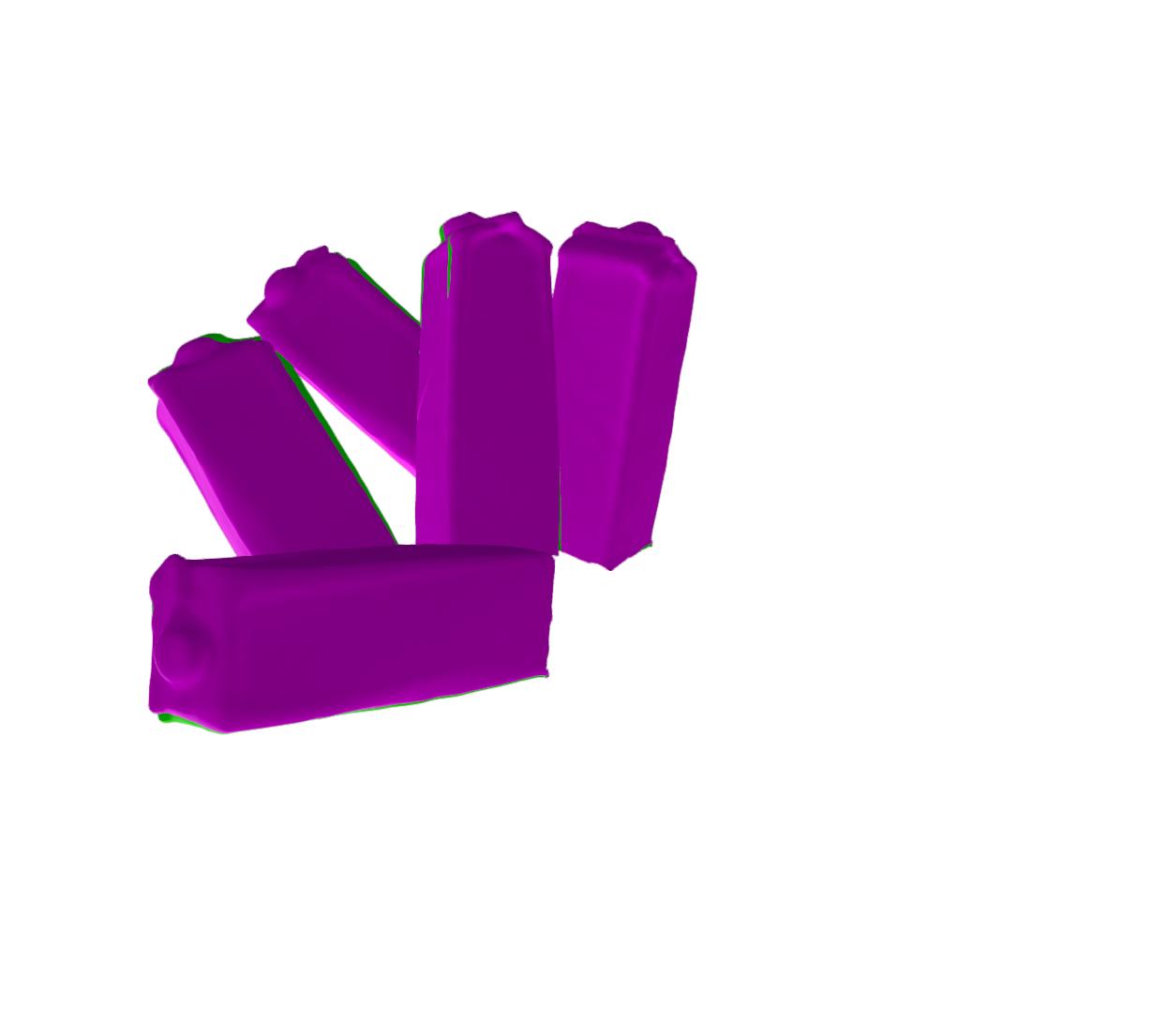}}\quad
  \subfloat{\includegraphics[height=.2\textwidth]{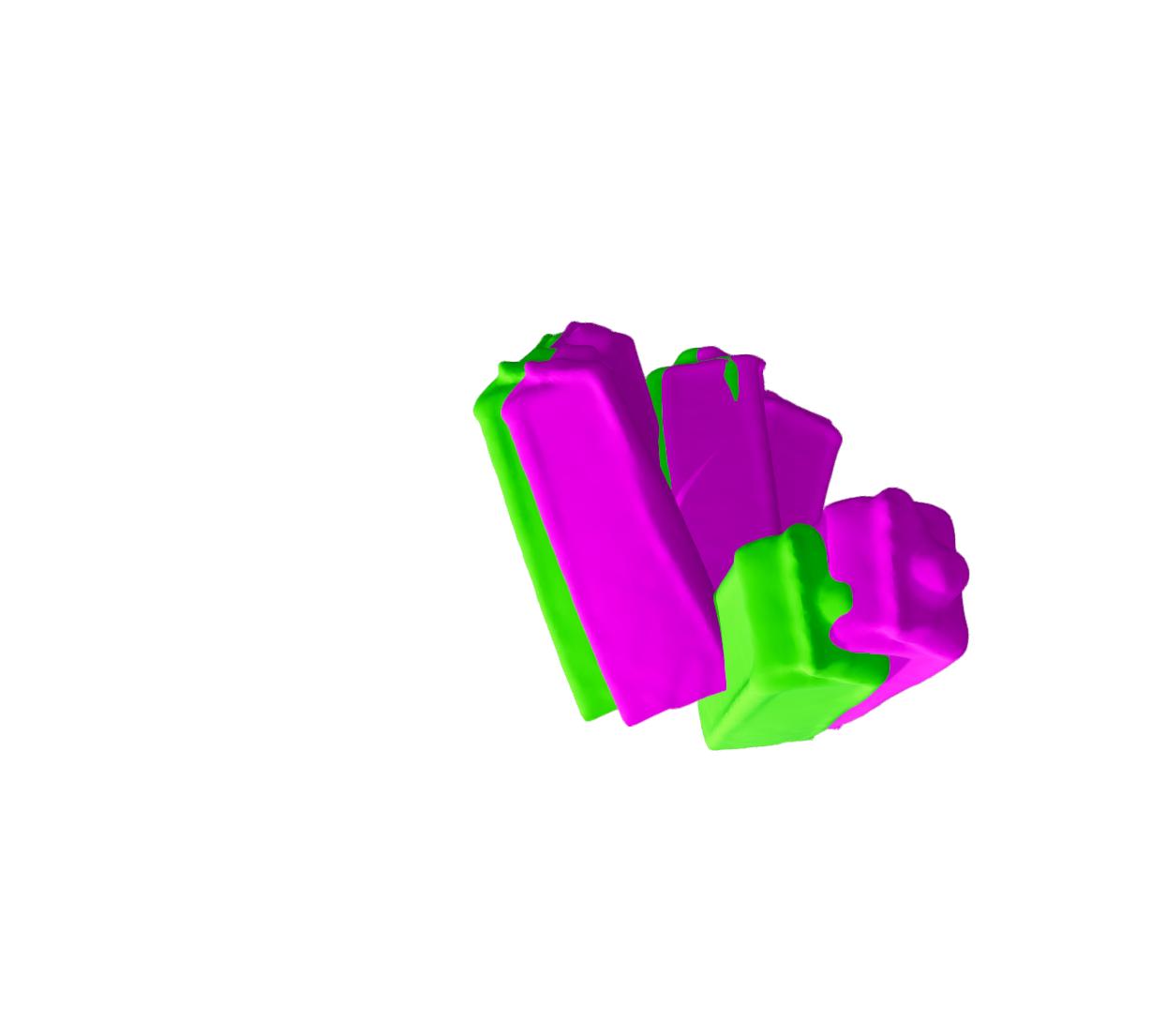}}\quad
  \subfloat{\includegraphics[height=.2\textwidth]{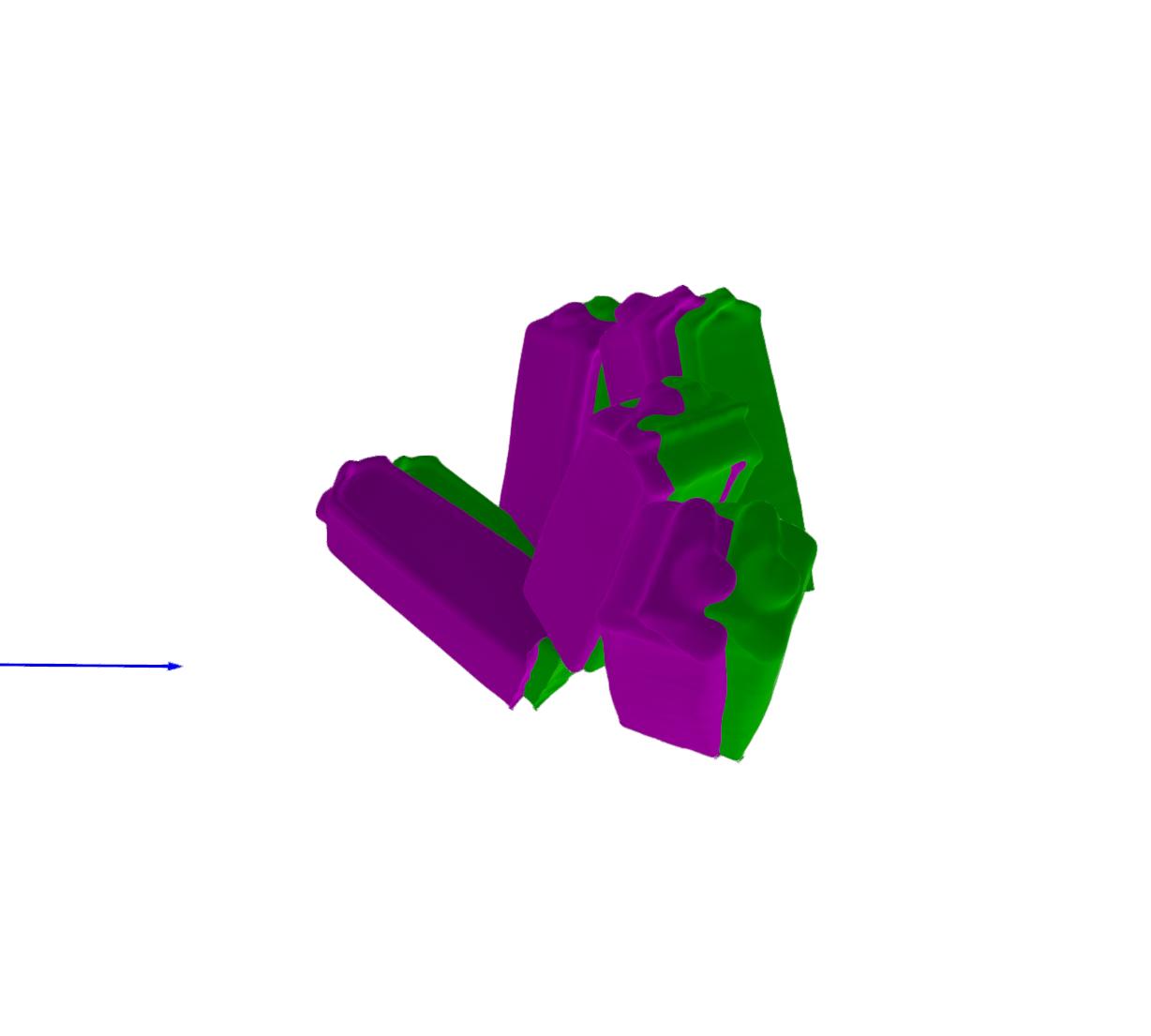}}\\
  
  \subfloat{\includegraphics[height=.2\textwidth]{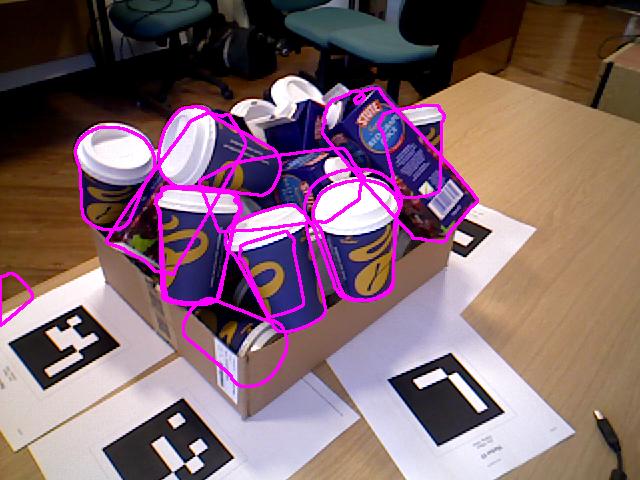}}\quad
  \subfloat{\includegraphics[height=.2\textwidth]{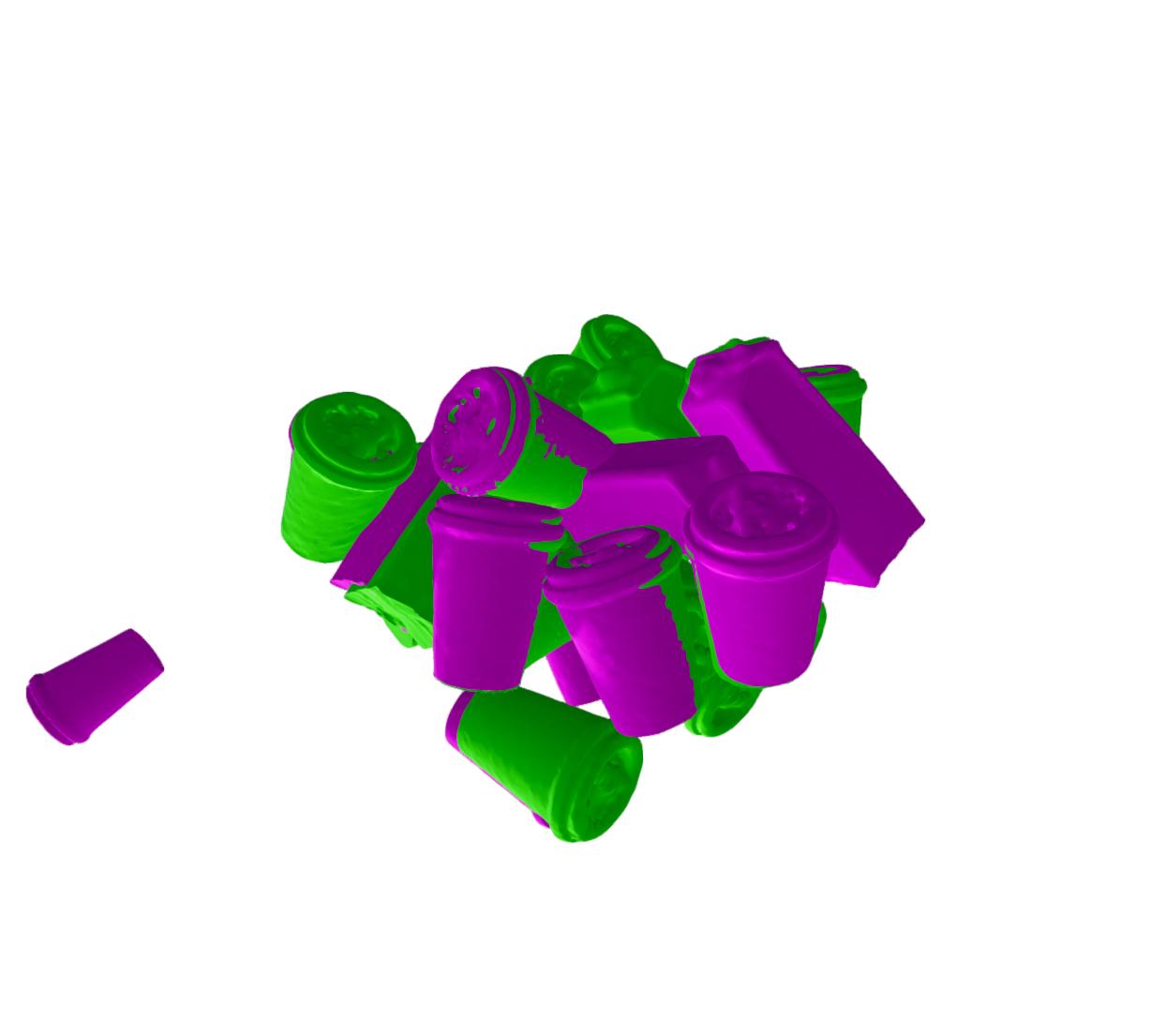}}\quad
  \subfloat{\includegraphics[height=.2\textwidth]{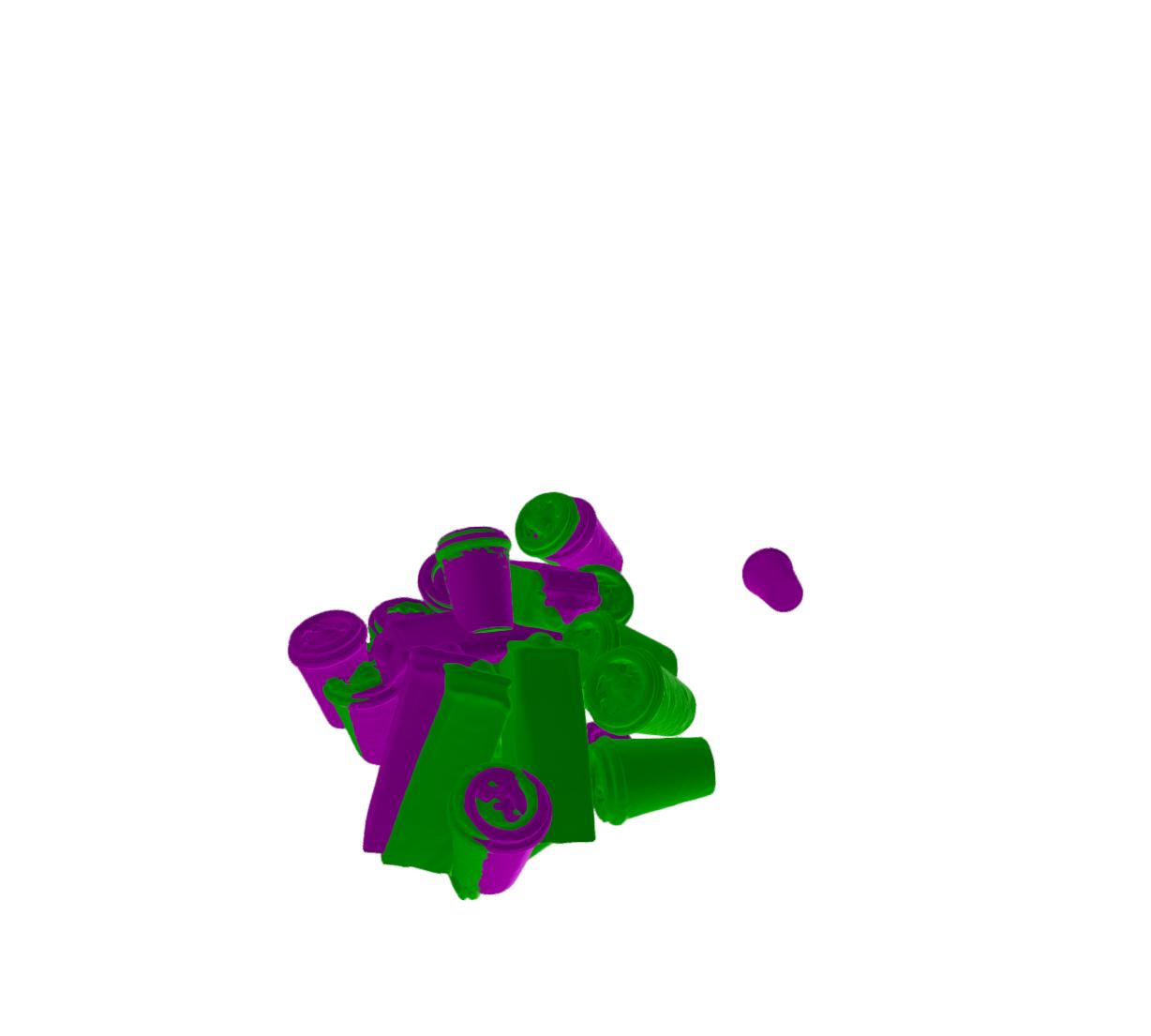}}\quad
  \subfloat{\includegraphics[height=.2\textwidth]{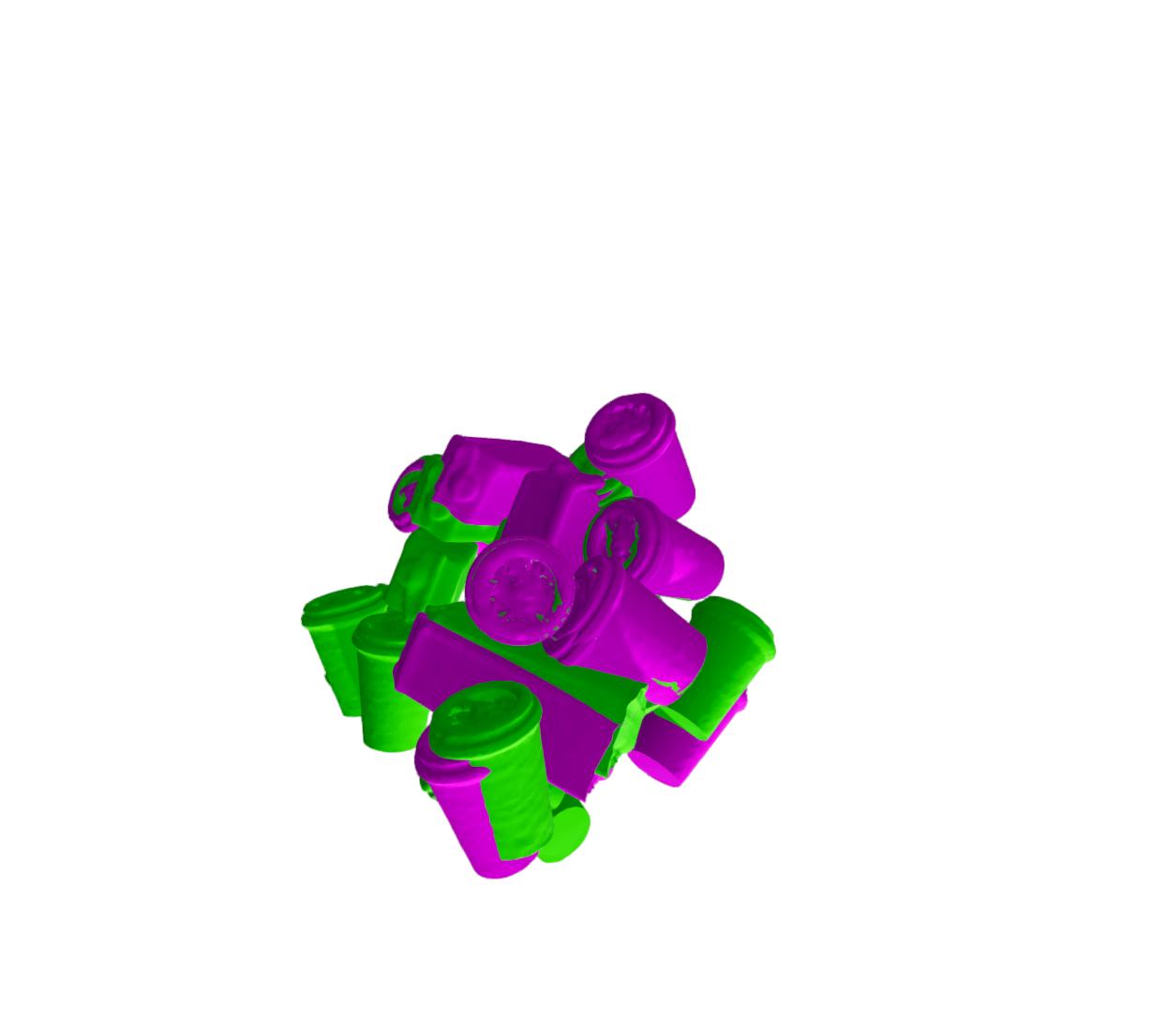}}\\
  
  \subfloat{\includegraphics[height=.2\textwidth]{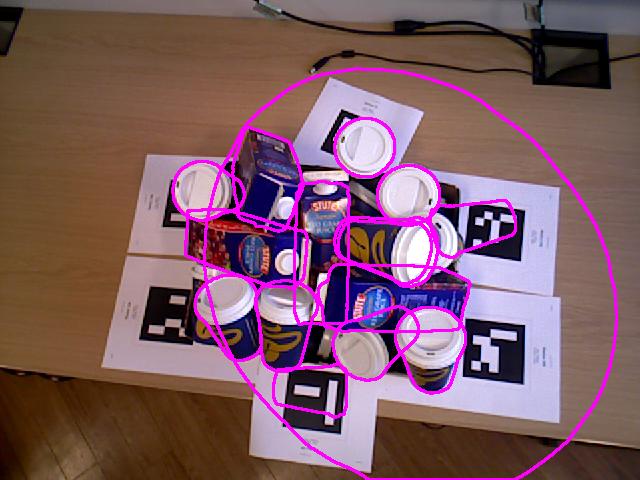}}\quad
  \subfloat{\includegraphics[height=.2\textwidth]{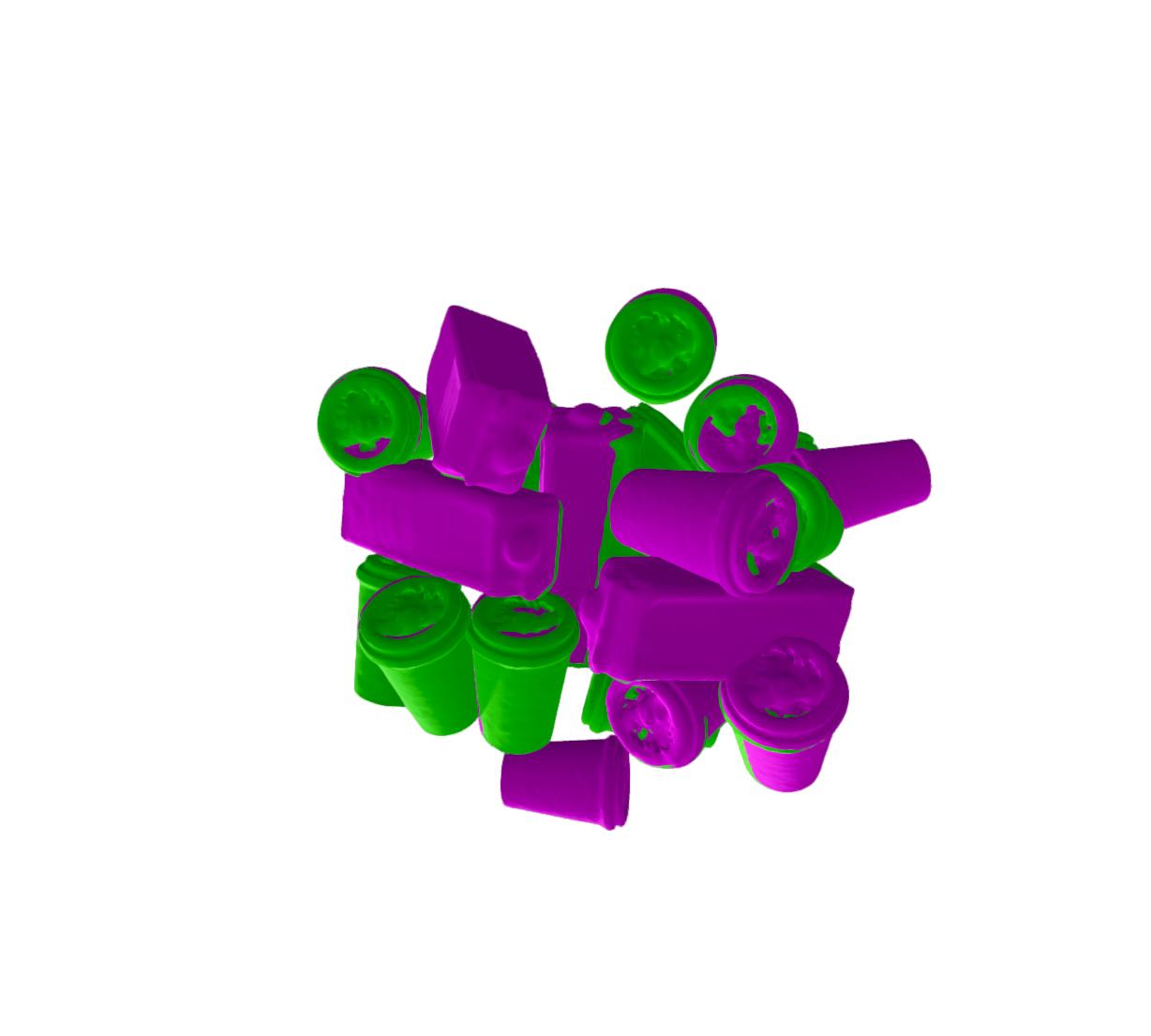}}\quad
  \subfloat{\includegraphics[height=.2\textwidth]{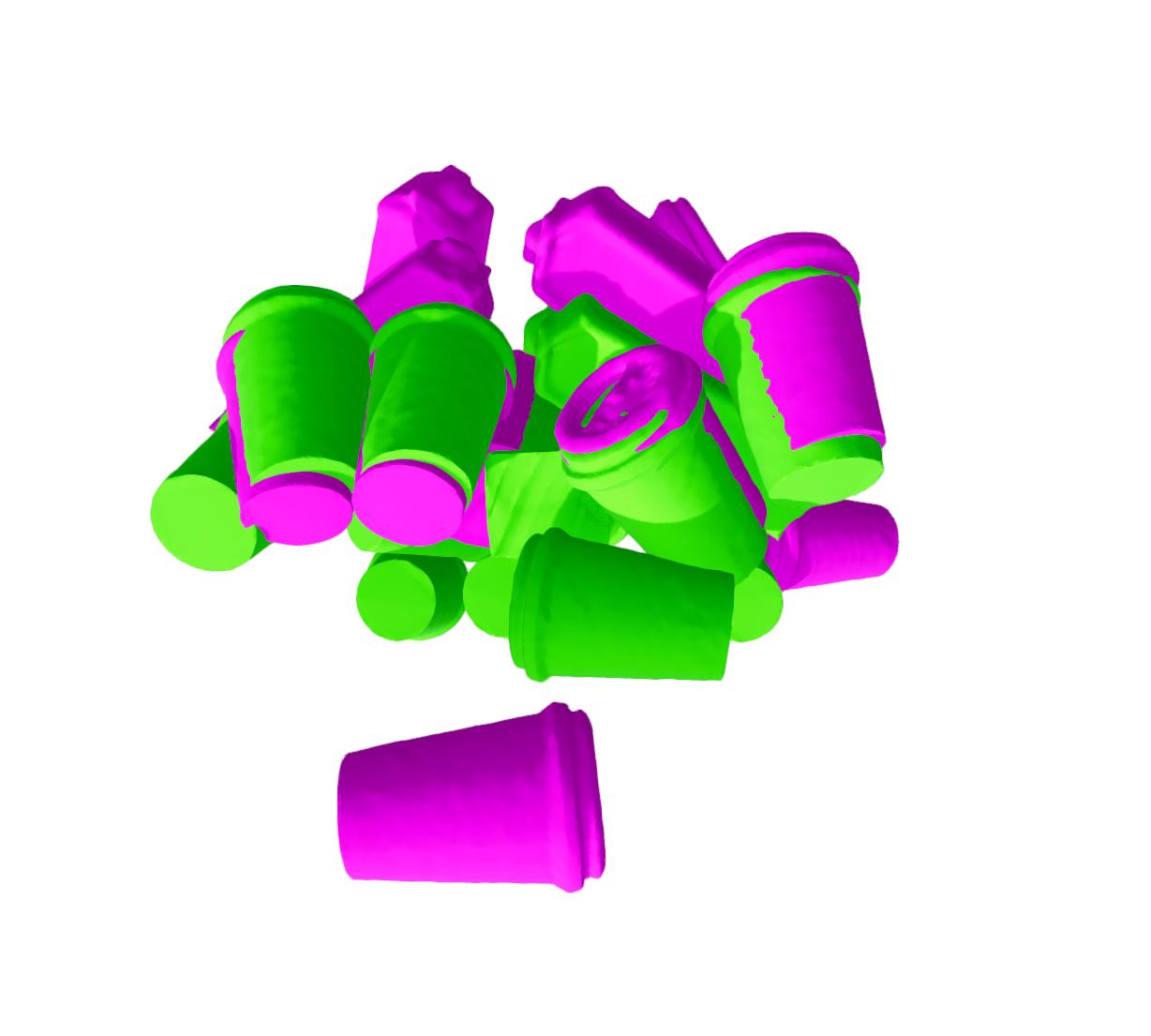}}\quad
  \subfloat{\includegraphics[height=.2\textwidth]{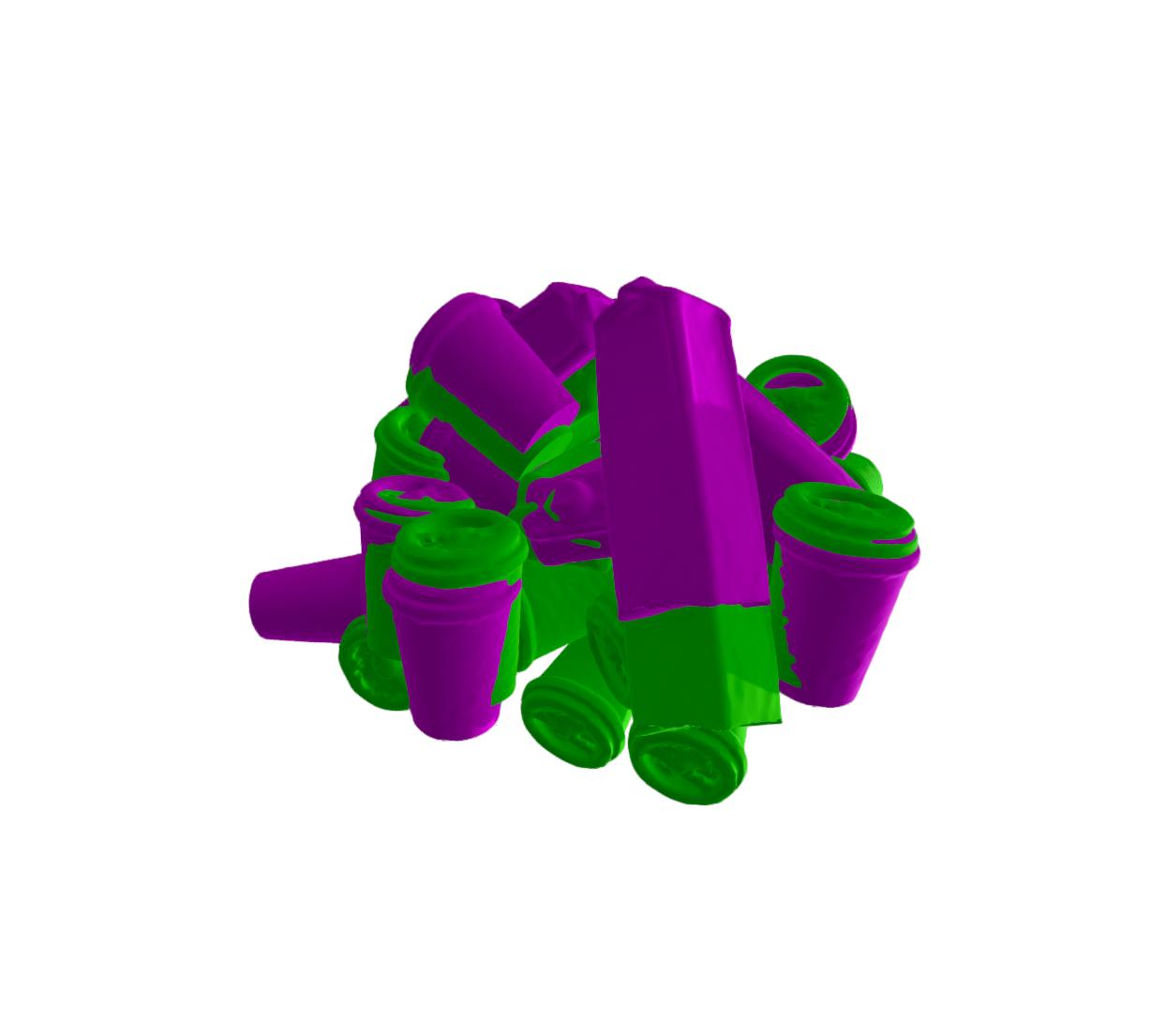}}\\
  
    \subfloat{\includegraphics[height=.2\textwidth]{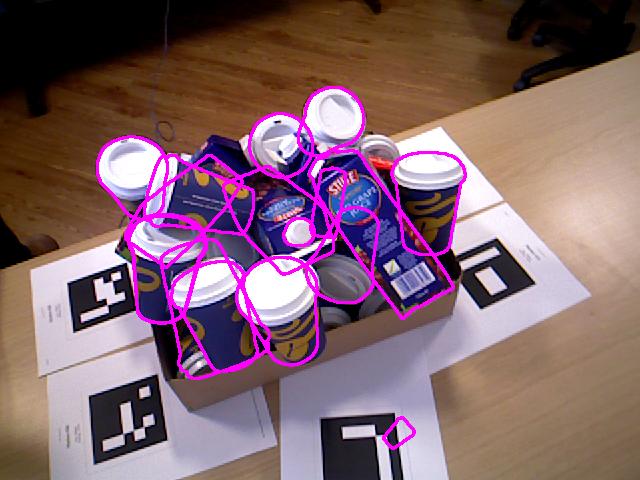}}\quad
  \subfloat{\includegraphics[height=.2\textwidth]{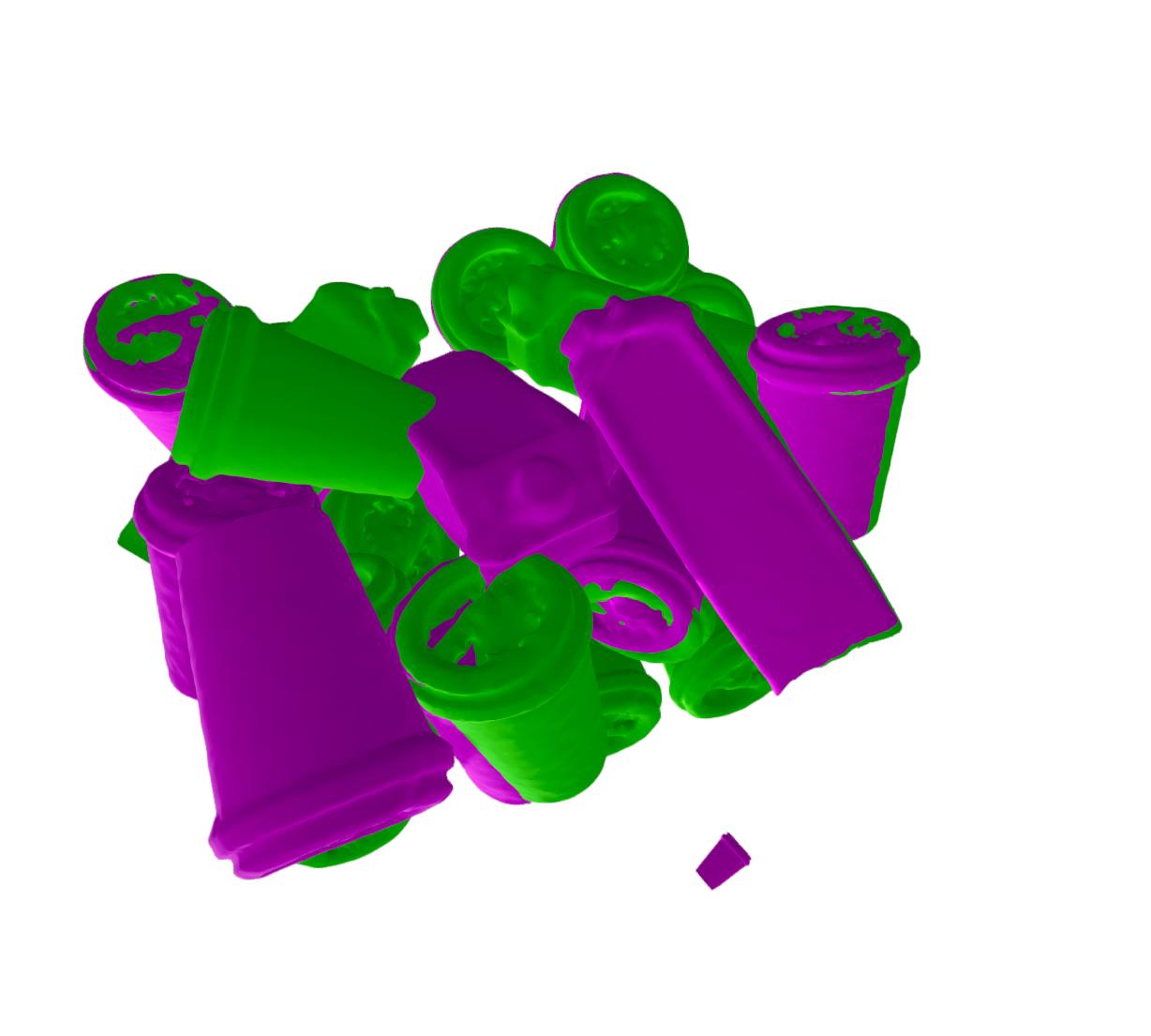}}\quad
  \subfloat{\includegraphics[height=.2\textwidth]{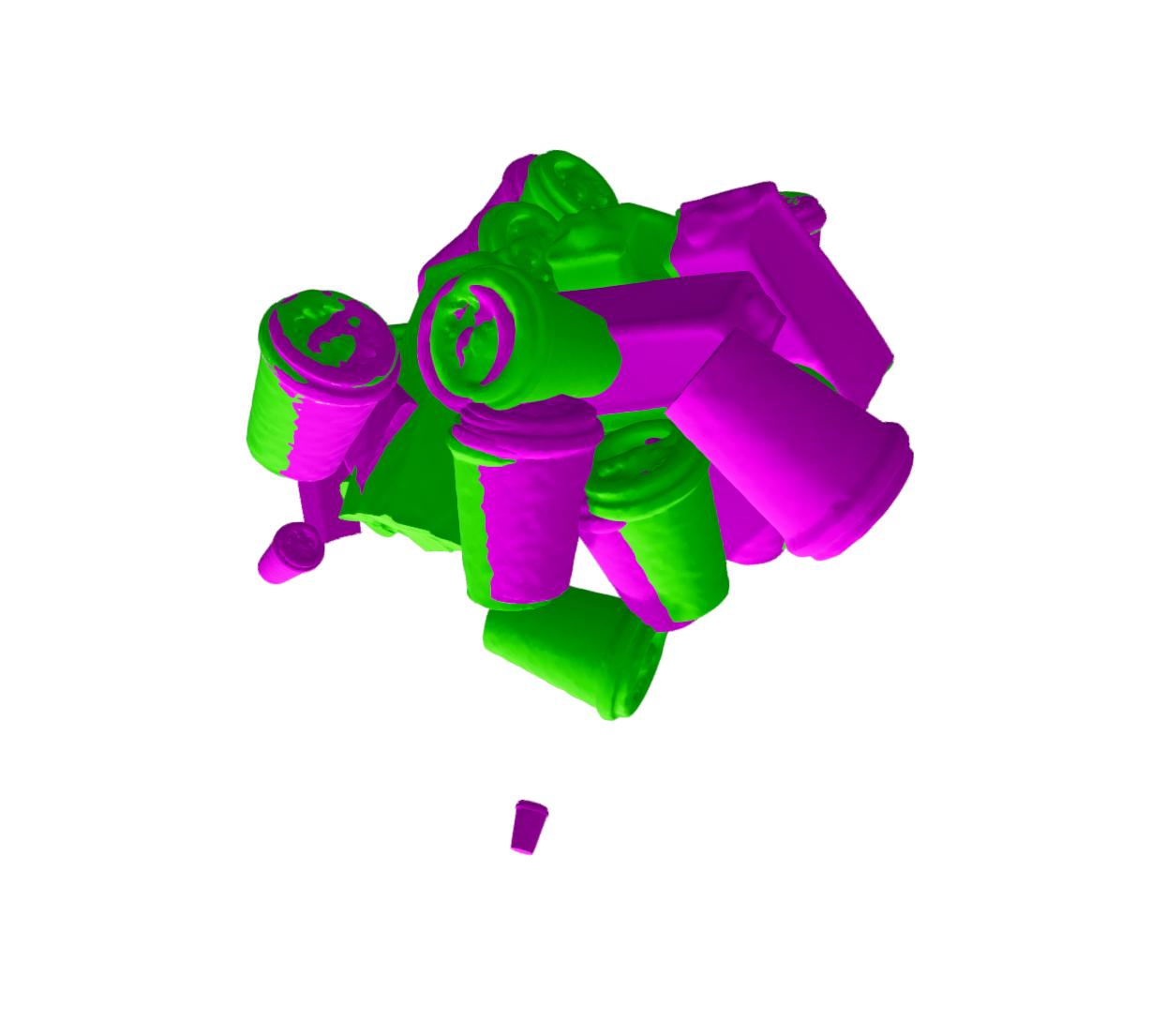}}\quad
  \subfloat{\includegraphics[height=.2\textwidth]{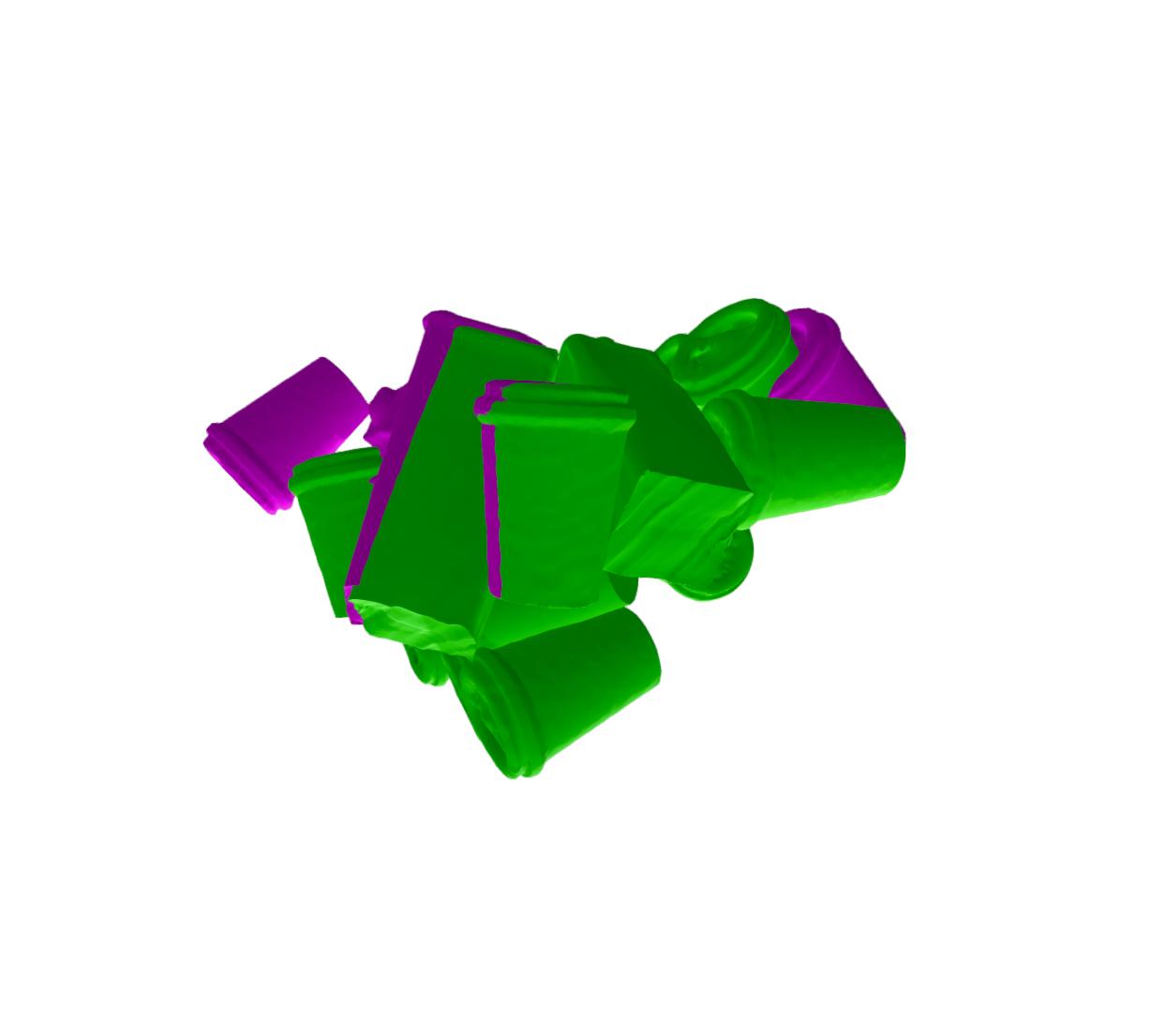}}\\
  
  \caption{IC-BIN dataset visualization with \textcolor{magenta}{estimated} (magenta) 6D pose of the meshes and \textcolor{green}{ground-truth} (green) poses. The first column shows the test image with a contour of the projection made by the predicted pose. The other three columns show the corresponding 3D view from different viewing angles. The first is taken from approximately the same viewing angle as the image was taken.}
  \label{fig:A0_ICBIN}
\end{figure*}

\begin{figure*}
  \centering
  \subfloat{\includegraphics[height=.2\textwidth]{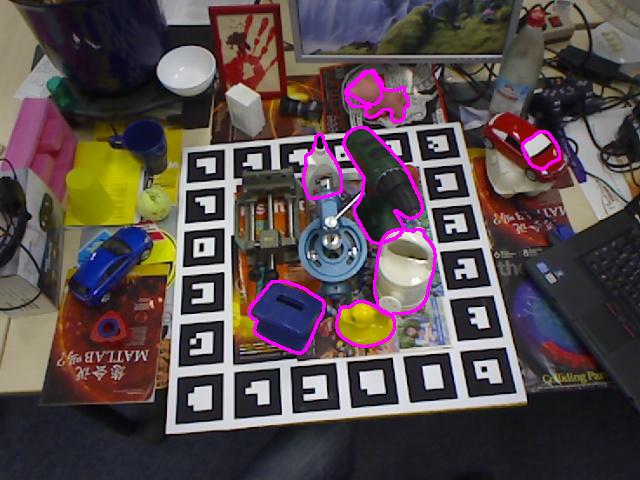}}\quad
  \subfloat{\includegraphics[height=.2\textwidth]{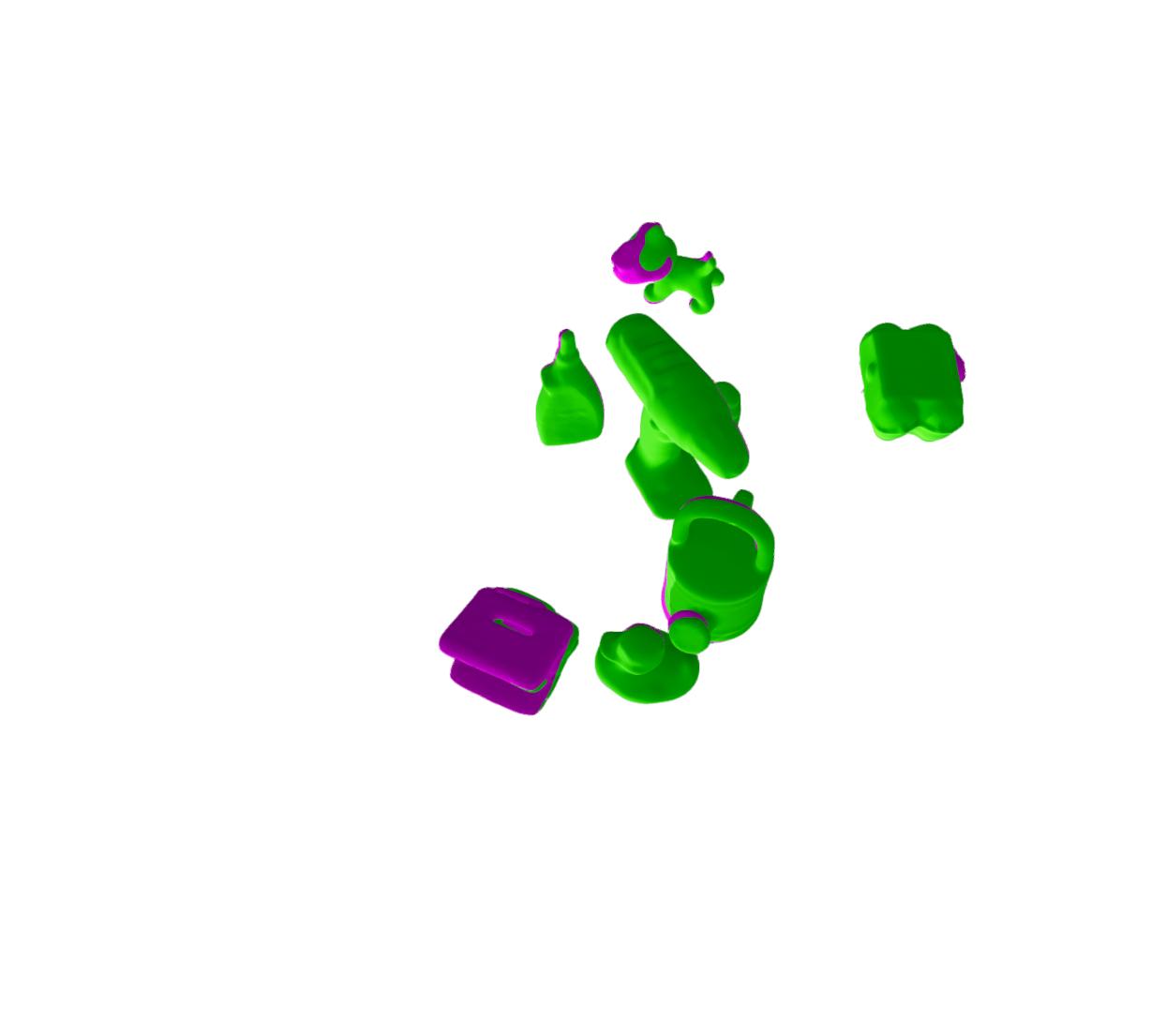}}\quad
  \subfloat{\includegraphics[height=.2\textwidth]{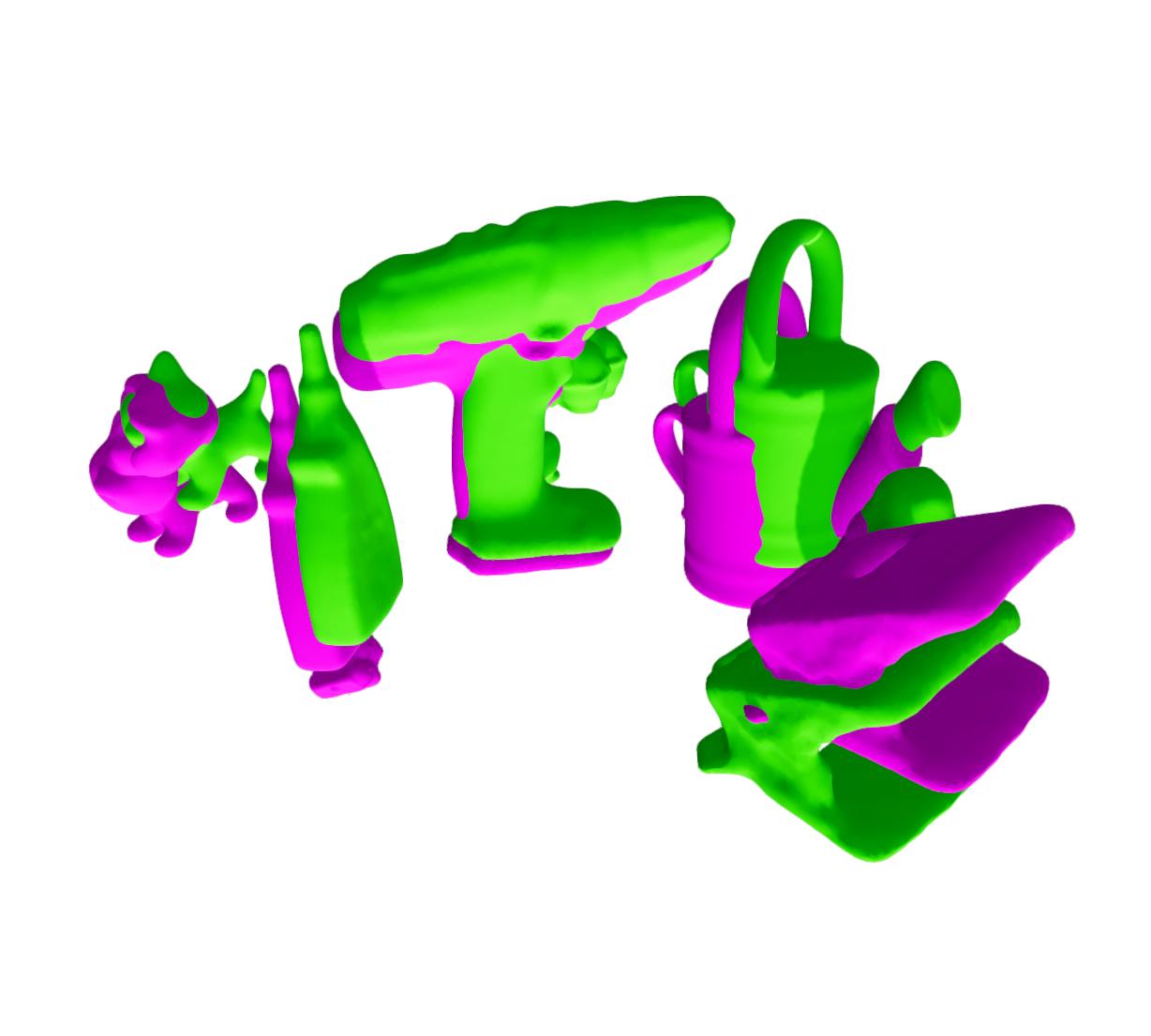}}\quad
  \subfloat{\includegraphics[height=.2\textwidth]{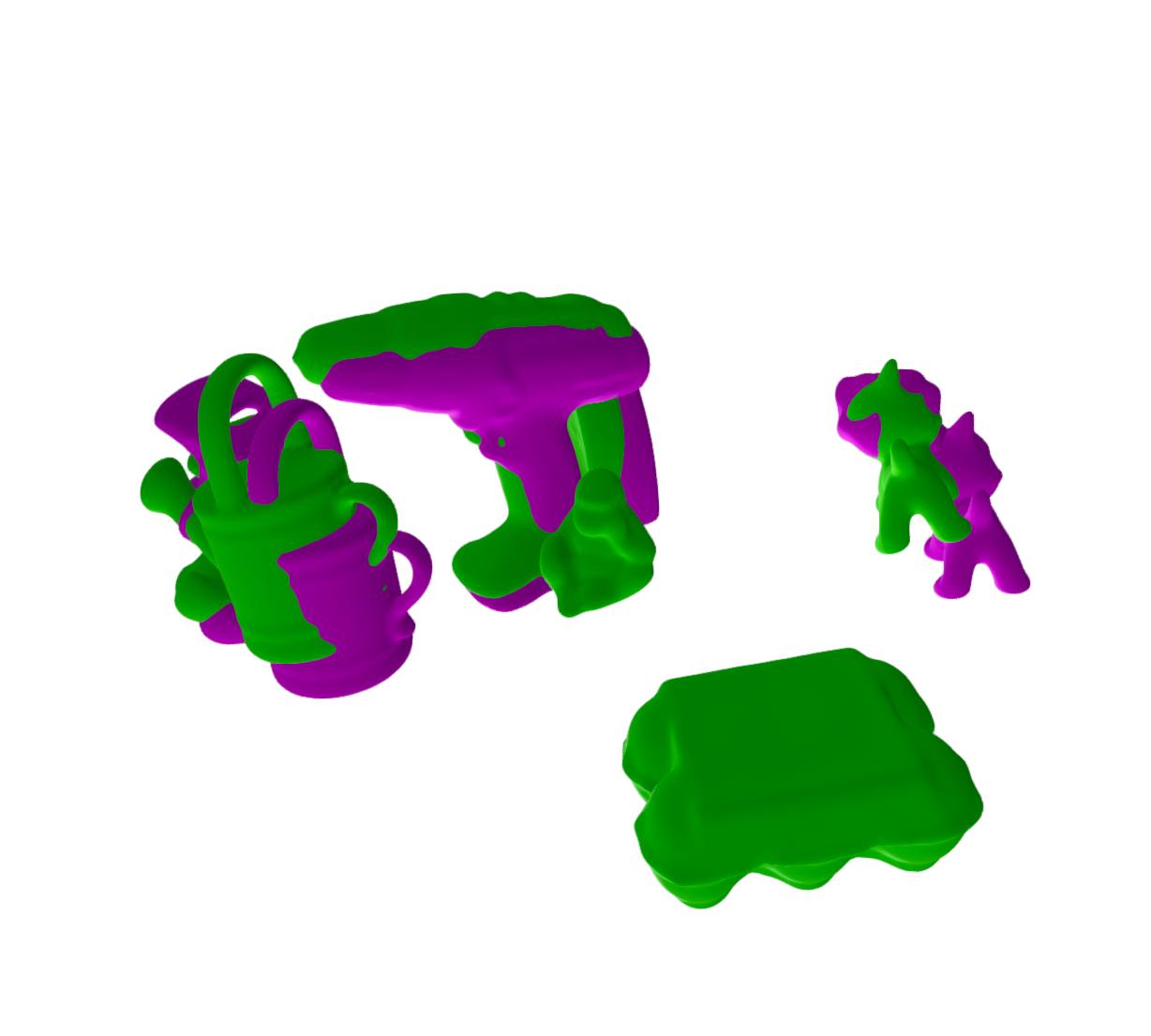}}\\
  
  \subfloat{\includegraphics[height=.2\textwidth]{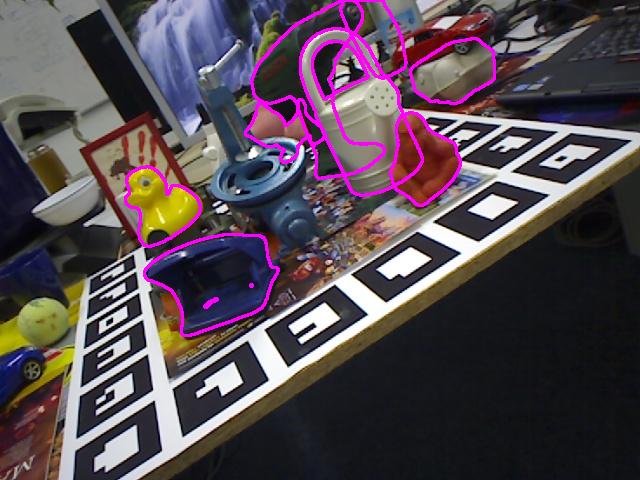}}\quad
  \subfloat{\includegraphics[height=.2\textwidth]{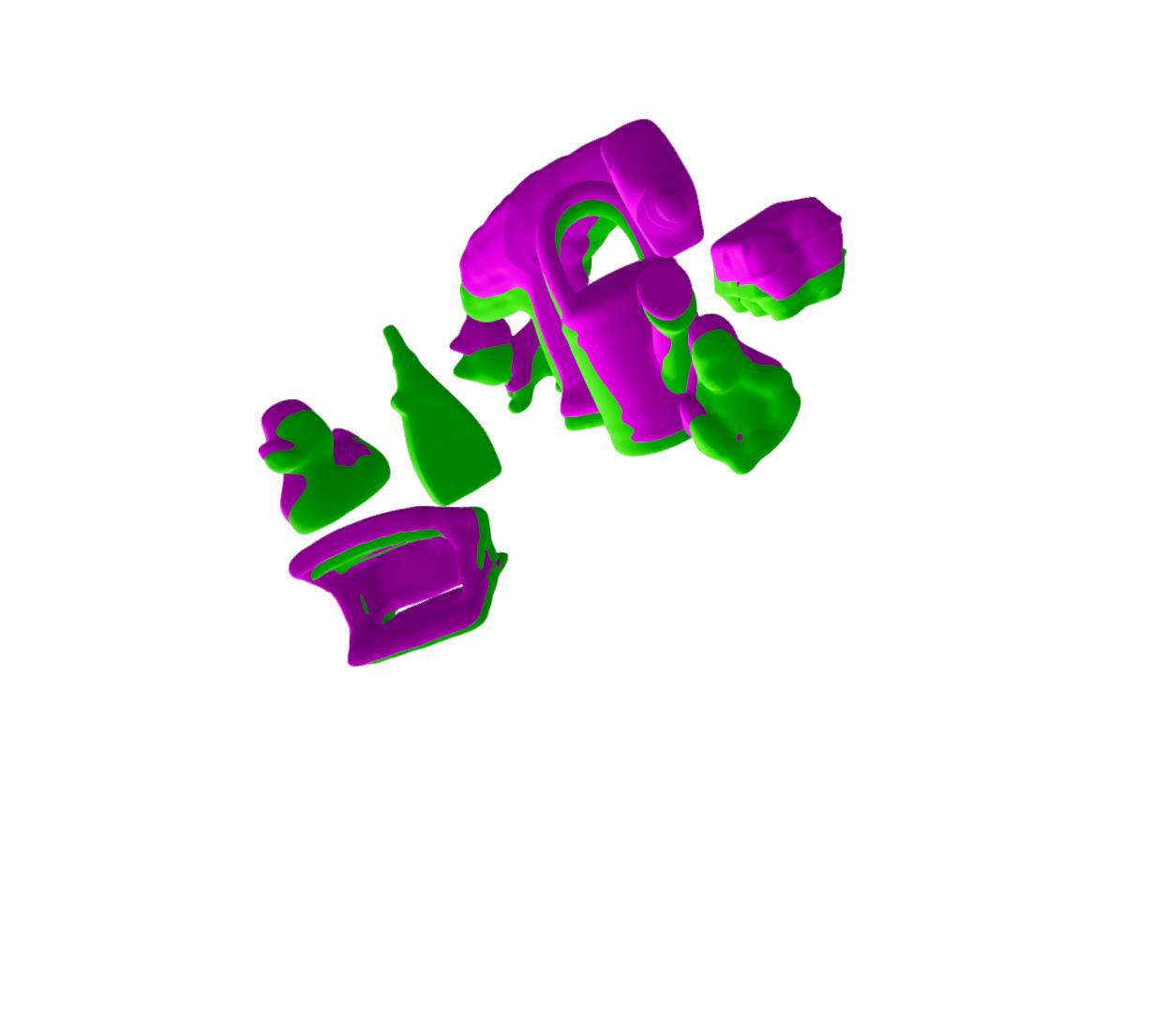}}\quad
  \subfloat{\includegraphics[height=.2\textwidth]{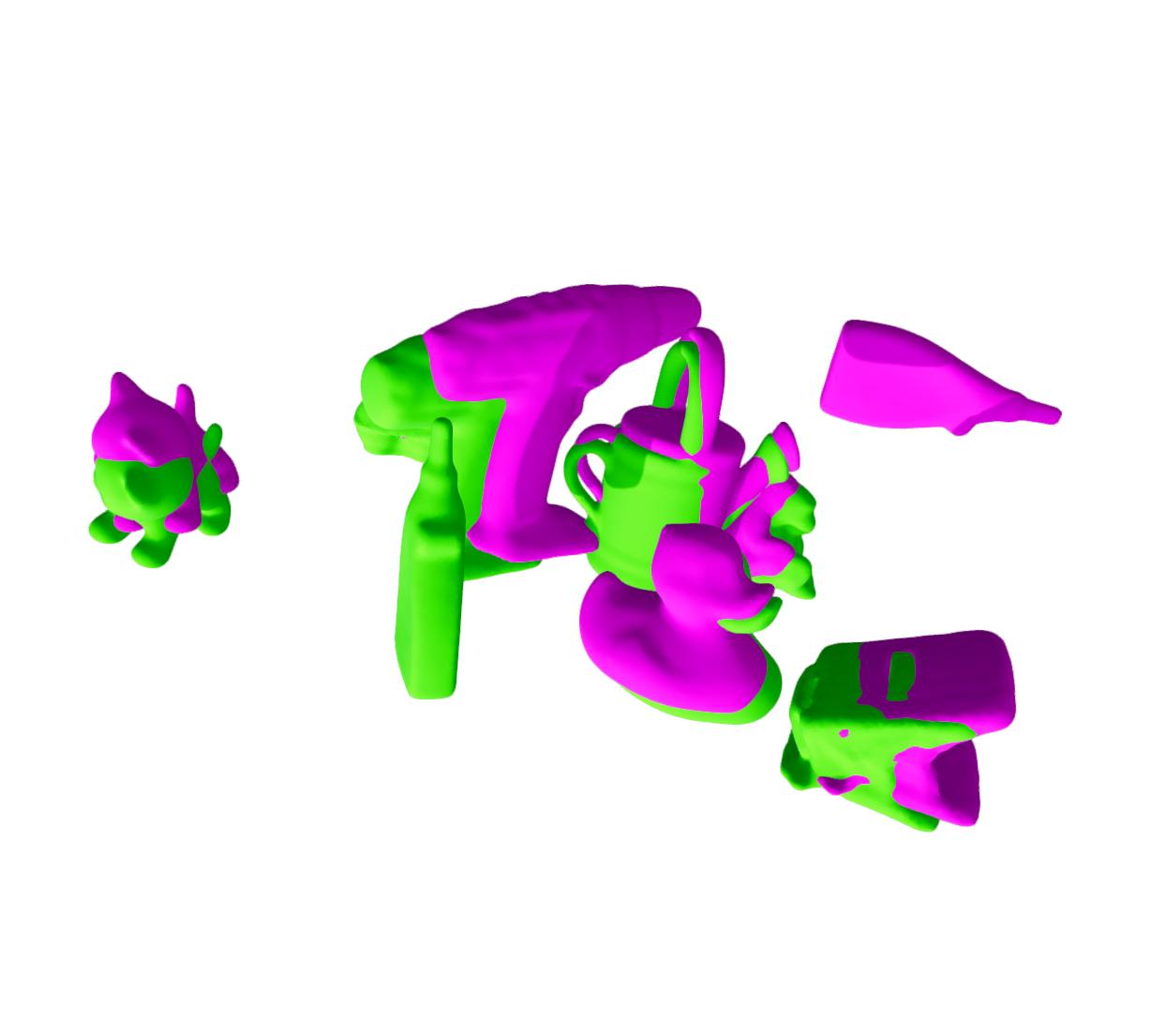}}\quad
  \subfloat{\includegraphics[height=.2\textwidth]{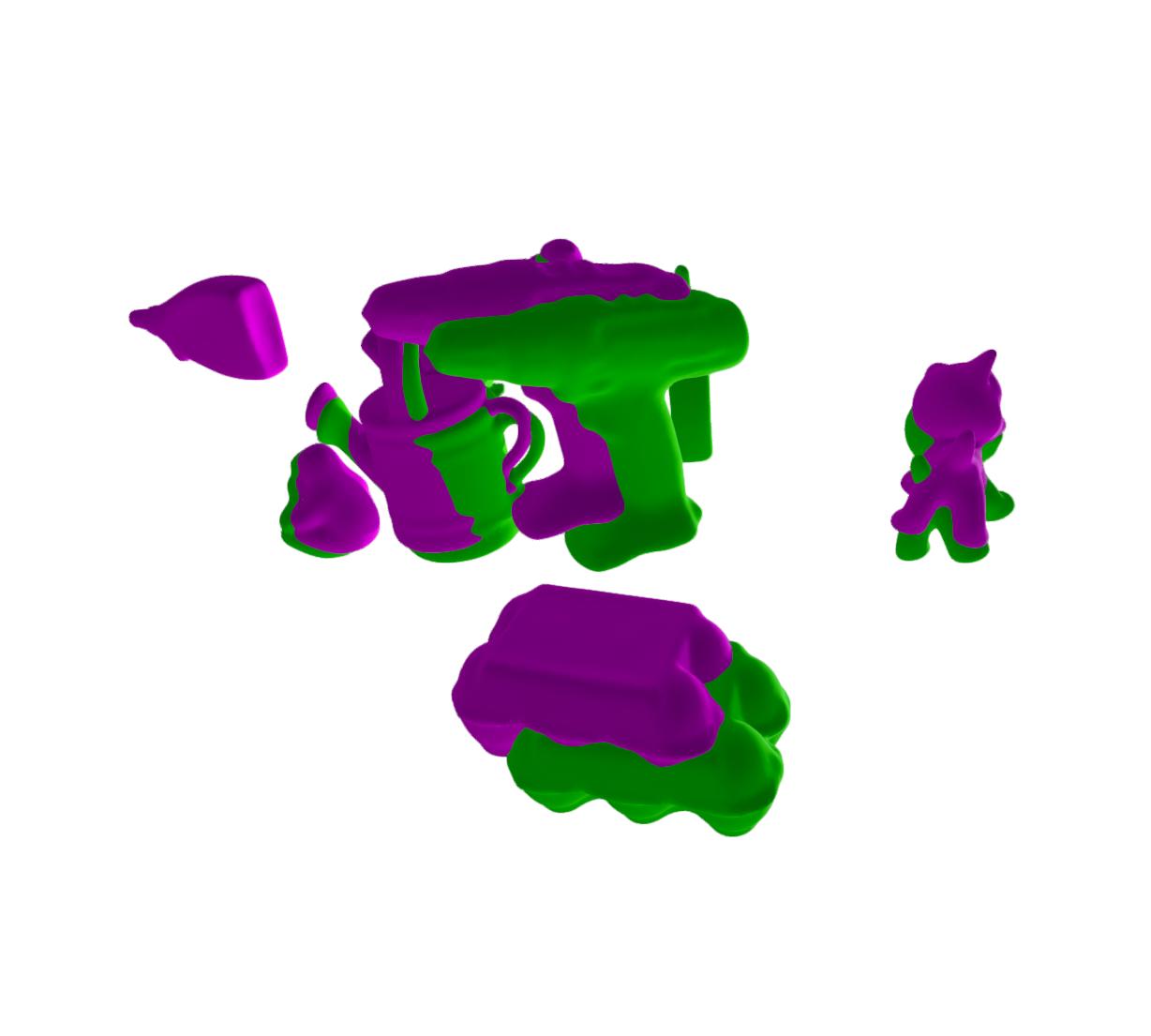}}\\
  
  \subfloat{\includegraphics[height=.2\textwidth]{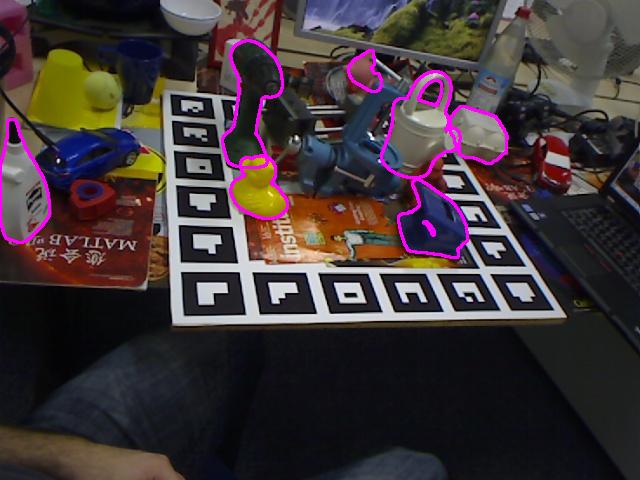}}\quad
  \subfloat{\includegraphics[height=.2\textwidth]{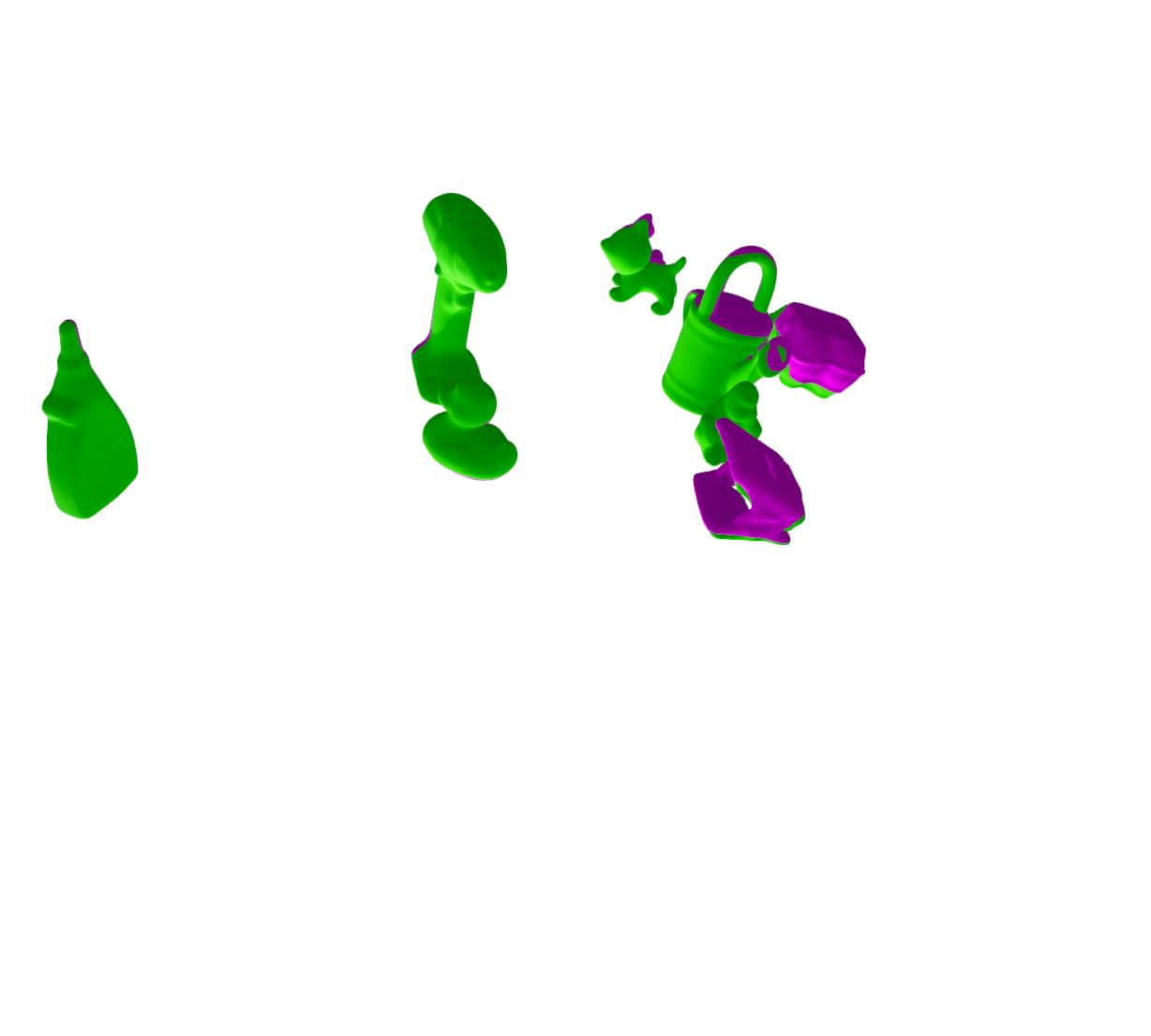}}\quad
  \subfloat{\includegraphics[height=.2\textwidth]{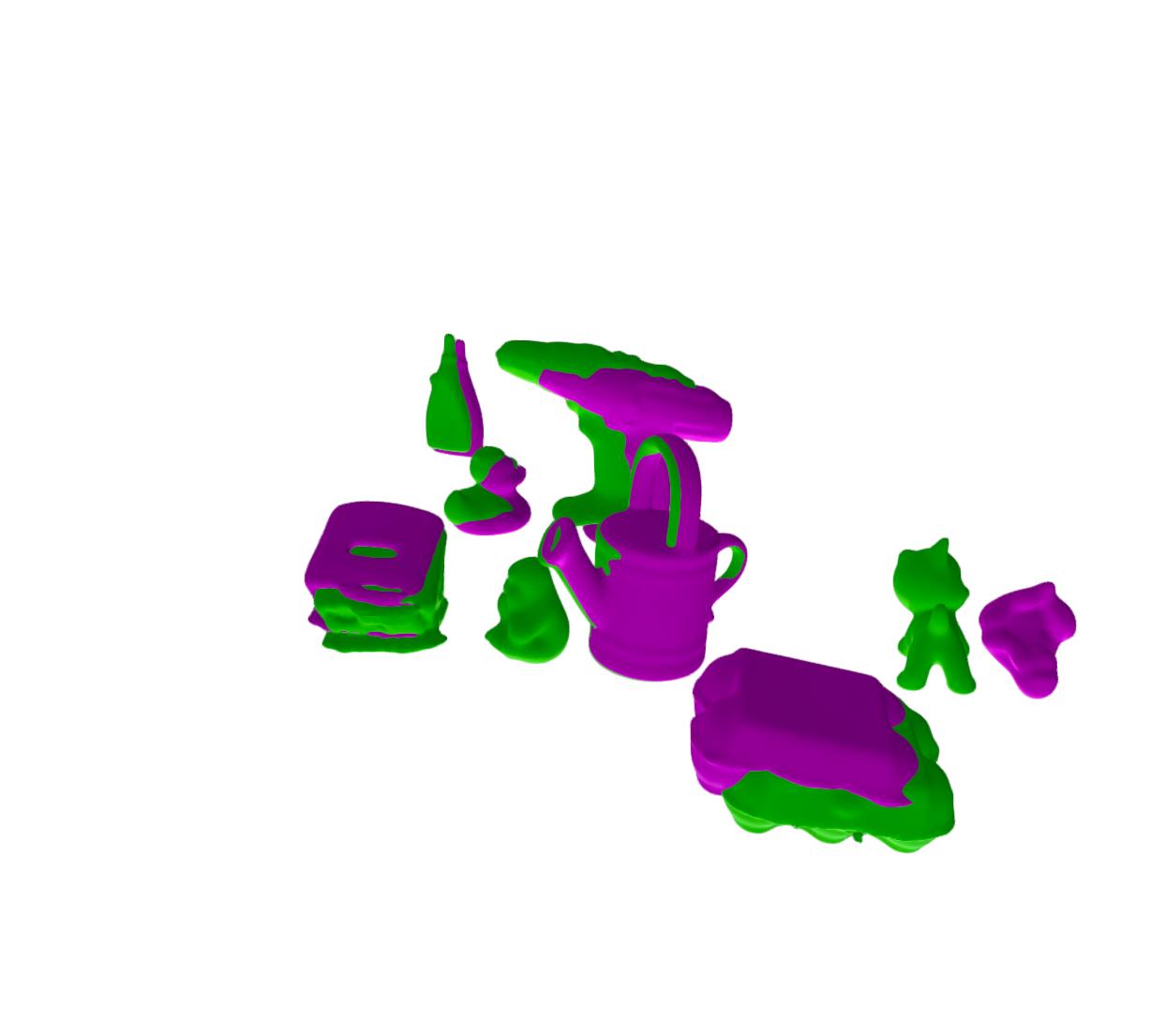}}\quad
  \subfloat{\includegraphics[height=.2\textwidth]{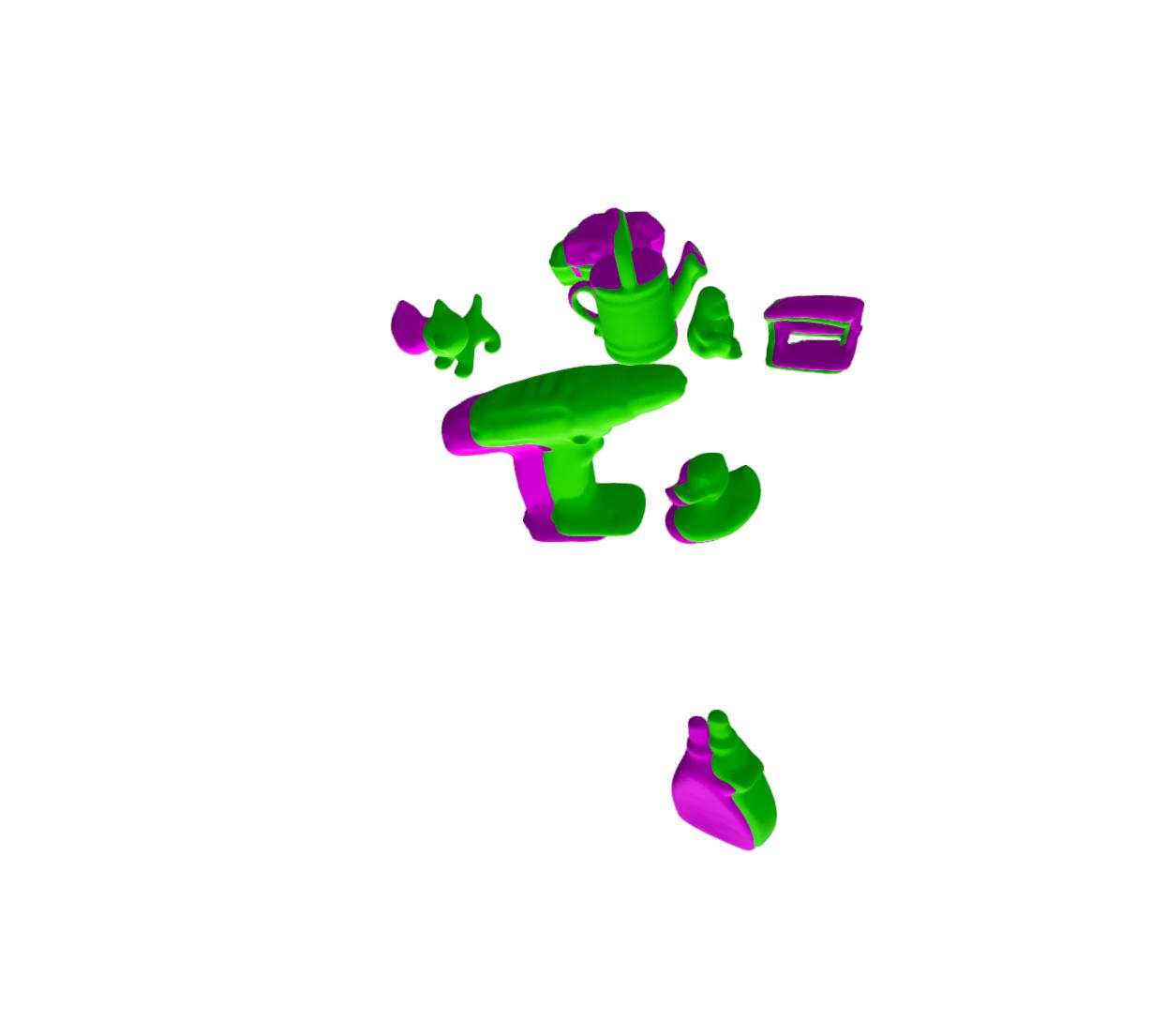}}\\
  
  \subfloat{\includegraphics[height=.2\textwidth]{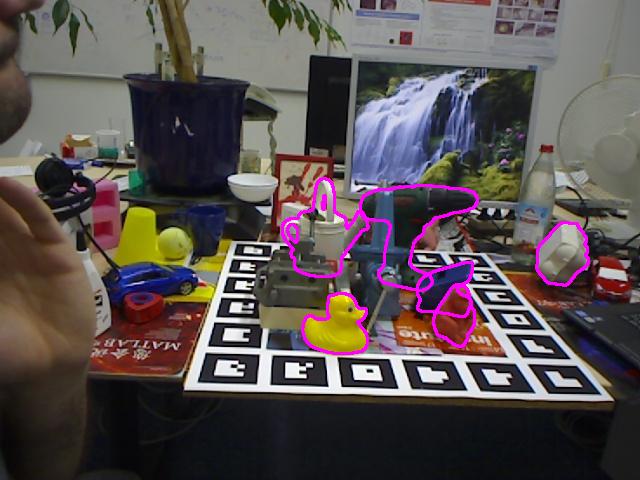}}\quad
  \subfloat{\includegraphics[height=.2\textwidth]{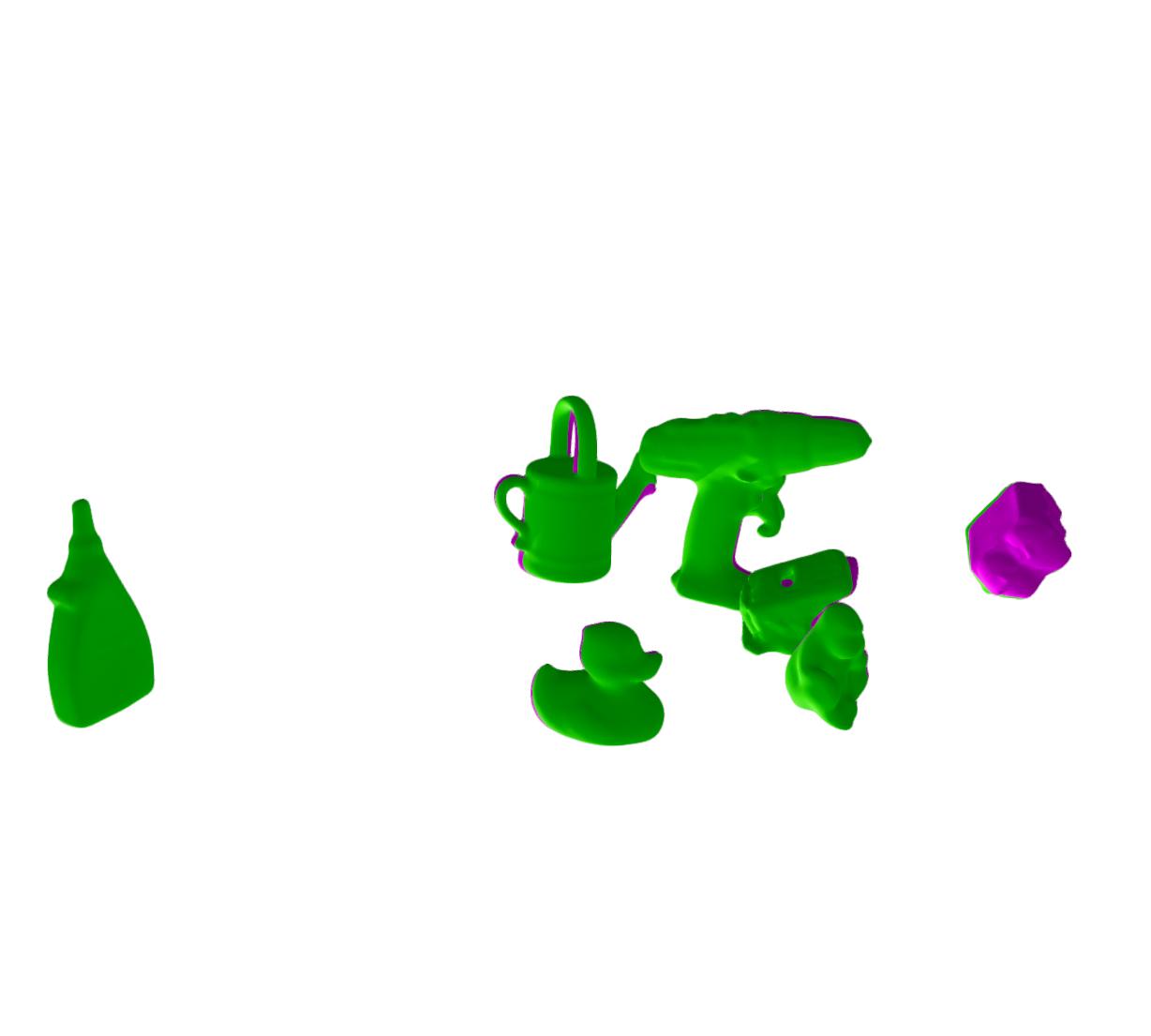}}\quad
  \subfloat{\includegraphics[height=.2\textwidth]{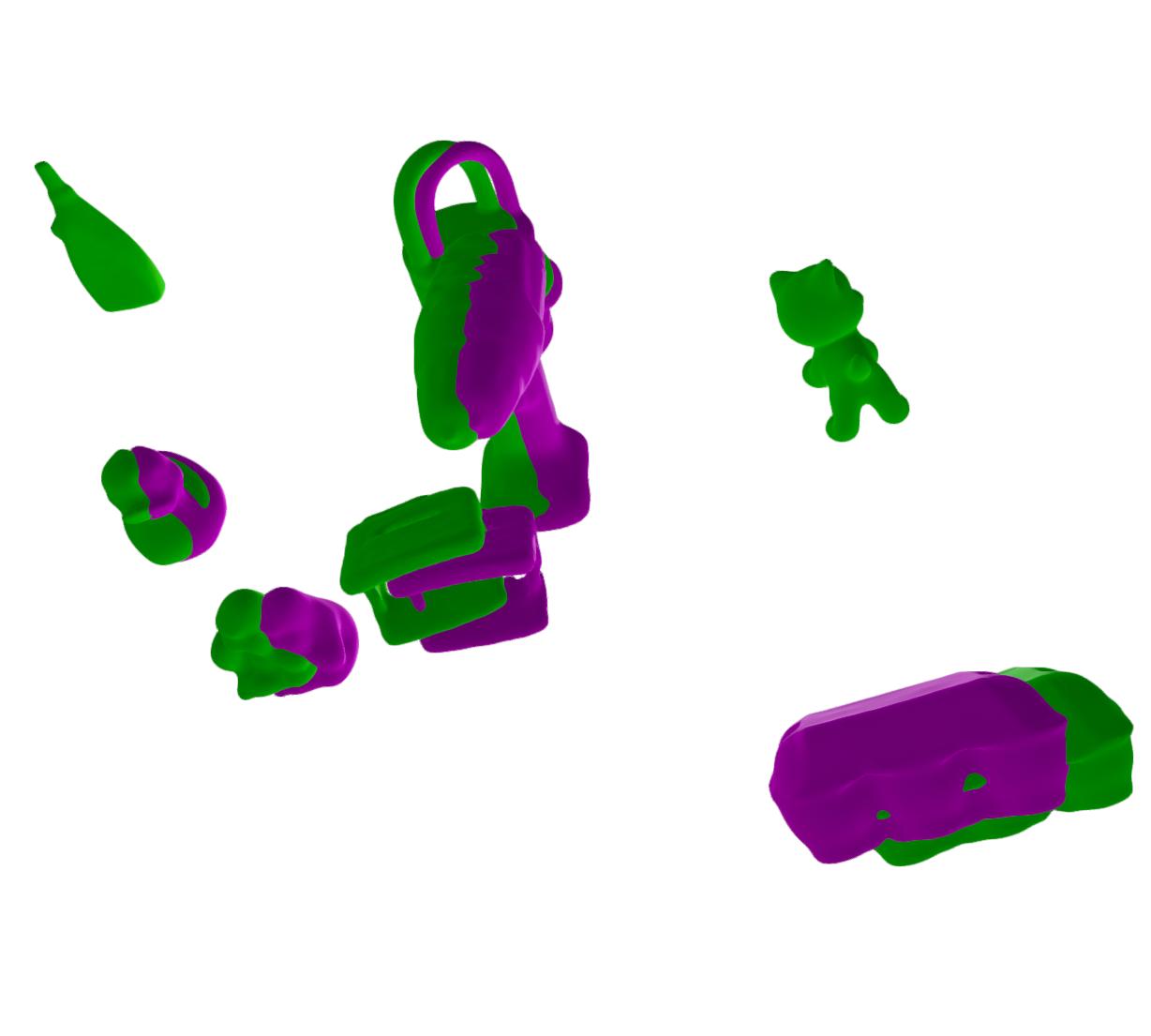}}\quad
  \subfloat{\includegraphics[height=.2\textwidth]{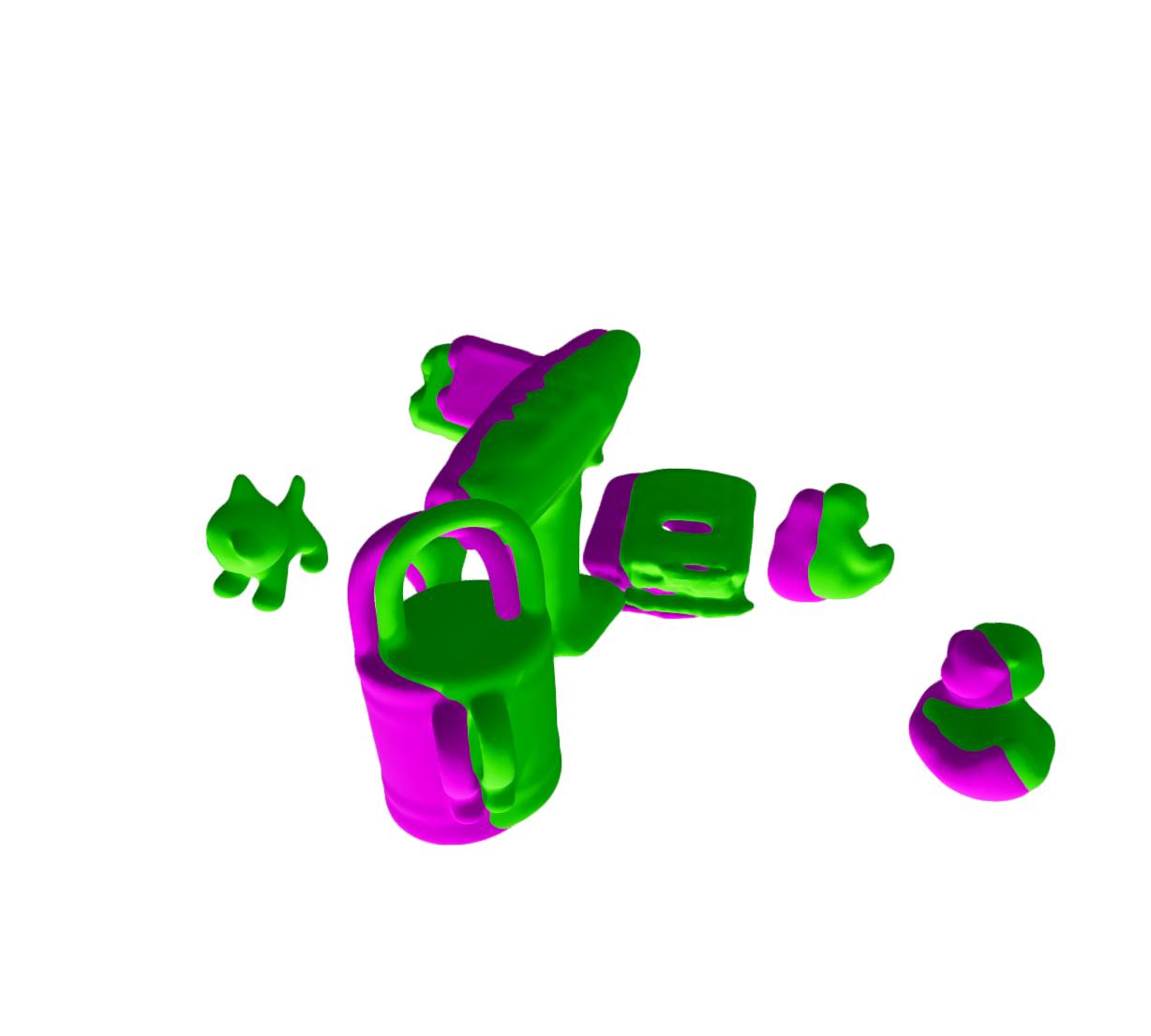}}\\
  
  \subfloat{\includegraphics[height=.2\textwidth]{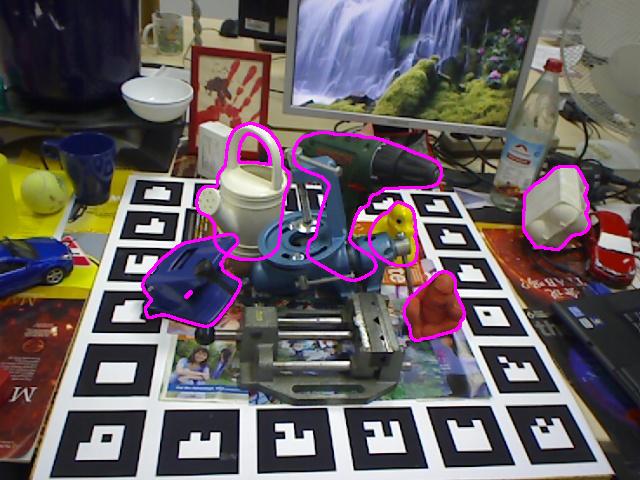}}\quad
  \subfloat{\includegraphics[height=.2\textwidth]{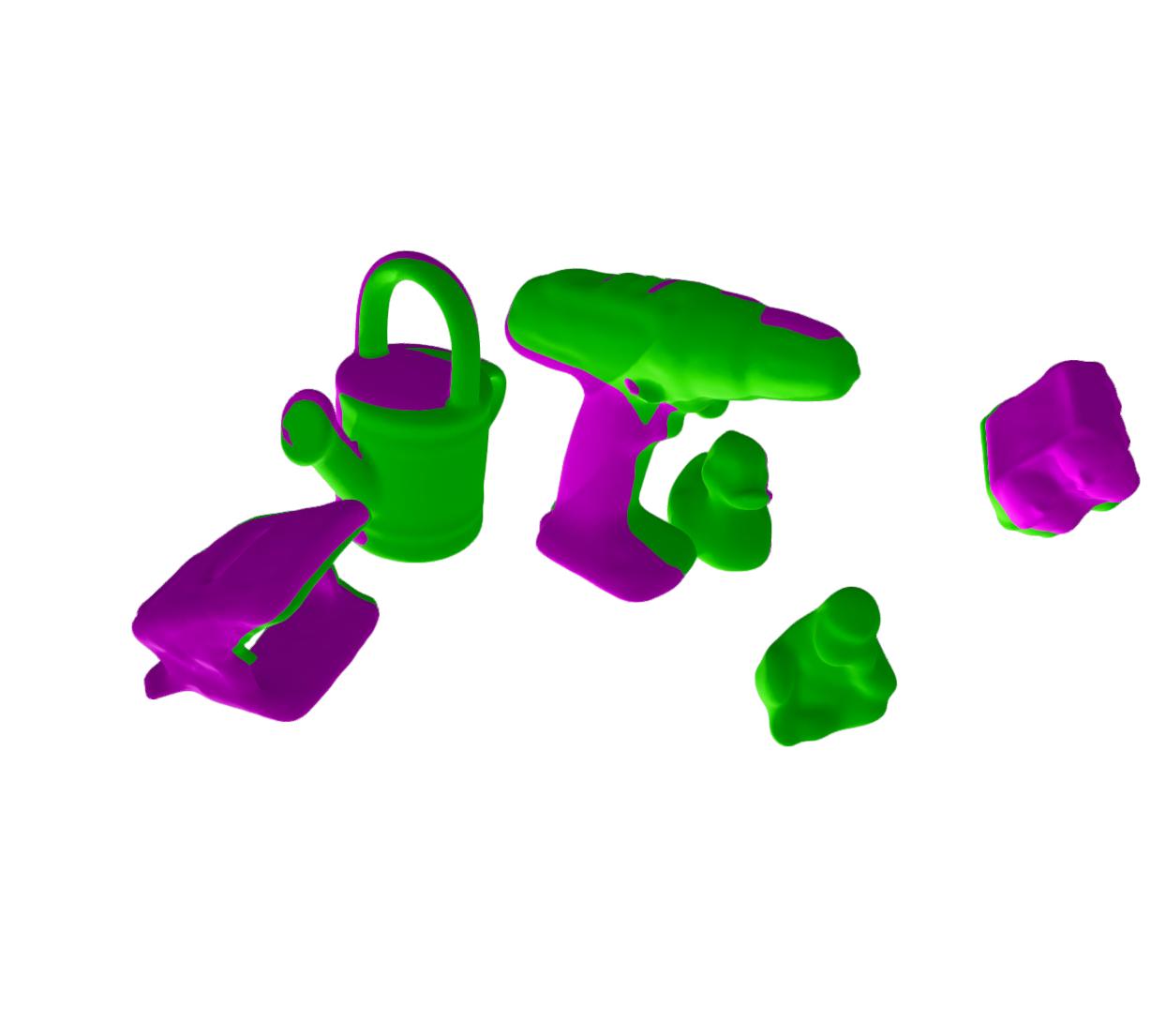}}\quad
  \subfloat{\includegraphics[height=.2\textwidth]{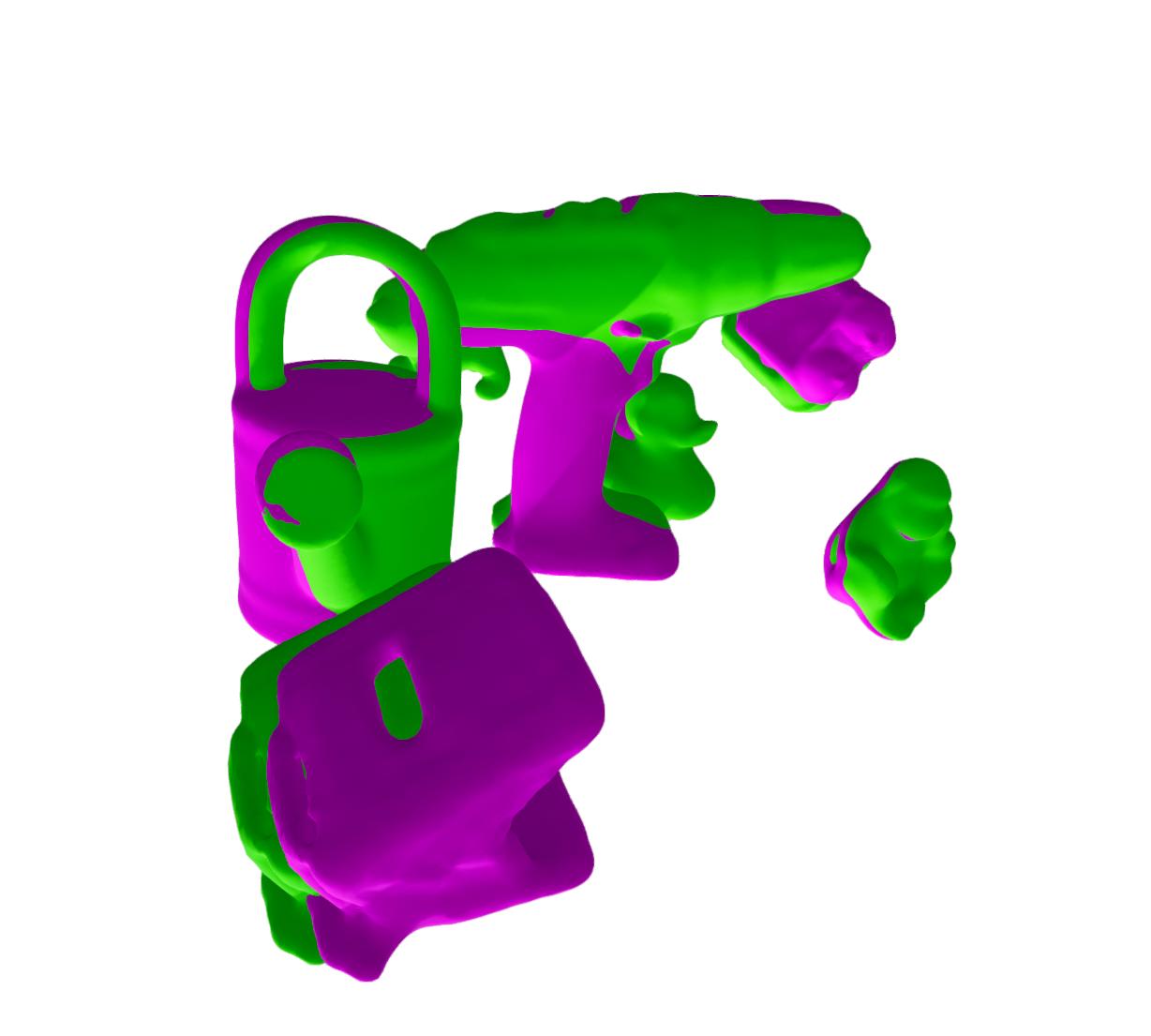}}\quad
  \subfloat{\includegraphics[height=.2\textwidth]{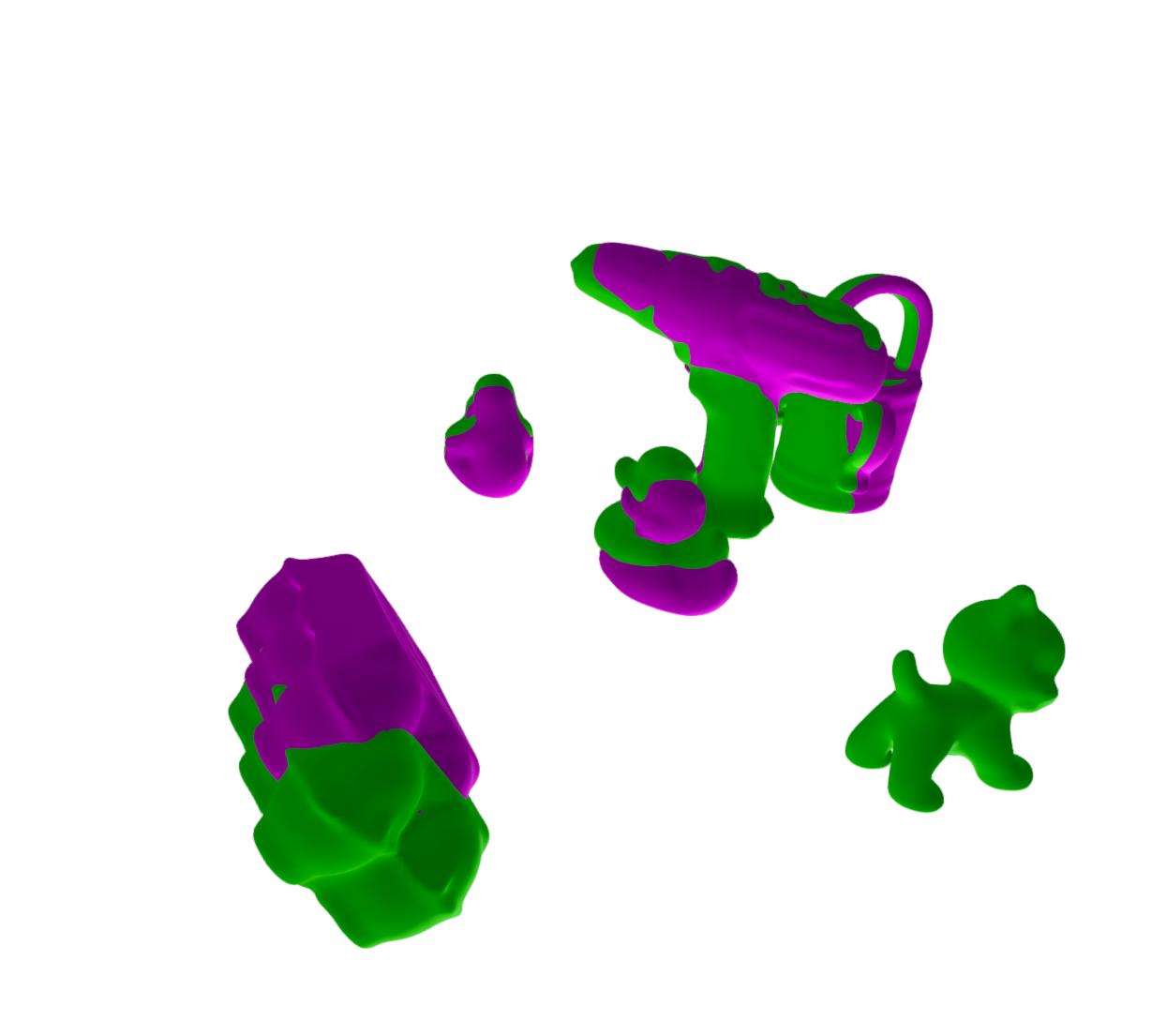}}\\
  
  \caption{LM-O dataset visualization with \textcolor{magenta}{estimated} (magenta) 6D pose of the meshes and \textcolor{green}{ground-truth} (green) poses.. The first column shows the test image with a contour of the projection made by the predicted pose. The other three columns show the corresponding 3D view from different viewing angles. The first is taken from approximately the same viewing angle as the image was taken.}
  \label{fig:A0_LMO}
\end{figure*}

\begin{figure*}
  \centering
  \subfloat{\includegraphics[height=.2\textwidth]{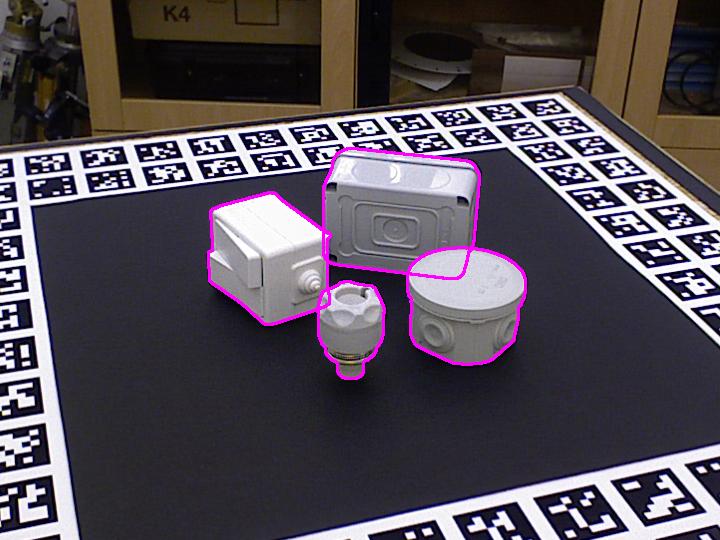}}\quad
  \subfloat{\includegraphics[height=.2\textwidth]{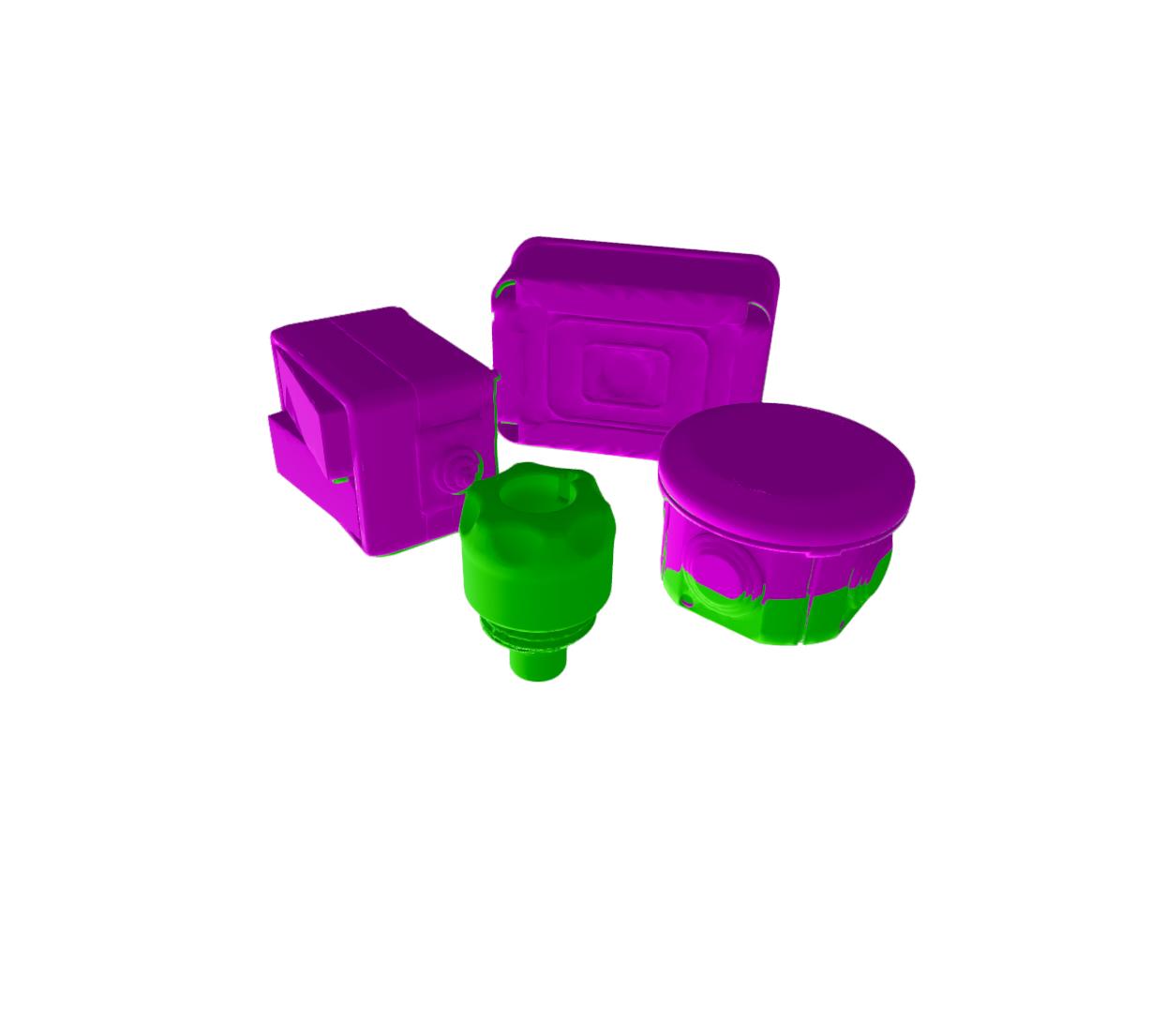}}\quad
  \subfloat{\includegraphics[height=.2\textwidth]{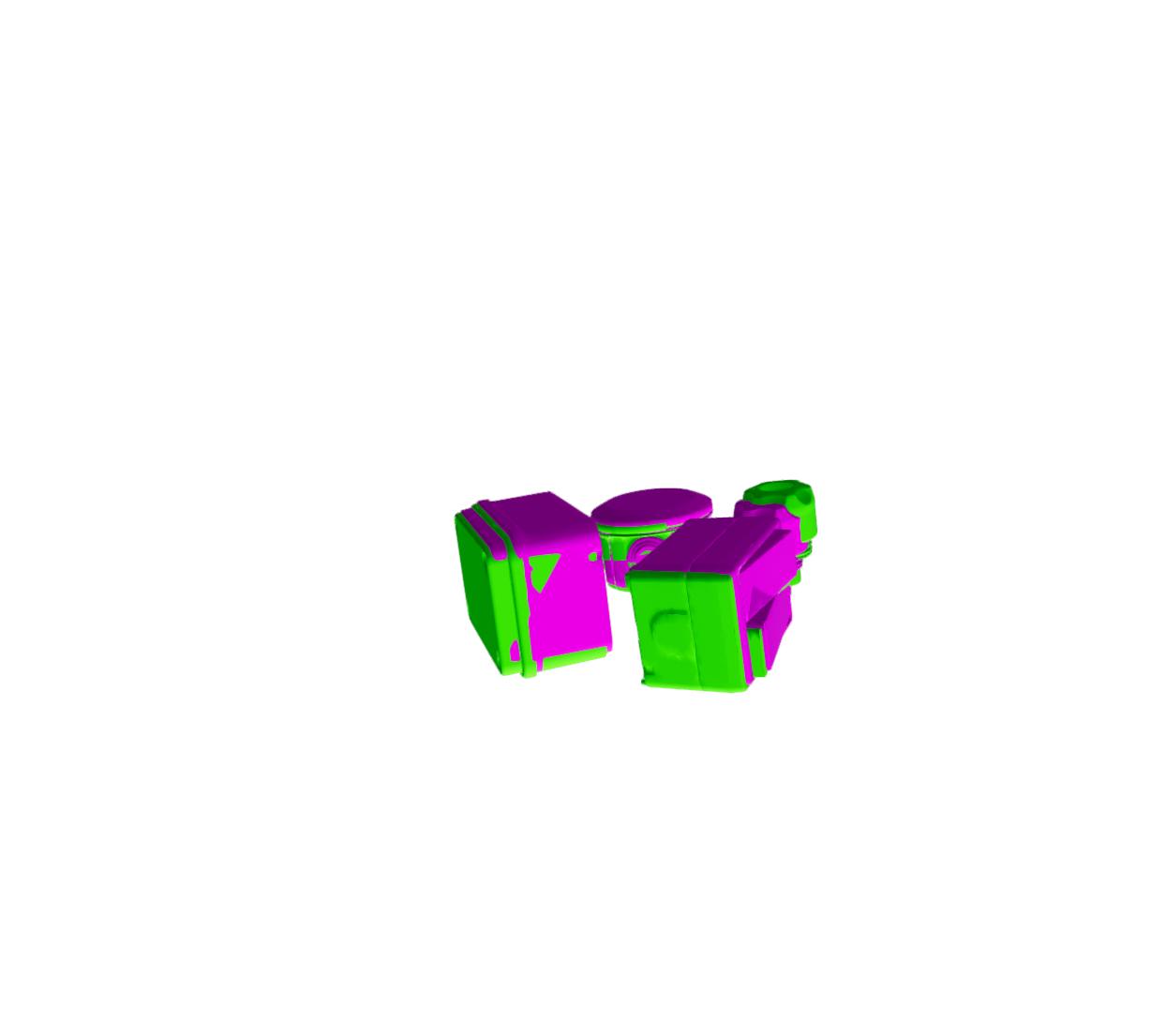}}\quad
  \subfloat{\includegraphics[height=.2\textwidth]{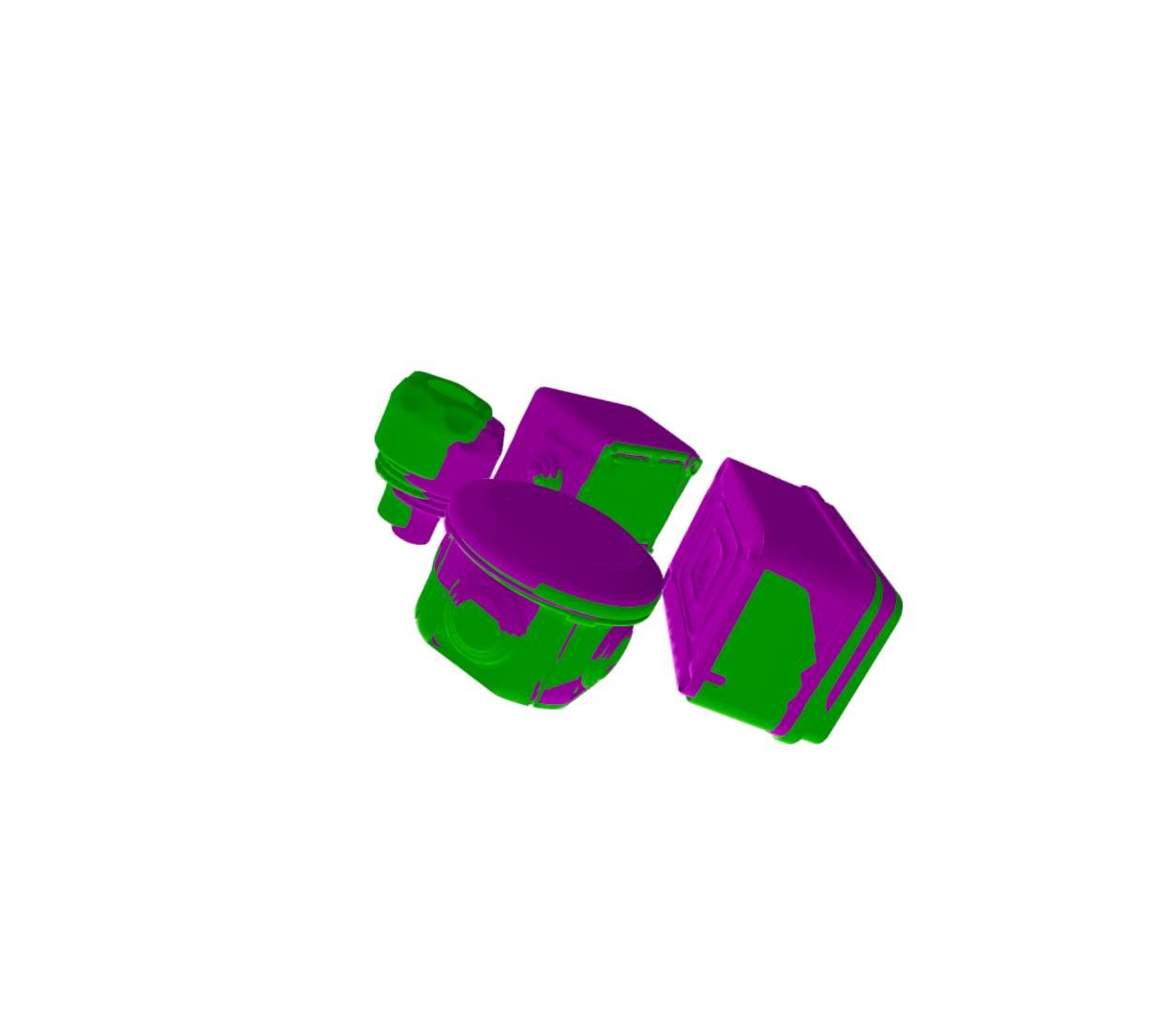}}\\
  
  \subfloat{\includegraphics[height=.2\textwidth]{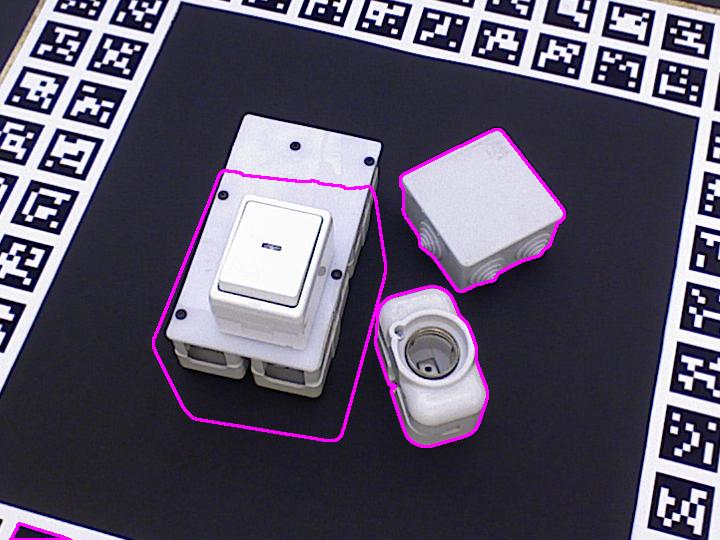}}\quad
  \subfloat{\includegraphics[height=.2\textwidth]{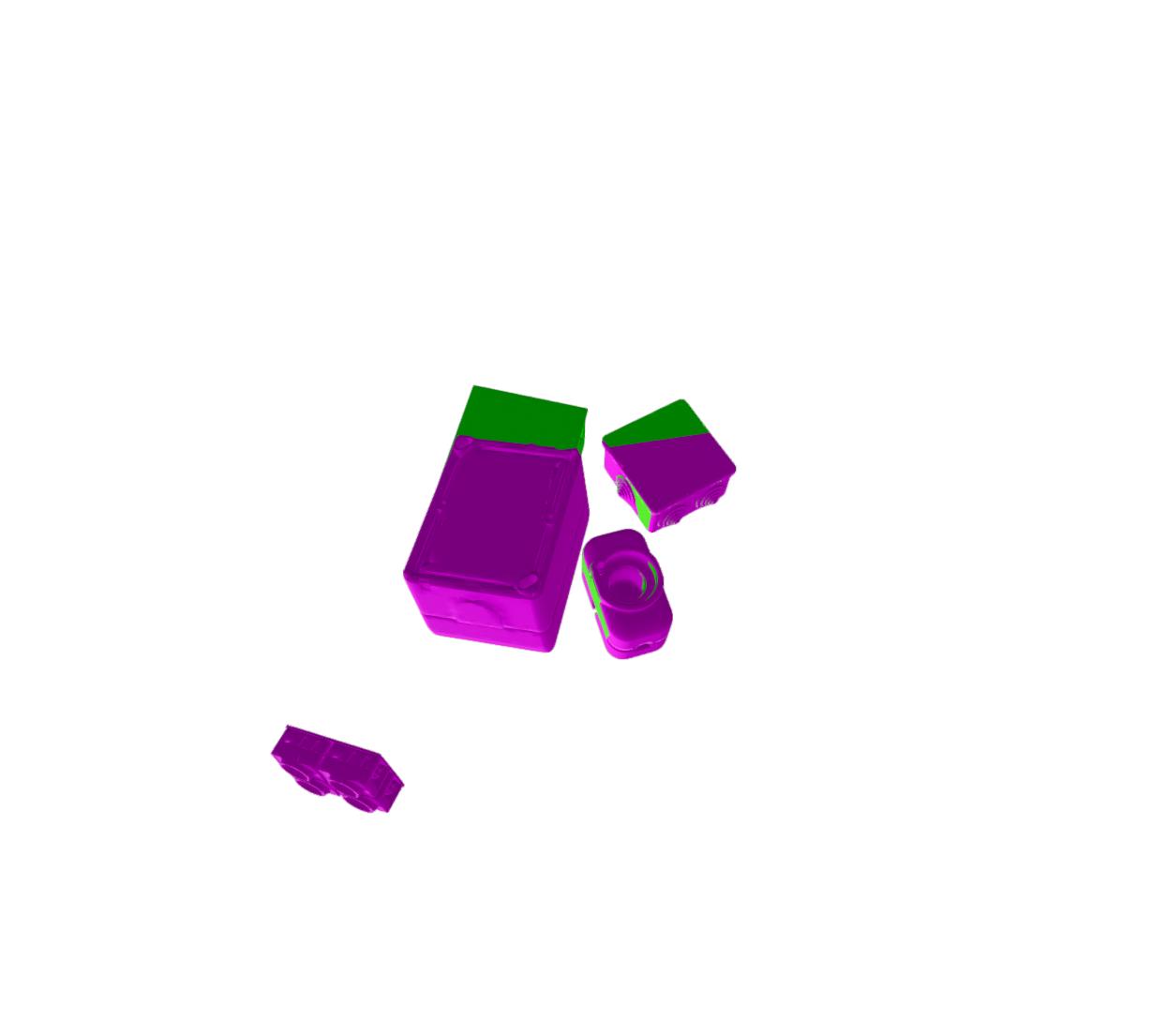}}\quad
  \subfloat{\includegraphics[height=.2\textwidth]{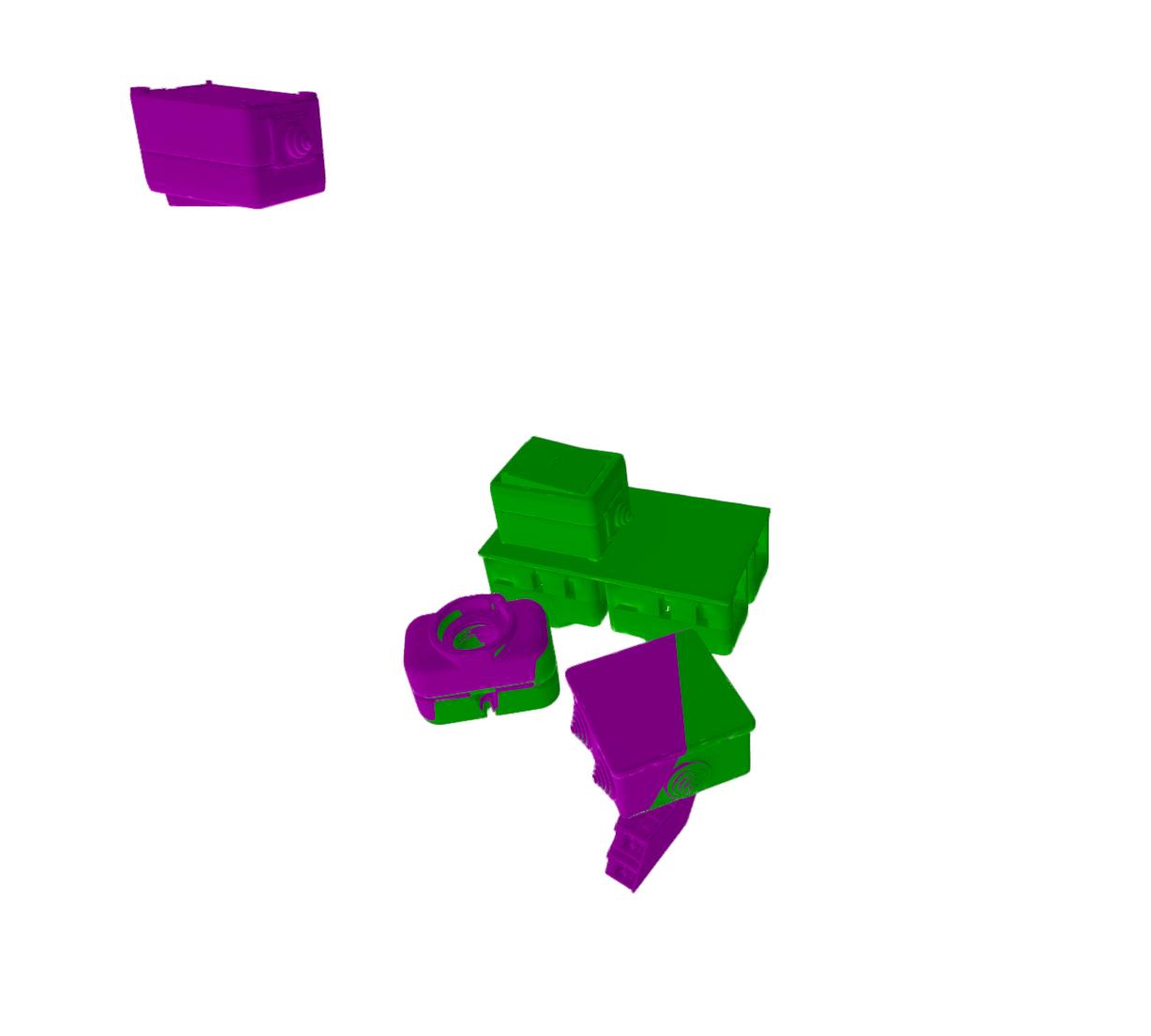}}\quad
  \subfloat{\includegraphics[height=.2\textwidth]{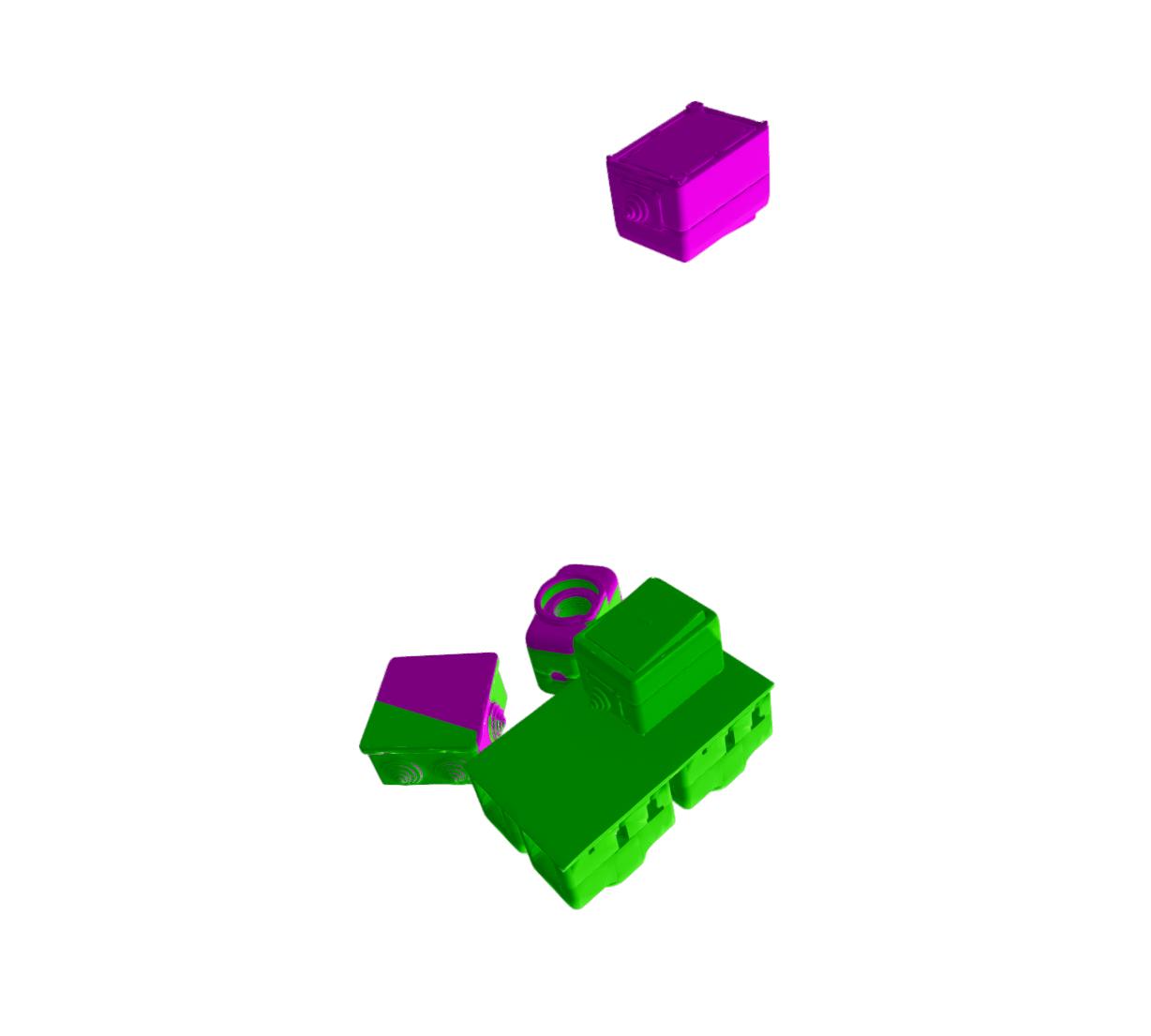}}\\
  
  \subfloat{\includegraphics[height=.2\textwidth]{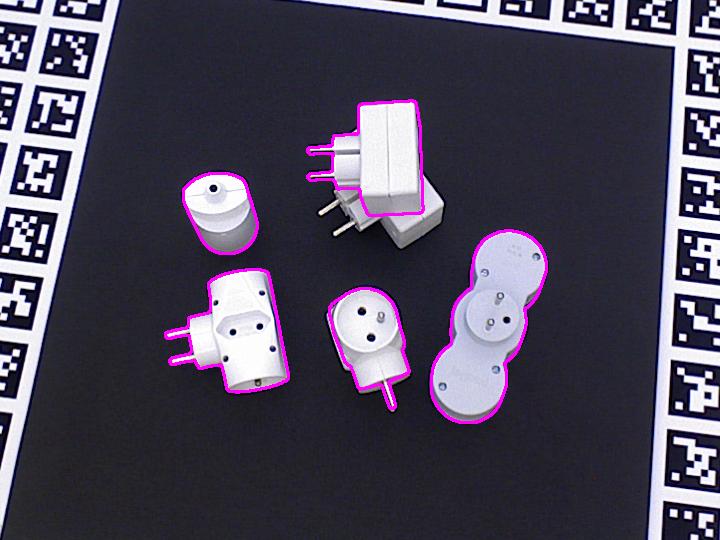}}\quad
  \subfloat{\includegraphics[height=.2\textwidth]{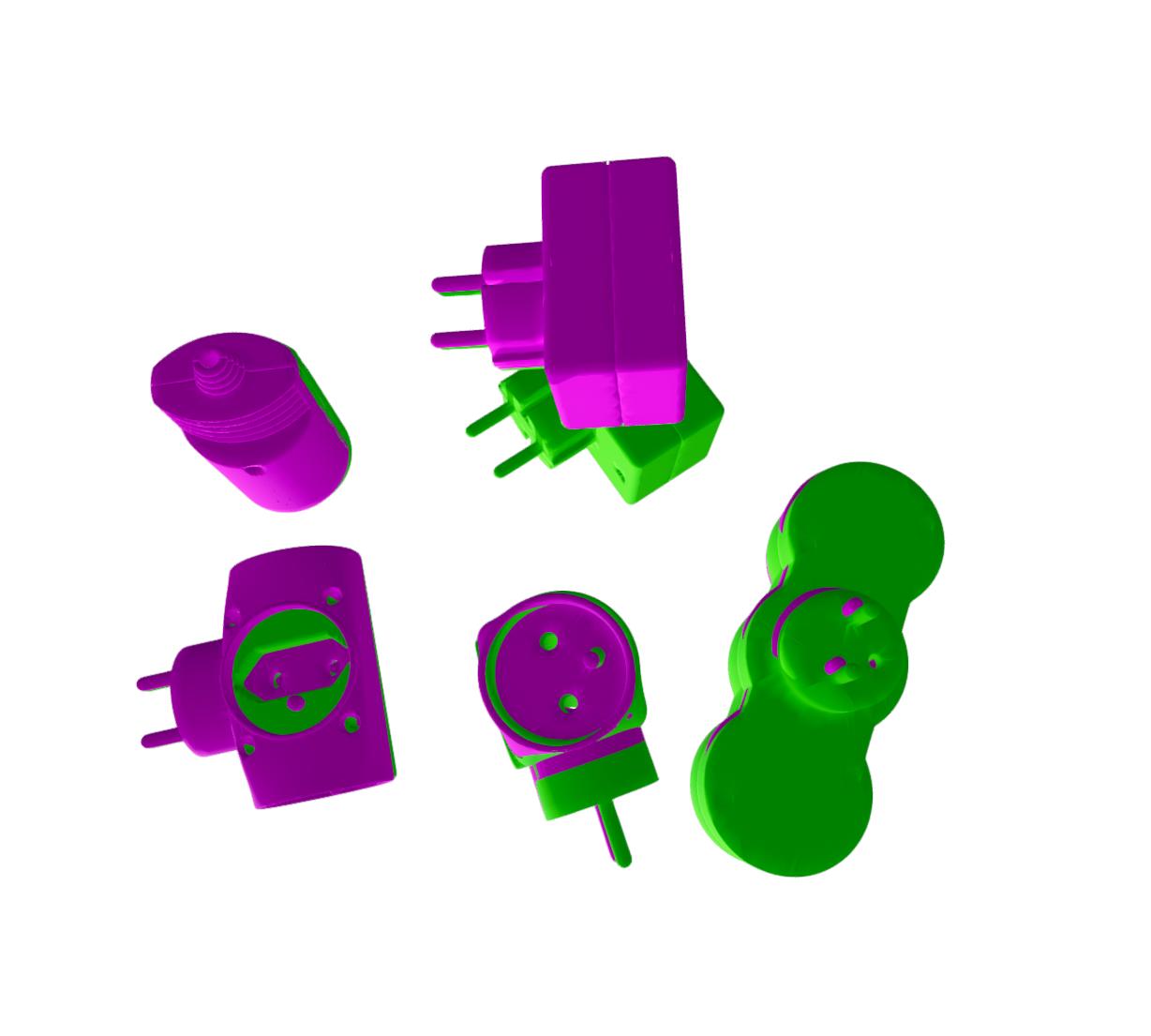}}\quad
  \subfloat{\includegraphics[height=.2\textwidth]{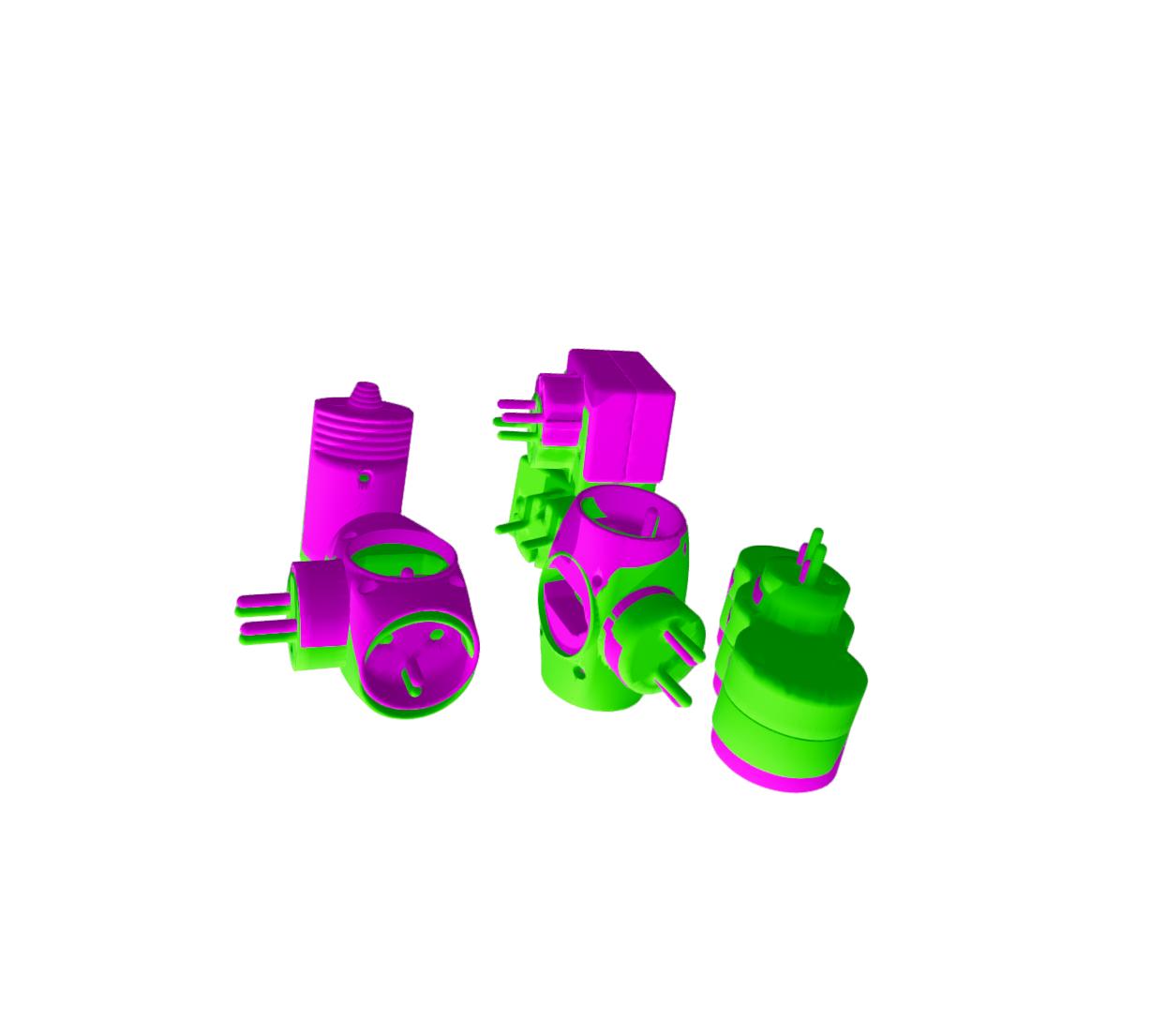}}\quad
  \subfloat{\includegraphics[height=.2\textwidth]{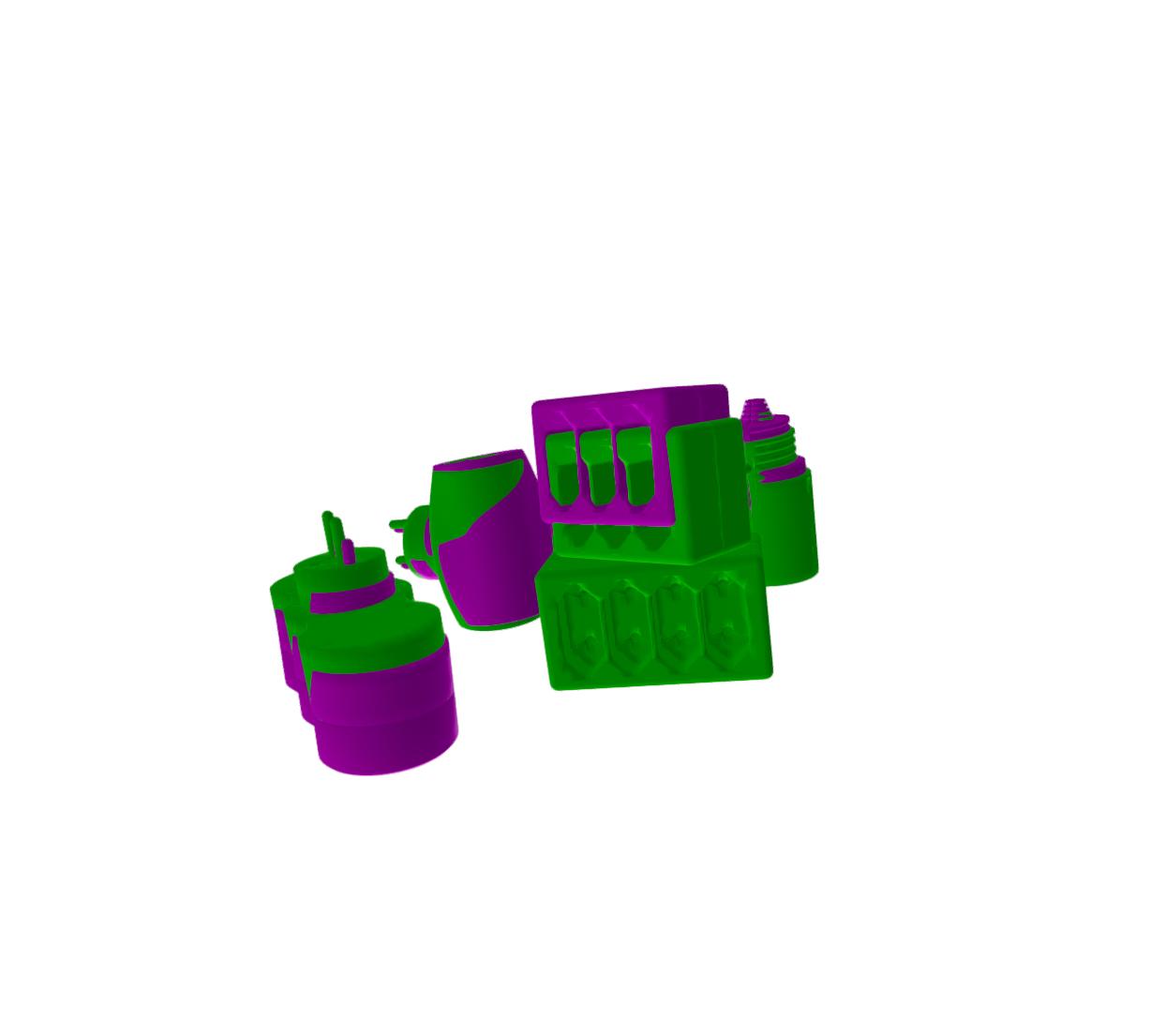}}\\
  
  \subfloat{\includegraphics[height=.2\textwidth]{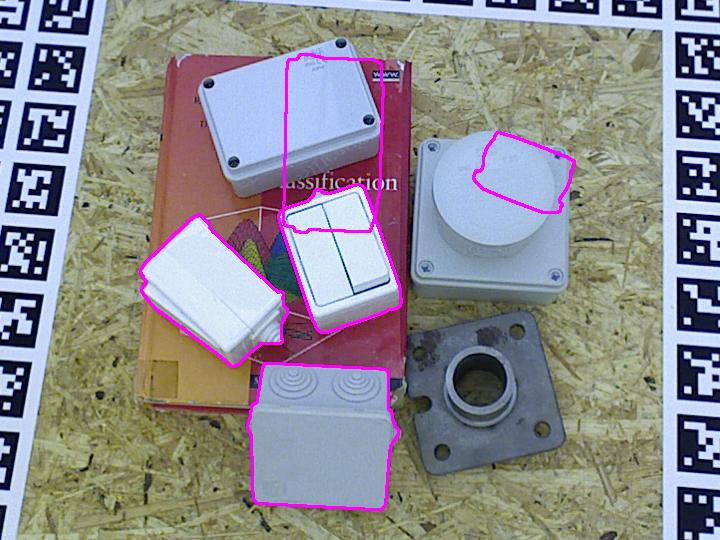}}\quad
  \subfloat{\includegraphics[height=.2\textwidth]{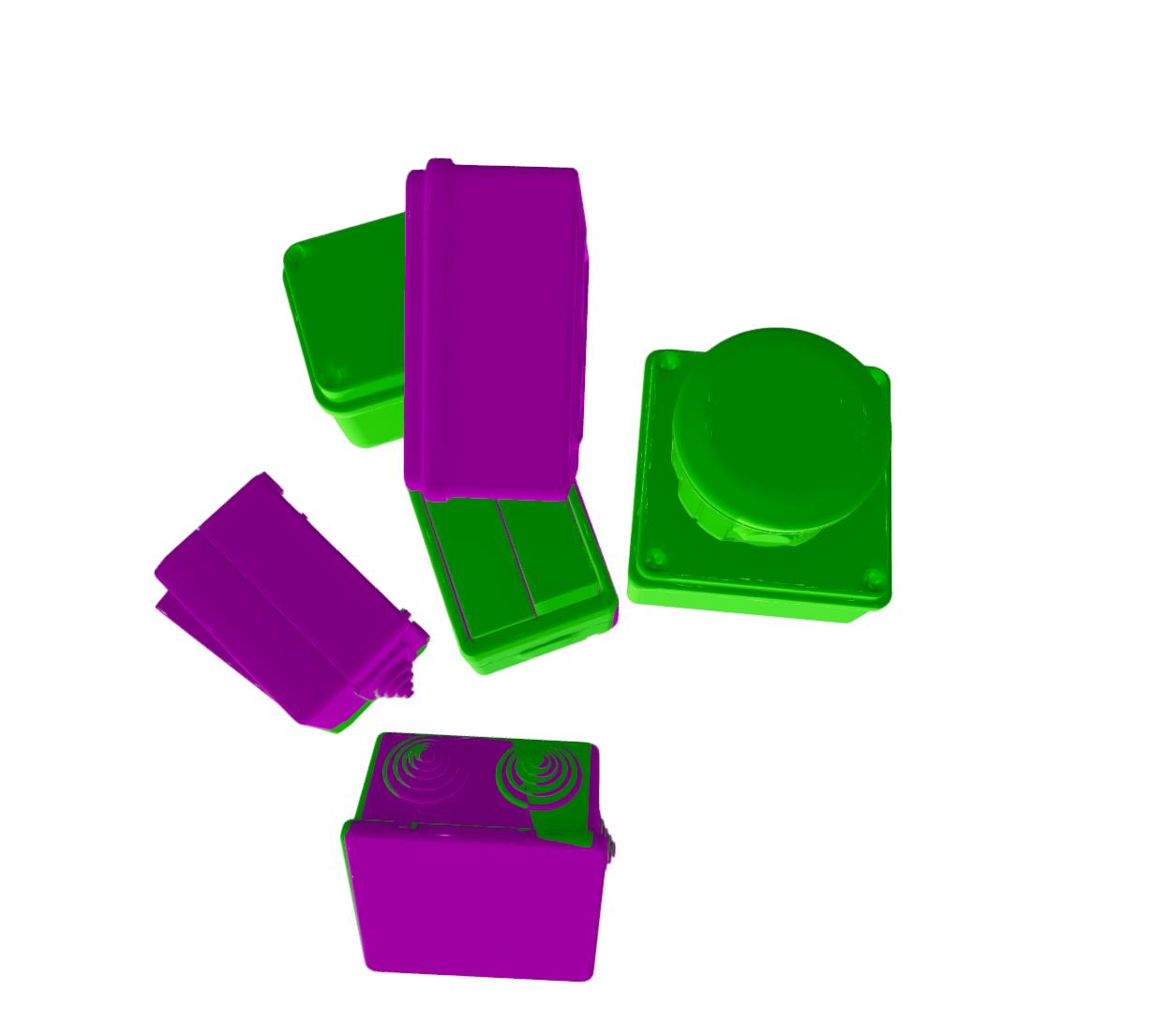}}\quad
  \subfloat{\includegraphics[height=.2\textwidth]{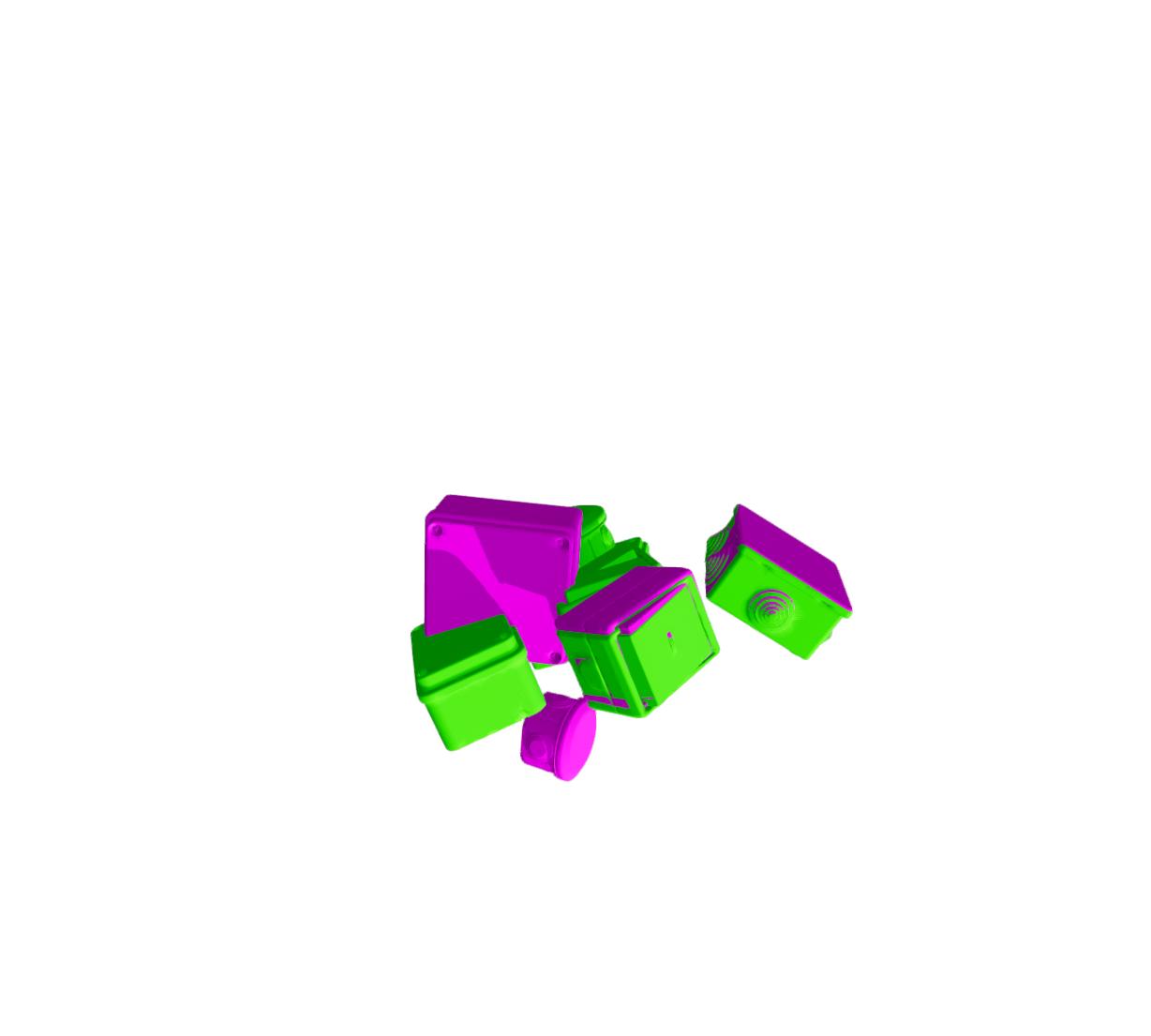}}\quad
  \subfloat{\includegraphics[height=.2\textwidth]{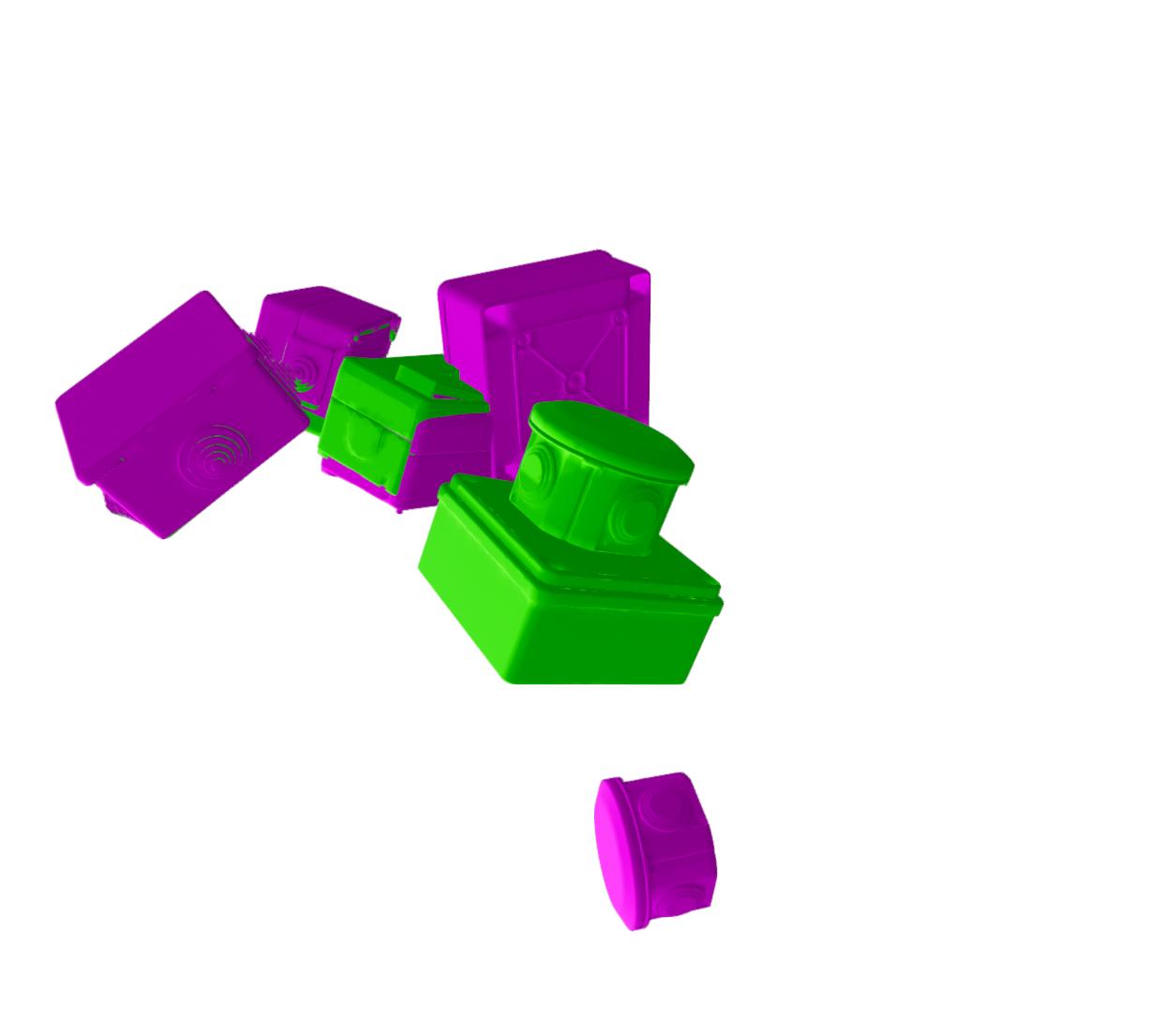}}\\
  
  \subfloat{\includegraphics[height=.2\textwidth]{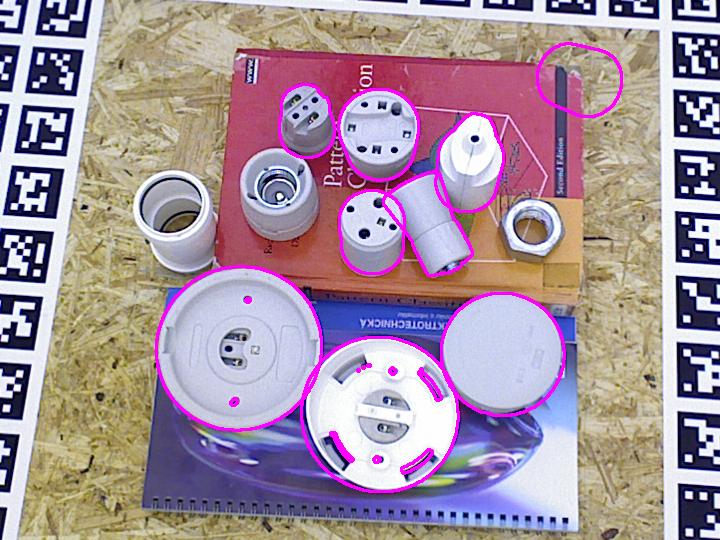}}\quad
  \subfloat{\includegraphics[height=.2\textwidth]{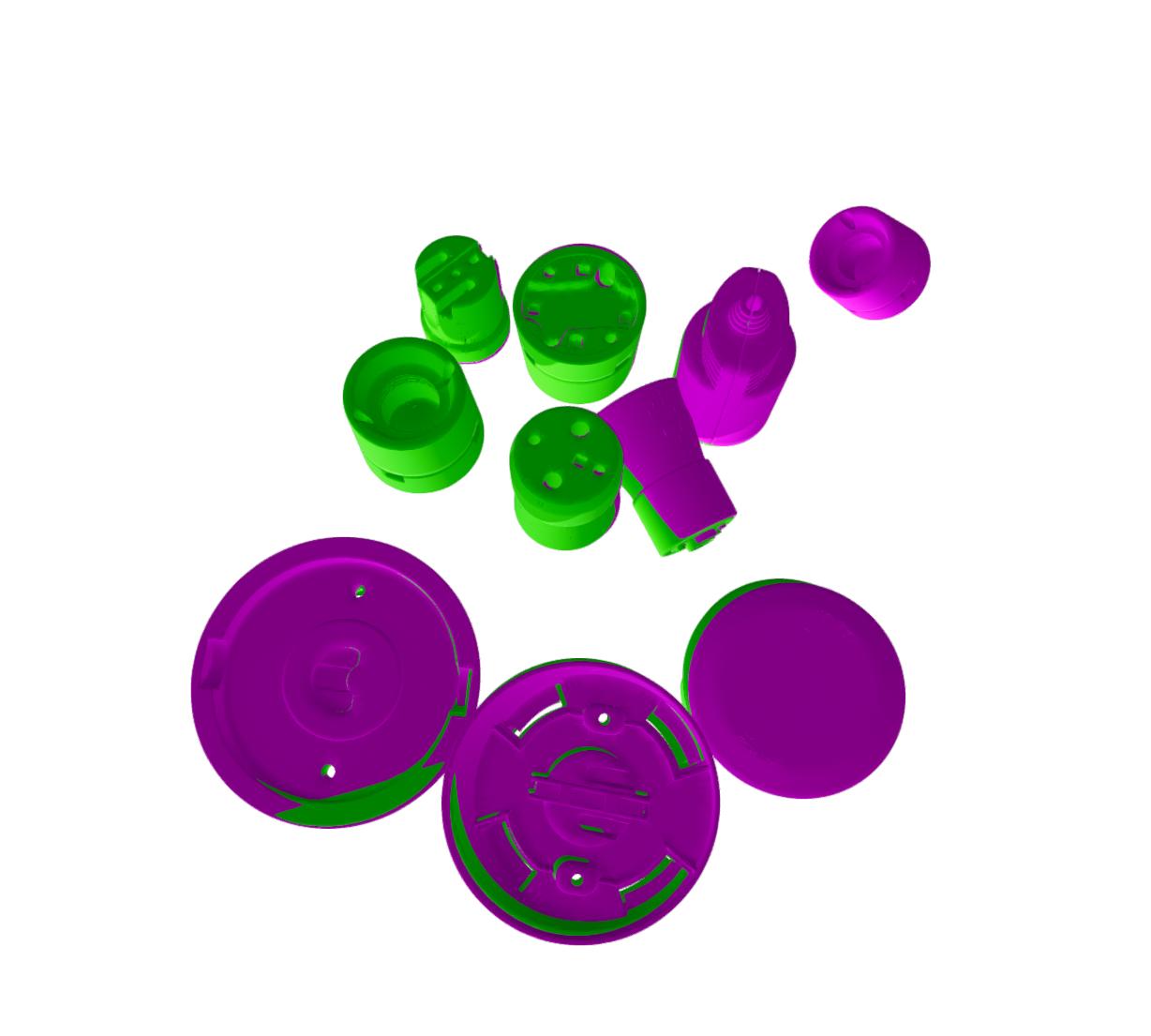}}\quad
  \subfloat{\includegraphics[height=.2\textwidth]{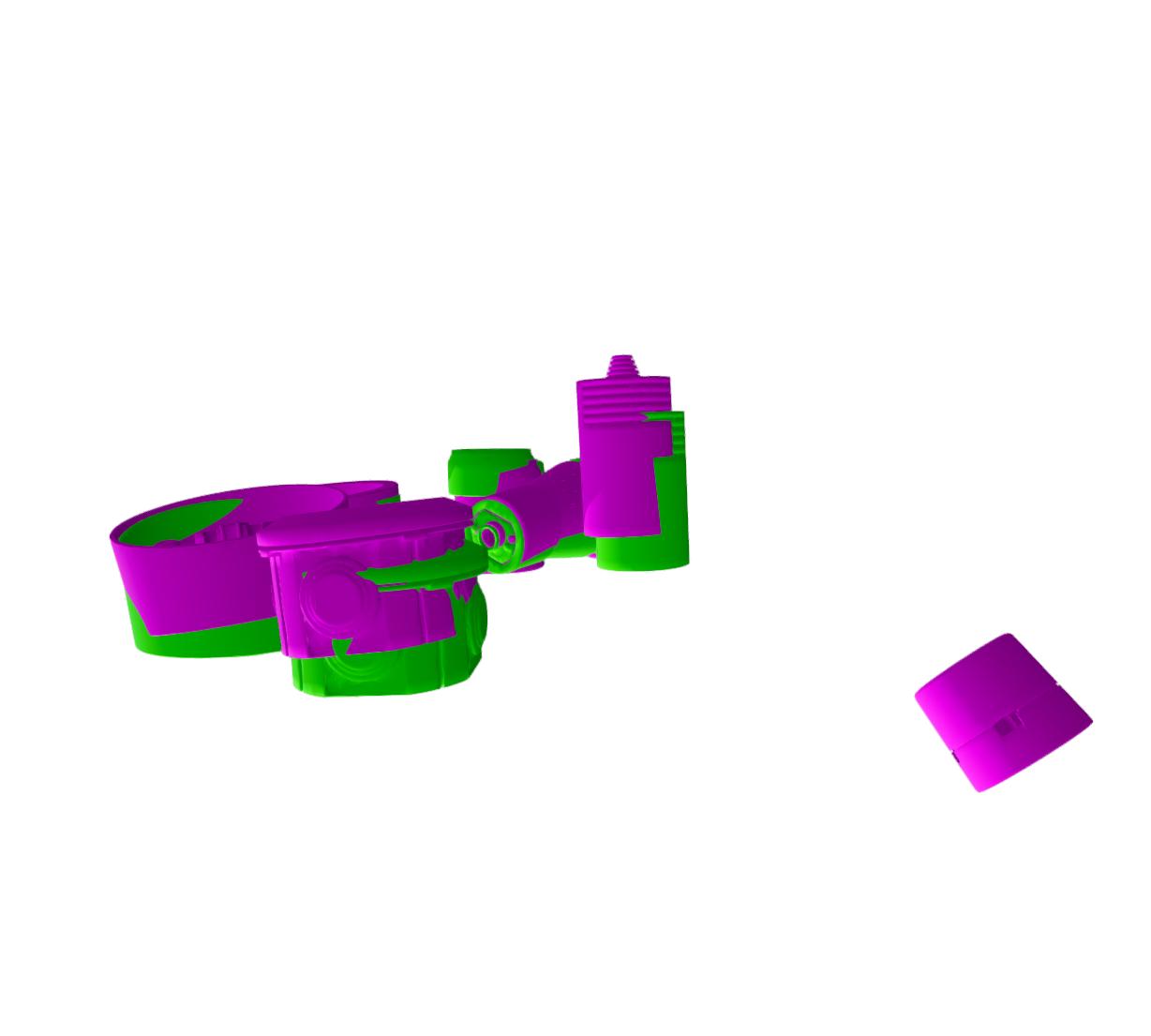}}\quad
  \subfloat{\includegraphics[height=.2\textwidth]{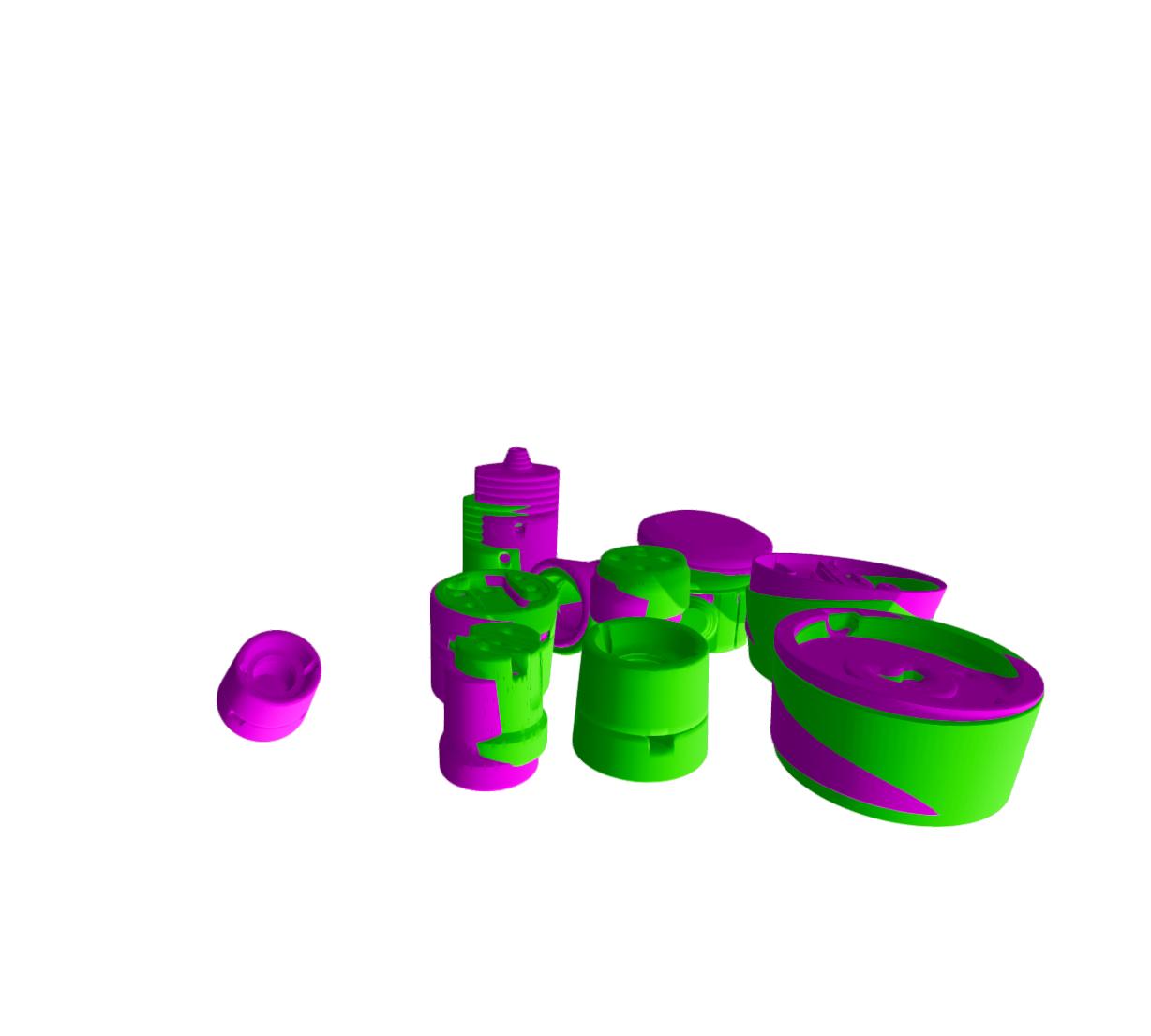}}\\

  \caption{T-LESS dataset visualization with \textcolor{magenta}{estimated} (magenta) 6D pose of the meshes and \textcolor{green}{ground-truth} (green) poses. The first column shows the test image with a contour of the projection made by the predicted pose. The other three columns show the corresponding 3D view from different viewing angles. The first is taken from approximately the same viewing angle as the image was taken.}
  \label{fig:A0_TLESS}
\end{figure*}

\begin{figure*}
  \centering
  \subfloat{\includegraphics[height=.2\textwidth]{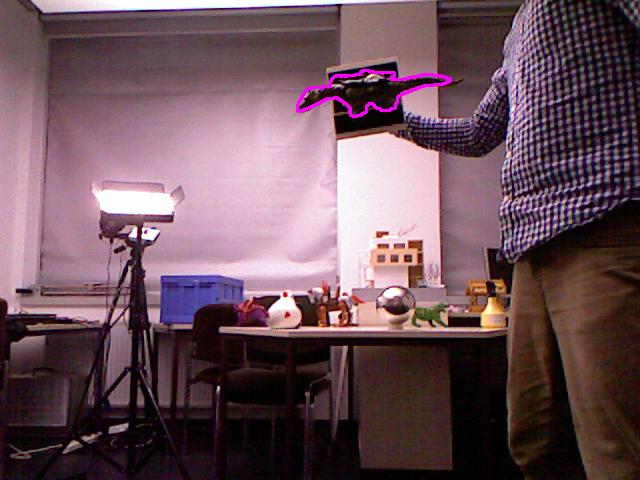}}\quad
  \subfloat{\includegraphics[height=.2\textwidth]{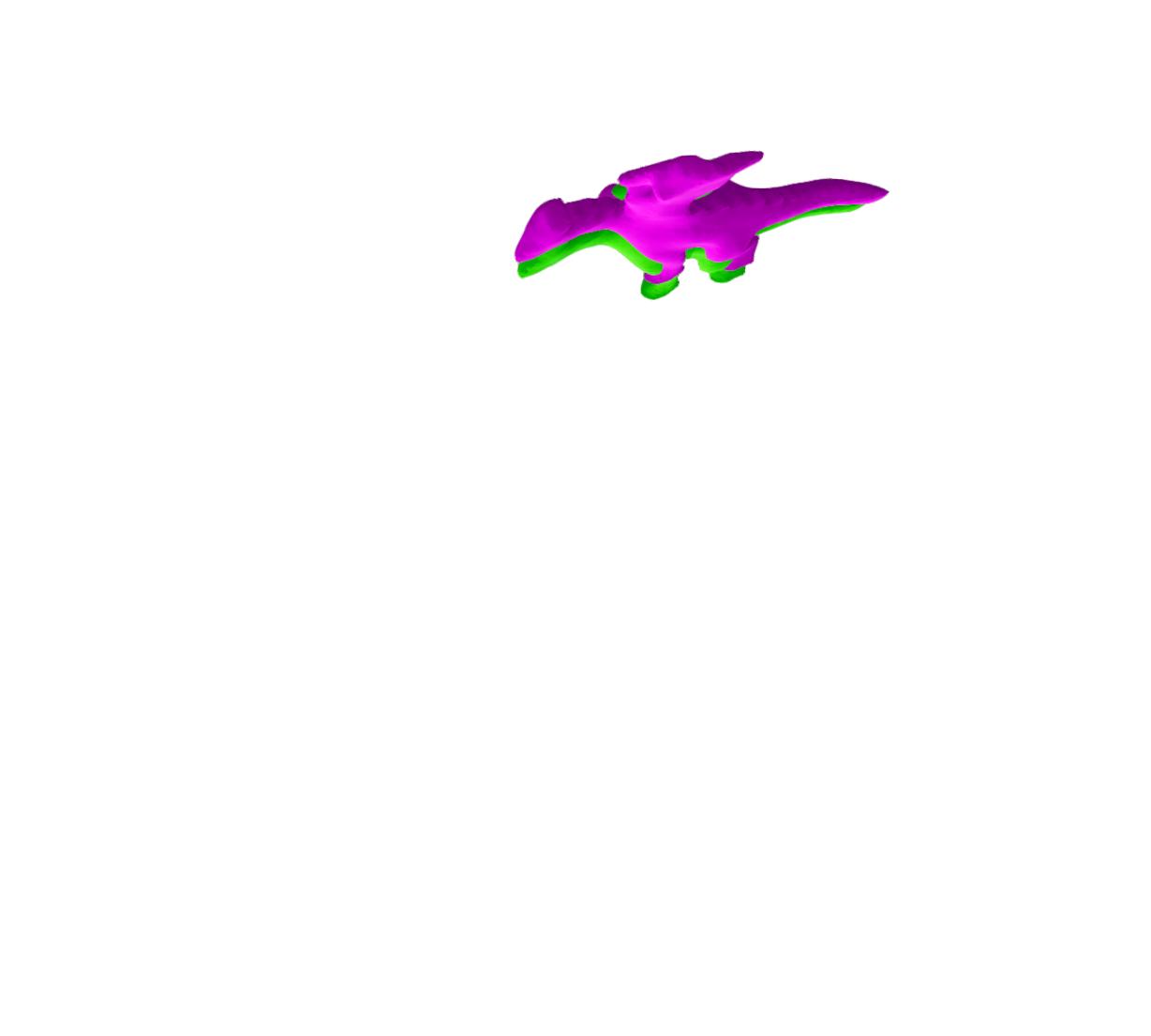}}\quad
  \subfloat{\includegraphics[height=.2\textwidth]{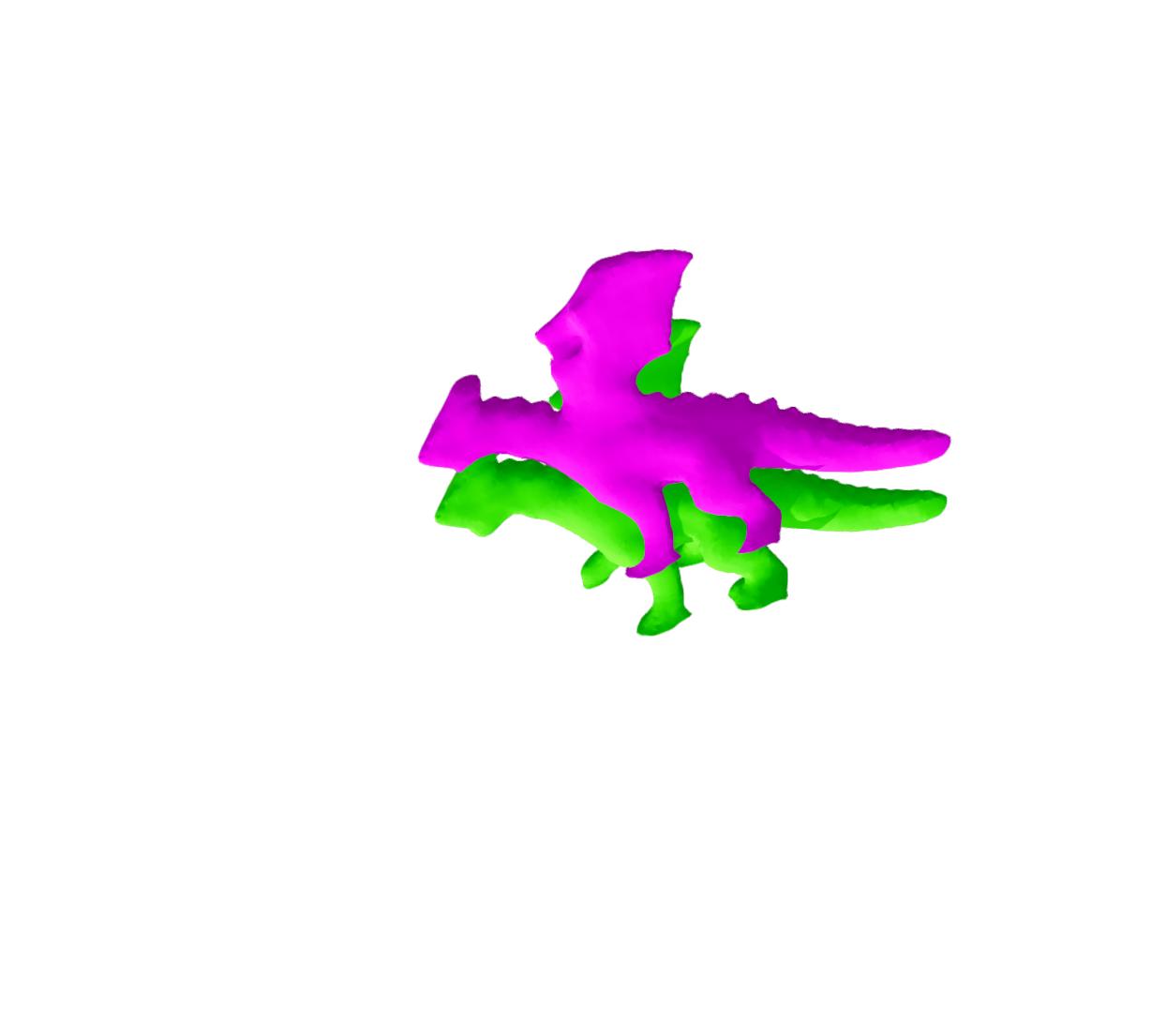}}\quad
  \subfloat{\includegraphics[height=.2\textwidth]{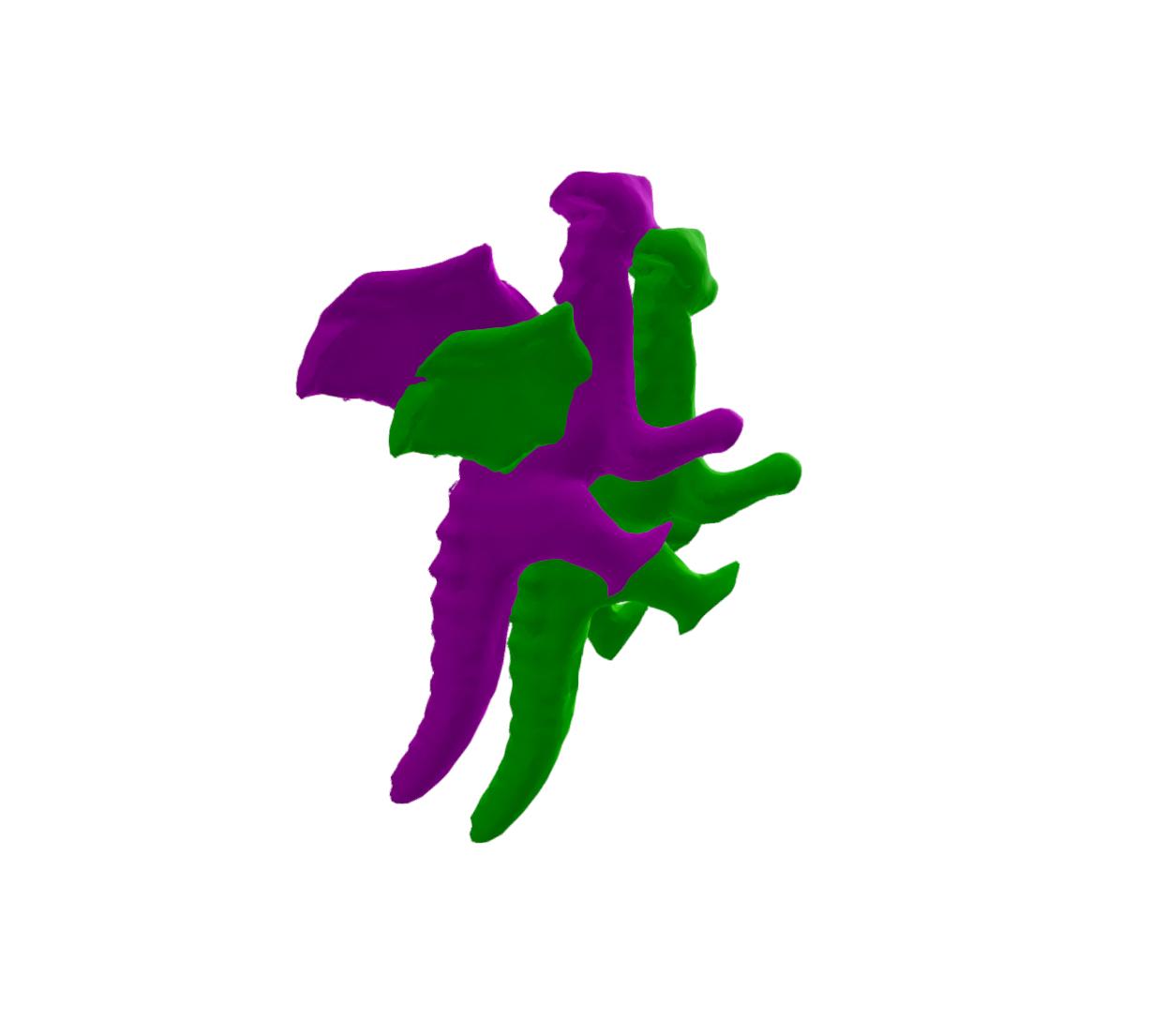}}\\
  
  \subfloat{\includegraphics[height=.2\textwidth]{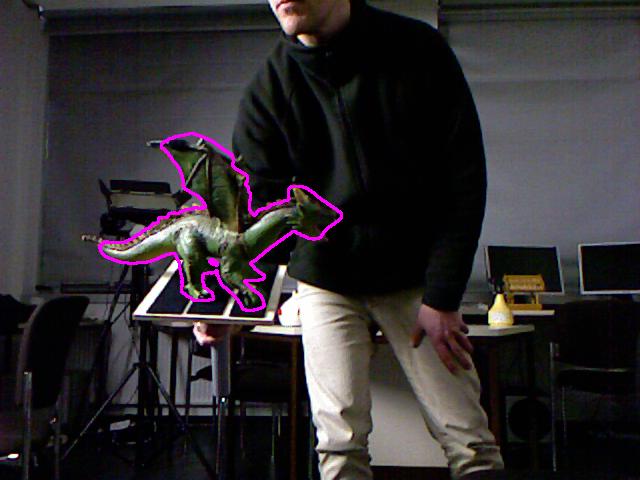}}\quad
  \subfloat{\includegraphics[height=.2\textwidth]{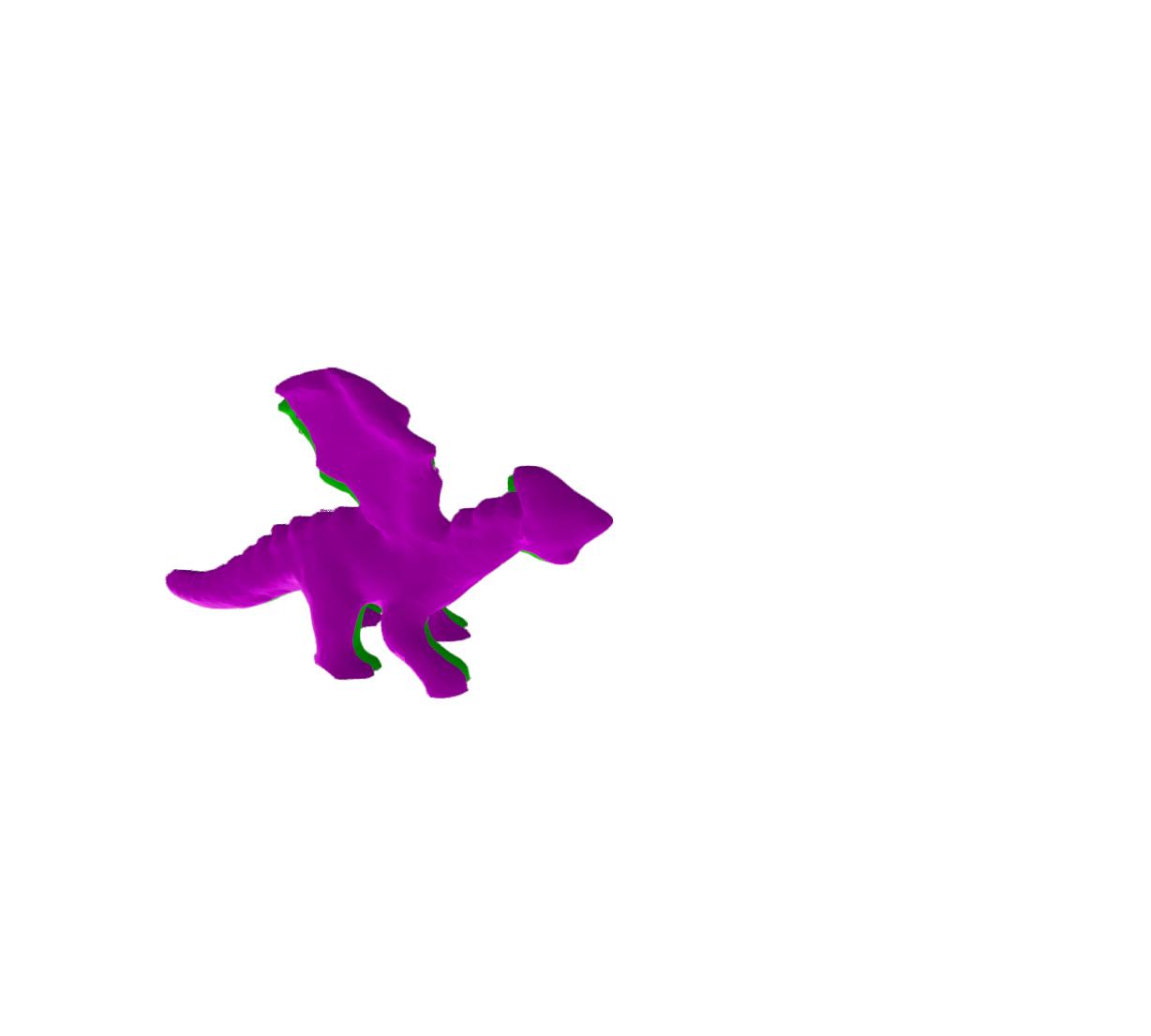}}\quad
  \subfloat{\includegraphics[height=.2\textwidth]{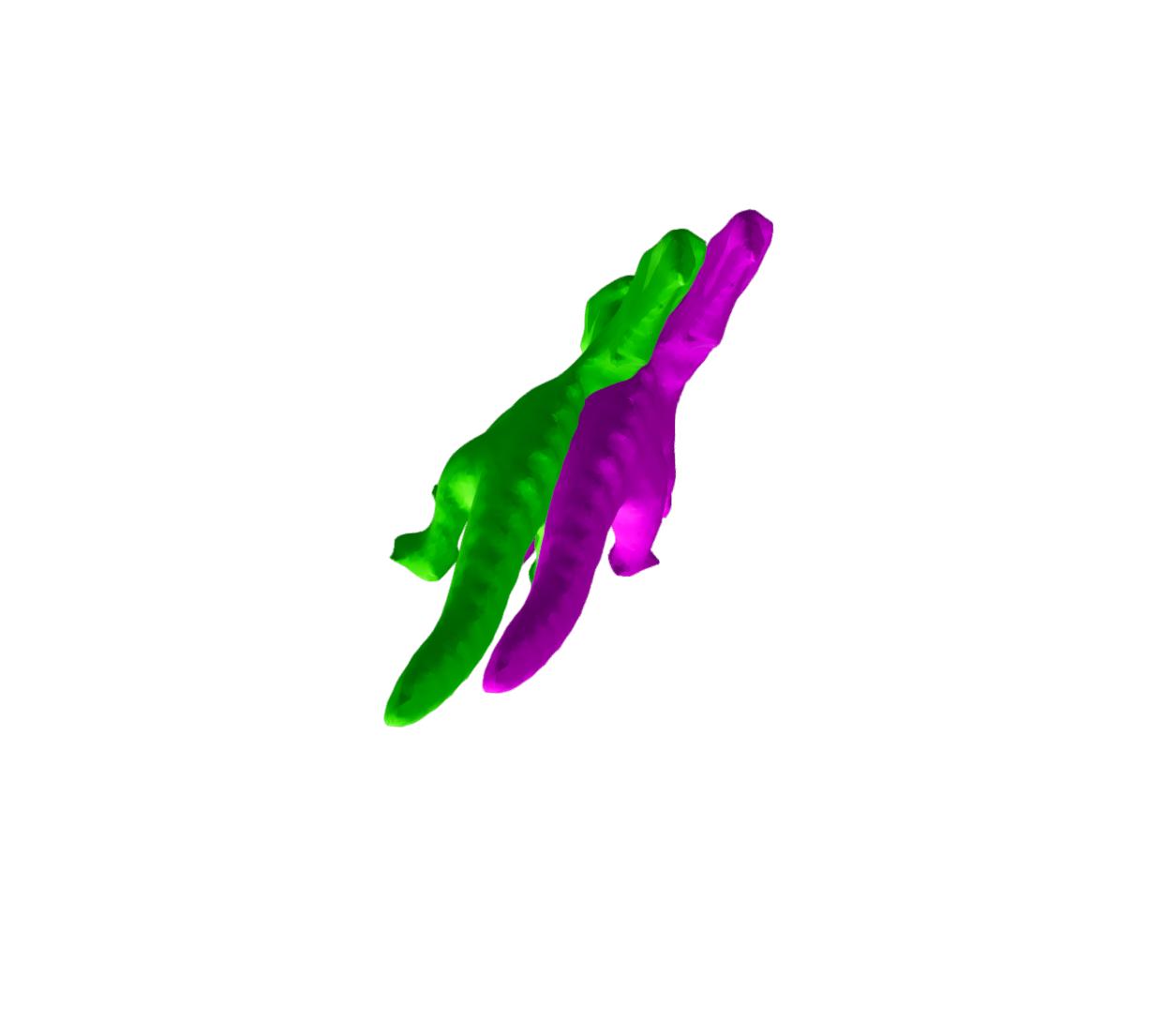}}\quad
  \subfloat{\includegraphics[height=.2\textwidth]{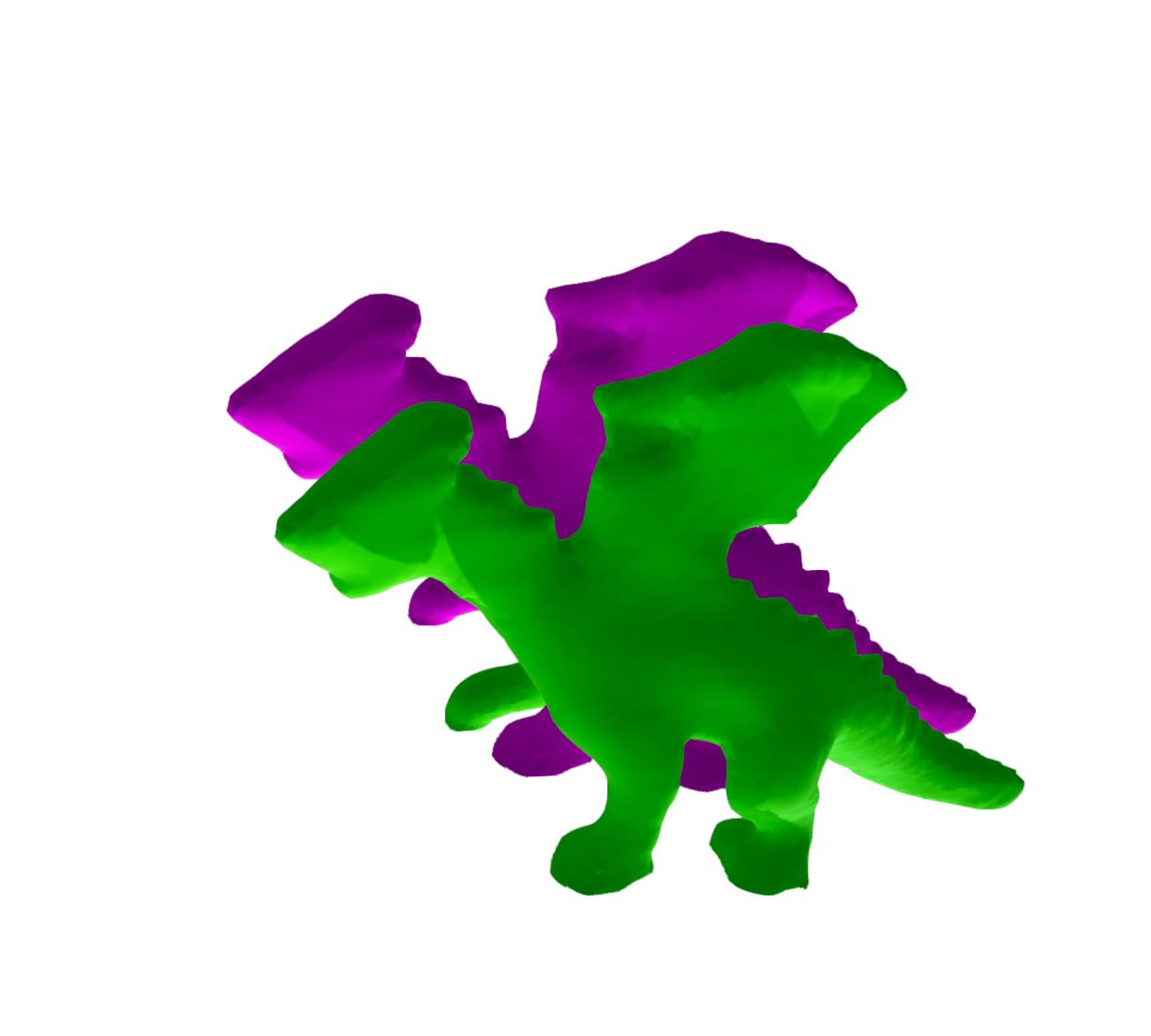}}\\
  
  \subfloat{\includegraphics[height=.2\textwidth]{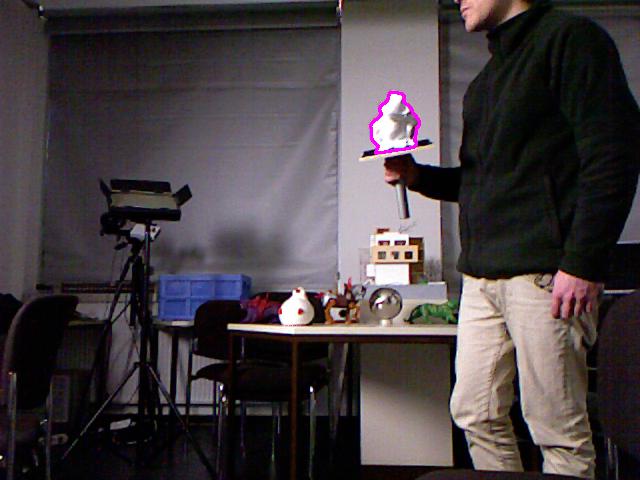}}\quad
  \subfloat{\includegraphics[height=.2\textwidth]{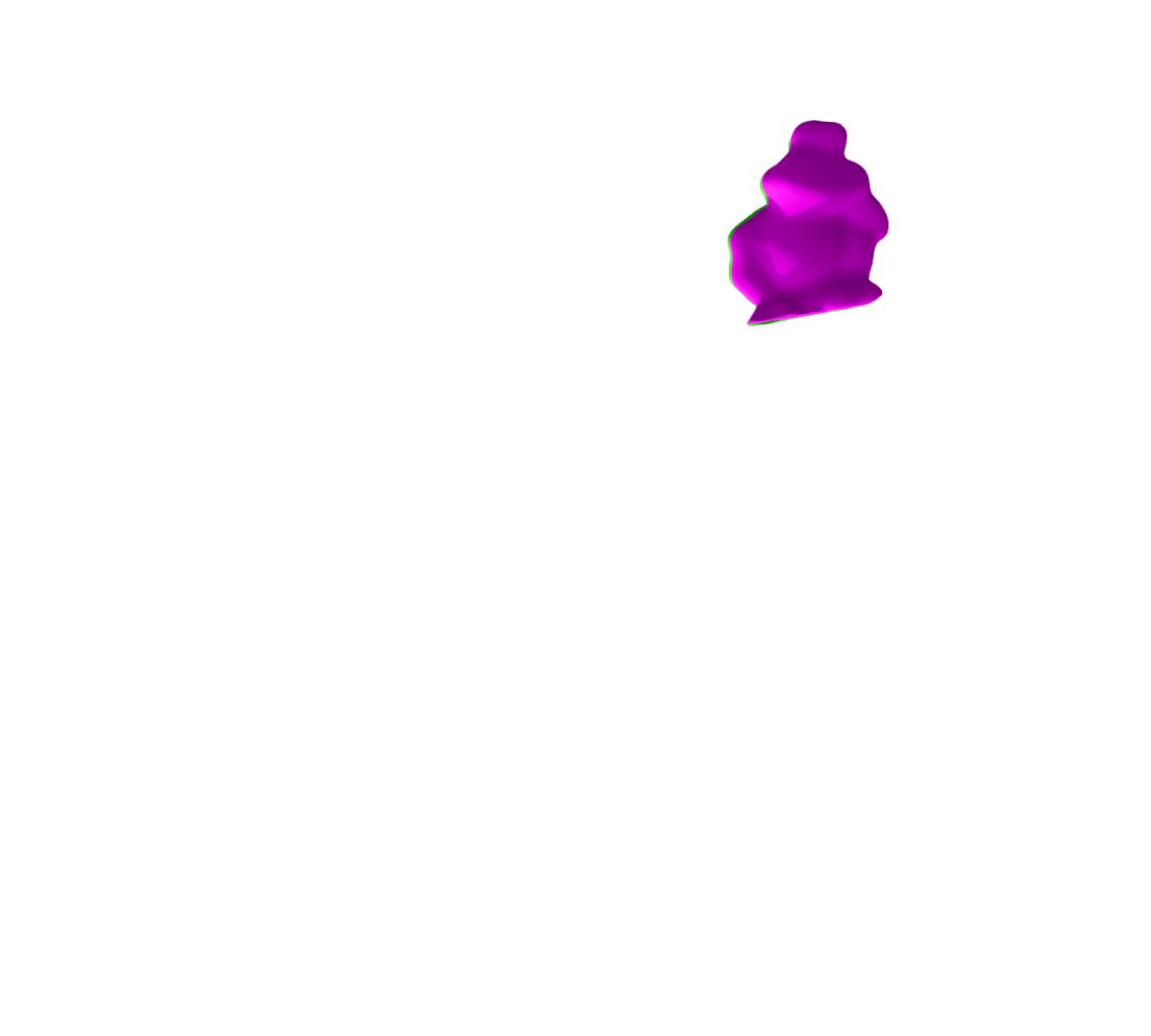}}\quad
  \subfloat{\includegraphics[height=.2\textwidth]{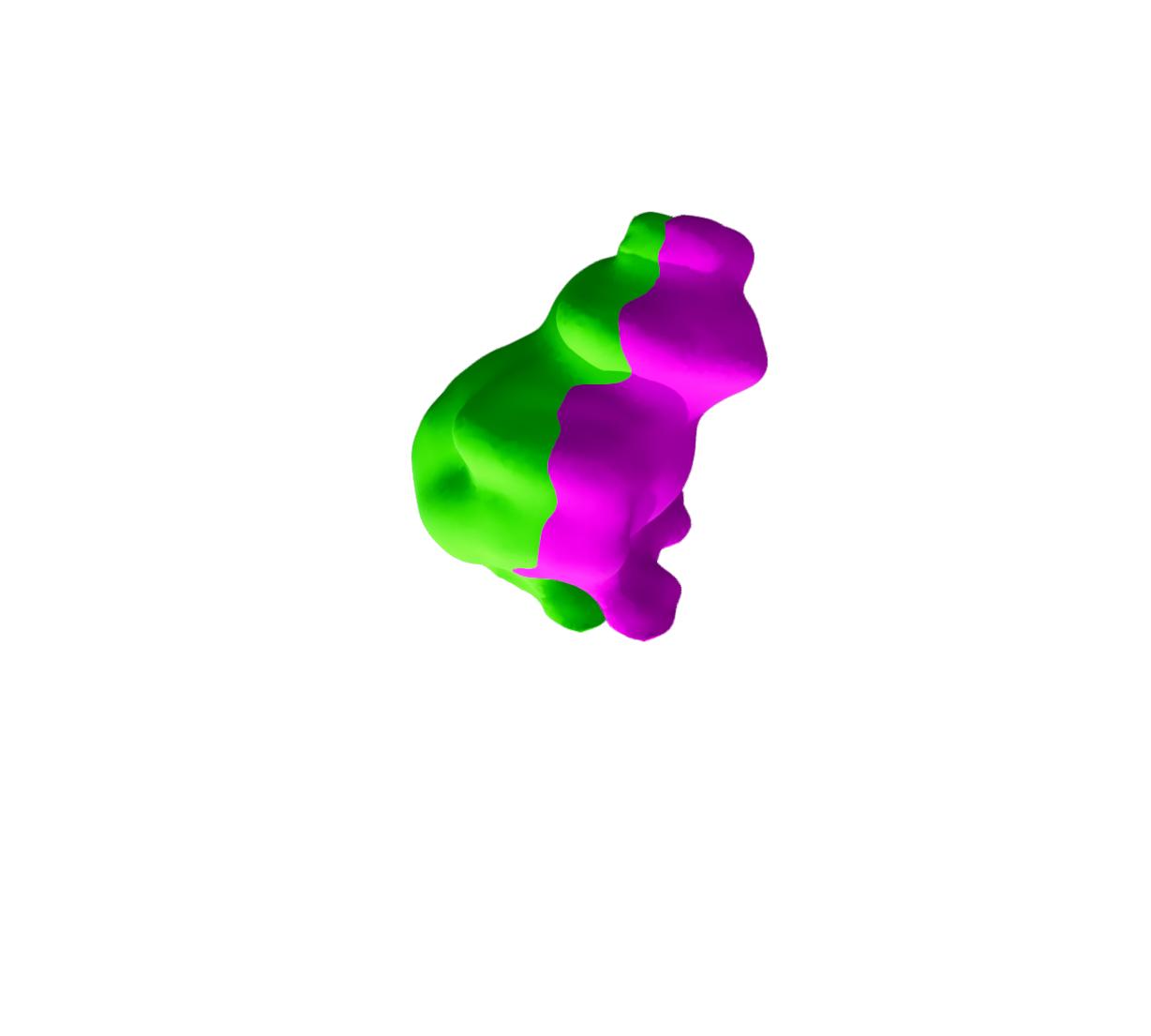}}\quad
  \subfloat{\includegraphics[height=.2\textwidth]{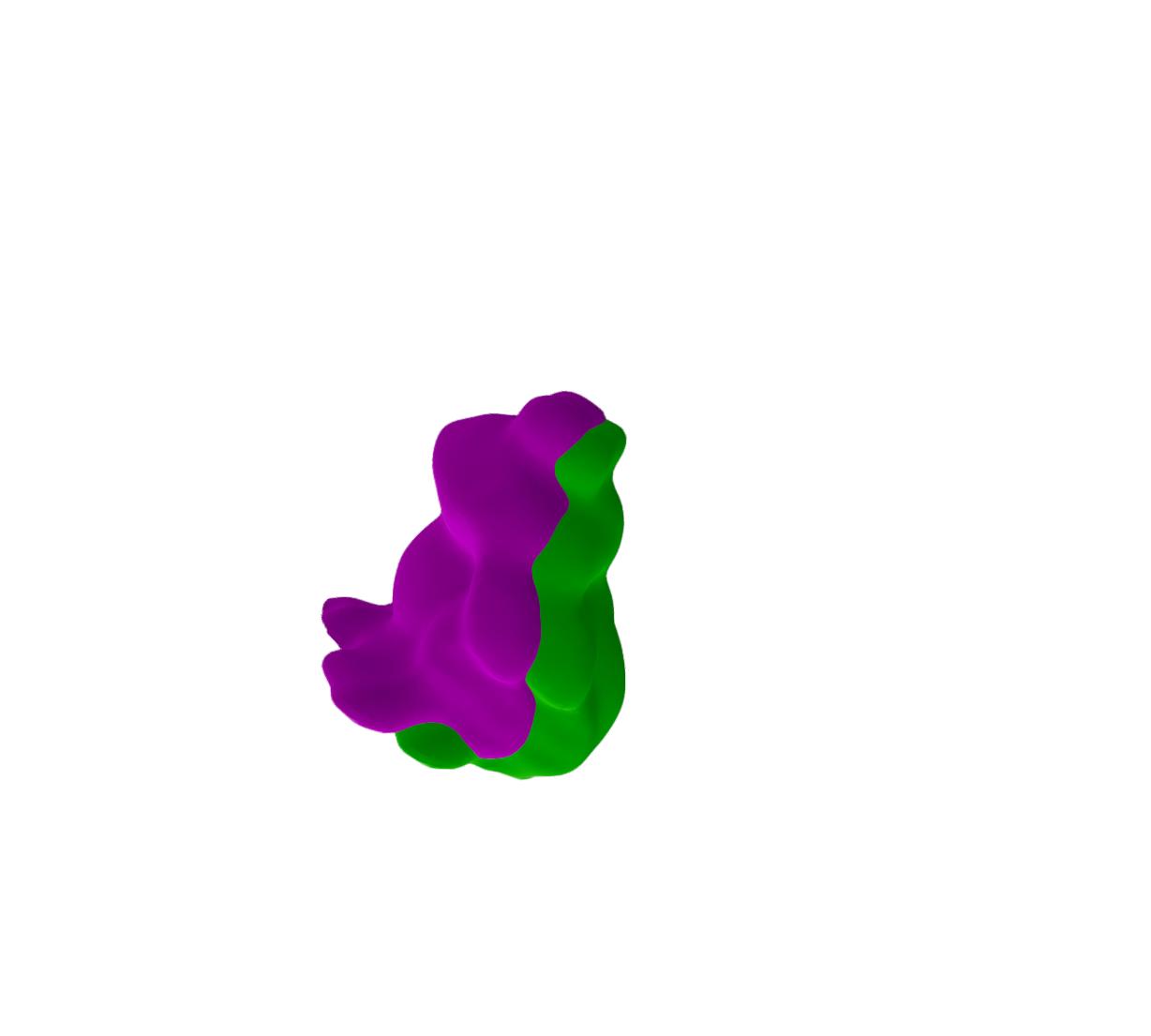}}\\
  
  \subfloat{\includegraphics[height=.2\textwidth]{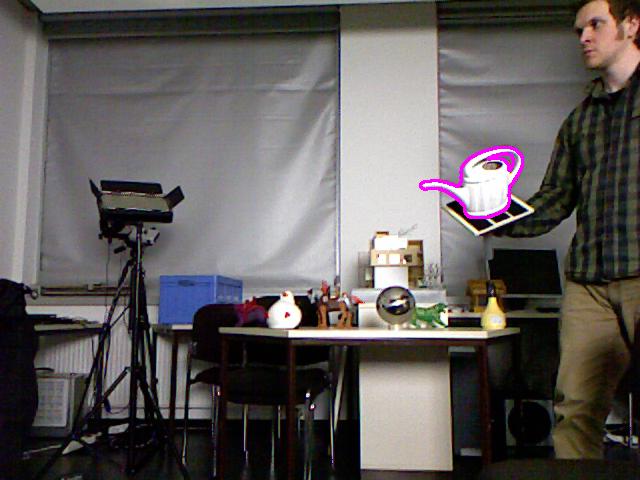}}\quad
  \subfloat{\includegraphics[height=.2\textwidth]{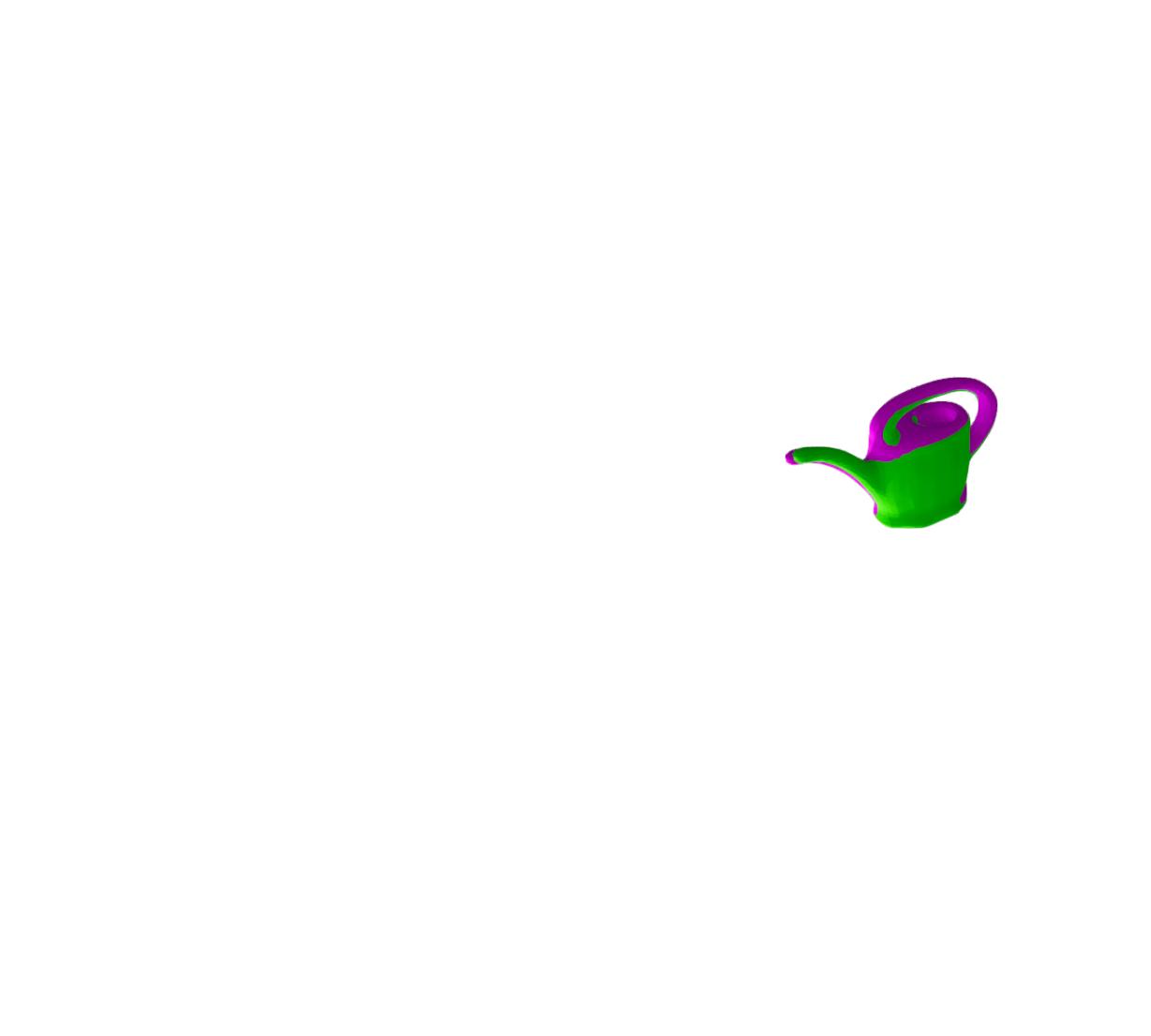}}\quad
  \subfloat{\includegraphics[height=.2\textwidth]{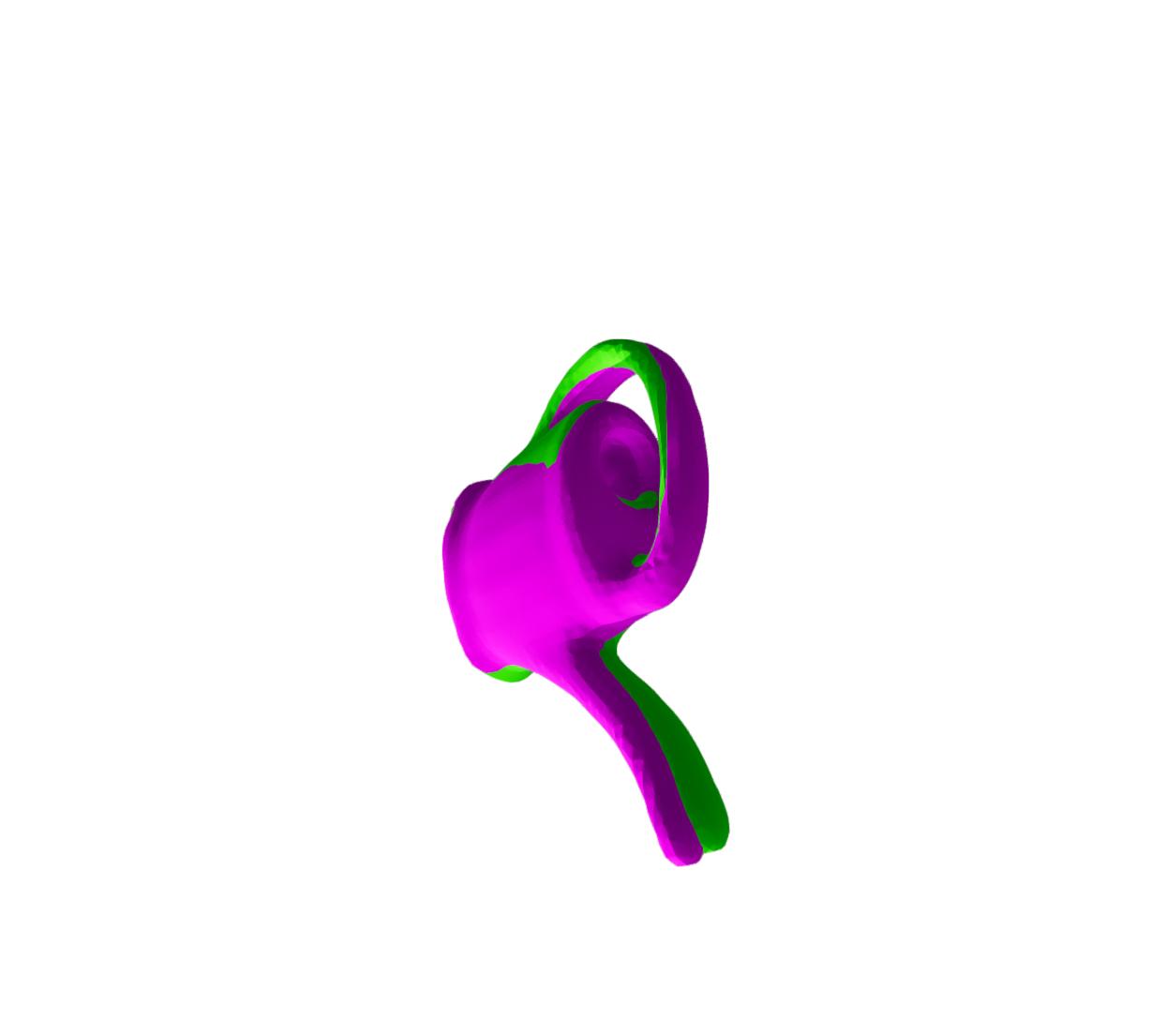}}\quad
  \subfloat{\includegraphics[height=.2\textwidth]{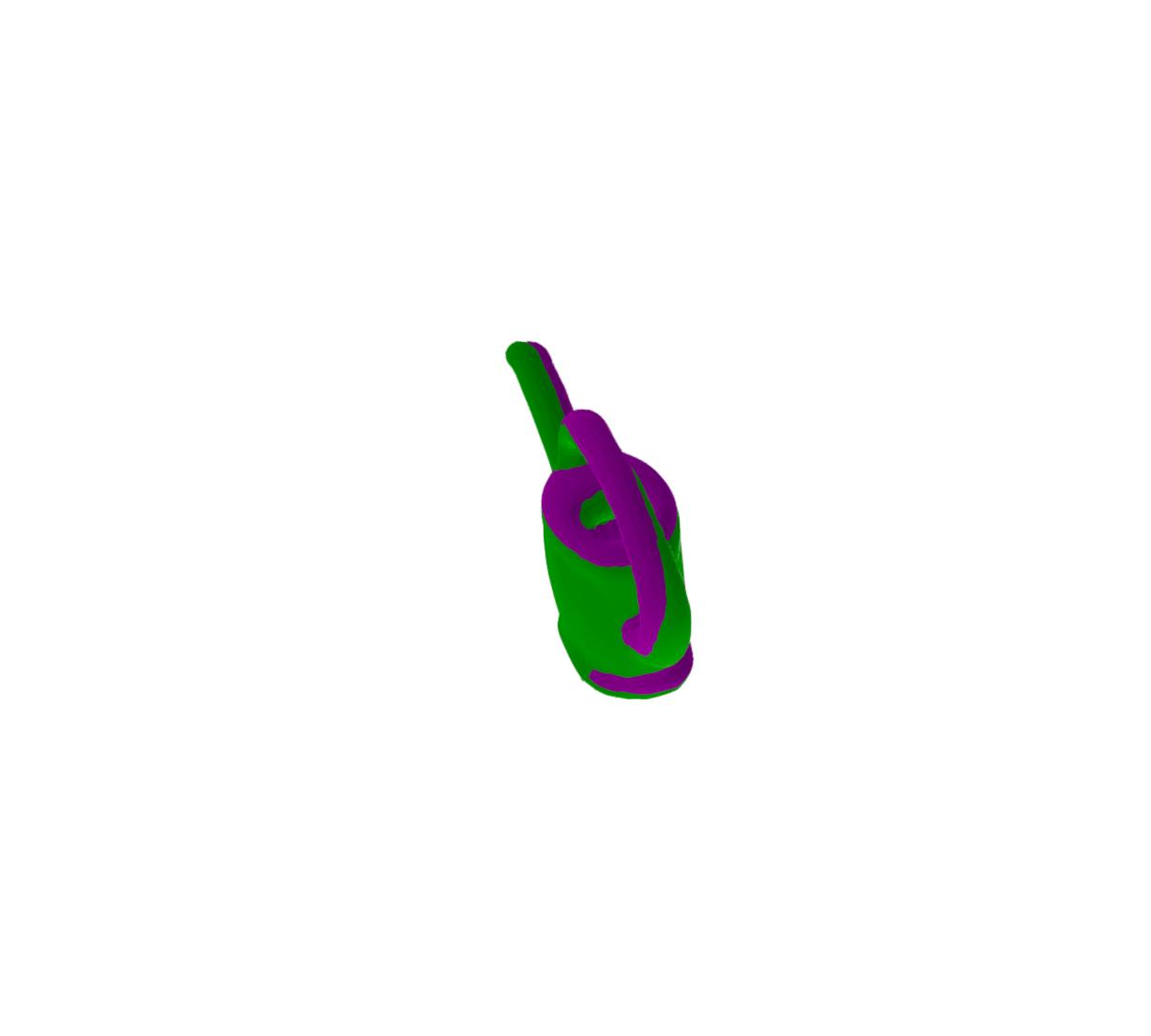}}\\
  
  \subfloat{\includegraphics[height=.2\textwidth]{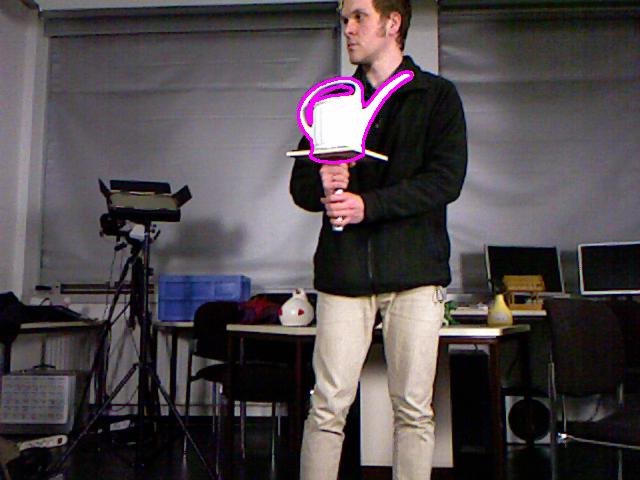}}\quad
  \subfloat{\includegraphics[height=.2\textwidth]{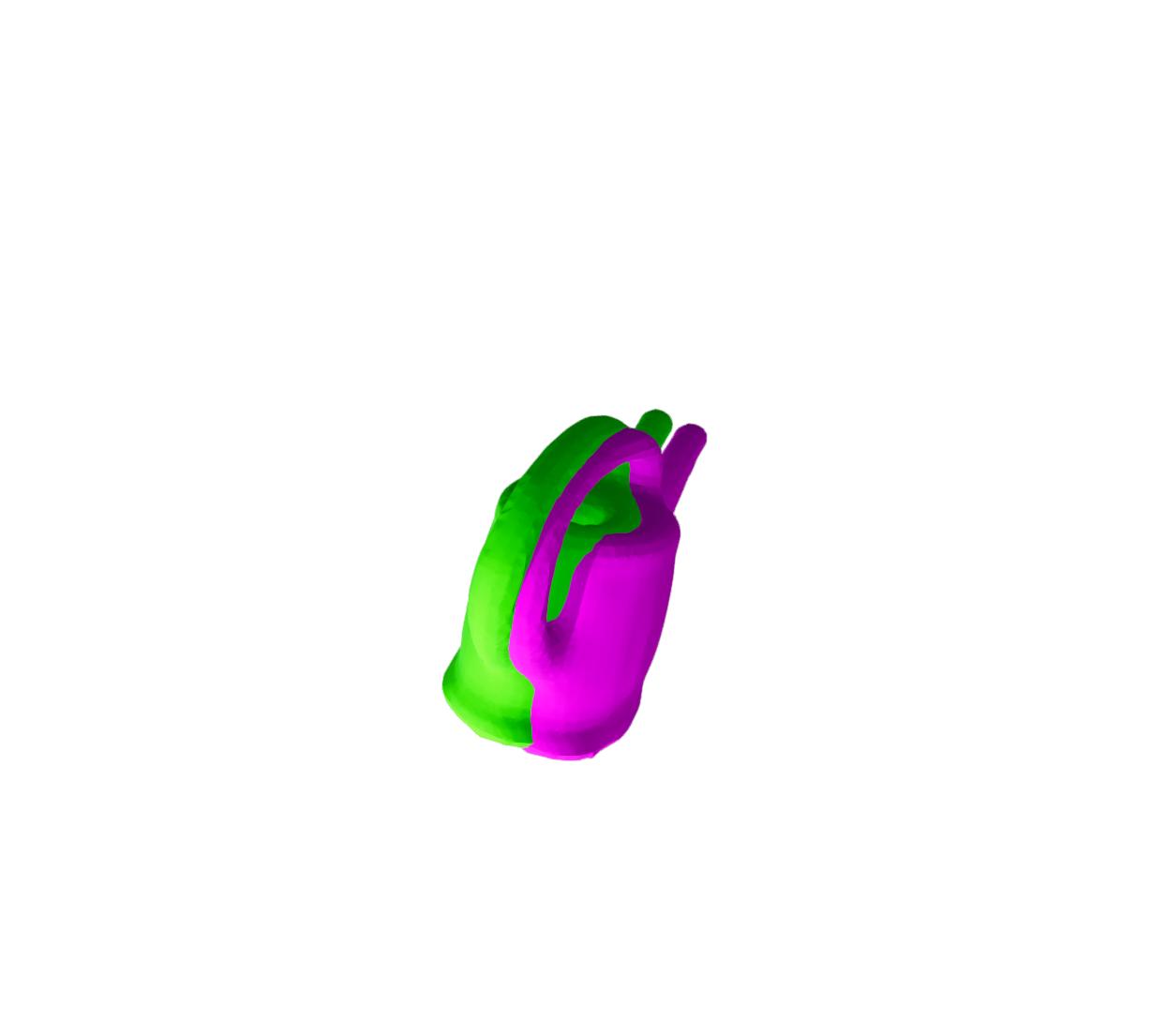}}\quad
  \subfloat{\includegraphics[height=.2\textwidth]{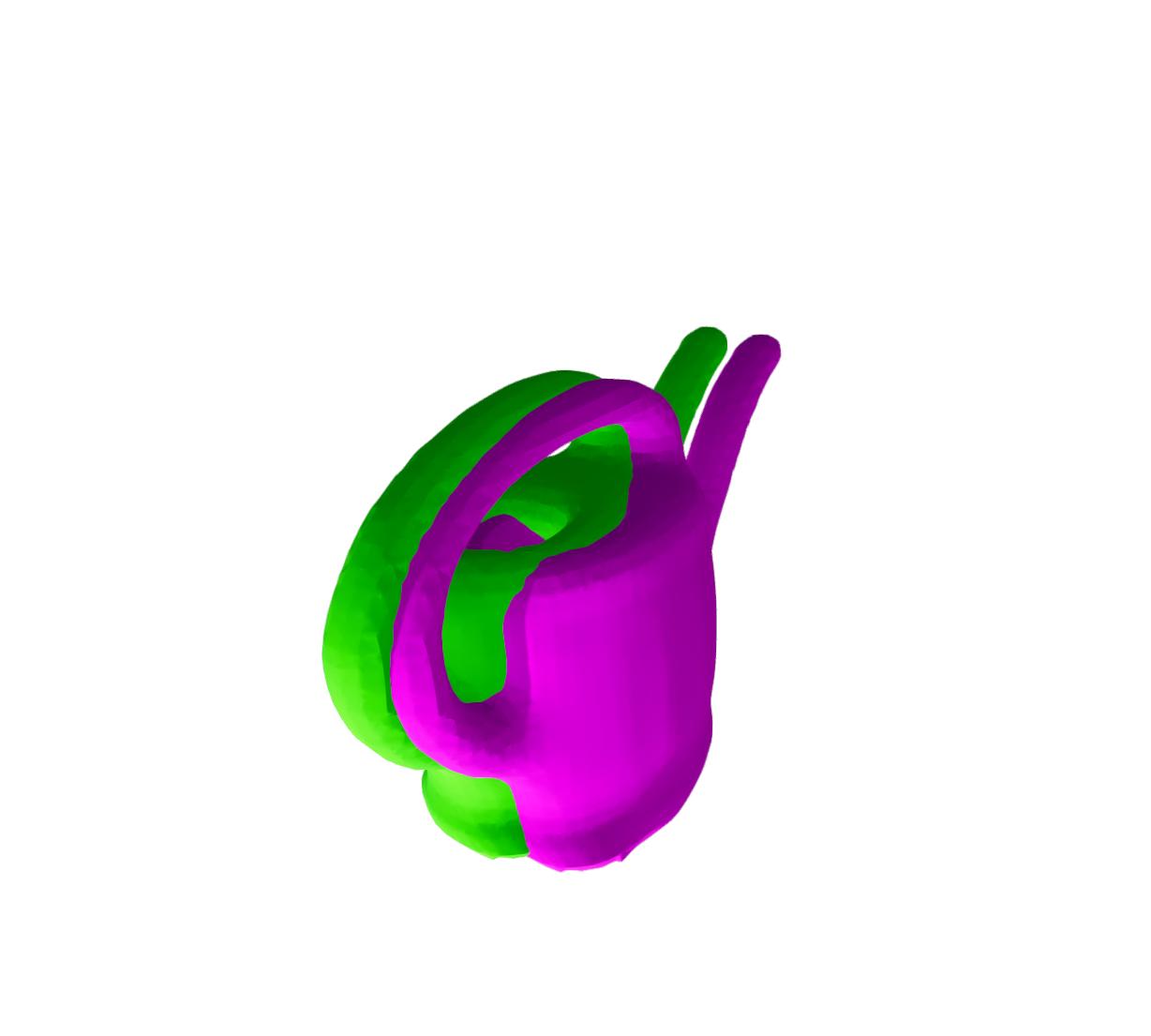}}\quad
  \subfloat{\includegraphics[height=.2\textwidth]{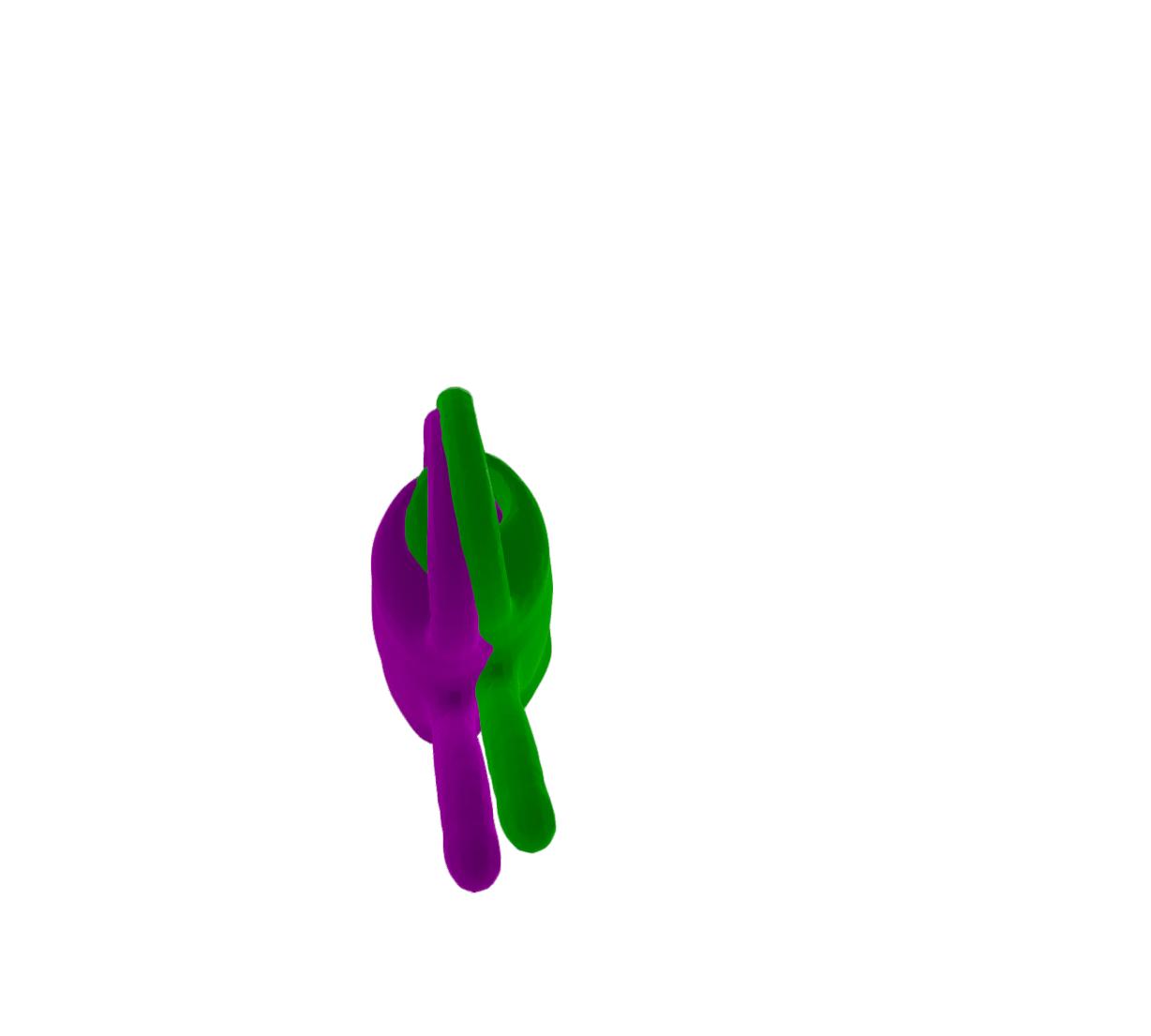}}\\

  \caption{TUD-L dataset visualization with \textcolor{magenta}{estimated} (magenta) 6D pose of the meshes and \textcolor{green}{ground-truth} (green) poses. The first column shows the test image with a contour of the projection made by the predicted pose. The other three columns show the corresponding 3D view from different viewing angles. The first is taken from approximately the same viewing angle as the image was taken.}
  \label{fig:A0_TUDL}
\end{figure*}

\begin{figure*}
  \centering
  \subfloat{\includegraphics[height=.2\textwidth]{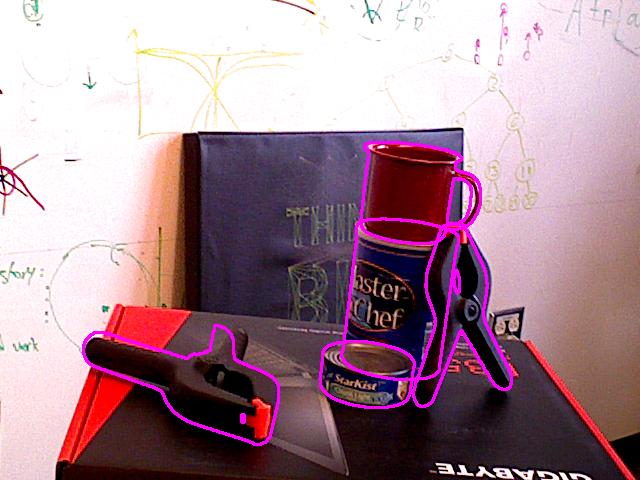}}\quad
  \subfloat{\includegraphics[height=.2\textwidth]{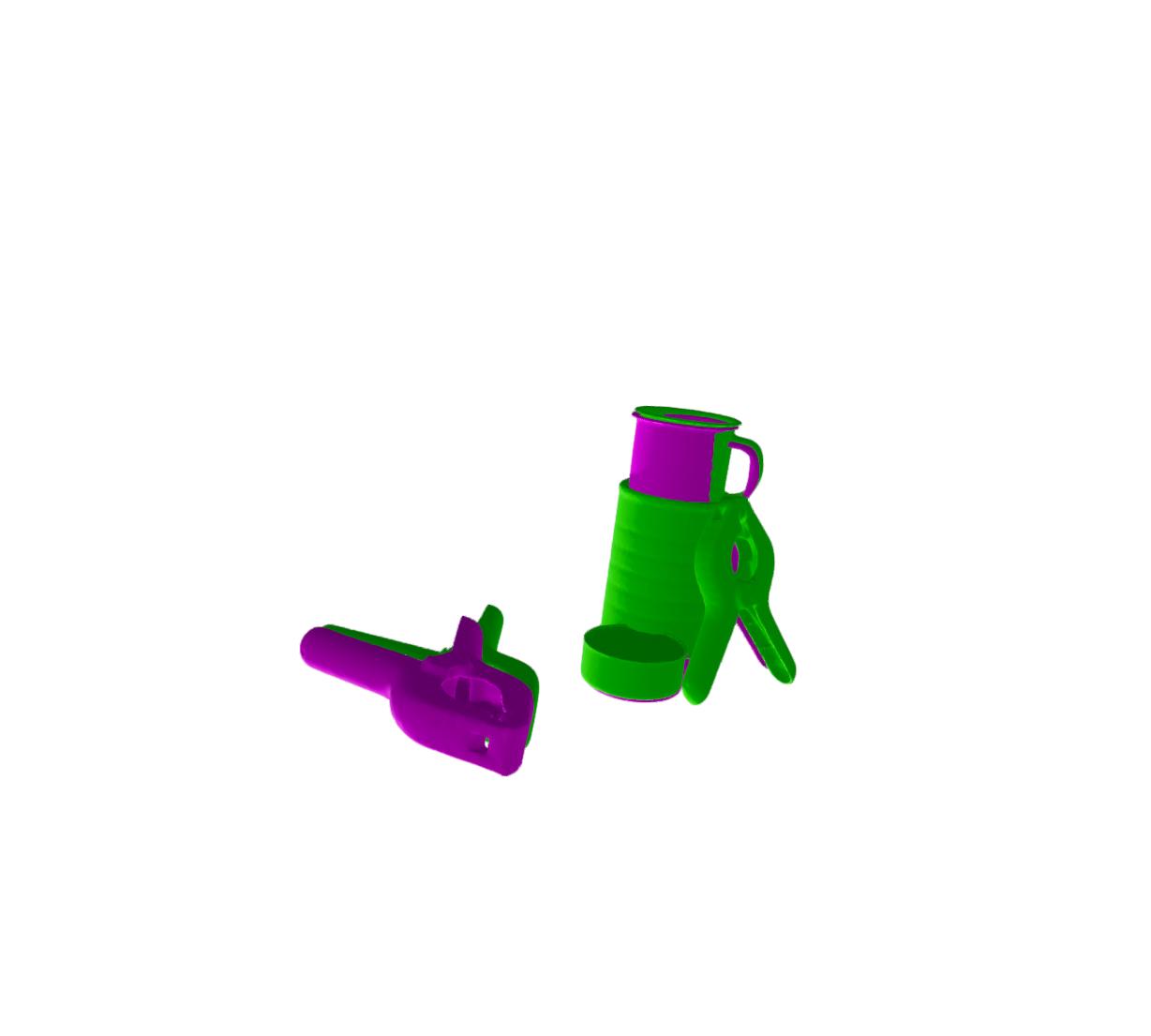}}\quad
  \subfloat{\includegraphics[height=.2\textwidth]{imgs/A0_imgs/ycbv_3Dvis_s000048_i001572_v00.jpg}}\quad
  \subfloat{\includegraphics[height=.2\textwidth]{imgs/A0_imgs/ycbv_3Dvis_s000048_i001572_v00.jpg}}\\
  
  \subfloat{\includegraphics[height=.2\textwidth]{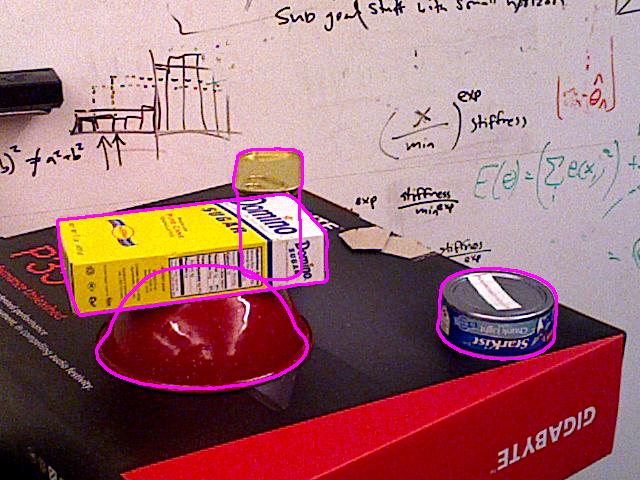}}\quad
  \subfloat{\includegraphics[height=.2\textwidth]{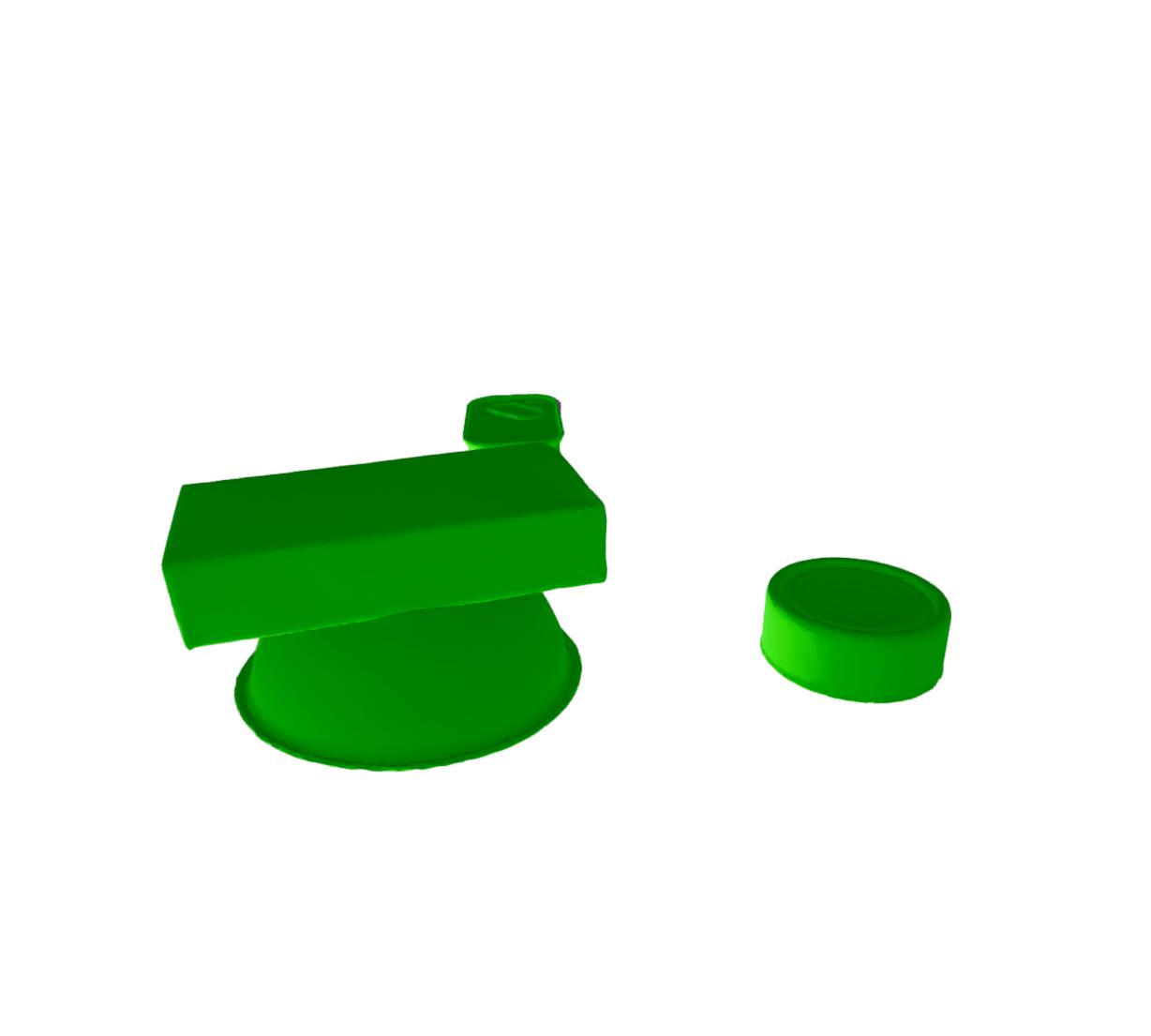}}\quad
  \subfloat{\includegraphics[height=.2\textwidth]{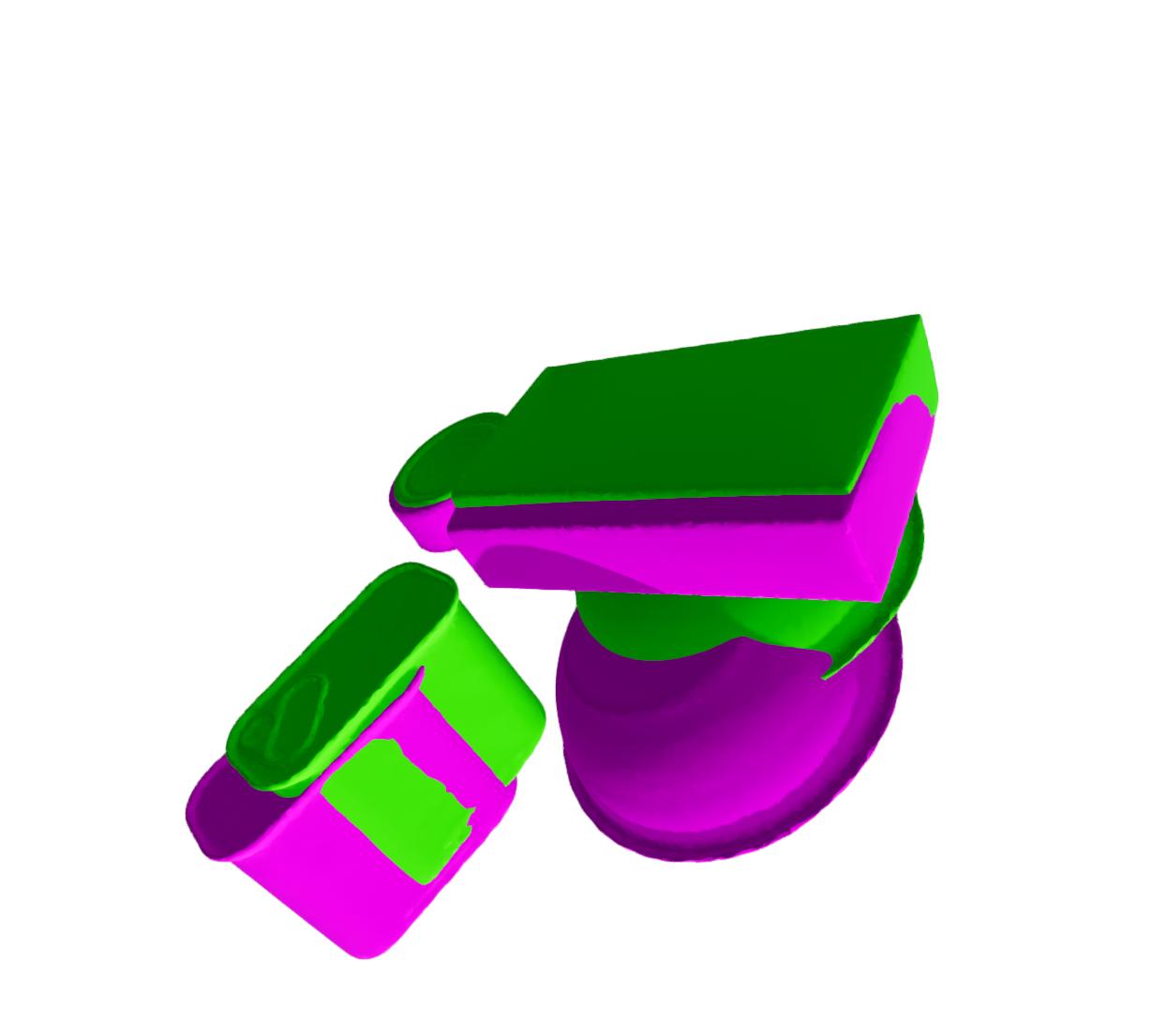}}\quad
  \subfloat{\includegraphics[height=.2\textwidth]{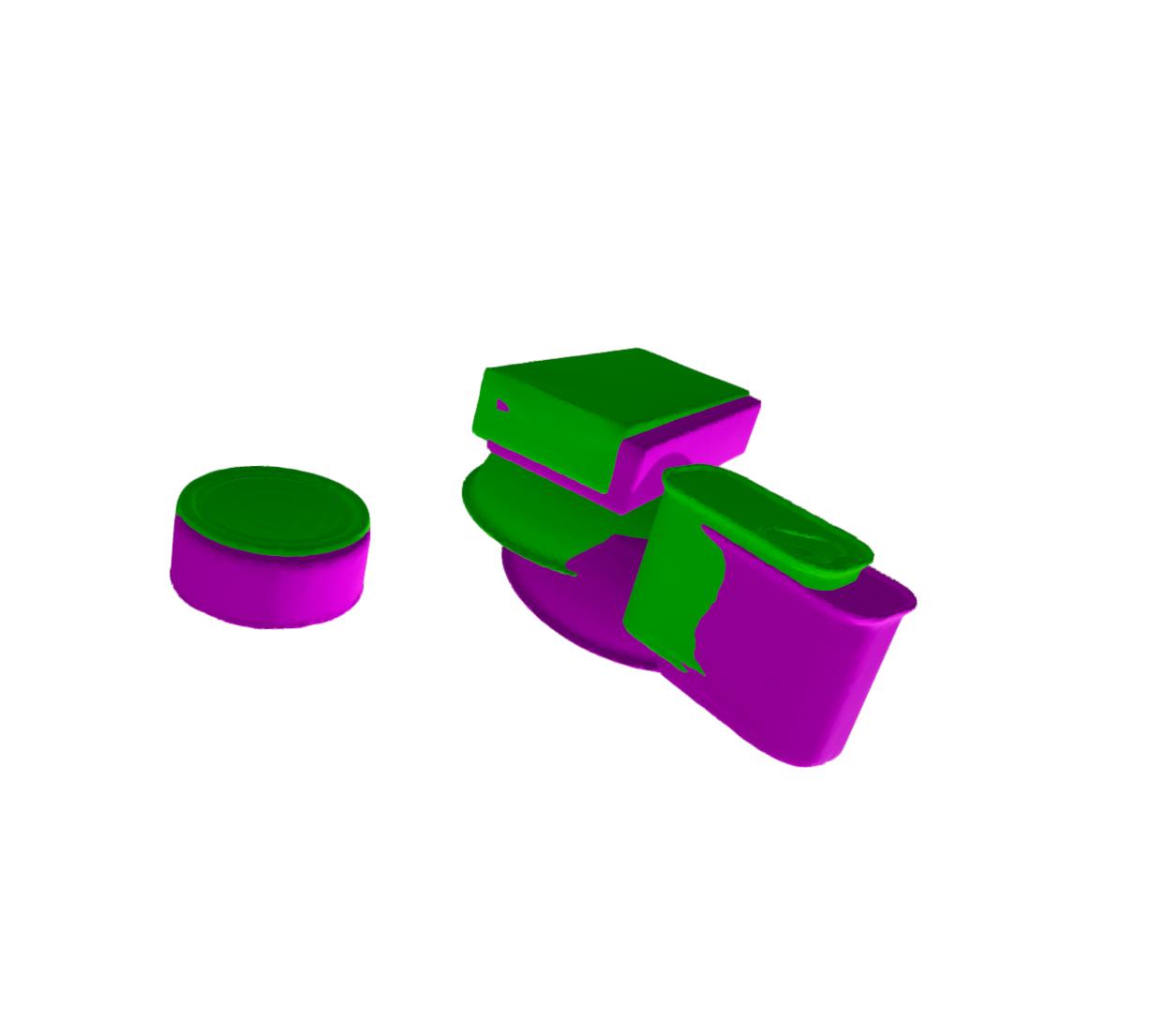}}\\
  
  \subfloat{\includegraphics[height=.2\textwidth]{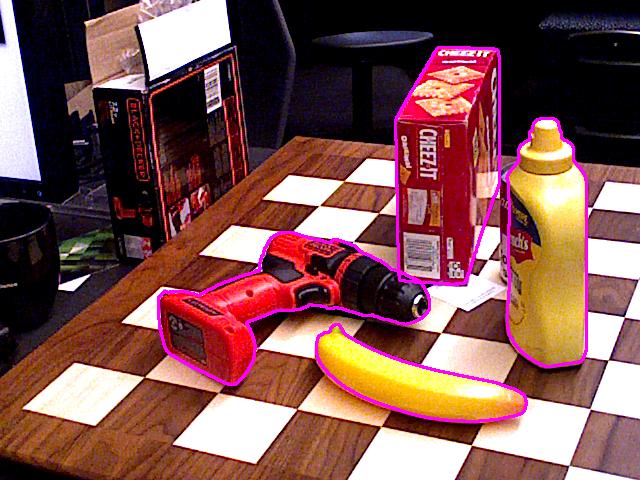}}\quad
  \subfloat{\includegraphics[height=.2\textwidth]{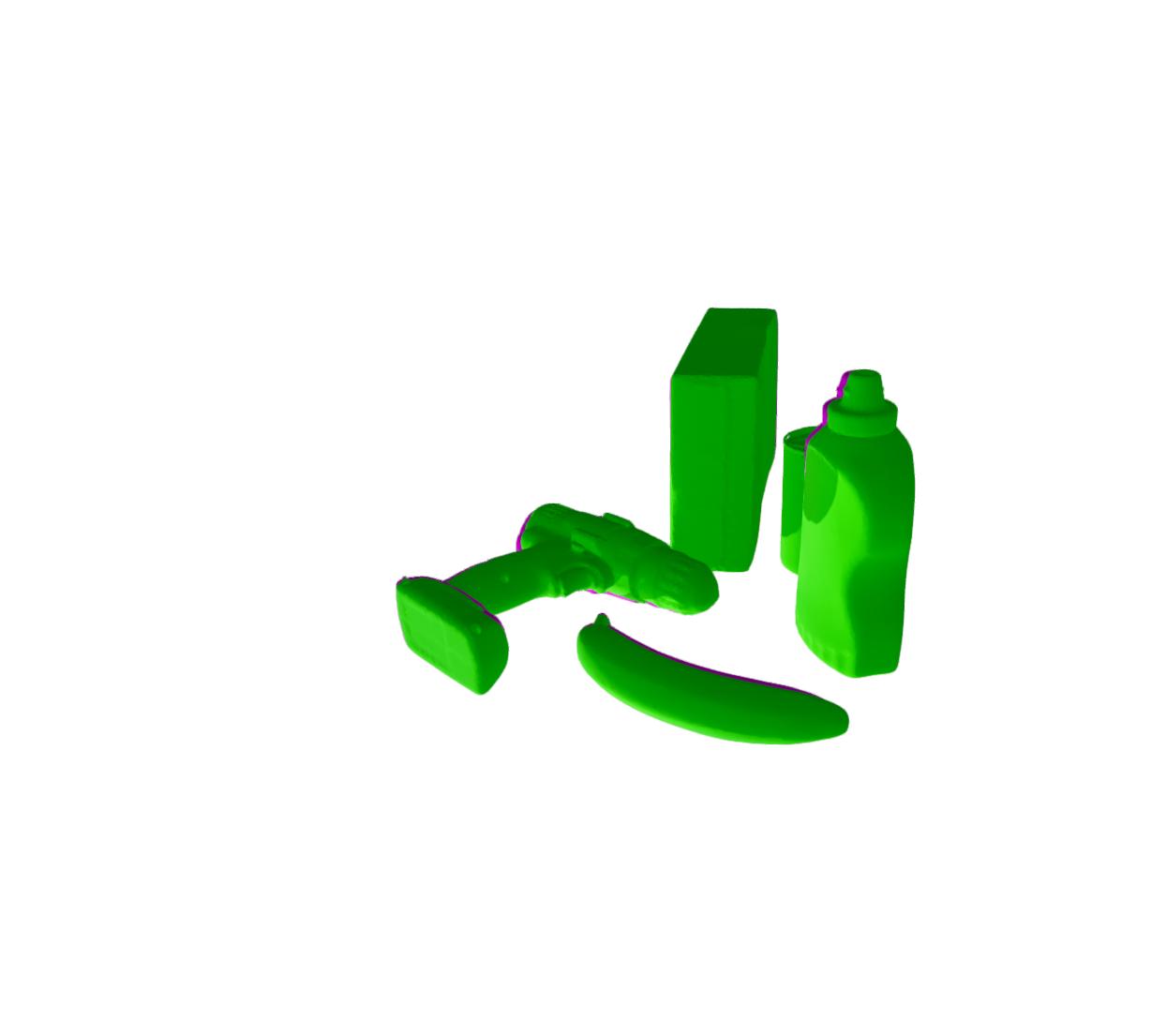}}\quad
  \subfloat{\includegraphics[height=.2\textwidth]{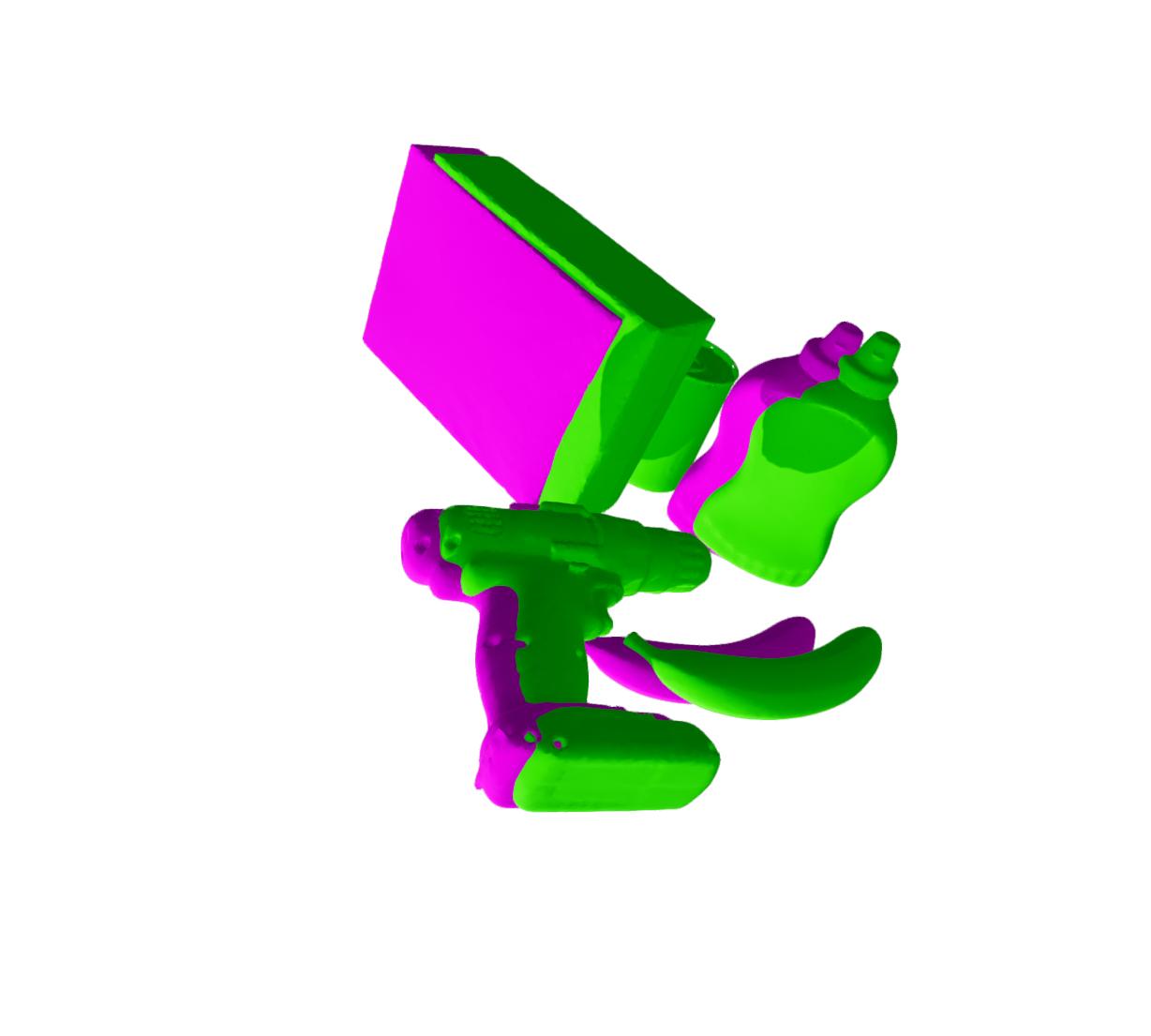}}\quad
  \subfloat{\includegraphics[height=.2\textwidth]{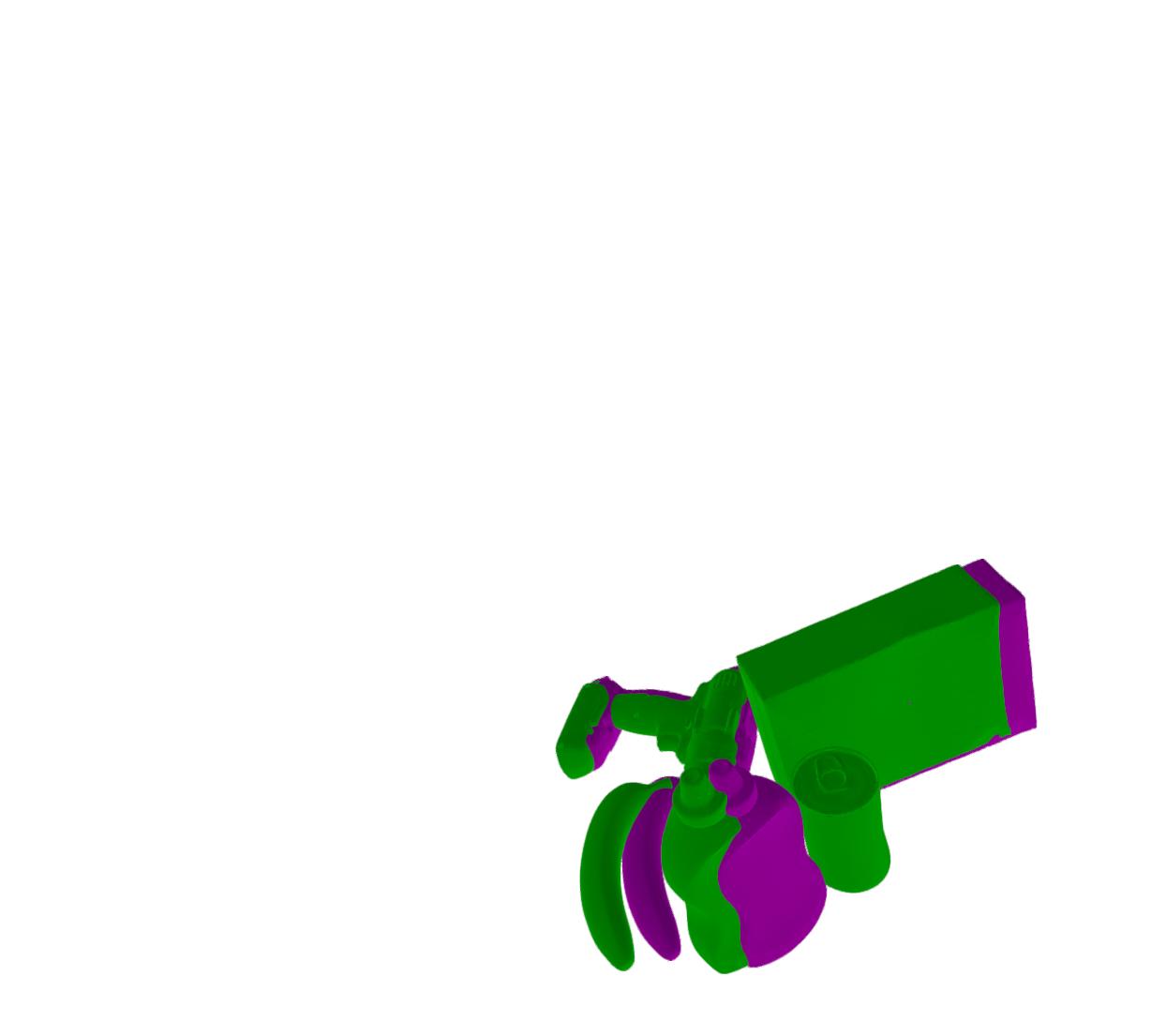}}\\
  
  \subfloat{\includegraphics[height=.2\textwidth]{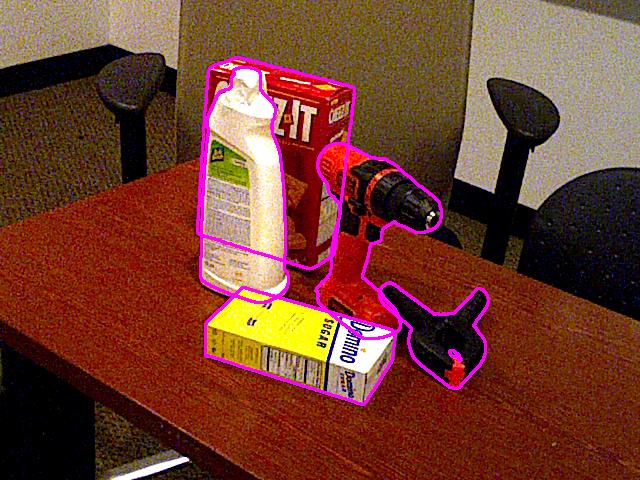}}\quad
  \subfloat{\includegraphics[height=.2\textwidth]{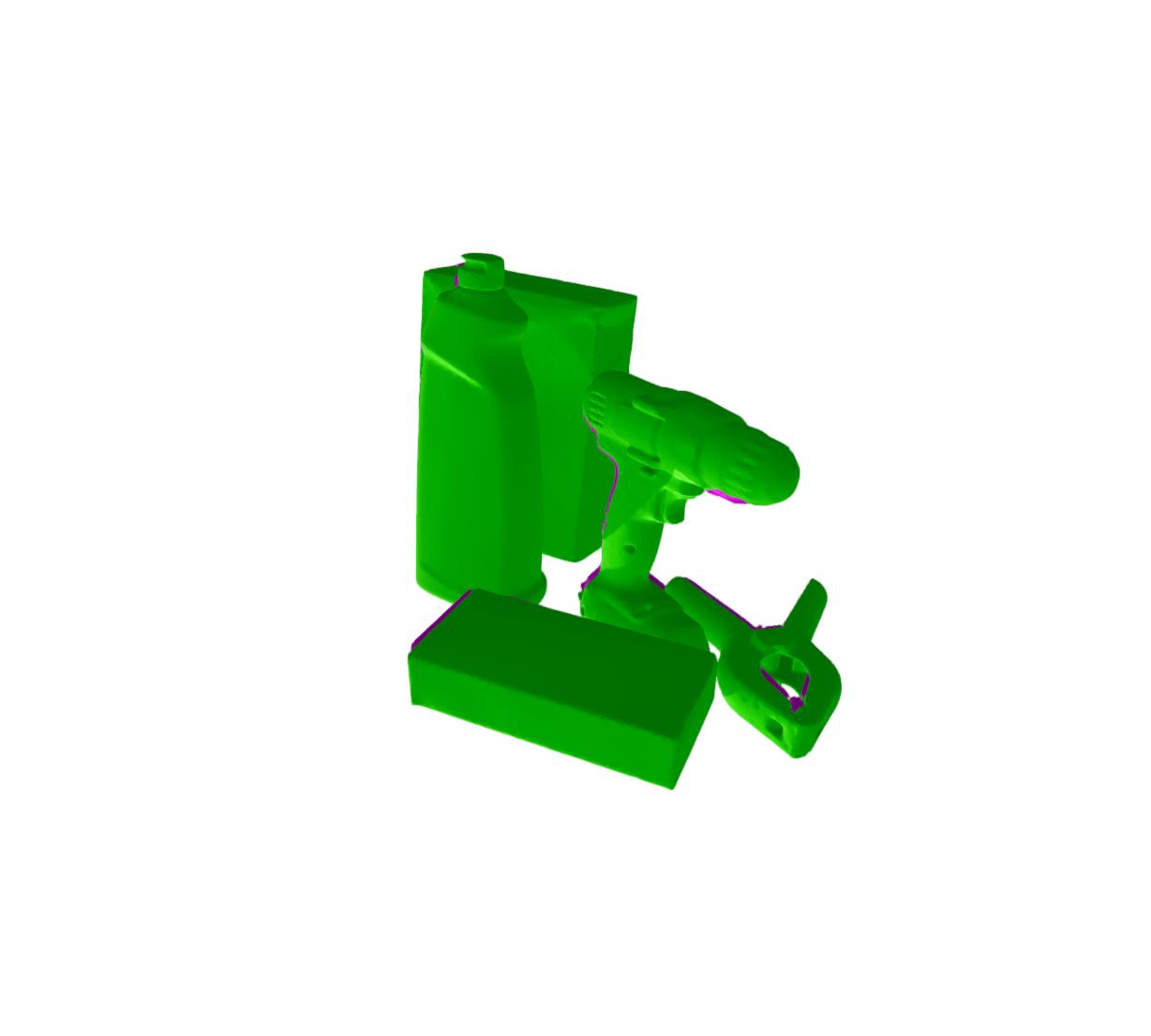}}\quad
  \subfloat{\includegraphics[height=.2\textwidth]{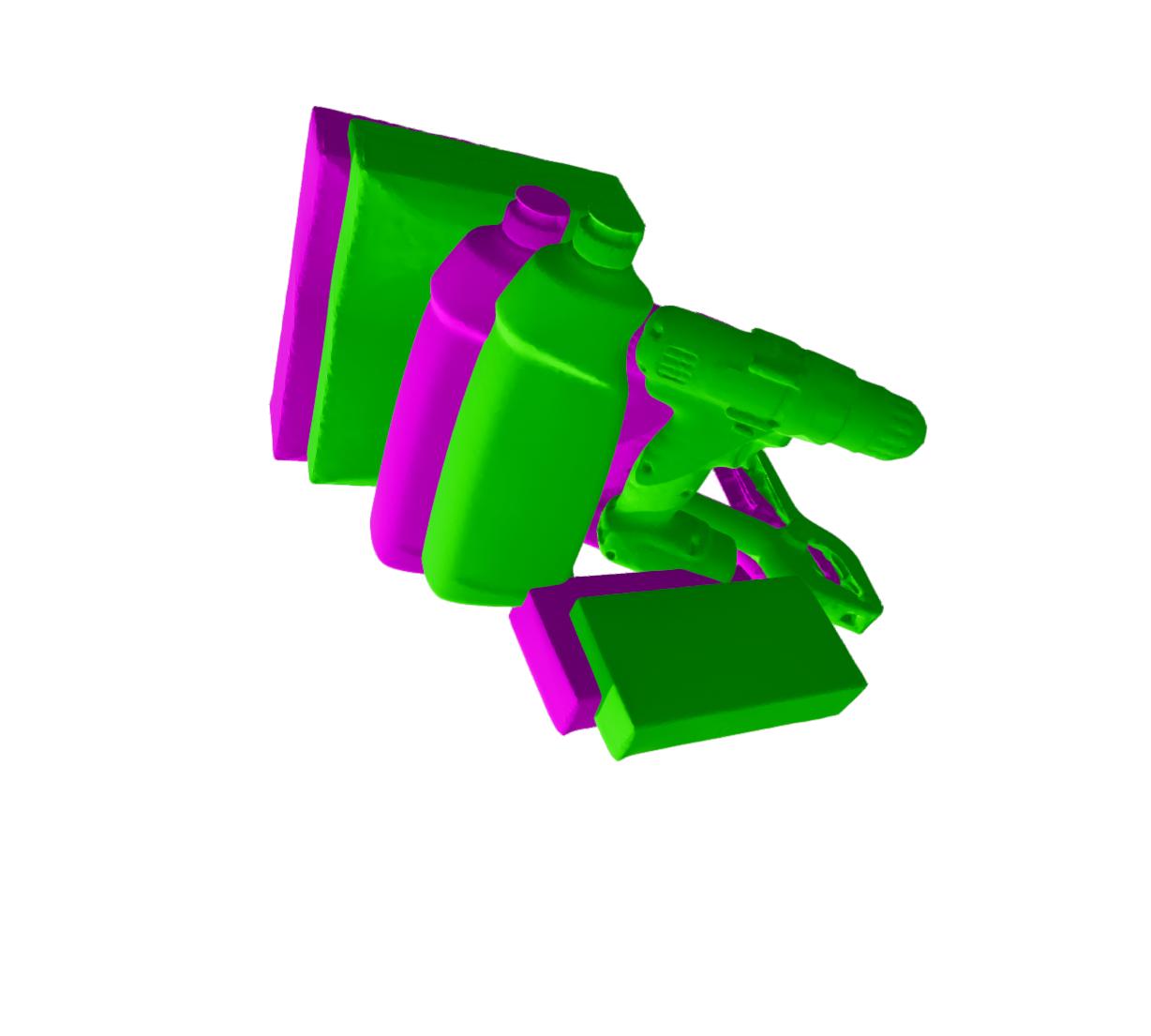}}\quad
  \subfloat{\includegraphics[height=.2\textwidth]{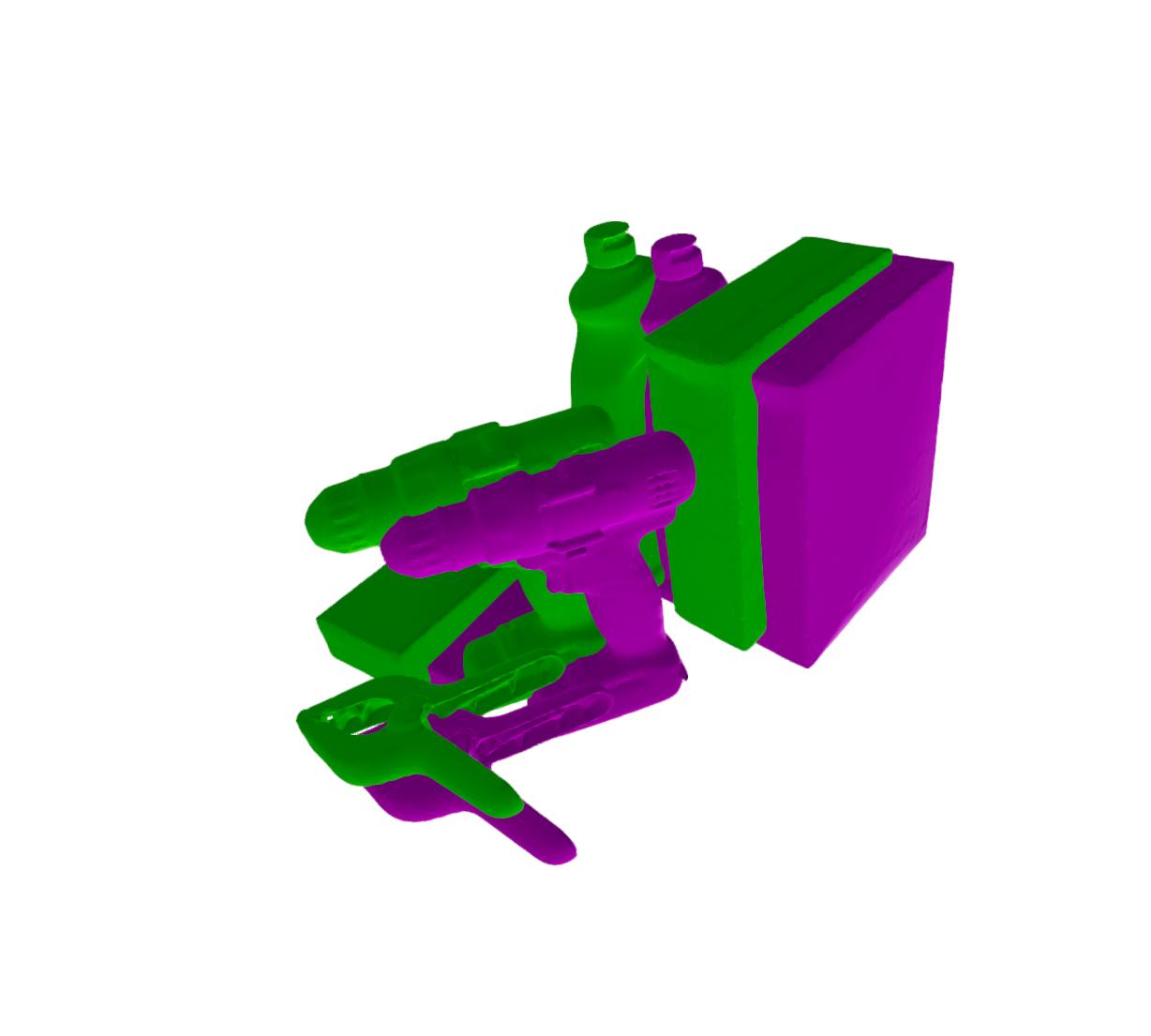}}\\
  
  \subfloat{\includegraphics[height=.2\textwidth]{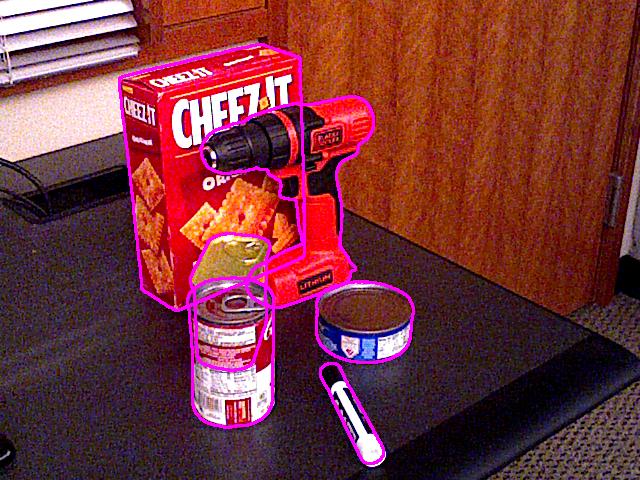}}\quad
  \subfloat{\includegraphics[height=.2\textwidth]{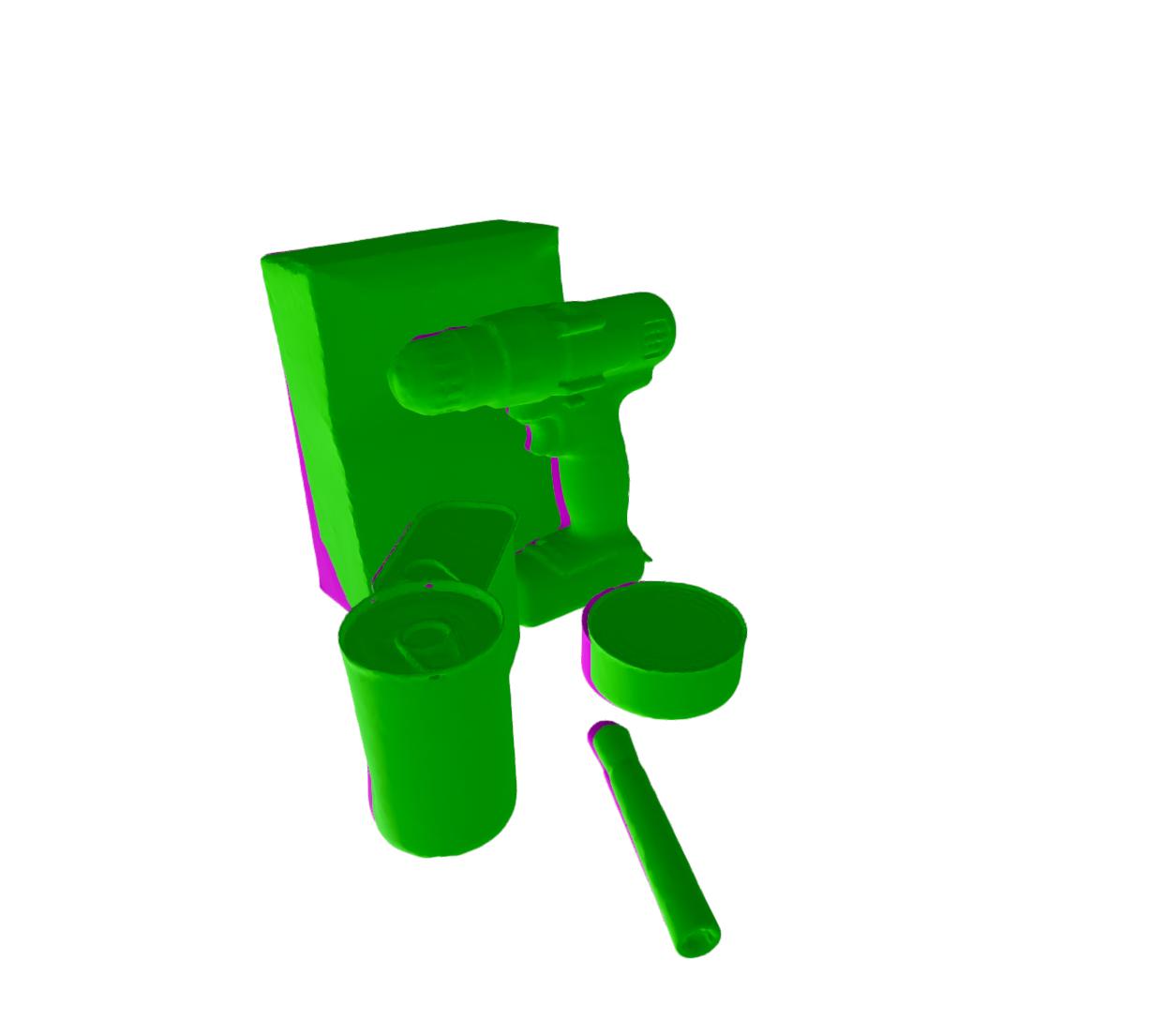}}\quad
  \subfloat{\includegraphics[height=.2\textwidth]{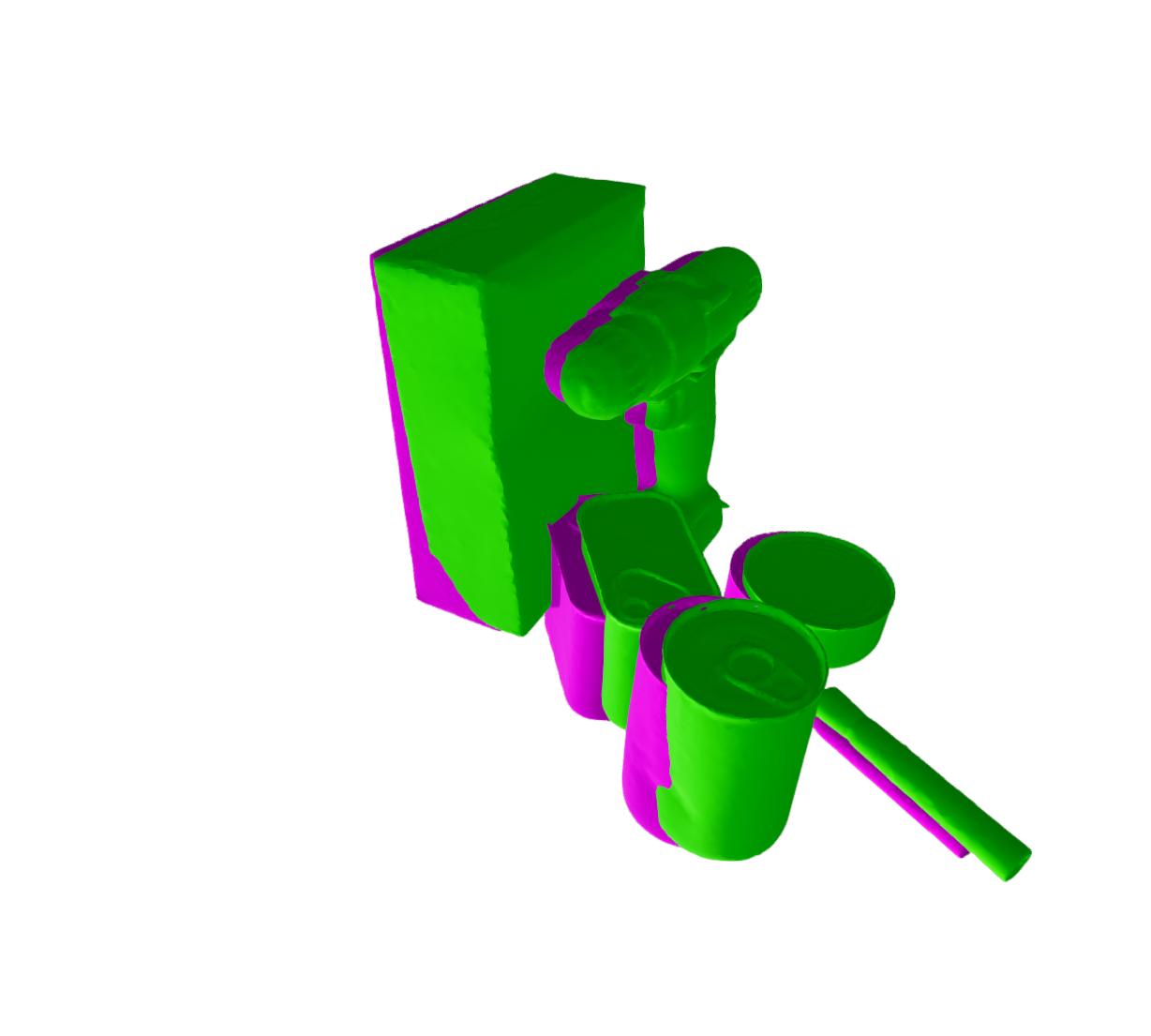}}\quad
  \subfloat{\includegraphics[height=.2\textwidth]{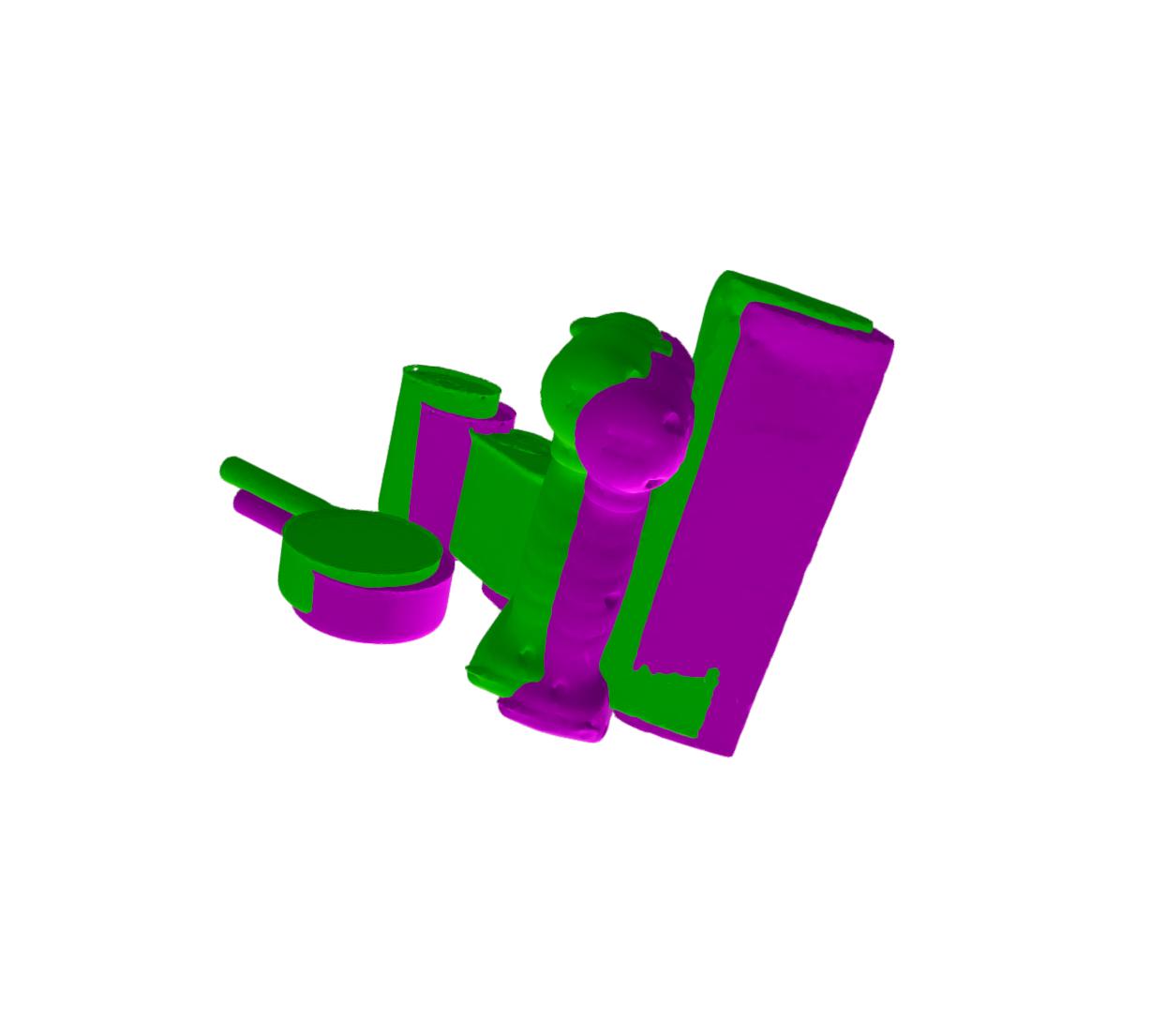}}\\

  \caption{YCB-V dataset visualization with \textcolor{magenta}{estimated} (magenta) 6D pose of the meshes and \textcolor{green}{ground-truth} (green) poses. The first column shows the test image with a contour of the projection made by the predicted pose. The other three columns show the corresponding 3D view from different viewing angles. The first is taken from approximately the same viewing angle as the image was taken.}
  \label{fig:A0_YCBV}
\end{figure*}

We provide additional visualization results for all BOP-Classic-Core datasets and the HANDAL and HOPEv2 datasets from BOP-H3 set, for which we report the results in \cref{sec:results-res}.
Since the HANDAL dataset lacks both depth and the ground-truth poses, providing 3D visualization would be uninformative, as only the predictions would be presented.
Therefore, we present only the 2D projection  visualization, as shown in \cref{fig:A0_HANDAL}.
In case of the HOPEv2 (\cref{fig:A0_HOPEv2}), HB (\cref{fig:A0_HB}), and ITODD (\cref{fig:A0_ITODD}) datasets, which do not provide ground-truth 6D pose but do contain depth, we visualize the estimated poses with the colored point cloud. 
The remaining BOP-classic core datasets, which provide the ground-truth poses, are visualized in figures \ref{fig:A0_ICBIN} - \ref{fig:A0_YCBV}.

In these figures, all ground-truth poses present in each scene are presented, and non-visible or highly occluded models in the inference image are not omitted.
This explains, in part, why, especially in the bin scenarios, there are more ground-truth models present than in the estimated meshes.
 Another reason is erroneous 2D detection, as mentioned in \cref{sec:A_2D},  which also leads to estimations where the detected mesh is positioned far from others (e.g., cases where the mesh is predicted to be under the floor).
In a few images, we observe the mesh being directly in front of camera; this is caused by the final stage of our 6D pose estimation pipeline, where RANSAC together with PnP algorithm fails and returns default prediction.

Notably, the visualization of both the 2D projections and the 3D views demonstrates the complexity of the 6D pose estimation from a single RGB image (mentioned in \cref{sec:A_Limits}), where minor pixel distance variations can result in more substantial Euclidean distance errors.
This observation is supported by findings from \cref{tab:A_BOP23} and \cref{tab:A_BOP24}, as MSPD and MSSD scores, although not directly comparable, suggest a preference for the projection error metric.



\end{document}